國立臺灣大學電機資訊學院資訊工程學系

博士論文

Department of Computer Science and Information Engineering

College of Electrical Engineering and Computer Science

National Taiwan University

Doctoral Dissertation

賦能大型語言模型多領域資源挑戰

Generalizing Large Language Model Usability Across Resource-Constrained

蔡昀達

Yun-Da Tsai

指導教授: 林守德 博士

Advisor: Shou-De Lin, Ph.D.

中華民國 114 年 4 月

April, 2025

# 國立臺灣大學博士學位論文
# 口試委員會審定書
## PhD DISSERTATION ACCEPTANCE CERTIFICATE
## NATIONAL TAIWAN UNIVERSITY

## 賦能大型語言模型多領域資源挑戰

## Generalizing Large Language Model Usability Across Resource-Constrained

本論文係蔡昀達君（學號 F08946007）在國立臺灣大學資訊工程學系完成之博士學位論文，於民國 114 年 04 月 28 日承下列考試委員審查通過及口試及格，特此證明。

The undersigned, appointed by the Department of Computer Science and Information Engineering on 28 April 2025 have examined a PhD dissertation entitled above presented by YUNDA TSAI (student ID: F08946007) candidate and hereby certify that it is worthy of acceptance.

口試委員 Oral examination committee:

_______________________   _______________________   _______________________
（指導教授 Advisor）

_______________________   _______________________   _______________________
Shang-Tse Chen (Apr 28, 2025 16:58 GMT+8)

系主任/所長 Director: _______________________

# 誌謝

在完成這篇論文的此刻，心中充滿了喜悅與不捨。博士求學的旅程，有挑戰、有掙扎，也有成長與收穫，而這一路走來，若沒有許多人的陪伴與支持，我無法走到今天。

首先，我要感謝我的指導教授林守德老師。老師在學術上的引領與人生上的鼓勵，始終是我堅持下去的重要力量。老師不僅教會我如何追尋學問，更教會我如何成為一位真正的研究者。

感謝我的口試委員林軒田老師、李宏毅老師、陳縕儂老師、陳尚澤老師、孫劭華老師與廖耿德學長，感謝各位撥冗審閱我的論文，並給予我許多寶貴的建議與啟發，讓這份工作能夠更加完善。

感謝所有在這段旅程中與我同行的夥伴們。特別感謝 MSLab 的郭沛孚、顏廷聿、吳耘安與蔡育哲，謝謝你們在研究上一起努力奮鬥，讓實驗室成為一個充滿活力的地方。感謝在 NVIDIA Research 實習期間遇到的每一位同事，特別是導師刘鸣杰、主管任昊星、邓晨晖、何嘉桐、劉冠廷、杨浩宇與劉文皓，謝謝你們在研究領域的指導與啟發。也感謝在 Appier 實習時的主管陳耀男，謝謝你在研究上的指導。

最深的感謝，獻給我的家人。感謝你們無條件的愛與支持，是你們讓我有勇氣面對每一個挑戰。特別感謝我的太太王郁婷，在每一段艱難時光中，總是默默

陪伴在我身旁，給予我最深的安慰與力量。能夠與你一同走過這段旅程，是我最珍貴的幸運。

最後，感謝所有曾經在生命中留下溫暖足跡的朋友們。謝謝你們在我迷惘時給予鼓勵，在我低潮時給予陪伴，在我努力時為我加油。因為有你們，這段旅程才如此豐富而有意義。

這篇論文不只是我的努力結晶，更是許多人愛與支持的累積。感謝你們成為我人生中重要的一部分。

# Acknowledgements

Completing this dissertation has been a challenging yet immensely rewarding journey, and it would not have been possible without the support and encouragement of many individuals.

First and foremost, I would like to express my deepest gratitude to my advisor, Shou-De Lin, for his invaluable guidance, unwavering support, and insightful advice throughout my doctoral studies. The mentorship has profoundly shaped both my research and my professional growth.

I am also deeply thankful to my committee members, Hsuan-Tien Lin, Hung-Yi Lee, Yun-Nung Chen, Shang-Tse Chen, Shao-Hua Sun, and Keng-Te Liao for their constructive feedback, encouragement, and time. Their expertise and perspectives have greatly strengthened this work.

I am grateful to my collaborators and colleagues, whose discussions, suggestions, and camaraderie have greatly enriched my research experience. In particular, I would like to thank my MSLab lab mates, Pei-Fu Guo, Ting-Yu Yen, Yun-Ang Wu, Yu-Che Tsai, and many others, for their generous support and hard work during collaboration. I am also thankful to my colleagues at NVIDIA Research during my internship, especially my mentor Mingjie Liu, my manager Haoxing Ren, and colleagues Chenhui Deng, Mark Ho, Danny Liu, Haoyu Yang, and Wen-Hao Liu for their valuable guidance and encouragement on LLM research. In addition, I would like to thank my colleagues at Appier, particularly my manager Yao-Nan Chen, for their guidance on bandit algorithm research during my internship.

On a personal note, I am forever indebted to my family for their endless love and patience. I am especially grateful to my wife Yu-Ting Wang for standing by me through every high and low. Their support has been my greatest source of strength.

Finally, to my friends, near and far, who kept me grounded, encouraged me during difficult times, and celebrated my milestones —thank you for making this journey all the more meaningful.

This dissertation is as much a reflection of the contributions of those around me as it is of my own efforts. I am deeply grateful to all who have been part of this chapter of my life.

# 摘要

大型語言模型（Large Language Models, LLMs）在各類自然語言任務上取得了顯著成果，近年來亦積極拓展至多模態領域與資源受限環境。然而，現有方法多仰賴高成本的監督式微調，或假設訓練與推論條件相同，因而在面對未見模態、有限資料或計算資源受限情境時，泛化能力仍存在顯著侷限。

本論文系統性地探討提升大型語言模型在現實環境中可用性的途徑，聚焦於泛化能力與資源限制下的適應性。首先，提出一套以文字為中心的多模態對齊框架，將文本、圖像、表格及波形等異質模態轉換為自然語言描述，使模型能夠透過即時提示學習（in-context learning）應對未見或動態變化的模態組合，無需重新訓練。為強化模型在面對噪聲或缺失模態時的魯棒性，本論文亦設計出對抗式提示（adversarial prompting）技術，於提示層級生成語意挑戰性高的擾動資料，以提升模型韌性。

除多模態對齊外，論文亦探討推論階段最佳化策略，透過提示搜尋與不確定性量化，於無需額外訓練的情況下提升模型效能，為傳統擴大參數規模或重訓練以外，提供另一種高效途徑。同時，本研究針對資源稀缺領域如 Verilog 程式碼生成，設計出正確性保證的合成資料生成流程及邏輯增強型推理模型，於有限資料條件下達成最新最佳表現。

綜合上述，本論文提出的方法在對齊、最佳化與合成資料生成三大面向上，

皆展現了在不同模態、資源限制與應用場景下，顯著提升大型語言模型適用性、擴展性與效率的潛力。

# Abstract

Large Language Models (LLMs) have achieved remarkable success across a wide range of natural language tasks, and recent efforts have sought to extend their capabilities to multimodal domains and resource-constrained environments. However, existing approaches often rely on costly supervised fine-tuning or assume fixed training conditions, limiting their generalization when facing unseen modalities, limited data, or restricted compute resources.

This dissertation presents a systematic study toward generalizing LLM usability under real-world constraints. First, it introduces a robust text-centric alignment framework that enables LLMs to seamlessly integrate diverse modalities—including text, images, tables, and any modalities —via natural language interfaces. This approach supports in-context adaptation to unseen or dynamically changing modalities without requiring retraining. To enhance robustness against noisy and missing modalities, an adversarial prompting technique is proposed, generating semantically challenging perturbations at the prompt level to stress-test model reliability.

Beyond multimodal setting, the dissertation investigates inference-time optimization strategies for LLMs, leveraging prompt search and uncertainty quantification to improve performance without additional model training. This perspective offers an efficient alternative to scaling model parameters or retraining from scratch. Additionally, the work addresses low-resource domains such as Verilog code generation by designing correct-by-construction synthetic data pipelines and logic-enhanced reasoning models, achieving state-of-the-art performance with minimal data.

Together, these contributions form a unified effort to enhance the adaptability, scalability, and efficiency of large language models under practical constraints. The results demonstrate that principled methods for alignment, optimization, and synthetic data generation can significantly broaden the usability of LLMs across modalities, resource regimes, and application domains.

# Contents

# List of Figures

# List of Tables

# Chapter 1   Introduction

Large language models (LLMs) have achieved remarkable success across a wide range of tasks, from natural language understanding and code generation to vision-language reasoning and symbolic computation. However, deploying these models in real-world settings remains challenging due to practical resource constraints. LLMs are often trained on massive datasets with substantial computational budgets, yet real-world applications demand flexibility: operating across unseen modalities, generalizing from limited or noisy data, and adapting to low-resource domains where fine-tuning is infeasible.

This dissertation addresses the central question: How can we enhance the generalization and usability of LLMs under constrained data, compute, and modality conditions? Through a unified investigation spanning multimodal alignment, inference-time optimization, code generation, and reinforcement learning for hardware-oriented tasks, this work proposes scalable strategies for building more adaptable, robust LLMs capable of operating effectively even when ideal training conditions are absent.

## 1.1 Motivation

As LLMs become integral components of intelligent systems, their limitations in flexibility and resource efficiency are increasingly exposed. Three major gaps motivate this dissertation:

First, multimodal models often assume fixed modality configurations at both training and inference time. However, real-world systems frequently encounter missing, noisy, or dynamically varying modalities—a problem known as modality mismatch. Developing LLMs that can generalize across arbitrary modality combinations without retraining remains an open challenge.

Second, while scaling laws indicate that model performance improves with increased parameters and data, achieving such scale demands immense resources. In many applications—such as hardware design, medicine, or low-resource languages—such resources are unavailable. Inference-time optimization methods offer an alternative pathway by improving robustness and reasoning ability without retraining.

Third, the domain of structured symbolic reasoning, particularly in hardware logic and Verilog code generation, presents unique difficulties. High functional correctness is required, but labeled data is scarce, and error analysis is complex. Reinforcement learning and reasoning model techniques, successful in math and programming, remain underexplored for hardware-oriented domains.

Together, these challenges reveal a critical need: to rethink how LLMs can generalize effectively, not through scale alone, but through methods that adapt intelligently to resource constraints.

## 1.2 Research Scope

This dissertation focuses on enhancing LLM generalization and usability in three complementary resource-constrained dimensions:

- Modality Generalization under Limited or Mismatched Inputs: Designing text-centric multimodal alignment frameworks that allow LLMs to handle missing, noisy, or dynamically structured modality inputs via in-context learning.

- Robustness Optimization without Retraining: Developing inference-time strategies—including adversarial prompting, uncertainty calibration, and chain-of-thought augmentation—to improve model resilience without incurring additional training costs.

- Low-Resource Code and Hardware Logic Generation: Addressing the scarcity of high-quality data for Verilog and RTL design tasks by generating correct-by-construction synthetic datasets, targeted code repair data, and reinforcement-learning-based symbolic reasoning models.

The dissertation bridges across domains—multimodal learning, natural language processing, symbolic logic, and hardware-oriented reasoning—to propose methods that prioritize adaptability and resource efficiency as first-class goals.

## 1.3 Contributions

The contributions of this dissertation can be summarized as follows:

- Text-Centric Multimodal Alignment Framework: Proposed a robust in-context learning framework for modality mismatch, enabling LLMs to generalize across missing, noisy, and unseen modality configurations without retraining.

- Inference-Time Optimization Techniques: Introduced scalable methods such as adversarial prompting and uncertainty calibration to improve model robustness under data noise and dynamic modality conditions at inference.

- Efficient Code Generation via Data Curation: Developed correct-by-construction synthetic data pipelines and targeted code repair strategies to enhance Verilog code generation in low-resource settings.

- Reinforcement Learning for RTL Reasoning: Created the first RTL-oriented reasoning framework by adapting reinforcement learning techniques to structured logic sub-tasks, significantly improving performance in hardware logic synthesis.

- Unified View of Generalization under Constraints: Across all domains, this dissertation demonstrates that strategic data design, inference-time reasoning, and symbolic task decomposition provide viable alternatives to large-scale retraining, laying foundations for more practical LLM deployment.

## 1.4 Chapter Outlines

The remainder of this paper is organized as follows:

- Chapter 2 reviews the foundations of large language models, scaling laws, generalization across modalities, resources, and symbolic domains, and surveys related research.

- Chapter 3 introduces the robust text-centric multimodal alignment framework, detailing the modality mismatch problem, adversarial prompting techniques, experimental evaluations, and efficiency analysis.

- Chapter 5 presents methods for efficient data pruning and synthetic data generation to improve code LLMs, focusing on Verilog generation tasks.

- Chapter 6 develops an RTL-oriented reasoning framework using reinforcement learning, introducing structured symbolic sub-tasks and evaluating improvements in hardware code generation.

- Chapter 7 concludes the dissertation with a summary of contributions, lessons learned, limitations, and future research directions toward adaptable, resource-efficient LLMs.

# Chapter 2  Foundations of Generalization and Resource Constraints in Large Language Models

In this chapter, we begin by briefly reviewing the foundations of large language models (LLMs), particularly focusing on their architectural design and scaling behaviors. We then discuss the multidimensional nature of generalization in LLMs, including across modalities, resources, and symbolic domains. Finally, we survey related research in each of these dimensions, providing context for the methods proposed in later chapters.

## 2.1  Large Language Models and Scaling Laws

Large Language Models (LLMs) have achieved significant advancements by leveraging scale. Empirical studies have shown that model performance improves in a predictable manner as three fundamental factors increase: the number of model parameters ($N$), the size of the training dataset ($D$), and the amount of training compute ($C$). These trends are formalized in what are now widely referred to as scaling laws [45, 52].

### 2.1.1 Training-Time Scaling Laws

Training-time scaling laws model how the training or validation loss $L$ decreases with increasing scale. The relationship can be expressed as a power-law:

$$L(N, D) = L_\infty + aN^{-\alpha} + bD^{-\beta} \qquad (2.1)$$

Here, $L_\infty$ represents the irreducible loss (the loss floor), $N$ is the model size (number of non-embedding parameters), and $D$ is the number of training tokens. The constants $a, b$ are fitted coefficients, while $\alpha, \beta$ are scaling exponents that depend on the architecture and domain.

Key observations from the original studies include:

- Increasing either $N$ or $D$ improves performance, but only up to the limits imposed by the other. For example, increasing model size without sufficient data leads to overfitting.

- Training compute $C \sim ND$ must be budgeted carefully—optimal scaling lies on a compute frontier where both model and data size are balanced to maximize performance.

- Scaling behavior is consistent across language modeling, vision, and multi-modal settings.

These findings guide how modern LLMs are designed and trained under approximately compute-optimal scaling conditions.

## 2.1.2 Inference-Time Scaling Laws

More recently, Snell et al. [117] introduced the concept of inference-time scaling, showing that increasing computation at test time—without changing model parameters—can improve performance significantly.

Inference-time enhancements include:

- Sampling multiple completions and aggregating results (e.g., majority voting in self-consistency).

- Chain-of-thought prompting to elicit intermediate reasoning steps [149].

- Longer prompt contexts, including few-shot demonstrations or task instructions.

- Beam search, reranking, or ensembling at generation time.

While these techniques increase token-level compute or latency, they can yield improvements comparable to larger models trained from scratch. This makes inference-time scaling an attractive alternative in resource-constrained settings.

## 2.1.3 Implications for Generalization

The practical insight from scaling laws is that generalization is not solely a function of parameter count. Strategic use of data, compute, and inference methods can yield competitive or even superior results at a fraction of the cost. This dissertation leverages these ideas across multiple chapters:

- Inference-time optimization via prompt search and uncertainty calibration (Chapter 4).

- Data-efficient fine-tuning using pruning and synthetic data (Chapter 5).

- Symbolic and reasoning-based generalization without massive pretraining (Chapter 6).

In summary, scaling laws not only offer a principled understanding of model performance, but also directly inform the design of efficient, domain-adapted strategies explored in this dissertation—enabling generalization under limited data, compute, and modality constraints.

## 2.2 Dimensions of Generalization and Resource Constraints

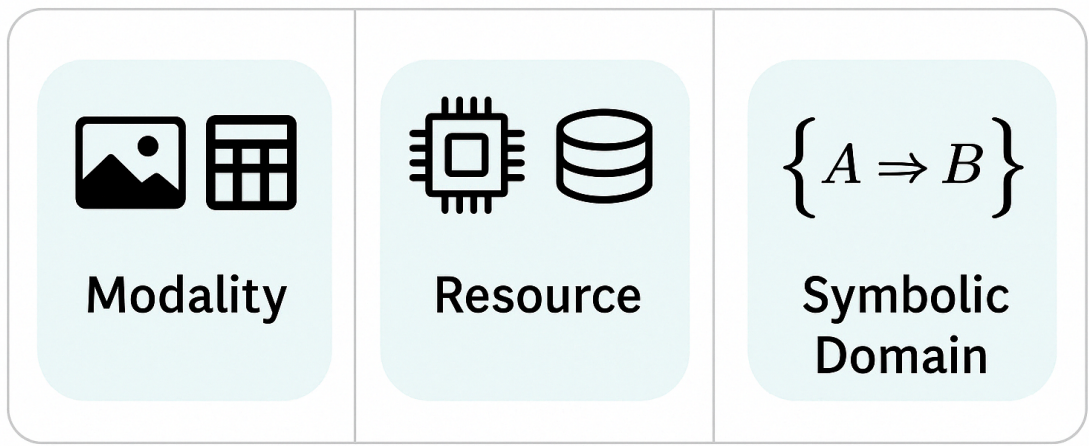

Figure 2.1: Three axes of generalization considered in this dissertation: modality, resource, and symbolic domain.

LLMs are increasingly deployed in diverse real-world applications that require them to generalize across different input modalities, tasks, and resource conditions. This dissertation explores generalization across three complementary axes: modality, resource, and symbolic domain.

### 2.2.1 Cross-Modality Generalization

The ability of LLMs to operate on non-textual modalities (e.g., images, tables, graphs) requires either fine-tuning with paired data or clever representations that unify modalities. Text-centric alignment has become a popular paradigm [61, 68, 138], wherein all modalities are translated into natural language, allowing the use of pre-trained language-only models in a zero-shot setting. This enables generalization to unseen modality combinations without retraining, as demonstrated in Chapter 3.

### 2.2.2 Resource-Constrained Generalization

In many domains (e.g., hardware, medicine, or low-resource languages), labeled data and compute resources are scarce. To address this, recent work explores:

- Data pruning to reduce fine-tuning cost without sacrificing model quality [19, 24, 62, 63, 74, 105, 144, 159].

- In-context learning and prompt engineering to bypass training entirely [12, 150].

- Synthetic data generation to bootstrap fine-tuning with minimal human annotation [33, 67, 116, 147, 152, 161, 170].

These strategies are discussed further in Chapters 5 and 4.

### 2.2.3 Generalization to Symbolic and Logical Domains

Despite impressive capabilities in natural language tasks, LLMs struggle with precise symbolic reasoning, especially in domains like formal logic, hardware description languages (HDLs), and structured programming. To generalize effectively, LLMs must not only output syntactically valid programs but also exhibit functional correctness. Recent approaches employ:

- Synthetic Data [130].

- Reward-based fine-tuning for correctness [7, 32, 108].

- Reinforcement learning for structured reasoning [39, 115, 151, 160, 172].

Understanding and enhancing generalization along these axes enables LLMs to be more robust and adaptable across real-world scenarios.

### 2.2.4 Summary and Implications

This section outlined three principal dimensions of generalization that this dissertation seeks to address: cross-modality generalization, resource-constrained generalization, and generalization to symbolic and logical domains. Each dimension reflects a key practical challenge in applying LLMs beyond benchmark datasets and large-scale pretraining environments.

Cross-modality generalization enables models to process diverse input types by aligning them in a unified semantic space—often via text-centric representations. Resource-constrained generalization focuses on improving model utility in low-data or low-compute scenarios, using methods such as prompt engineering, data pruning, and synthetic data generation. Finally, generalization in symbolic domains emphasizes structural correctness and logical reasoning, pushing beyond fluency toward functional reliability in tasks like code generation and formal logic.

Together, these three axes form a conceptual backbone for the methods developed in this dissertation. The subsequent chapters build upon this framing to propose scalable, domain-adaptable solutions that enhance LLM performance across these generalization boundaries.

## 2.3 Related Works

This section reviews prior work that underpins the research directions explored in this dissertation. We cover related studies in six primary areas: multimodal alignment, robustness, adversarial prompting, Verilog code generation, LLM-based reasoning, and data efficiency techniques such as pruning and fine-tuning.

### 2.3.1 Text-Centric Multimodal Alignment

Multimodal alignment has been a central challenge in extending the generalization capacity of LLMs beyond pure-text inputs. Recent research has explored using natural language as an intermediate representation to unify disparate input modalities. Models such as LLaVA [68] and VideoChat-Text [61] use captioning and transcription techniques to convert visual or video inputs into textual descriptions. In specialized domains, such as healthcare, systems like OphGLM [36] and ChatCAD [146] demonstrate the power of converting medical imagery into report-style text to facilitate LLM reasoning. TAMML [138] extends this idea to broader settings, proposing a text-centric alignment strategy for handling diverse modality combinations, especially in zero-shot inference scenarios. While effective, the performance of these systems is tightly coupled to the fidelity and diversity of their text representations.

### 2.3.2 Robustness in Multimodal Learning

A growing body of work highlights the importance of robustness in multimodal systems, particularly in the face of noise or missing modalities. Ma et al. [83] introduced a taxonomy for modality robustness, emphasizing realistic perturbations like noisy inputs, dropped modalities, or dynamic changes. However, even in text-aligned setups, robustness issues persist. Wang et al. [144] observed that LLMs trained on synthetic captions generated by captioning models often suffer from "captioning collapse," wherein output diversity is significantly reduced [142, 145]. This collapse undermines alignment quality and leads to overfitting to trivial or overly generic textual inputs [167].

### 2.3.3 Adversarial Prompting and Attacks

Adversarial prompting has emerged as a critical method for both stress-testing and improving the robustness of LLMs. Prompt injection [75], prompt leaking [48, 104], and jailbreaking techniques [76, 82] exploit the vulnerabilities of autoregressive models to override intended behaviors or extract sensitive content. Beyond direct attacks, methods like PromptAttack [29, 162, 164] and PAT generate adversarial examples via masked language modeling, producing fluently deceptive inputs. These insights have inspired frameworks that proactively improve LLM robustness via adversarial training in the prompt space.

### 2.3.4 LLMs for Verilog and Hardware-Oriented Code Generation

Code generation with LLMs has rapidly expanded across programming languages, with models such as Codex [20], CodeLlama [111], and GitHub Copilot [34] leading the charge. However, hardware description languages (HDLs) like Verilog pose unique challenges due to their symbolic structure, correctness requirements, and data scarcity. VerilogEval [71] and RTLLM [79] have benchmarked Verilog-specific capabilities, while VeriGen [127], RTLFixer [132], and Chip-Chat [11] explore synthetic data generation and compiler feedback loops to enhance functional correctness. Recent efforts also integrate self-reflection [22] and preference-guided reward modeling [7, 108] to refine learning signals in hardware design contexts.

### 2.3.5 Reasoning and Planning with LLMs

LLMs excel not only in generation but also in structured reasoning and interactive decision-making. Chain-of-thought prompting [149] decomposes tasks into intermediate logical steps, dramatically improving performance on complex QA and math tasks. Beyond static reasoning, models such as WebGPT [94], Toolformer [112], and ToolLLM [106] demonstrate how LLMs can use tools dynamically. The ReAct framework [165] integrates reasoning traces with action planning, enabling LLMs to interact with external environments in real-time. These methods form the concep-tual foundation for RL-enhanced reasoning in symbolic domains, as elaborated in Chapter 6.

### 2.3.6 Instructional Fine-Tuning and Data Pruning

Instruction tuning is central to aligning LLMs with human intent. From Self-Instruct [147] to OSS-Instruct [152], fine-tuning approaches now leverage both synthetic prompts and curated open-source code. Scaling both task diversity and data quality—via filtering [74, 163], self-verification [153], and reinforcement learning [173]—has proven effective in enhancing generalization. Complementing this, data pruning methods improve efficiency by selecting training examples based on difficulty [120], influence [105, 159], or diversity [24, 144]. For instance, AlphaCode [65] applies clustering-based pruning to retain high-quality code exemplars. These strategies are key to achieving scalable, low-resource training as described in Chapter 5.

### 2.3.7 Retrieval-Augmented Generation (RAG)

RAG techniques augment LLMs with access to external memory, typically via document retrieval or codebase querying [60]. This improves factual accuracy and allows models to dynamically integrate new or domain-specific knowledge. ReACC [78] and RepoCoder [171] show how RAG pipelines enhance code generation by retrieving related code snippets or documentation. In this dissertation, similar ideas are used in the RTLFixer framework to guide iterative repairs with compiler context.

Synthetic Data Generation

Synthetic data generation has become a key technique for fine-tuning large language models (LLMs), especially in domains with limited human-labeled data. These methods leverage LLMs themselves to generate instruction-following data, reducing annotation cost while increasing coverage and diversity.

Self-Instruct [147] pioneered this approach by using seed prompts to iteratively expand training sets. WizardLM [161] introduced evolutionary refinement, while OSS-Instruct [152] generated instructions from open-source code with minimal supervision. To improve data quality, recent work applies output selection [147], trace-based scoring [67], and teacher-model filtering [170]. Self-training methods further extend this by iteratively fine-tuning on model-generated data [33, 116].

In this dissertation, synthetic generation is central to enabling generalization in low-resource domains such as Verilog. As explored in Chapters 5 and 6, we adapt these methods to support both textual and non-textual input formats—including FSM diagrams, Karnaugh maps, and waveform tables—broadening model capabilities beyond standard natural language tasks.

Together, these prior works frame the foundation upon which this dissertation builds: generalization in LLMs that is efficient, robust, and adaptable across domains as introduced in Section 2.2.

# Chapter 3   Robust In-context Multimodal Alignment with Text-Centric Interfaces

Large Language Models (LLMs) have demonstrated remarkable capabilities across a wide range of tasks, and recent advances have extended these models to handle multimodal inputs spanning vision, speech, tables, and time-series signals. A particularly promising approach to enabling generalization across modalities is text-centric multimodal alignment, in which all input modalities are converted into natural language representations. This enables a single, pretrained LLM to serve as the core reasoning engine—without modality-specific retraining.

In real-world deployment, however, two key challenges emerge: modality mismatch, where the modalities encountered during inference differ from those seen during training, and modality robustness, where inputs may be noisy, corrupted, or incomplete. Existing systems often struggle under these conditions due to assumptions of static, clean modality configurations.

This chapter proposes a unified, text-centric framework to address both challenges using prompt-based in-context adaptation. Rather than modifying model weights or building modality-specific fusion modules, our approach operates entirely through structured prompt construction, built on two components:

- Modality-to-text transformation: Each modality is converted into structured, interpretable natural language using templates or pretrained captioning models.

- Adversarial prompting: Structured perturbations are applied to simulate noise, absence, or reordering of modalities during inference, enhancing robustness without retraining.

Extensive experiments show that this strategy achieves state-of-the-art performance under mismatched and noisy modality settings, outperforming fusion-based and embedding-aligned baselines across multiple tasks.

## 3.1 Modality Mismatch and Robustness

In this section, we define the two core challenges that motivate our approach: modality mismatch, where inference-time modality configurations differ from training, and modality robustness, where individual modalities may be corrupted, noisy, or missing.

### 3.1.1 Types of Modality Mismatch

We categorize modality mismatch based on the relationship between the training modality set $M_T$ and the inference-time modality set $M_I$, following the formalism in [138]:

**Training**

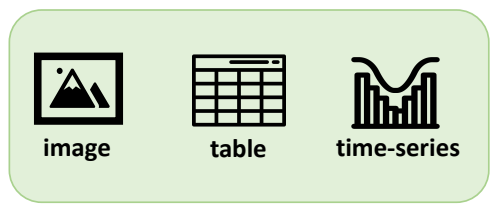

**Inference**

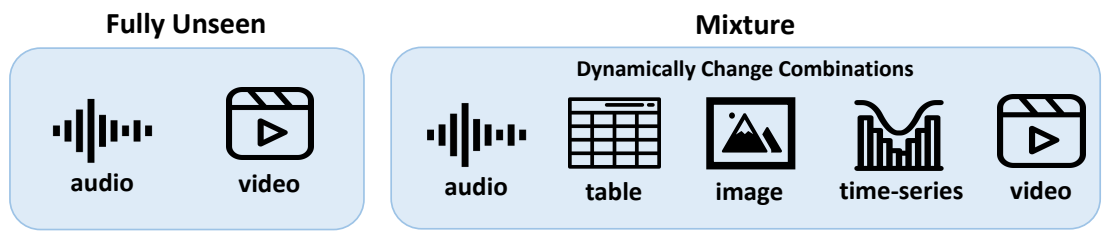

Figure 3.1: Illustration of the modality mismatch problem. During training, the model observes a fixed modality configuration. At inference, it must generalize across all combinations, including fully seen, fully unseen, and mixture setups.

- Fully Seen: $M_I = M_T$. The inference modality set exactly matches the training configuration. This serves as the baseline.

- Fully Unseen (Disjoint Modality): $M_T \cap M_I = \emptyset$. Inference uses entirely different modalities than training, requiring generalization through in-context reasoning.

- Mixture (Partial Mismatch): Inference modality deviates through one or more of:

  - Subset (Dropping): $M_I \subset M_T$.

  - Augmentation: $M_I \supset M_T$.

  - Reordering: Same set but different sequence/layout.

### 3.1.2 Types of Modality Robustness

Modality robustness refers to the model's ability to maintain performance when inputs are imperfect or corrupted. We identify three primary robustness stressors:

- Noisy Modality: Semantic-preserving perturbations (e.g., Gaussian image noise, word dropout, feature masking).

- Dynamic Modality (Permutation): Reordering of modalities or prompt segments.

- Missing Modality: Complete absence of one or more modalities.

## 3.2 Text-Centric Alignment for Multi-Modal Learning

Text-centric alignment constitutes a design paradigm wherein all input modalities—irrespective of their origin—are transformed into textual representations before processing by a language model. This section presents a comprehensive exposition of the motivation, architecture, and implementation of the proposed text-centric pipeline. The methodology addresses challenges in modality mismatch and robustness without depending on retraining or complex fusion architectures.

### 3.2.1 Motivation and Background

Traditional multimodal learning pipelines typically rely on fixed modality-specific encoders followed by fusion networks. While such systems perform adequately when all expected modalities are present and well-formed, they exhibit significant vul-nerabilities when confronted with missing, noisy, or dynamically composed inputs. Furthermore, fusion methods necessitate substantial retraining overhead whenever new modalities are introduced or input distributions shift.

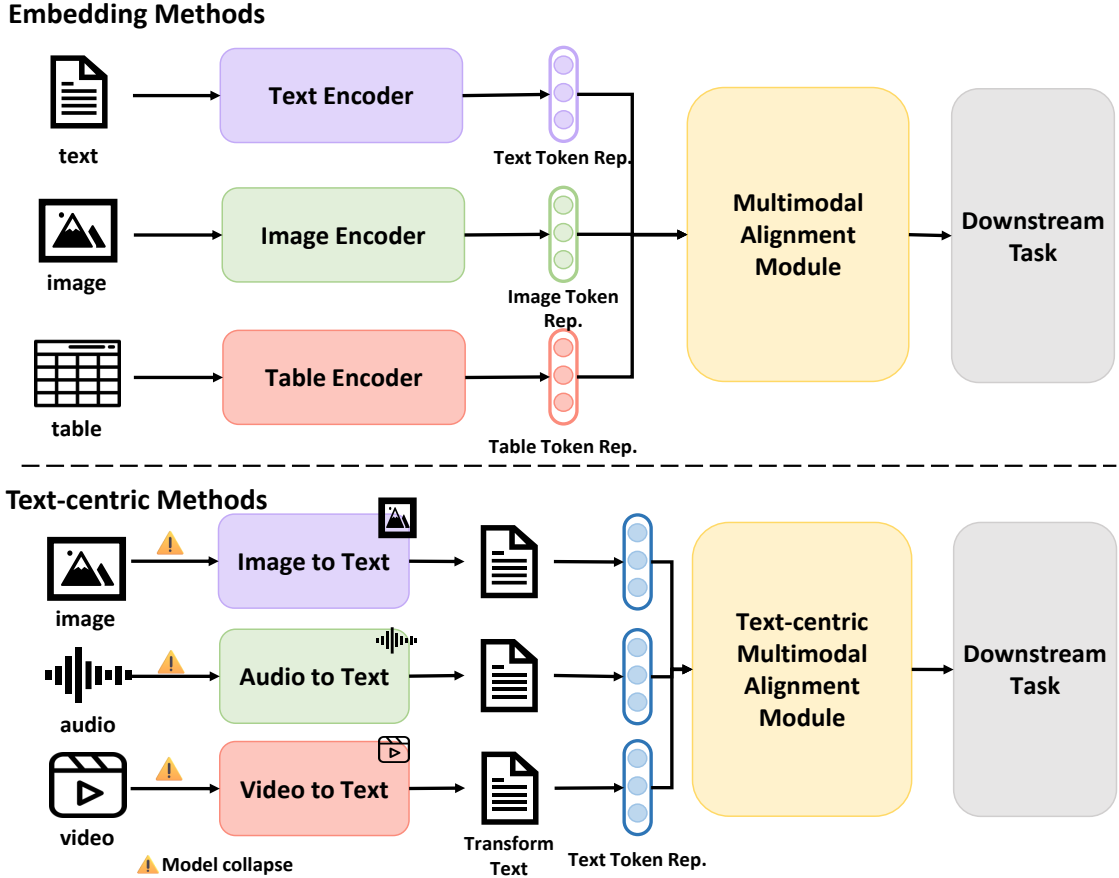

Figure 3.2: Architectural comparison between embedding-based and text-centric multimodal alignment approaches. Text-centric methods transform all inputs into a common linguistic representation, enabling more flexible processing and fusion.

The proposed approach leverages the emerging paradigm of text-centric multimodal processing, which transforms all non-textual inputs into language using pre-trained captioners or summarizers. These textual representations are then composed into prompts and processed by a frozen or instruction-tuned LLM. Unlike previous approaches in retrieval-augmented generation or vision-language transformers, this pipeline effectively decouples modality-specific processing from the core reasoning engine and systematically treats all inputs as structured descriptions.

### 3.2.2 Overview of the Text-Centric Pipeline

The comprehensive pipeline comprises four fundamental steps:

1. Modality-to-text transformation: Each input modality (image, table, waveform, etc.) is converted into a semantically faithful natural language representation using a specialized converter or summarization model.

2. Text-style translation: To ensure compatibility across diverse modalities, each modality's output undergoes normalization to match the language style employed during training via few-shot prompting.

3. Multimodal summarization: A unified cross-modality summary is generated from all textual representations to minimize redundancy and focus the LLM's attention on salient information.

4. Reasoning augmentation: Through chain-of-thought prompting, the model generates intermediate reasoning paths conditioned on the summary, which subsequently inform the final task execution.

This pipeline enables effective generalization to previously unseen modality compositions while maintaining high levels of interpretability and modularity.

### 3.2.3 Modality-to-Text Transformation

The initial stage transforms raw inputs into natural language text. Each modality is processed using a dedicated summarization function $\mathcal{F}_m$:

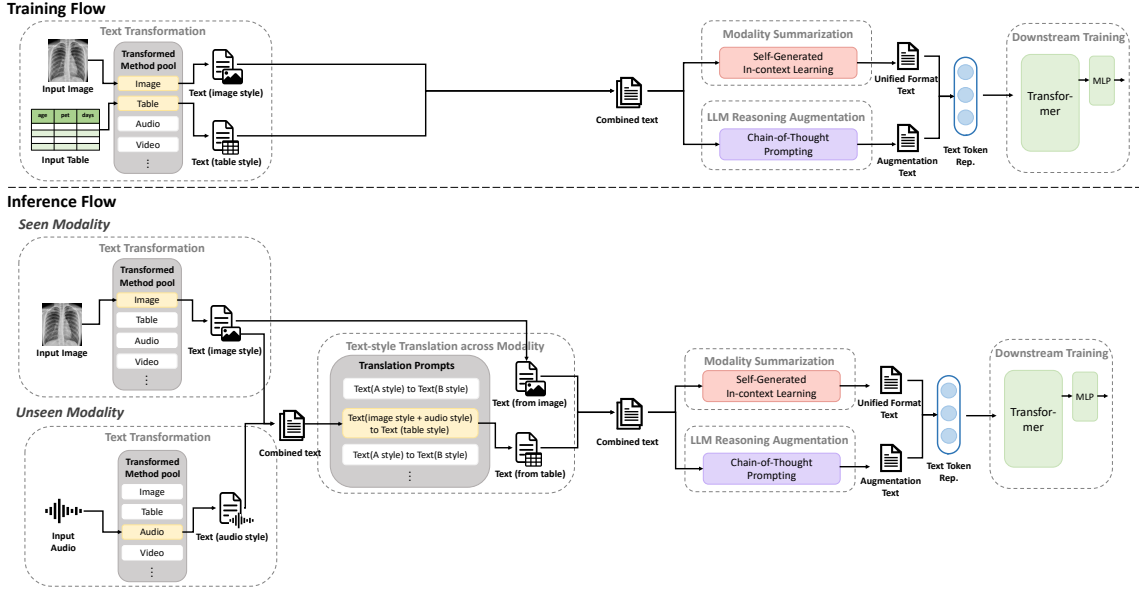

Figure 3.3: Text-centric multimodal alignment pipeline architecture. Each modality undergoes transformation into text via specialized summarization functions, followed by text-style translation, multimodal summarization, and reasoning augmentation to produce a unified prompt for a frozen language model.

$$\mathcal{F}_m(x) = \text{NL string summarizing modality input } x \qquad (3.1)$$

For example:

- An image of a congested road: "A 4-way intersection with heavy traffic on all lanes."

- A table of latency measurements: "Design A has 110 cycles, while Design B has 85 cycles."

- A waveform input: "Signal clk transitions from low to high at timestep 12."

State-of-the-art transformers such as BLIP-2 or Whisper may be employed for this purpose, while templates with slot-filling mechanisms might be preferred for highly structured inputs like tables.

| Modality | Example Transformed Text |
|---|---|
| Image | "A congested intersection with vehicles in all directions." |
| Table | "Design A has 83% accuracy while Design B achieves 76%." |
| Waveform | "Signal clk transitions from low to high at timestep 12." |
| FSM | "State A transitions to State B on input '1' with output '0'." |

Table 3.1: Modality-to-text transformation examples across different input types. Each modality is converted to a natural language representation using appropriate conversion techniques while preserving key semantic information.

### 3.2.4 Text-Style Translation

Since different summarization mechanisms produce stylistically inconsistent text (e.g., image captioning versus table serialization), a text-style normalization stage becomes essential. Through few-shot in-context prompting, all input summaries are converted to align with a reference style (typically corresponding to that of the training corpus):

"Convert the following description to match the formal style used in previous inputs."

This systematic normalization reduces distributional shift between modalities and significantly enhances alignment during downstream reasoning processes.

### 3.2.5 Multimodal Summarization

Given multiple text-modality inputs, a specialized summarization model condenses the information into a coherent, unified paragraph. This critical step:

- Integrates shared or overlapping information (e.g., when both image and table

describe congestion).

- Emphasizes critical signals (e.g., "The table indicates a sharp drop in performance.").

- Eliminates noisy or redundant phraseology that might otherwise impede comprehension.

The summarizer can be efficiently implemented using a GPT-style prompt with few-shot exemplars:

"Given the following information from multiple sources, write a single coherent summary."

This stage substantially improves alignment between different combinations of modalities, particularly when certain modalities are missing or contain noise.

### 3.2.6 Reasoning-Augmented Alignment

To further enhance performance, an additional step incorporates model-generated reasoning traces based on the summary. This utilizes techniques analogous to chain-of-thought prompting [150]:

"Based on the summary above, explain why the design is likely to fail and what action should be taken."

This intermediate reasoning functions as an implicit regularizer and substantially improves robustness, particularly in tasks involving complex logic or planning (e.g., congestion prediction, failure analysis). Additionally, it facilitates attribution and interpretability of the model's decision-making process.

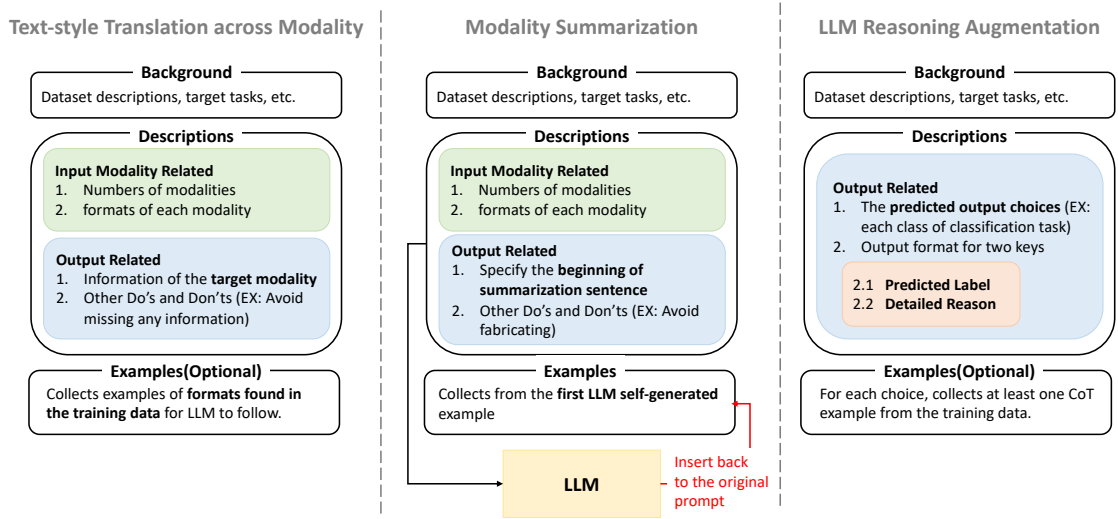

Figure 3.4: Prompt templates for different pipeline modules, illustrating the text transformation, translation, summarization, and reasoning processes employed in the multimodal alignment framework.

### 3.2.7 Discussion

This pipeline offers several compelling advantages:

- Modular extensibility: New modalities can be seamlessly integrated by defining an appropriate $\mathcal{F}_m$ function and corresponding few-shot translation prompt.

- Robust generalization: By conditioning exclusively on natural language, the model effectively handles any subset of modalities present during inference.

- Zero retraining: Once the downstream model is properly prompt-aligned, no additional retraining is necessary to accommodate new modality types or combinations.

- Interpretability: All intermediate processing stages remain human-readable,

facilitating comprehensive debugging and verification.

This architectural design establishes a robust foundation for in-context multimodal alignment via strategic prompting and summarization. The following section presents advanced adversarial prompting methods that further enhance the model's resilience to corrupted or potentially misleading modality inputs.

## 3.3   Text-Centric Adversarial Training and Prompting

To improve the robustness of text-centric multimodal alignment, we introduce a targeted perturbation strategy termed adversarial prompting. Rather than relying on conventional robust training paradigms such as noise injection or dropout, our approach leverages LLMs to construct semantically challenging prompts that stress-test and enhance the resilience of the downstream model.

### 3.3.1   Overview of the Adversarial Training Pipeline

Our adversarial prompting method integrates within a broader text-centric pipeline. As illustrated in Figure 3.5, the workflow includes:

1. Text Transformation: Each raw input modality (e.g., image, table) is converted into a natural language description via expert captioning or serialization models.

2. Modality Summarization: These descriptions are merged using an LLM-guided summarization step that unifies linguistic style and eliminates redundancy.

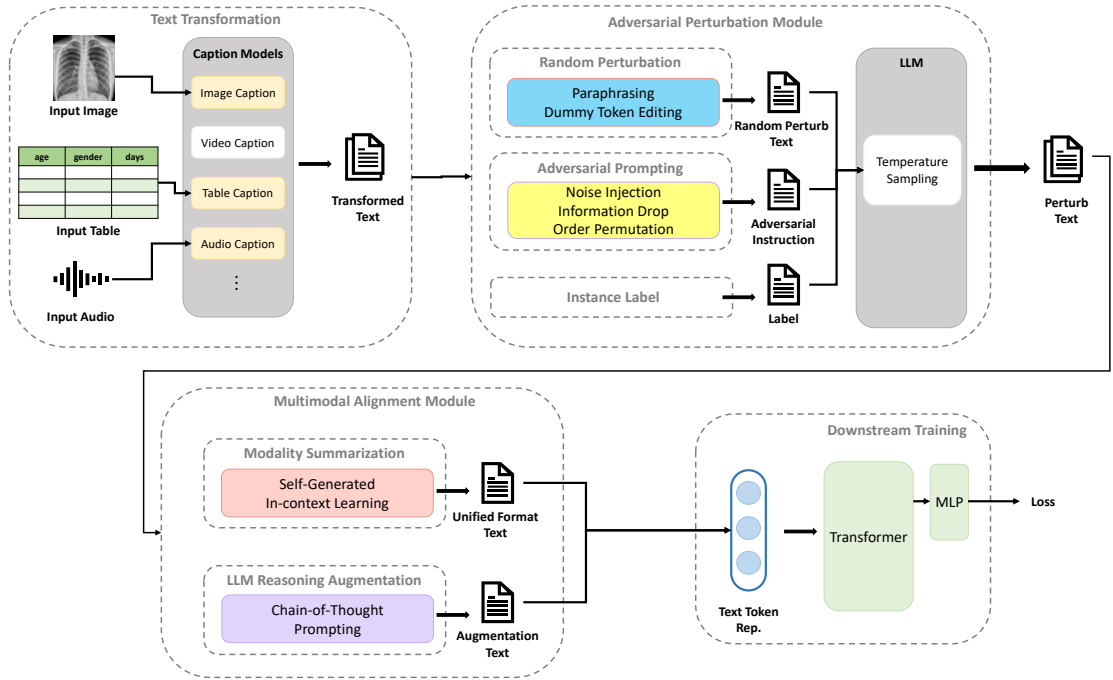

Figure 3.5: Transformation pipeline for multimodal alignment. Raw inputs are converted to text representations using foundation models, then processed through parallel modality summarization and LLM reasoning. The resulting text outputs are concatenated for downstream prediction. During inference, a one-shot in-context learning approach adapts linguistic style to match training expectations.

3. Reasoning Augmentation: Chain-of-thought prompting enhances the descriptions with LLM-generated reasoning steps.

4. Perturbation: Before final input to the model, prompts are perturbed via one of two methods:

    - Random Perturbation: Naive alterations such as paraphrasing, dummy token insertion, or reordering.

    - Adversarial Perturbation: Guided prompts instruct an LLM to generate adversarial examples that shift semantics toward incorrect labels.

This pipeline supports modular training and generalization, as each component operates in prompt space without requiring model weight modifications.

### 3.3.2 Text-centric Perturbation Strategies

Random Perturbation. This baseline strategy produces alternative versions of the input prompt using rule-based edits and lightweight LLM queries. It consists of three subtypes:

- Paraphrasing: Using prompts such as "Suggest $k$ ways to paraphrase the text above. Maintain the same semantics.", the LLM generates $k$ semantically equivalent variants $\{x_i\}$ of the input $x_0$, capturing lexical diversity.

- Dummy Token Insertion: We randomly insert tokens $d_i$ such as newlines, punctuation, or padding into the input (e.g., $x_i = x_0 + d_i$), which minimally affect semantics while testing the model's input invariance.

- Edit Operations: Deletion, insertion, and substitution of tokens are applied stochastically to create varied inputs.

Adversarial Perturbation. We introduce semantically disruptive perturbations guided by instruction prompts. These simulate challenging real-world scenarios where modality inputs might be misleading or corrupted. The process follows these steps:

1. Random Initialization: Apply random perturbation to generate $x' = \text{RandomPerturb}(x)$.

2. Instruction Sampling: Select a domain-specific adversarial instruction inst from a pool targeting noise injection, information dropout, or permutation.

3. Adversarial Prompting: Direct the LLM to produce an adversarial sample $x_{adv}$ by completing the instruction in a way that maximally shifts semantics toward

the incorrect label:

$$x_{\text{adv}} = \text{LLM}(x', \text{inst}, \text{label}, T)$$

where $T$ controls generation randomness.

This modular perturbation strategy enhances robustness by exposing the model to diverse failure cases without requiring retraining or explicit gradient-based adversarial attacks. The strategies can be iteratively refined to support comprehensive robustness evaluation across all input modalities.

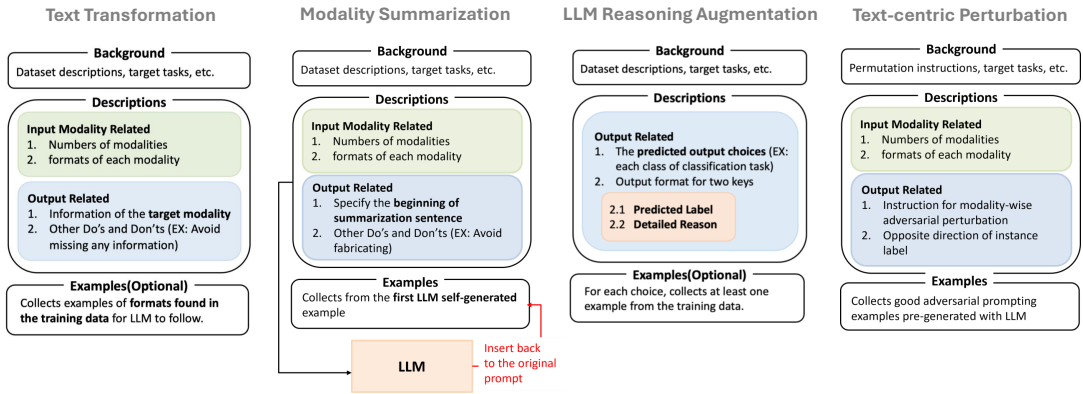

Figure 3.6: Example prompt templates for each module in the adversarial training pipeline.

## 3.4 Experimental Evaluation

### 3.4.1 Evaluation Setup

Evaluation Metrics   The evaluation uses two primary metrics:

- Absolute performance: Measured via accuracy (for classification) or mean squared error (MSE) for regression under varied modality inputs.

- Effective robustness: Performance degradation compared to the clean or matched-modality setting, quantifying resilience to modality shifts.

Datasets and Tasks   The experiments are conducted on three multimodal datasets:

- PetFinder.my: A classification task predicting pet adoption speed from tabular, text, and image data.

- Airbnb Pricing: A regression task estimating housing prices using image, text, and tabular information.

- Avito Demand: A regression task predicting advertisement demand using multimodal online listing data.

Image Caption Model   All experiments primarily do image captioning using GPT-4-Vision, unless specified otherwise. For additional image caption models results, please refer to Table 3.8. Furthermore, the performance enhancement is all calculated using the formula: (measured performance - baseline performance) / baseline performance.

Additional details on datasets, hyperparameters, foundation models, and model checkpoints are provided in Appendix 7.4. For prompt designs and implementation specifics, refer to Appendix A.4 and Appendix A.4, respectively.

### 3.4.2  Modality Mismatch Baselines

The experimental evaluation compares TAMML against multiple established zero-shot cross-modality translation methodologies. These comparative approaches are built upon embedding-based downstream model training, alongside a naive zero-shot transformation technique utilizing GPT-4-vision without additional fine-tuning. Figure 3.7 presents a high-level illustration of conventional embedding-based downstream training architecture.

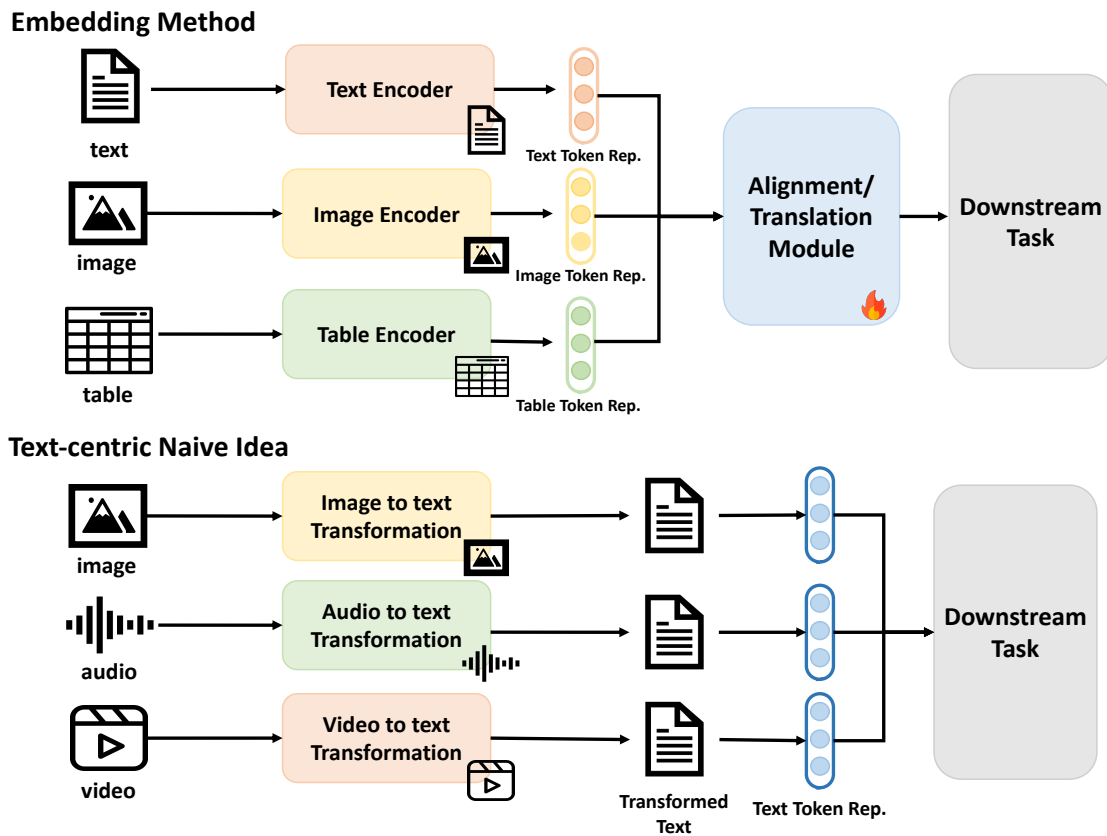

Figure 3.7: Comparison between traditional and proposed approaches for multi-modal learning. Traditional downstream training (top) relies on embeddings from modality-specific foundation models, limiting adaptation to unseen modalities without complete retraining. Previous research (bottom) has addressed this limitation by implementing zero-shot cross-modality translations during inference.

The comparative baseline methods encompass two perturbation-diffusion-based approaches—SDEdit [88] and DDRM [54]—as well as one GAN inversion technique, In-domain Inversion GAN (Idinvert) [177]. Each of these methodologies incorporates a generative model (either diffusion-based or GAN-based) capable of transforming input embedding distributions into representations that closely approximate training modality distributions. During inference, these trained generative models syn-thesize modality embeddings for various unseen modalities, designed to mimic the distribution characteristics of modalities encountered during training.

For the diffusion-based models, the implementation utilized a score-based generative model [46] alongside a dedicated backbone model for each modality combination. DDIM [118] served as the pre-trained foundation for both SDEdit and DDRM implementations. In the case of GAN-based approaches, StyleGAN [53] architecture was employed as the backbone implementation.

Across all embedding-based methodologies, specialized foundation models were applied as modality-specific encoders, each extracting embedding representations from their designated modality types. These representations were subsequently processed through alignment layers for integration and concatenation. Following these modality-specific encoding stages, the combined representational outputs from all modalities were concatenated and processed through a transformer encoder. After appropriate fine-tuning procedures, the model generated cross-modality alignment outputs suitable for downstream tasks.

### 3.4.3 Modality Mismatch Experiments

The evaluation of the proposed TAMML framework was conducted through a comprehensive series of test-time modality mismatch scenarios, encompassing both classification and regression tasks. The primary research objective was to determine whether in-context text transformation could provide a robust and effective alignment solution capable of generalizing across diverse and previously unseen modality configurations.

The investigation was structured around the following central research questions, systematically examining various aspects of the TAMML approach:

- Q1: Does TAMML demonstrate superior performance compared to embedding-based state-of-the-art zero-shot cross-modality translation methods under test-time unseen modality scenarios?

- Q2: Beyond complete modality mismatch, how effective is TAMML in scenarios where testing modalities were seen during training (e.g., training with all modalities, testing with a subset) and other partial mismatch combinations?

- Q3: In cross-modality translation tasks, does text representation offer inherently greater robustness than embedding representation?

- Q4: When training and testing modalities are identical (no modality mismatch), what performance differences emerge between text-based and embedding-based solutions?

- Q5: How does TAMML 's zero-shot performance compare with non-zero-shot transfer methods such as domain adaptation?

- Q6: How does TAMML compare to approaches that rely on extensive fine-tuning for modality alignment?

Q1: Comparative Performance of TAMML Against State-of-the-Art Zero-shot Cross-Modality Translation Methods   This research question examines scenarios in which training and testing modalities exhibit complete dissimilarity, focusing on the comparative efficacy of various zero-shot cross-modality translation methodologies.

| Training | Testing | SDEdit | DDRM | Idinvert | Naive MLLM Transformation | TAMML Methods | | | | |
| --- | --- | --- | --- | --- | --- | --- | --- | --- | --- | --- |
| | | | | | | LLaMa 3 8B | Mistral 7B | Mixtral 8x7B | Mixtral 8x22B | GPT-3.5 |
| PetFinder \| Accuracy ↑ | | | | | | | | | | |
| text+image | tabular | 0.282 | 0.291 | 0.279 | 0.310 | 0.309 | 0.301 | 0.317 | 0.332 | <u>0.348</u> |
| text+tabular | image | 0.289 | 0.277 | 0.286 | 0.329 | 0.306 | 0.322 | 0.335 | 0.323 | <u>0.380</u> |
| image+tabular | text | 0.281 | 0.297 | 0.279 | 0.305 | 0.303 | 0.304 | 0.320 | 0.313 | <u>0.355</u> |
| text | image+tabular | 0.291 | 0.283 | 0.289 | 0.282 | 0.315 | 0.304 | 0.330 | <u>0.368</u> | 0.344 |
| text | image | 0.289 | 0.276 | 0.287 | 0.293 | 0.355 | 0.341 | 0.307 | 0.360 | <u>0.374</u> |
| text | tabular | 0.293 | 0.259 | 0.277 | 0.297 | 0.295 | 0.286 | 0.325 | 0.341 | <u>0.357</u> |
| image | text+tabular | 0.290 | 0.297 | 0.284 | 0.314 | 0.310 | 0.322 | <u>0.346</u> | 0.342 | 0.341 |
| image | text | 0.288 | 0.282 | 0.280 | 0.306 | 0.323 | 0.325 | 0.329 | <u>0.330</u> | 0.319 |
| image | tabular | 0.291 | 0.287 | 0.284 | 0.300 | 0.322 | 0.302 | 0.341 | 0.319 | <u>0.348</u> |
| tabular | text+image | 0.290 | 0.271 | 0.285 | 0.194 | 0.314 | 0.309 | 0.327 | 0.333 | <u>0.360</u> |
| tabular | text | 0.289 | 0.265 | 0.280 | 0.193 | 0.295 | 0.294 | 0.302 | 0.317 | <u>0.364</u> |
| tabular | image | 0.289 | 0.263 | 0.277 | 0.196 | 0.294 | 0.305 | 0.338 | 0.311 | <u>0.364</u> |
| Average | | 0.289 | 0.279 | 0.282 | 0.277 | 0.312 | 0.310 | 0.326 | 0.332 | 0.355 |
| Airbnb \| MSE ↓ | | | | | | | | | | |
| text+image | tabular | 0.935 | 0.600 | 0.799 | 0.365 | 0.303 | 0.371 | 0.326 | <u>0.313</u> | 0.367 |
| text+tabular | image | 0.656 | 0.778 | 0.643 | 0.957 | 0.626 | 0.466 | 0.451 | <u>0.447</u> | 0.508 |
| image+tabular | text | 0.514 | 0.565 | 0.781 | 0.695 | 0.413 | 0.325 | <u>0.312</u> | 0.359 | 0.332 |
| text | image+tabular | 1.548 | 0.914 | 0.915 | 0.438 | 0.315 | 0.368 | 0.323 | <u>0.284</u> | 0.421 |
| text | image | 1.513 | 0.895 | 1.010 | 0.524 | 0.537 | 0.521 | 0.439 | <u>0.404</u> | 0.520 |
| text | tabular | 1.061 | 0.824 | 0.931 | 0.759 | 0.308 | 0.348 | 0.345 | <u>0.297</u> | 0.448 |
| image | text+tabular | 0.556 | 0.530 | 0.602 | 0.457 | 0.431 | <u>0.368</u> | 0.382 | 0.392 | 0.395 |
| image | text | 0.678 | 0.589 | 0.759 | 0.423 | 0.439 | 0.389 | <u>0.375</u> | 0.421 | 0.391 |
| image | tabular | 0.592 | 0.538 | 0.516 | 0.668 | 0.459 | 0.452 | 0.487 | <u>0.405</u> | 0.414 |
| tabular | text+image | 0.637 | 0.675 | 0.662 | 0.480 | 0.467 | 0.347 | 0.310 | 0.379 | <u>0.280</u> |
| tabular | text | 0.569 | 0.693 | 0.707 | 0.477 | 0.481 | 0.341 | 0.313 | 0.339 | <u>0.301</u> |
| tabular | image | 0.609 | 0.715 | 0.615 | 0.913 | 0.627 | 0.461 | <u>0.431</u> | 0.535 | 0.551 |
| Average | | 0.822 | 0.693 | 0.745 | 0.596 | 0.451 | 0.396 | <u>0.375</u> | 0.381 | 0.411 |
| Avito \| MSE ↓ | | | | | | | | | | |
| text+image | tabular | 0.103 | 0.113 | 0.126 | 0.051 | 0.045 | 0.045 | 0.043 | <u>0.041</u> | 0.044 |
| text+tabular | image | 0.130 | 0.133 | 0.142 | 0.051 | 0.048 | 0.048 | 0.047 | 0.048 | <u>0.046</u> |
| image+tabular | text | 0.113 | 0.125 | 0.137 | <u>0.040</u> | 0.045 | 0.045 | 0.045 | <u>0.043</u> | 0.046 |
| text | image+tabular | 0.124 | 0.123 | 0.131 | 0.050 | 0.046 | 0.047 | 0.046 | 0.046 | <u>0.045</u> |
| text | image | 0.124 | 0.122 | 0.129 | 0.052 | 0.048 | 0.050 | <u>0.047</u> | 0.048 | <u>0.047</u> |
| text | tabular | 0.127 | 0.124 | 0.134 | 0.052 | 0.045 | 0.046 | 0.046 | 0.046 | <u>0.044</u> |
| image | text+tabular | 0.123 | 0.126 | 0.134 | <u>0.044</u> | <u>0.044</u> | <u>0.044</u> | <u>0.044</u> | <u>0.044</u> | <u>0.044</u> |
| image | text | 0.118 | 0.124 | 0.129 | 0.045 | 0.045 | 0.046 | 0.047 | 0.046 | <u>0.045</u> |
| image | tabular | 0.119 | 0.126 | 0.134 | 0.049 | 0.044 | 0.044 | 0.045 | <u>0.043</u> | 0.044 |
| tabular | text+image | 0.128 | 0.139 | 0.137 | 0.044 | 0.046 | 0.045 | 0.046 | <u>0.044</u> | 0.046 |
| tabular | text | 0.124 | 0.131 | 0.138 | 0.046 | 0.046 | <u>0.044</u> | 0.047 | <u>0.044</u> | 0.045 |
| tabular | image | 0.126 | 0.137 | 0.140 | <u>0.044</u> | 0.050 | 0.047 | 0.048 | 0.046 | 0.048 |
| Average | | 0.122 | 0.127 | 0.135 | 0.048 | 0.046 | 0.046 | 0.046 | <u>0.045</u> | <u>0.045</u> |

Table 3.2: Comprehensive performance comparison of TAMML against various baseline models under modality mismatch scenarios across three datasets. PetFinder results are reported using accuracy (higher is better), while Airbnb and Avito results are reported using Mean Squared Error (lower is better).

The empirical evidence presented in Table 3.2 demonstrates the significant performance advantages of TAMML across diverse modality combinations and foundation model architectures. Notably, the GPT-3.5 implementation of TAMML on the PetFinder dataset achieved an average accuracy improvement of approximately <u>21%</u> relative to the strongest baseline approaches. Similarly, on the Airbnb dataset, TAMML reduced mean square error by an impressive <u>54%</u> on average, substantially outperforming alternative approaches which achieved a maximum error reduction of only 16%.

| Training | Testing | Pet | | Acc ↑ | | Airbnb | | MSE ↓ | | Avito | | RMSE ↓ | |
|---|---|---|---|---|---|---|---|---|---|---|---|---|---|
| | | SDEdit | DDRM | Idinvert | TAMML | SDEdit | DDRM | Idinvert | TAMML | SDEdit | DDRM | Idinvert | TAMML |
| all | tabular | 0.282 | 0.269 | 0.252 | 0.338 | 0.428 | 0.621 | 0.732 | 0.270 | 0.108 | 0.123 | 0.133 | 0.041 |
| all | image | 0.285 | 0.286 | 0.267 | 0.356 | 0.566 | 0.649 | 0.711 | 0.486 | 0.114 | 0.123 | 0.136 | 0.044 |
| all | text | 0.284 | 0.284 | 0.274 | 0.349 | 0.502 | 0.601 | 0.695 | 0.253 | 0.113 | 0.123 | 0.131 | 0.044 |
| all | image+tabular | 0.307 | 0.276 | 0.256 | 0.382 | 0.394 | 0.556 | 0.683 | 0.251 | 0.118 | 0.124 | 0.129 | 0.042 |
| all | text+tabular | 0.315 | 0.306 | 0.283 | 0.377 | 0.353 | 0.470 | 0.544 | 0.185 | 0.124 | 0.124 | 0.134 | 0.041 |
| all | text+image | 0.292 | 0.286 | 0.244 | 0.378 | 0.489 | 0.537 | 0.673 | 0.212 | 0.110 | 0.115 | 0.125 | 0.043 |
| all | all | 0.334 | 0.304 | 0.281 | 0.395 | 0.345 | 0.463 | 0.542 | 0.178 | 0.109 | 0.114 | 0.123 | 0.042 |
| | Average | 0.300 | 0.287 | 0.265 | 0.368 | 0.440 | 0.557 | 0.654 | 0.262 | 0.112 | 0.121 | 0.130 | 0.042 |
| all | comb | 0.294 | 0.285 | 0.263 | 0.362 | 0.455 | 0.572 | 0.673 | 0.299 | 0.115 | 0.122 | 0.131 | 0.043 |
| text+image | comb | 0.282 | 0.290 | 0.277 | 0.320 | 0.643 | 0.623 | 0.747 | 0.331 | 0.108 | 0.120 | 0.132 | 0.043 |
| text+tabular | comb | 0.290 | 0.278 | 0.282 | 0.341 | 0.517 | 0.645 | 0.674 | 0.318 | 0.120 | 0.127 | 0.134 | 0.044 |
| image+tabular | comb | 0.296 | 0.296 | 0.269 | 0.358 | 0.452 | 0.524 | 0.666 | 0.236 | 0.116 | 0.121 | 0.132 | 0.042 |
| | Average | 0.291 | 0.287 | 0.273 | 0.345 | 0.517 | 0.591 | 0.690 | 0.286 | 0.115 | 0.123 | 0.132 | 0.043 |

Table 3.3: Performance comparison between TAMML and embedding-based translation baselines when models are trained on all modalities and evaluated on different subset modalities. Results demonstrate TAMML 's effectiveness even when testing modalities were seen during training, using Mixtral 8x7B as the foundation model.

Further analysis examining the influence of foundation model architectures within the TAMML framework revealed meaningful correlations between model scale and generation quality. For instance, the Mixtral 8x22B model demonstrated a 7% accuracy improvement on the PetFinder dataset compared to the smaller Mistral 7B model. This performance differential was particularly pronounced for complex tasks such as summarization and translation. Nevertheless, even models of more modest scale exhibited substantial improvements over baseline approaches in modality mismatch conditions. These findings provide compelling evidence that the proposed methodology, which integrates in-context learning capabilities of LLMs with foundation models, offers definitive advantages over existing alternative approaches.

Q2: Efficacy of TAMML Under Partial and Non-Existent Modality Mismatch Conditions This research question examines the performance characteristics of TAMML in scenarios where testing modalities were either partially or fully represented during the training phase. The empirical evidence presented in Table 3.3 demonstrates that TAMML maintains its performance advantages even in the absence of complete modality mismatching.

The quantitative analysis reveals substantial performance differentials across multiple datasets. On the PetFinder dataset, TAMML achieved an average accuracy improvement of approximately 22.6% compared to the most effective embedding-based methodologies. The performance enhancement was even more pronounced for regression tasks. For the Airbnb dataset, TAMML reduced mean squared error by approximately 40.5%, representing a significant advancement in prediction fidelity. Most notably, the Avito dataset experiments yielded dramatic improvements, with mean squared error reductions of approximately 62.5% when applying the TAMML framework.

These consistent performance advantages across diverse experimental conditions demonstrate that the text-based representation strategy inherent to TAMML provides fundamental advantages over embedding-based alternatives, regardless of whether the modality configuration at inference time precisely matches that encountered during training. This suggests that the text-centric alignment approach offers inherent representational benefits beyond merely addressing modality mismatch challenges.

Q3: Comparative Robustness of Text versus Embedding Representations in Cross-Modality Translation  This section investigates the inherent trade-offs between performance and flexibility when transforming various modalities from embedding-based to text-based representations, particularly under conditions of modality mismatch.

Multimodal LLM Experimental Design  The experimental analyses presented in Sections 3.4.3 and 3.4.3 do not facilitate direct comparison between text and embedding representations due to the heterogeneity of foundation models employed in text-based modality conversion. Each foundation model introduces distinct capabilities and characteristics that could confound comparative analysis. To establish methodological validity, this section employs Multimodal Language Models (MLLMs) as a controlled experimental framework, ensuring that all modality conversions are processed through identical foundation models.

The experimental protocol utilized two state-of-the-art MLLMs: Kosmos-2 [103] and Flamingo [3]. These models served dual functions as pre-trained feature extractors and text decoders. The feature representations were aggregated through mean pooling operations and subsequently processed by a Multi-Layer Perceptron (MLP) functioning as the task-specific prediction model. This methodological approach enabled precise quantification of performance differentials between downstream models trained on raw image representations versus those trained on image-derived caption representations.

Empirical Findings  The quantitative results presented in Table 3.4 provide compelling evidence that downstream models trained on caption-based representations exhibit significantly reduced performance degradation compared to their image embedding-trained counterparts when evaluated under modality mismatch conditions. This finding demonstrates remarkable consistency across the entire spectrum of state-of-the-art multimodal language models examined in this study.

The empirical evidence strongly indicates that cross-modality translation uti-lizing text-based representations, as implemented in the TAMML framework, offers substantially enhanced robustness compared to embedding-based approaches when confronting modality mismatch challenges. This suggests fundamental representational advantages inherent to linguistic encoding of multimodal information in cross-domain translation tasks.

| Pet \| Acc ↑ Test/Train | Flamingo caption | Flamingo image | Kosmos2 caption | Kosmos2 image |
|---|---|---|---|---|
| text | -0.07 | -0.10 | -0.09 | -0.11 |
| tabular | -0.08 | -0.10 | -0.12 | -0.21 |
| text+tabular | -0.08 | -0.11 | -0.10 | -0.15 |
| Air \| MSE ↓ Test/Train | Flamingo caption | Flamingo image | Kosmos2 caption | Kosmos2 image |
| text | -0.00 | -0.06 | -0.01 | -0.03 |
| tabular | -0.03 | -0.07 | -0.05 | -0.05 |
| text+tabular | -0.02 | -0.07 | -0.04 | -0.03 |

Table 3.4: Performance comparison between text and embedding representations for cross-modality translation. Both representations are derived from the same Multimodal LLMs (Flamingo and Kosmos2) for fair comparison. Results demonstrate that text representations consistently exhibited less performance degradation during modality shifts, despite the slight reduction in performance when transforming from image to caption.

Q4: Comparative Analysis of Text-based versus Embedding-based Approaches Under Modality-Matched Conditions  Having established the advantages of text-based representations in modality mismatch scenarios, this section examines performance characteristics when training and inference modalities are identical. This analysis provides critical insights into potential trade-offs between robustness to modality shifts and optimal performance under matched conditions.

Table 3.5 presents a comprehensive empirical comparison of performance metrics under conditions where no modality mismatch exists between training and inference phases. The results reveal several noteworthy patterns across different modality

configurations:

| Train & Test | Regular (Embedding) | TAMML (Text) |
|---|---|---|
| text | 0.352 | 0.382 |
| image | 0.273 | 0.369 |
| tabular | 0.429 | 0.394 |
| text+image | 0.286 | 0.400 |
| text+tabular | 0.411 | 0.404 |
| image+tabular | 0.403 | 0.408 |

Table 3.5: Performance analysis under matched modality conditions where training and testing modalities are identical. Under these conditions, TAMML maintains competitive performance and even outperforms embedding-based approaches in multiple modality combinations. Note that these results differ from those in Table 3.4 due to the use of different foundation models for generating text and embedding representations.

1. Text Modality: The TAMML approach demonstrates superior performance (0.382 vs. 0.352) when both training and testing utilize textual modalities exclusively.

2. Image Modality: Similarly, under image-only conditions, TAMML achieves substantially improved accuracy (0.369 vs. 0.273) compared to embedding-based methods.

3. Tabular Modality: For tabular data, embedding-based approaches exhibit a moderate advantage (0.429 vs. 0.394), suggesting that the structured nature of tabular information may benefit from direct embedding representations.

4. Multimodal Configurations: In combined modality scenarios, TAMML outperforms embedding-based approaches in text+image (0.400 vs. 0.286) and image+tabular (0.408 vs. 0.403) configurations, while demonstrating competitive but slightly lower performance in text+tabular scenarios (0.404 vs. 0.411).

These findings indicate that contrary to potential expectations, the text-centric approach rarely incurs significant performance penalties even under matched modality conditions. In fact, for most modality configurations, TAMML maintains performance advantages compared to traditional embedding-based methods. This suggests that the linguistic representational framework offers inherent advantages beyond merely addressing modality mismatch challenges, potentially due to improved semantic alignment and the rich contextual representation capabilities of language models.

Q5: Comparative Analysis of TAMML Against Domain Adaptation and Supervised Learning Approaches   This research question extends beyond zero-shot methodologies to compare TAMML against adaptation and supervised learning techniques that have access to target domain information or labels. Table D.3 presents a systematic comparative evaluation across multiple adaptation paradigms:

- Zero-shot embedding-based methods: Standard embedding approaches without fine-tuning, representing the complete modality mismatch scenario identical to the main experimental conditions.

- Zero-shot text-based method (TAMML ): The proposed approach utilizing text transformation without requiring any adaptation to target modalities.

- Unsupervised domain adaptation: A methodology that fine-tunes the downstream model with inference modality information but without access to target domain labels. For this experimental condition, the Adversarial Discriminative Domain Adaptation (ADDA) framework [141] was implemented.

- Fully supervised training: The upper-bound reference condition where the model is trained with paired source/target modality information and corresponding labels, effectively incorporating the testing modality during the training phase.

The empirical results presented in Table D.3 reveal several significant findings. Most notably, TAMML consistently outperforms unsupervised domain adaptation approaches despite requiring no fine-tuning or optimization for specific target domains. For example, when examining text-to-image translation, TAMML achieves an accuracy of 0.374 compared to only 0.195 for unsupervised domain adaptation— a substantial performance differential of 17.9 percentage points.

Furthermore, in four of the six modality transfer scenarios examined, TAMML demonstrates superior performance even when compared to fully supervised training approaches. This is particularly evident in tabular-to-text and tabular-to-image translations, where TAMML achieves accuracies of 0.364 in both cases, substantially outperforming the fully supervised approaches (0.306 and 0.338, respectively).

These findings are especially significant considering that TAMML requires no adaptation phase whatsoever, while the comparative methods necessitate either unlabeled target domain data (unsupervised adaptation) or completely labeled target domain data (supervised training). The ability of TAMML to outperform methods with privileged access to target domain information underscores the effectiveness of text-centric modality alignment and suggests fundamental advantages in the linguistic representational framework for cross-modality transfer tasks.

Q6: Evaluating TAMML Against Specialized Fine-tuning-based Alignment Systems

This analysis extends the comparative evaluation to include specialized modality alignment systems that are specifically designed for multimodal integration through extensive fine-tuning procedures. Specifically, this investigation considers approaches inspired by ImageBind [37] and LanguageBind [176] architectures, which represent the current state-of-the-art in supervised multimodal alignment. These methodologies offer several theoretical advantages:

- Optimization of representational alignment through shared anchoring (typically text-based), which reduces the combinatorial complexity of paired datasets from exponential $O(2^n)$ to quadratic $O(n^2)$.

- Potential for optimal performance characteristics through comprehensive supervised training across all anticipated modality configurations.

- Establishment of empirical performance upper bounds under conditions where ideal training data is available for all modality combinations.

In contrast, the TAMML framework represents a fundamentally different paradigm, utilizing pretrained modality-to-text encoders within an in-context learning framework that requires no additional fine-tuning. This architectural distinction offers potentially significant advantages in low-resource environments or scenarios involving dynamic modality configurations.

To establish a methodologically sound comparison, three distinct approaches were evaluated:

- TAMML : Text-aligned inference-time prompting without retraining require-

ments.

- LanguageBind-style Fine-tuning: Comprehensive supervised alignment utilizing paired cross-modal datasets.

- Direct Input Baseline: Standard processing without alignment optimization.

To ensure experimental validity, a controlled LanguageBind-style training environment was implemented using the PetFinder dataset with identical modality encoders across all conditions. These encoders were systematically fine-tuned across all modality pair combinations. Table 3.6 presents the comprehensive performance metrics across all training and testing modality configurations.

Table 3.6: Comparative performance analysis of LanguageBind-style fine-tuning approaches across diverse modality configurations on the PetFinder dataset. While this approach consistently achieved strong performance (average accuracy 0.382), it required extensive labeled data and comprehensive retraining for each novel modality configuration.

| Train Modalities | Test Modalities | Accuracy |
| --- | --- | --- |
| Text+Image | Tabular | 0.369 |
| Text+Tabular | Image | 0.387 |
| Image+Tabular | Text | 0.374 |
| Text | Image+Tabular | 0.371 |
| Text | Image | 0.386 |
| Text | Tabular | 0.383 |
| Image | Text+Tabular | 0.379 |
| Image | Text | 0.384 |
| Image | Tabular | 0.385 |
| Tabular | Text+Image | 0.389 |
| Tabular | Text | 0.388 |
| Tabular | Image | 0.386 |
| Average | — | 0.382 |

Performance and Resource Trade-offs   The experimental findings reveal that fine-tuning-based models achieve marginally superior accuracy (0.382 on average) com-pared to the in-context TAMML approach (0.355 on equivalent data splits). However, this modest 2.7 percentage point advantage comes with significant practical constraints—specifically, the requirement for comprehensive labeled data across all potential modality combinations and the necessity for complete retraining whenever new modality types are introduced.

Practical Implications for Deployment    From an implementation perspective, TAMML offers several substantial advantages:

- True zero-shot generalization capabilities for entirely unseen modality configurations without requiring anticipatory training.

- Elimination of retraining requirements and cross-modal supervised dataset dependencies, significantly reducing computational and data collection overhead.

- Favorable performance-flexibility trade-off (0.355 vs. 0.382 accuracy) that prioritizes deployment versatility and adaptation capabilities over marginal performance gains.

These findings suggest that while specialized fine-tuning approaches may establish theoretical performance upper bounds under ideal conditions, the TAMML framework offers compelling practical advantages in real-world deployment scenarios where system flexibility, resource constraints, and generalization capabilities represent critical considerations.

### 3.4.4 Modality Mismatch Ablation Studies

This section investigates the contributions of individual components within TAMML through systematic ablation studies. Components were sequentially incorporated to assess their performance impact, with results documented in a comprehensive ablation analysis. For this evaluation, the TAMML framework utilized GPT-3.5 as its underlying large language model.

Text Transformation  Our initial ablation experiment evaluated the efficacy of converting modality features into textual representations. This transformation yielded an approximate 2% performance enhancement compared to the embedding-based SDEdit method, demonstrating reduced modality discrepancy between training and inference phases relative to embedding representations. This improvement manifested consistently across most data modalities, with the exception of tabular data, which exhibited a performance decline of approximately 10%. This anomaly can be attributed to the structured nature of tabular text transformation, which creates a notable stylistic divergence from more natural, human-like prose, particularly impacting the inference performance of tabular data.

Modality Summarization  The implementation of modality summarization substantially enhanced tabular data accuracy, increasing from 0.277 to 0.321 on average. Following this enhancement, TAMML surpassed even the strongest baseline competitor, SDEdit. These results suggest that summarization effectively normalizes diverse text formats into a unified stylistic framework, reducing heterogeneity in text transformation and strengthening alignment across varied data formats.

Reasoning Augmentation   The incorporation of reasoning augmentation elevated average performance from 0.321 to 0.334. Moreover, it contributed to more consistent performance across diverse scenarios. The variance observed with augmentation was notably lower than without it, signifying enhanced reliability and consistency across different modality combinations.

Text-Style Translation across Modality   Text-style translation effectively narrows the gap between training and inference phases, yielding approximately 6% improvement from 0.334 to 0.355. This enhancement was particularly evident in scenarios where the textual style discrepancy remained consistent across phases, as exemplified in image-to-table transformations. Such consistency facilitated the model's ability to establish more precise mapping functions.

### 3.4.5   Modality Mismatch Analysis

This section presents supplementary analyses and discussions that yield valuable insights from our experimental findings. We provide supporting evidence through visualization techniques and distribution distance measurements.

Visualization for Distribution Alignment   In Section 3.4.3, we empirically validated the effectiveness of text transformation in TAMML through performance metrics. Additionally, we visualized 1,400 data points across multiple modalities with their position-aware embeddings using UMAP [86] in Figure 3.8. The left visualization depicts the original distributions of image and text embeddings, while the right visualization illustrates the corresponding distributions after applying TAMML 's summarization module. We observed that the distributional boundaries between image and text modalities became significantly less pronounced, indicating closer semantic alignment. Quantitatively, TAMML reduced the average instance Euclidean distance between image and text in the semantic space from 10.213 to 0.411, as documented in Table 3.7.

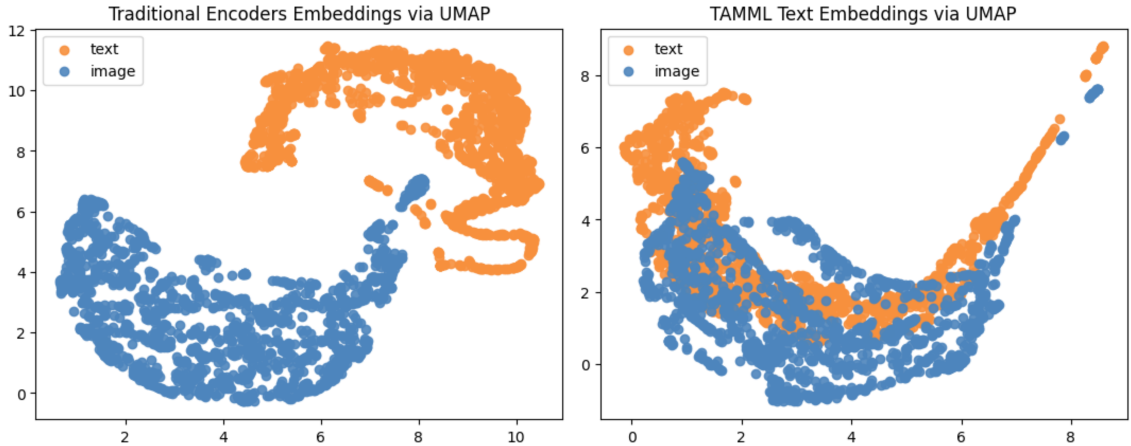

Figure 3.8: Visualization of embedding distributions before and after applying the proposed approach. The left panel shows the original distinct clustering of image and text embeddings, while the right panel demonstrates how the method brings these distributions closer in the semantic space, reducing modality boundaries.

|  | w/o | Normalization | Standardization |
| --- | --- | --- | --- |
| Embedding | 10.213 | 5.444 | 39.151 |
| Text | 0.411 | 0.101 | 0.584 |

Table 3.7: Average Euclidean distances between modality embeddings under different normalization conditions. Text representations consistently demonstrated significantly smaller inter-modality distances compared to embedding representations, confirming better alignment across modalities in the semantic space.

Effects of the Image Caption Models  To investigate whether improvements in text transformation stem exclusively from GPT-4's capabilities, we conducted experiments substituting it with smaller open-source alternatives. Our ablation studies evaluated performance across four distinct image foundation models.

Table 3.8 displays results averaged across twelve training-inference modality combinations. The evidence indicates that employing smaller image caption models maintained comparable performance with TAMML , confirming the approach's resilience across various foundation models.

|  | Image Caption Models | | | |
|---|---|---|---|---|
| Pet  \|  Acc ↑ | Blip2 | Kosmos2 | Flamingo | GPT4 |
| Average | 0.303 | 0.299 | 0.293 | 0.307 |

Table 3.8: Comparative analysis of various image caption models on the PetFinder dataset, with results averaged across twelve training-inference modality combinations. Despite GPT-4 achieving slightly superior results, all models demonstrated comparable performance, indicating the foundation model's minimal impact on TAMML 's effectiveness.

In-context Modality Transfer Outperforms Zero-shot Learning Based Methods    TAMML implements text-style translation across modalities by transforming training modality combinations into testing modality combinations, thereby reducing the semantic gap between them through LLMs. Zero-shot learning baselines employ comparable concepts by developing generative models for modality translation. For comprehensive comparison, we collected various training-testing data pairs and generated visualizations for each combination.

In our visualizations, orange denotes the source modality, blue represents the target modality, and purple indicates the transformed source modality. Figure 3.9 illustrates visualization results from our proposed approach. Figure 3.10 presents visualization results using SDEdit. The results demonstrate that our translation method effectively aligns source modalities with target modalities in semantic space.

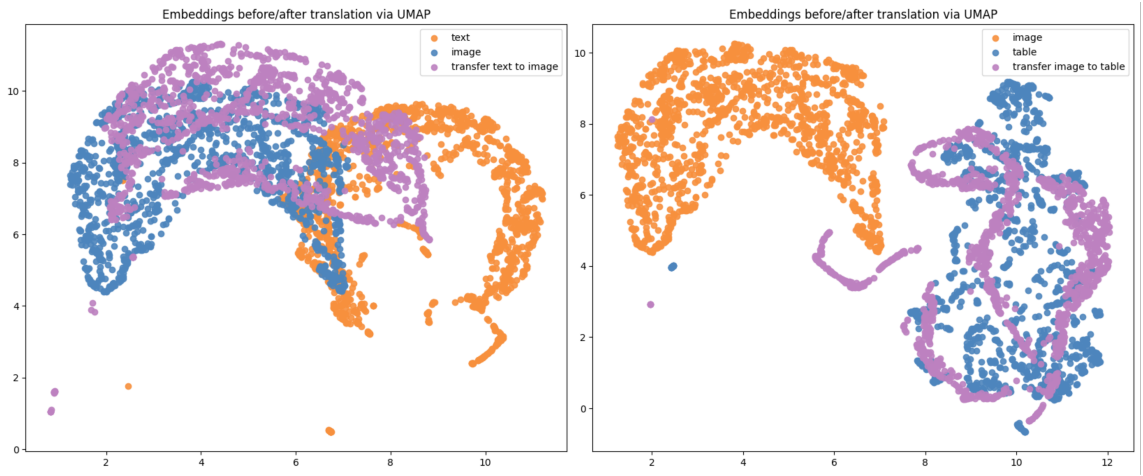

Figure 3.9: Visualization of the cross-modality translation approach, showing how source modality data (orange) is effectively transformed (purple) to align closely with the target modality distribution (blue) in semantic space.

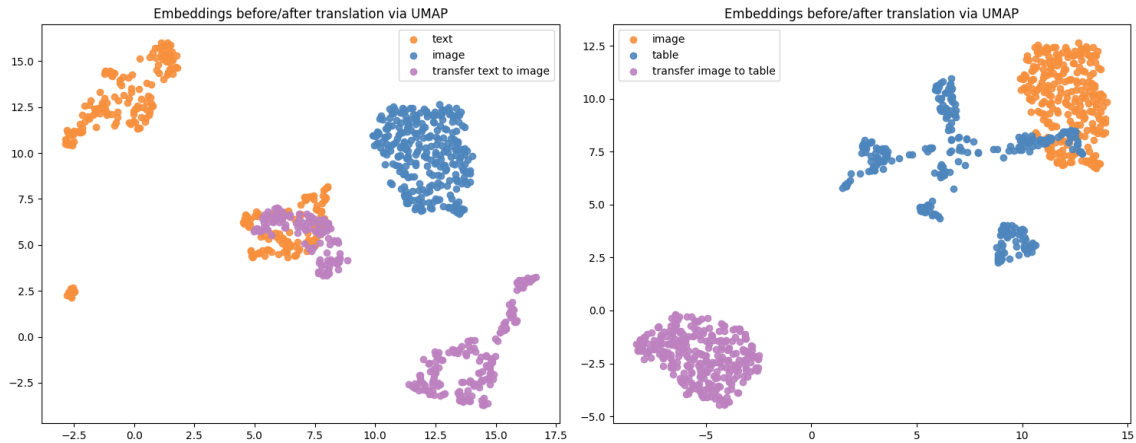

Figure 3.10: Visualization of SDEdit-based cross-modality translation on the PetFinder dataset, demonstrating less effective alignment between transformed source modality (purple) and target modality distribution (blue) compared to the proposed approach.

### 3.4.6 Modality Robustness Baselines

- MLLMs: For robustness comparison, we selected two cutting-edge open-source Multimodal Language Models (MLLMs): Kosmos-2 [103] and Flamingo [3]. This comparison highlights performance differences between large foundation models lacking specific robust training techniques.

- Robust Training: To assess our text-centric approach against conventional

methods, we implemented several robust training techniques for downstream models, including Gaussian noise injection, dropout, and adversarial training with Projected Gradient Descent (PGD) [85]. These baselines demonstrate whether our LLM-leveraging text-based method provides enhanced robustness compared to traditional embedding-based approaches.

- Text-Centric Approaches: We compared the effects of naive transformation (text conversion without perturbation), random perturbation, and adversarial perturbation to determine whether adversarial prompting enhances robustness beyond merely diversifying inputs and basic text transformation.

### 3.4.7 Modality Robustness Baselines

- Multimodal Large Language Models (MLLMs): Two state-of-the-art open-source MLLMs, Kosmos-2 [103] and Flamingo [3], are selected for comparative resilience analysis. This evaluation highlights performance differences among large foundation models that have not been explicitly trained with robustness as a primary objective.

- Robust Training Baselines: To benchmark the resilience of the proposed text-centric framework against conventional methods, several standard robust training techniques are employed on downstream models. These include Gaussian noise injection, dropout regularization, and adversarial training using Projected Gradient Descent (PGD) [85]. These baselines provide a comprehensive reference for evaluating the effectiveness of the text-based approach, particularly in contrast to traditional embedding-based strategies.

- Text-Centric Strategies: A comparative analysis is conducted across three categories: basic text transformation (without perturbation), random perturbation, and adversarial perturbation. This investigation assesses whether adversarial prompting offers superior robustness relative to simpler strategies that rely on enhancing input diversity through text variation alone.

### 3.4.8 Modality Robustness Experiments

Evaluation Protocol  For comprehensive robustness assessment, methodologies aligned with the MULTIBENCH [66] framework were adopted. The evaluation encompassed three distinct scenarios:

1. Noisy Modality: Gaussian noise was systematically introduced to images at five graduated levels from 10% to 90%; words were selectively removed from text descriptions with probabilities ranging from 10% to 50%; and column features were strategically eliminated from tabular data with probabilities from 10% to 90%.

2. Dynamic Modality: The sequence of input modalities was systematically varied to assess robustness. Given the structure of text-centric alignment and the token-based nature of transformer architectures, these models are expected to exhibit invariance to changes in modality order within the prompt.

3. Missing Modality: Random modalities were strategically excluded during test-time evaluation, substituting zero vectors for absent modalities when implementing robust training methodologies.

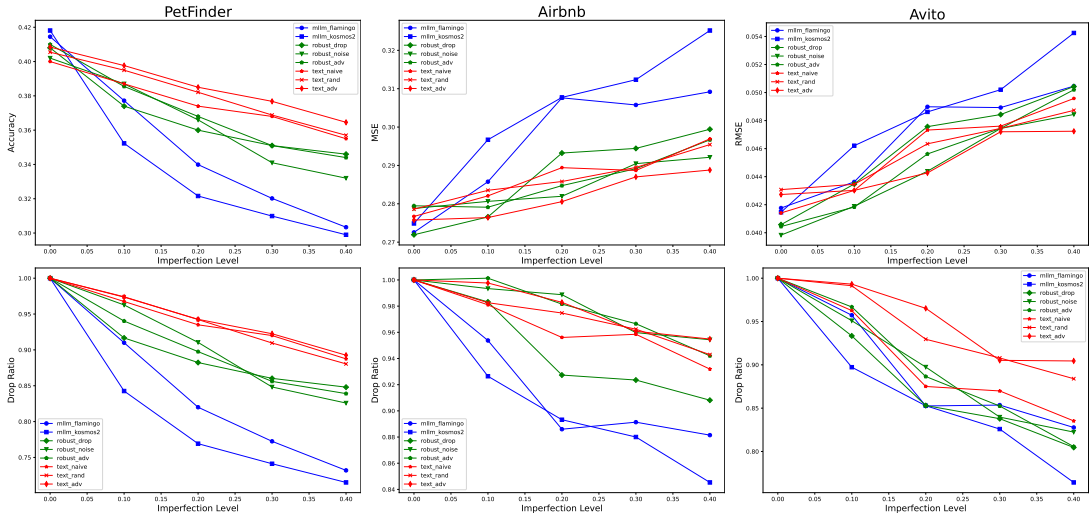

Figure 3.11: Robustness evaluation under noisy conditions across three datasets. The top row displays relative accuracy while the bottom row shows effective robustness (performance drop ratio). Text-centric methods consistently exhibit superior resilience to noise compared to baseline approaches, particularly as noise levels intensify.

Noisy Modality Results  As illustrated in Figure 3.11, our proposed methodology consistently exhibits the minimal performance degradation ratio when subjected to corrupted modality conditions, outperforming all comparative approaches, particularly when noise intensities are elevated. In the context of the Petfinder dataset, the text-centric adversarial technique experienced only an 11.3% reduction in performance, considerably outpacing robust training methodologies (15.2% reduction) and MLLMs (28.5% reduction). Similar patterns are evident in both the Airbnb and Avito datasets, where our proposed approach consistently surpasses all baseline methodologies. Furthermore, Figure 3.12, which applies specific noise types across various modalities, indicates differential vulnerability patterns according to modality type, suggesting a promising avenue for future exploration regarding text-centric modality collapse phenomena.

Dynamic Modality Results  Examination of model behavior under varying input order configurations required testing and averaging performance across all permutations of input sequencing. Analysis of Table 3.9 confirms that our proposed methodology yields the most favorable performance retention ratio when compared against all alternative approaches. In particular, for the Petfinder dataset, the text-centric adversarial approach displayed minimal performance deterioration of just 5.5%, markedly superior to MLLMs (11.1% decline) and vastly outperforming traditional robust training techniques, which exhibited accuracy levels approximating random chance selection. This performance advantage persists consistently throughout both the Airbnb and Avito datasets, where our proposed framework maintains its superiority over all comparative methodologies. The inherent design of token-based processing combined with text-centric representation naturally facilitates exceptional stability when confronted with variations in input sequencing, underscoring its adaptability across numerous practical applications.

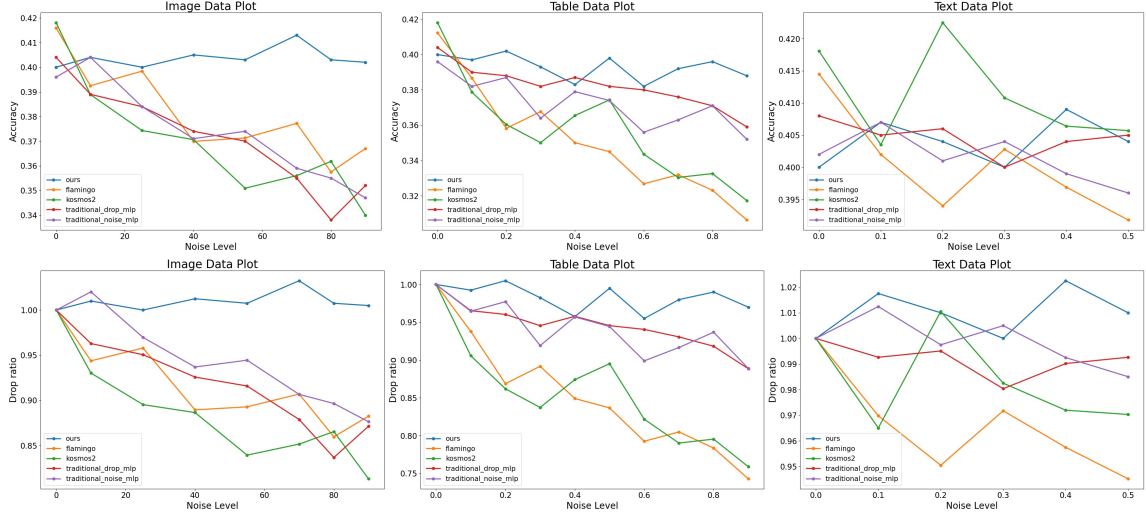

Figure 3.12: Modality-specific noise impact on accuracy (top) and performance drop ratio (bottom) for Image (left), Table (center), and Text (right). Results reveal varying degrees of sensitivity to noise across different modalities, suggesting opportunities for targeted robustness enhancements.

Missing Modality Results  To investigate performance under conditions of incomplete modality availability during inference, we conducted comprehensive evaluations across diverse configurations of excluded modalities. As illustrated in Table 3.10, our proposed methodology consistently achieves the most favorable performance preservation ratio among all comparative approaches. Specifically for the Petfinder dataset, the text-centric adversarial technique exhibited a minimal performance reduction of merely 10%, dramatically outperforming both conventional robust training methodologies (22.5% reduction) and multimodal language models (26.2% reduction). This performance pattern is consistently replicated throughout both the Airbnb and Avito experimental datasets, where our proposed framework maintains clear superiority over all baseline approaches, definitively confirming its exceptional capacity to maintain prediction quality when confronted with incomplete modality inputs.

|  |  | Petfinder | | Airbnb | | Avito | |
|---|---|---|---|---|---|---|---|
|  |  | Drop | ACC↑ | Drop | MSE↓ | Drop | RMSE↓ |
| MLLM | Kosmos2 | 0.883 | 0.371 | 0.954 | 0.285 | 0.953 | 0.043 |
|  | Flamingo | 0.890 | 0.374 | 0.961 | 0.283 | 0.931 | 0.044 |
| Robust Training | Noise | 0.704 | 0.296 | 0.569 | 0.478 | 0.512 | 0.080 |
|  | Dropout | 0.745 | 0.313 | 0.632 | 0.430 | 0.611 | 0.067 |
|  | Adversarial | 0.719 | 0.302 | 0.578 | 0.470 | 0.594 | 0.069 |
| Text centric | Naive | 0.919 | 0.386 | 0.981 | 0.277 | 0.976 | 0.042 |
|  | Random | 0.928 | 0.390 | 0.871 | 0.280 | 0.953 | 0.043 |
|  | Adversarial | 0.945 | 0.397 | 0.992 | 0.274 | 0.977 | 0.042 |

Table 3.9: Performance assessment under Dynamic Modality conditions displaying comparative metrics for robustness (values on left) and effectiveness (values on right) across the three experimental datasets. The text-centric adversarial prompting approach consistently exhibits superior performance compared to all baseline methods and demonstrates exceptional stability when confronted with variations in input ordering.

Comparison with Traditional Robust Training  Unlike conventional robust training methods (e.g., projected gradient descent [85], Gaussian noise), the proposed approach offers several distinctive advantages:

- Operates in natural language space, rendering perturbations interpretable.

|        |             | Drop  ACC↑ | Drop  MSE↓ | Drop  RMSE↓ |
|        |             | Petfinder  | Airbnb     | Avito       |
| ------ | ----------- | ---------- | ---------- | ---------- |
| MLLM   | Kosmos2     | 0.719  0.302 | 0.851  0.320 | 0.824  0.050 |
|        | Flamingo    | 0.738  0.310 | 0.855  0.318 | 0.803  0.051 |
| Robust Training | Noise | 0.769  0.323 | 0.852  0.319 | 0.836  0.049 |
|        | Dropout     | 0.738  0.310 | 0.850  0.320 | 0.824  0.050 |
|        | Adversarial | 0.785  0.330 | 0.883  0.308 | 0.854  0.048 |
| Text centric | Naive | 0.861  0.362 | 0.880  0.309 | 0.872  0.047 |
|        | Random      | 0.881  0.370 | 0.877  0.310 | 0.871  0.047 |
|        | Adversarial | 0.900  0.378 | 0.899  0.302 | 0.891  0.046 |

Table 3.10: Missing Modality Evaluation results showing relative robustness (left values) and effective robustness (right values) across three datasets. Text-centric adversarial prompting significantly outperforms all baseline methods, demonstrating superior robustness to modality absence.

- Eliminates requirements for gradient computation or model access.

- Seamlessly integrates with in-context learning and frozen LLMs.

Experiments in Section 3.4 demonstrate that adversarial prompting surpasses traditional noise-based training and achieves exceptional robustness under noisy, missing, and dynamically ordered modality configurations.

### 3.4.9 Modality Robustness Ablation Studies

Module Ablation   An ablation study was conducted to evaluate the contributions of the two core components of the proposed framework: the Alignment Module and the Permutation Module. As shown in Table 3.11, both modules contribute comparably to the overall performance. The removal of both components leads to a substantial decline in performance, nearing that of baseline multimodal language models (MLLMs), indicating their critical role in the system's effectiveness.

| Model | Noisy | Dynamic | Missing |
|---|---|---|---|
| w/o both | 0.3342 | 0.373 | 0.302 |
| w/o alignment | 0.3727 | 0.383 | 0.363 |
| w/o perturbation | 0.3659 | 0.386 | 0.362 |
| Mixtral8x7b | 0.4033 | 0.397 | 0.378 |
| GPT-3.5-turbo | 0.4037 | 0.398 | 0.379 |
| GPT-4o | 0.4086 | 0.398 | 0.381 |

Table 3.11: Module contribution and LLM impact on PetFinder dataset performance. Both alignment and perturbation modules are essential for optimal results. While GPT-4o achieves the highest performance overall, differences between various LLMs remain relatively minimal (maximum 2% accuracy variation).

Language Model Ablation  To assess the generalizability of robustness across different language models, multiple LLMs were evaluated. Results in Table 3.11 demonstrate that GPT-4o achieves the highest accuracy among the tested models. Nonetheless, the variation in performance across model types and sizes is relatively small, with a maximum observed difference of approximately 2% in accuracy. These findings suggest that the robustness enhancements introduced by the proposed method are transferable across a range of language model architectures.

## 3.5 Scalability and Efficiency

While text-centric alignment and adversarial prompting offer strong robustness and generalization performance, they also introduce inference-time overhead due to longer prompts and multiple transformation stages. In this section, we analyze the scalability and computational efficiency of our proposed method, benchmarking its cost, latency, and throughput compared to traditional and fine-tuned baselines.

### 3.5.1 Latency Overhead

We compare average per-sample inference latency across four transformation stages in TAMML and three prior multimodal generation systems: SDEdit, DDRM, and Idinvert. As shown in Table 3.12, TAMML introduces a total latency of 1.76 seconds per sample. This is only slightly higher than zero-shot diffusion methods such as SDEdit (1.63s) and DDRM (1.89s), and remains well within practical limits for batch inference.

Importantly, TAMML requires no retraining or fine-tuning for new modalities, unlike diffusion or embedding-based baselines. Although its latency is higher than a direct input baseline (0.25s), the added transformation cost is a fixed, interpretable pipeline of text-based reasoning steps.

### 3.5.2 Extra Token Cost

Text-centric transformations also increase the number of tokens fed to the LLM. Table 3.12 shows that each alignment stage contributes cumulatively to a $3.21\times$ token expansion over the original input. The most significant growth comes from reasoning augmentation and summarization.

While this introduces computational cost, it also reflects a deeper enrichment of the model's understanding via structured, interpretable language. These tokens are not arbitrary padding but context-rich reasoning artifacts.

| Method / Operation | Avg. Latency (s) | Cumulative Extra Tokens (× Baseline) |
|---|---|---|
| w/o Alignment (Direct Input) | 0.25 | 0.00 |
| TAMML - Text Transformation | 0.42 | 1.00 |
| TAMML - Text-style Translation | 0.37 | 2.08 |
| TAMML - Modality Summarization | 0.45 | 2.73 |
| TAMML - LLM Reasoning Augmentation | 0.52 | 3.21 |
| TAMML (Total) | 1.76 | 3.21 |
| SDEdit | 1.63 | — |
| DDRM | 1.89 | — |
| Idinvert | 2.01 | — |

Table 3.12: Average Latency and Token Overhead per Sample. TAMML maintains competitive latency while introducing moderate token expansion through each alignment stage.

### 3.5.3 Throughput and Scalability

We benchmark throughput on an A6000 GPU for three methods:

- Direct input baseline (no transformation)

- TAMML (ours)

- Fine-tuned modality-specific encoders (e.g., LanguageBind)

While TAMML incurs some inference cost, it avoids the need for retraining, large-scale cross-modal datasets, or modality-specific adapters. As shown in Table 3.13, TAMML achieves a favorable trade-off between performance and efficiency.

Table 3.13: Comparison of inference-time scalability. TAMML strikes a balance between accuracy, token cost, and deployment practicality.

| Method | Accuracy (PetFinder) | Token Cost | Retraining Required |
|---|---|---|---|
| Direct Input | 0.346 | 1.00× | No |
| LanguageBind (Finetuned) | 0.382 | 1.00× | Yes |
| TAMML (Ours) | 0.355 | 3.21× | No |

### 3.5.4 Scalability Trade-Off Discussion

Although TAMML introduces a moderate increase in latency and token cost, it offers significant benefits in scalability. It eliminates the need for retraining, supports unseen modality compositions, and reduces dependence on expensive multimodal data curation.

In real-world settings—especially under constraints such as limited compute, privacy restrictions, or rapid task switching—TAMML remains a viable and efficient solution. The additional token cost and inference time represent a fixed cost amortized across flexible, dynamic modality handling.

In summary, our framework incurs modest overhead but yields significant practical advantages, including robustness, generalization, and deployment readiness in dynamic multimodal environments.

# Chapter 4  Inference-Time Optimization and Uncertainty-Aware Reasoning

Large Language Models (LLMs) have demonstrated exceptional capabilities in reasoning across a variety of natural language-based tasks. However, their potential extends far beyond multiple-choice questions or single-question answering. This chapter explores the optimization capabilities of LLMs across diverse tasks and problem dimensions, with a particular focus on inference-time optimization techniques and uncertainty-aware reasoning.

## 4.1  Motivations

Optimization involves iteratively generating and evaluating solutions to improve a given objective function. Our research assesses LLM performance in interactive optimization, where each step generates new solutions based on previous ones and their values. This approach is essential for enhancing LLM reasoning and decision-making capabilities, especially in complex tasks requiring multiple steps or iterations.

A key element in optimization algorithms is uncertainty estimation, which is vital for guiding decisions, balancing exploration and exploitation, and improving al-gorithm efficiency. Uncertainty estimation techniques, such as those used in bandit algorithms or Bayesian optimization, can dynamically adjust learning rates or hyperparameters. In combinatorial optimization (e.g., genetic algorithms or simulated annealing), uncertainty estimation informs heuristic decisions like mutation rates or temperature adjustments. Thus, developing robust methods to quantify uncertainty in LLMs, particularly for prompt optimization, is essential.

Previous approaches to measuring uncertainty in LLMs primarily rely on token-level or sentence-level generation likelihoods, often represented by metrics like token disparity probability, predictive entropy, and reciprocal of perplexity. However, we argue that such token-level uncertainty measurements are more indicative of model output confidence or output diversity, which may not align with the needs of multi-step reasoning or prompt optimization tasks.

We further justify our hypothesis in Figure 4.1 which illustrates the relationship between LLM correctness uncertainty, answer uncertainty (diversity), and response accuracy in the GSM8K dataset. A reliable uncertainty quantification metric targeting correctness of binary classification problem exhibits 50% accuracy when uncertainty is at its highest. On the contrary, answer uncertainty, designed to capture the confidence (diversity) of responses fails to be sufficient for prompt optimization tasks.

## 4.2 Problem Setting

We design four optimization tasks that require models to algorithmically search for optimal parameter values. These tasks—Gradient Descent, Hill Climbing, Grid

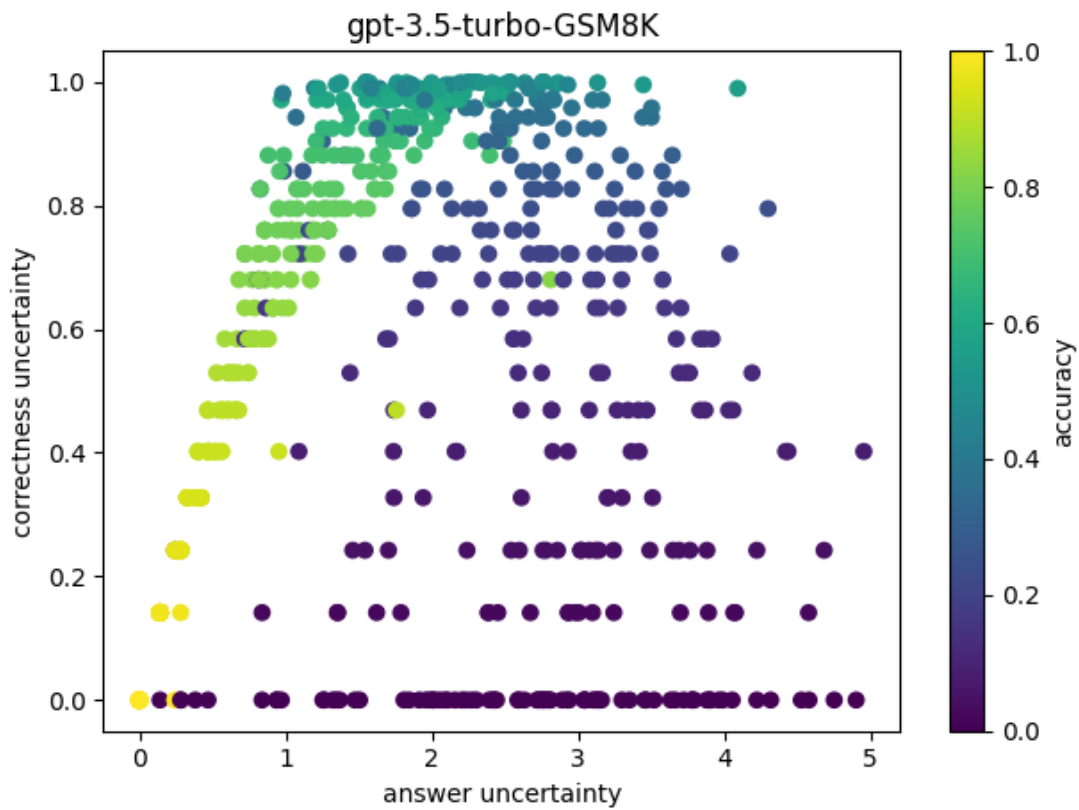

Figure 4.1: Relationship between LLM correctness uncertainty and response accuracy in the GSM8K dataset. A reliable uncertainty quantification metric exhibits 50% accuracy at maximum uncertainty, while standard confidence-based metrics fail to provide sufficient guidance for prompt optimization tasks.

Search, and Black-Box Optimization—span distinct optimization paradigms: gradient-based, meta-heuristic, decision-theoretic, and Bayesian methods. Grid Search and Hill Climbing operate over discrete search spaces, while Gradient Descent and Black-Box Optimization involve continuous spaces.

- Gradient Descent evaluates the model's ability to perform advanced calculations and apply gradient-based optimization. LLMs must define a loss function, compute gradients, and iteratively update parameters using the provided learning rate.

- Hill Climbing tests the model's capacity to follow unfamiliar, rule-based heuristics. Starting from an initial solution, the model explores neighbors by incrementally increasing or decreasing individual elements to find improvements.

- Grid Search assesses the model's systematic reasoning and exhaustive exploration skills. LLMs must enumerate all grid points within a predefined space and identify the one yielding the lowest loss.

- Black-Box Optimization measures the model's decision-making ability in abstract settings. Here, LLMs iteratively propose new solutions to minimize an unknown loss function without access to its internal structure.

## 4.3 Prompt Optimization with Uncertainty Feedback

### 4.3.1 Iterative Prompting Framework

We propose an iterative prompting framework that guides LLMs to progressively improve solutions within a given search space. As illustrated in Figure 4.2, the framework enables LLMs to perform optimization through a sequence of interactive and feedback-driven steps.

We adopt a prompting strategy that combines Chain-of-Thought reasoning and iterative prompting. The LLM generates intermediate reasoning steps to accomplish each stage of the optimization process. At the beginning of each task, the model is asked to define a loss function based on provided sample data. Each optimization iteration then consists of two steps:

1. Propose new candidate solutions based on algorithmic instructions and previ-

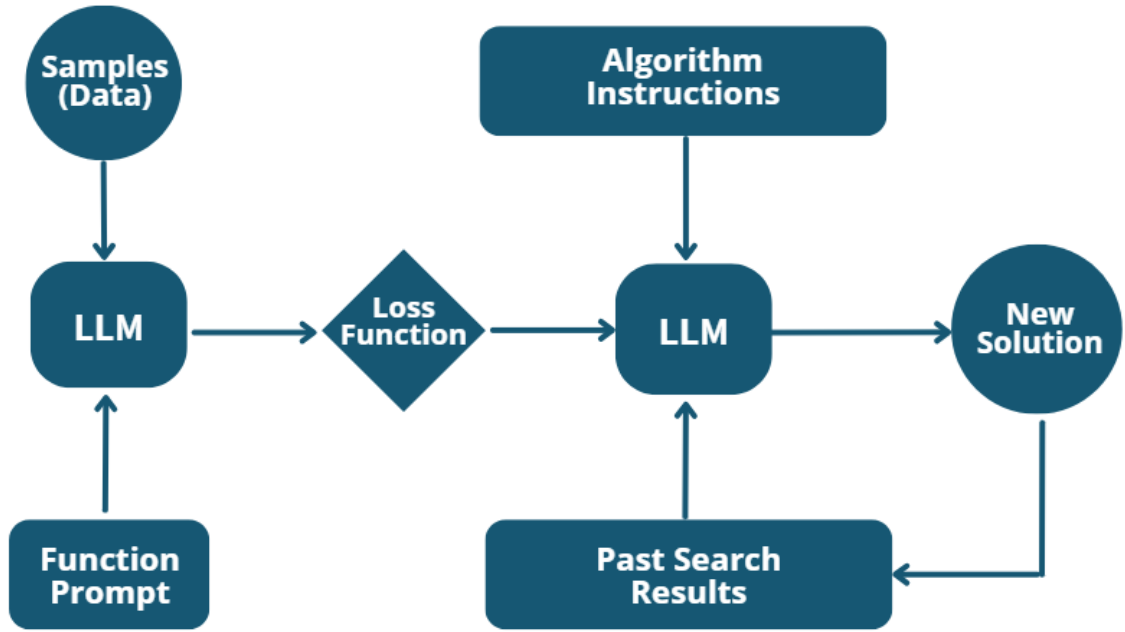

Figure 4.2: Prompting strategy for optimization with LLMs. The process involves: (1) loss function formulation from samples, (2) solution generation following algorithm instructions, (3) loss calculation and result tracking, and (4) iteration until stopping criteria are met.

ous search outcomes.

2. Evaluate the loss of the new solutions and incorporate the results into the prompt for the next iteration.

These steps are repeated until a stopping criterion is met. To support an interactive optimization process, we leverage the chat interface of LLMs, where the full conversation history serves as the evolving prompt. This setup allows the model to maintain memory of prior search results and reasoning trajectories.

### 4.3.2 Uncertainty Types in Optimization Tasks

Characterizing different types of uncertainty in LLM-driven optimization is essential for designing robust prompting strategies. We quantify the following un-

certainty types:

- Answer Uncertainty (AnsU) captures the model's confidence and output diversity. It is measured as the entropy of the predicted answer distribution:

$$\text{AnsU}(x) = -\sum_i p(y_i|x) \log p(y_i|x) \tag{4.1}$$

  where $p(y_i|x)$ denotes the probability of generating answer $y_i$ from perturbed inputs given prompt $x$.

- Correctness Uncertainty (CU) quantifies the model's uncertainty regarding the correctness of its outputs. It is computed as the entropy over the distribution of correctness labels:

$$\text{CU}(x) = -\sum_i p(c_i|x) \log p(c_i|x) \tag{4.2}$$

  where $p(c_i|x)$ represents the probability of correctness outcome $c_i$ (e.g., correct or incorrect) based on perturbed inputs.

- Aleatoric Uncertainty (AU) arises from inherent randomness or ambiguity in the data. It reflects uncertainty that cannot be reduced, even with more training data.

- Epistemic Uncertainty (EU) stems from the model's lack of knowledge and can be reduced through improved data coverage or model capacity.

We estimate Aleatoric and Epistemic uncertainty using the Deep Ensembles approach, where total predictive uncertainty is decomposed as follows:

$$H(q(Y|X)) = \underbrace{I(Y;\theta|X)}_{\text{Epistemic Uncertainty}} + \underbrace{\mathbb{E}_{q(\theta|D)}[H(q(Y|X,\theta))]}_{\text{Aleatoric Uncertainty}} \quad (4.3)$$

where $\theta$ represents the perturbed question instead of model parameters.

### 4.3.3 Uncertainty Quantification Metrics

We evaluate different types of black-box uncertainty metrics to determine their effectiveness in prompt optimization:

- Normalized Predictive Entropy (NPE) measures the uncertainty of generated text by calculating the average entropy of possible output sequences given an input context x:

$$\text{NPE(x)} = \frac{1}{N} \sum_n \sum_i \ln p(s_i|s_{<i}) \quad (4.4)$$

where $N$ is the number of generations and $s_i$ is the $i$-th token of sentence s.

- Length-Normalized Predictive Entropy (LNPE) adjusts for sentence length by normalizing the entropy with the number of tokens:

$$\text{LNPE(x)} = \frac{1}{N} \sum_n \left(\frac{1}{S_n}\right) \sum_i \ln p(s_i|s_{<i}) \quad (4.5)$$

where $S_n$ is the token length of sentence $n$.

- TopK-Token Disparity (Top-DISP) calculates the average difference in probability between the top-1 and top-2 tokens for each token within the output

sequence:

$$\text{Top-DISP(x)} = -\frac{1}{N}\sum_n \left(\frac{1}{S_n}\right)\sum_i \left|\ln\frac{p(s_{i,\text{top1}}|s_{<i})}{p(s_{i,\text{top2}}|s_{<i})}\right| \tag{4.6}$$

- Intra-Sample Similarity (Intra) computes the average of the uncertainties discerned individually for each sample output:

$$\text{Intra(x)} = -\frac{\sum_{i=0}^{k} c(x_i, y_i)}{k+1} \tag{4.7}$$

where $c(x_i, y_i)$ is the uncertainty articulated by the LLM for each perturbed input and output pair.

## 4.4 Benchmarking Prompt Uncertainty

### 4.4.1 Benchmark Construction

We propose a benchmark dataset designed to evaluate uncertainty types for uncertainty quantification metrics specifically for prompt optimization tasks. By performing extensive sampling on LLMs, we construct large, tree-structured reasoning traces from model outputs. Figure 4.3 illustrates our benchmark workflow.

The construction workflow consists of three steps:

1. Random Perturbation: We perturb each input question x by prompting the language model to regenerate M new questions $\{x_j\}_{j=1}^M$ that preserve the original meaning.

2. Random Sampling: For each perturbed question $x_j$, we perform $K$ rounds of

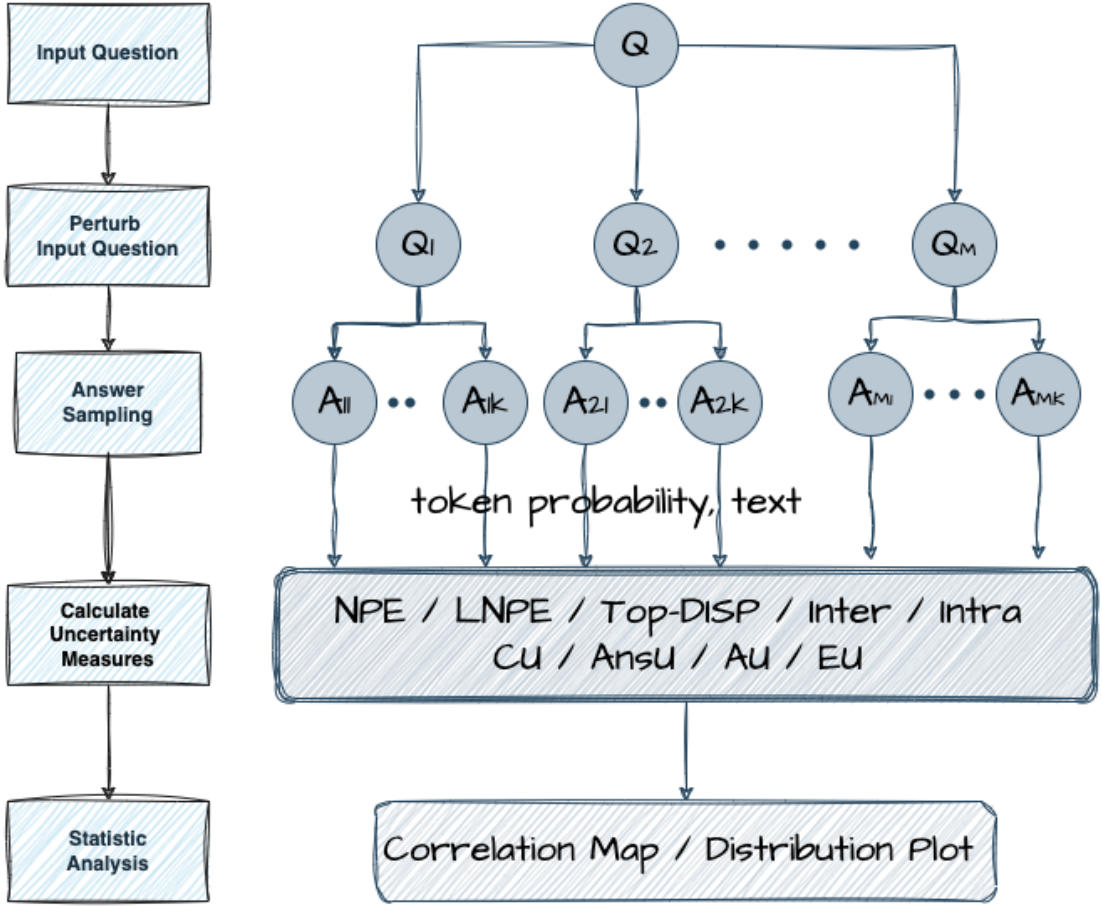

Figure 4.3: Benchmark workflow for evaluating uncertainty in prompt optimization. The three-step process at each level includes: 1) question perturbation, 2) model output sampling with temperature adjustments, and 3) uncertainty measurement using different metrics.

answer sampling with temperature, yielding a set of responses $\{y_{jk}\}_{k=1}^{K}$.

3. Calculate Uncertainty: We use the $j \times k$ responses to compute black-box uncertainty metrics and uncertainty ground truth for the input question.

### 4.4.2 Evaluation Metrics

Three novel metrics were designed for the comprehensive evaluation of LLM optimization capabilities. These metrics offer versatility in assessing LLM performance across diverse tasks, making concurrent evaluation easier. Their reliance on ratio measures, rather than differences, makes them less sensitive to sample variations.

- Goal Metric quantifies the optimization effectiveness by measuring the relative improvement from the initial solution:

$$G_j = \frac{1}{N} \sum_{i=1}^{N} \frac{loss_{LLM,init} - loss_{LLM,i}}{loss_{LLM,init}} \quad (4.8)$$

where $loss_{LLM,init}$ denotes the initial loss for sample $j$, $loss_{LLM,i}$ is the loss after the $i$-th LLM trial, and $N$ is the total number of trials.

- Policy Metric evaluates how closely the LLM's outputs approach the optimal (ground truth) solution:

$$P_j = \frac{1}{N} \sum_{i=1}^{N} \frac{loss_{LLM,i} - loss_{truth}}{loss_{truth}} \quad (4.9)$$

where $loss_{truth}$ is the ground truth loss for sample $j$.

- Uncertainty Metric captures the variance in the LLM's outputs across repeated trials under identical conditions:

$$U_j = \frac{1}{N} \sum_{i=1}^{N} \left( loss_{LLM,i} - \overline{loss_{LLM}} \right)^2 \quad (4.10)$$

where $\overline{loss_{LLM}}$ is the mean loss over the $N$ trials for sample $j$.

### 4.4.3 Datasets and Models

We evaluate uncertainty metrics using GSM8K and StrategyQA datasets, which involve multi-step math problems and complex reasoning, respectively. We selected these datasets because they have exact ground truth answers and involve different types of knowledge and problem-solving tasks.

For our optimization experiments, we create five synthetic datasets with dimension values chosen from the set $\{3, 6, 12, 24, 48\}$ and generate instances belonging to $[0, 10]^d$ in each dataset to examine sensitivity to the number of parameters, representing the dimension of the optimization problem.

We use two large language models, GPT-3.5-Turbo and Meta-Llama-3.1-8B-Instruct, to showcase that our benchmarking method is suitable for both commercial and open-source LLMs of varying sizes. For the optimization experiments, we primarily use GPT-3.5-Turbo and GPT-4.

## 4.5 Analysis

### 4.5.1 LLM Performance in Optimization Tasks

Our experiments test the comprehensive optimization capabilities of LLMs, with the following key findings:

- LLMs show strong optimization capabilities in small-scale problems: GPT-3.5-Turbo showcases considerable optimization capabilities across various scenarios. In the Gradient-Descent task, it even surpasses the ground truth, particularly for sample dimensions equal to six. The model also achieves respectable results in the Grid-Search task, despite the exponential increase in grid points as the problem dimension expands.

- LLMs can be black-box optimizers: Experimental results demonstrate that

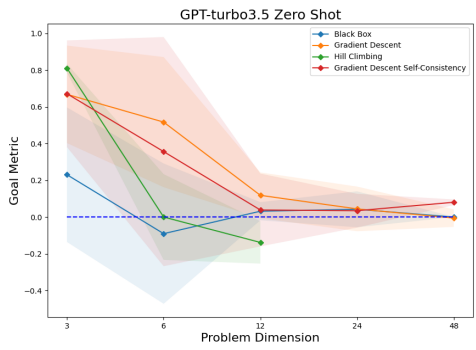 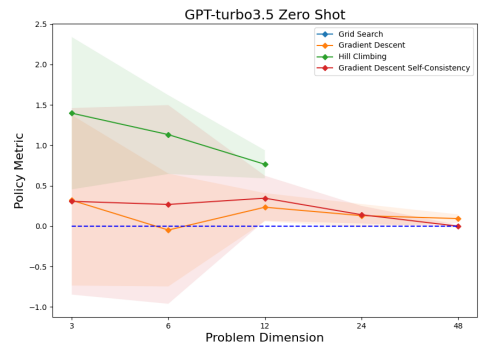

(a) Goal Metric - showing optimization capability

(b) Policy Metric - showing alignment with ground truth

Figure 4.4: Performance metrics for GPT-3.5-Turbo across optimization tasks. Goal Metric values (left) show optimization capability, while Policy Metric values (right) demonstrate alignment with ground truth.

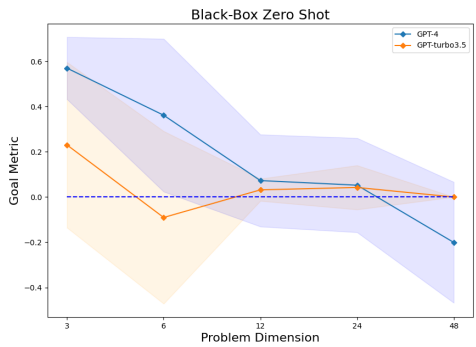 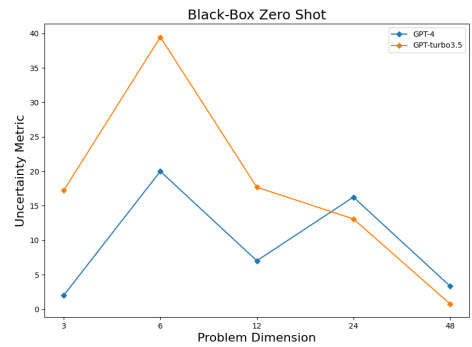

(a) Goal Metric - performance degrades with dimension increase

(b) Uncertainty Metric - stability declines with dimension increase

Figure 4.5: Black-Box optimization performance for GPT-3.5-Turbo and GPT-4. Performance degrades as dimensions increase, with significant drop-off beyond dimension 6 in both optimization capability (left) and solution stability (right).

LLMs can function effectively as black-box optimizers without explicit algorithmic guidance. GPT-3.5-Turbo performs well on 3-dimensional problems, while GPT-4 achieves strong results in both 3- and 6-dimensional settings. However, performance for both models declines as problem dimensionality increases.

- LLMs perform well on gradient descent tasks: The gradient descent experiment evaluates the model's capability in advanced mathematical reasoning.

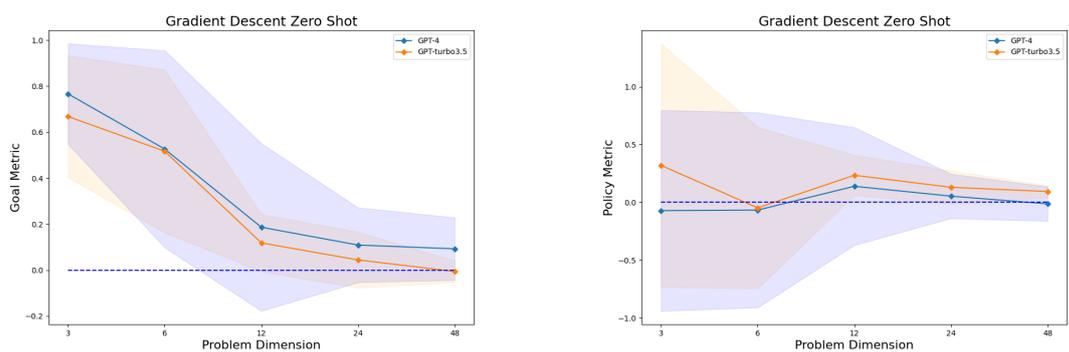

(a) Goal Metric - high values indicate strong optimization

(b) Policy Metric - near-zero values show alignment with ground truth

Figure 4.6: Gradient Descent performance metrics showing consistent optimization capability across problem dimensions. Near-zero Policy Metric values indicate strong alignment with ground truth, while high Goal Metric values demonstrate effective optimization.

The policy metric remains consistently near zero, indicating strong alignment between the LLM's outputs and the ground truth. Although the goal metric declines with increasing sample size, the stable and low policy metric highlights that GPT's performance closely matches that of the ground truth model in this task.

### 4.5.2 Uncertainty Analysis for Prompt Optimization

Our uncertainty benchmarking reveals important insights about the nature of uncertainty metrics in LLMs:

- Existing uncertainty quantification metrics fail to capture correctness uncertainty: As shown in Figure 4.7, the correlation maps across different models and datasets consistently indicate that current metrics align more closely with answer uncertainty (diversity) than with correctness uncertainty. This supports our hypothesis that these metrics are inadequate for evaluating prompt

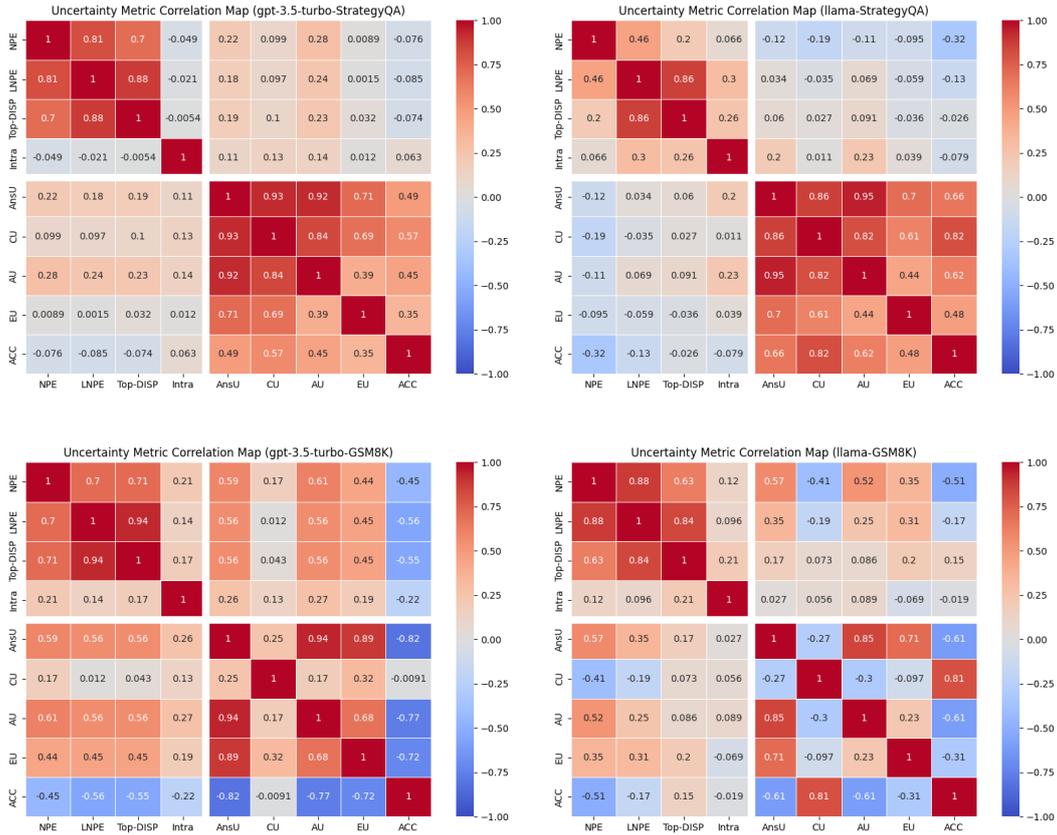

Figure 4.7: Correlation between uncertainty metrics and uncertainty types across models and datasets. The maps reveal that existing metrics correlate strongly with answer uncertainty (confidence/diversity) rather than correctness uncertainty, making them unsuitable for prompt optimization.

optimization effectiveness.

- Current uncertainty metrics primarily capture aleatoric uncertainty: The correlation patterns in Figure 4.7 suggest that existing approaches predominantly reflect aleatoric uncertainty rather than epistemic uncertainty. Aleatoric uncertainty stems from the inherent variability in natural language, where multiple valid outputs may exist for a given input, while epistemic uncertainty arises from model limitations or insufficient knowledge.

- LLMs exhibit a richer solution space in low-dimensional problems: We observe higher uncertainty metric values and greater variation in policy and goal metrics for low-dimensional instances. Interestingly, LLMs tend to perform better

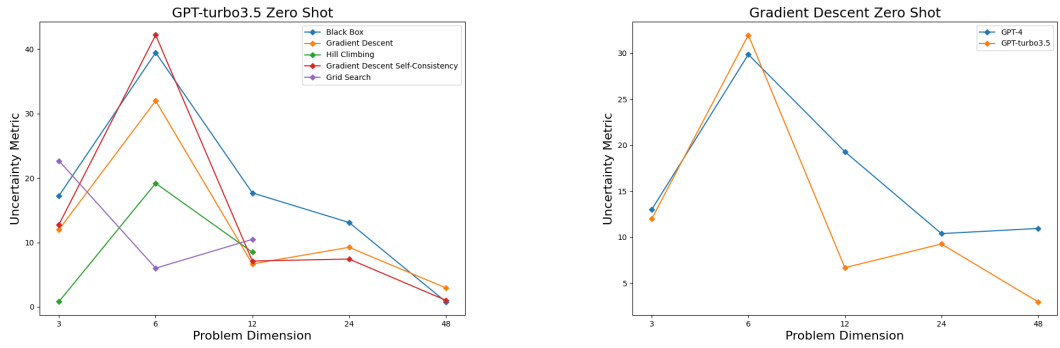

(a) Uncertainty across different optimization tasks

(b) Uncertainty for Gradient Descent optimization tasks

Figure 4.8: Uncertainty Metric results across optimization tasks and problem dimensions. Higher uncertainty values for smaller dimensions suggest LLMs possess a richer solution space for small-scale problems, consistent with their better performance on these problems.

in these settings, suggesting a link between greater uncertainty and improved outcomes. This consistent trend across tasks and models indicates that LLMs explore a richer solution space when solving small-scale problems.

### 4.5.3 Key Findings and Implications

Based on our comprehensive experiments, we draw several important conclusions:

- Pretrained knowledge drives LLM optimization performance: Across all tasks, gradient descent consistently yields the best results, while hill-climbing remains more challenging. This suggests that LLMs rely heavily on pretrained knowledge embedded in their parameters, rather than on user-provided contextual information, for optimization.

- LLMs show promise as hybrid optimizers: The consistently positive goal metric across tasks and datasets indicates that LLMs possess general optimization

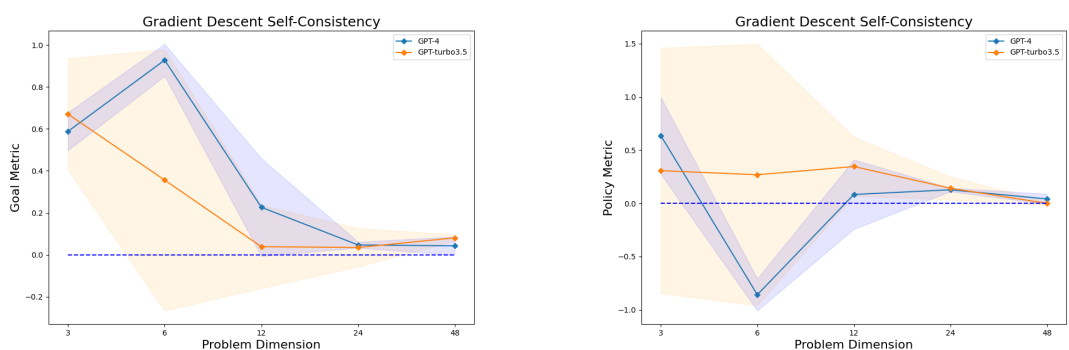

(a) Goal Metric - higher value with self-consistency

(b) Policy Metric - negative values show outperformance

Figure 4.9: Performance improvement with self-consistency in Gradient Descent optimization. Narrower confidence intervals indicate enhanced stability, while negative Policy Metric values with high Goal Metric values demonstrate significant outperformance compared to ground truth.

capabilities. This versatility suggests the potential to dynamically switch between optimization strategies within a single task, enabling better exploration of the solution space and avoidance of local optima.

- Self-consistency prompting enhances stability: In gradient descent tasks, applying self-consistency—repeating each iteration and selecting the most frequent outcome—substantially improves performance. GPT-4, in particular, benefits from this approach, with narrower confidence intervals in both policy and goal metrics, reflecting increased stability and reliability.

- Reliable uncertainty metrics are essential for prompt optimization: Existing uncertainty metrics primarily capture answer uncertainty—reflecting output diversity and confidence—rather than correctness uncertainty, which is critical for optimization tasks. This underscores the need for metrics that are aligned with optimization objectives. In binary classification settings, a reliable correctness-oriented uncertainty metric should yield 50% accuracy under maximum uncertainty, a property not exhibited by current answer-based met-rics, making them insufficient for guiding prompt optimization.

## 4.6 Summary

This chapter provides a comprehensive examination of large language models (LLMs) in the context of optimization and uncertainty-aware reasoning. The analysis spans four optimization tasks, each requiring an understanding of algorithmic instructions and the capacity to generate improved solutions based on prior outputs and their associated objective values.

Evaluation results indicate that LLMs exhibit notable optimization capabilities across a range of problem types, with the strongest performance observed in gradient descent tasks. In contrast, greater difficulty is encountered in meta-heuristic scenarios, which demand adherence to unfamiliar, predefined rules. Performance in grid search tasks further demonstrates the models' ability to execute exhaustive search strategies effectively. Additionally, LLMs perform well in black-box optimization settings, particularly in low-sample regimes, suggesting an inherent aptitude for optimization without explicit guidance.

To assess the reliability of uncertainty quantification methods, a benchmark was introduced for evaluating uncertainty metrics in the context of LLM reasoning. The results reveal that current metrics predominantly capture answer uncertainty —related to output diversity—rather than correctness uncertainty, which is more pertinent to optimization objectives. This observation highlights the need for improved, task-aligned uncertainty metrics that can more effectively inform prompt optimization strategies.

The findings presented in this chapter point to several open research directions. Potential avenues for future investigation include the development of uncertainty estimation techniques tailored to optimization-driven prompting, and the exploration of hybrid optimization strategies that combine the strengths of multiple algorithms within the LLM paradigm.

# Chapter 5  Efficient and Enhanced Code Generation in Low-Resource RTL Language

Hardware description languages (HDLs) like Verilog and VHDL present unique challenges for language models due to their specialized syntax, strict design requirements, and different programming paradigms compared to traditional software languages. This chapter explores novel approaches to improve code generation for these low-resource RTL (Register Transfer Level) languages, with particular focus on agent-based systems, high-quality data generation, and efficient fine-tuning methods.

## 5.1  Challenges in Hardware-Oriented Code Modeling

Hardware description languages serve as the bridge between abstract hardware designs and their physical implementations. Unlike software programming languages, HDLs describe the physical structure and behavior of electronic circuits, which introduces several unique challenges:

- Strict Timing and Resource Constraints: HDL designs must adhere to precise timing requirements and hardware resource limitations.

- Low-Level Parallel Execution Model: HDLs operate on a fundamentally different execution model where operations occur concurrently rather than sequentially.

- Limited Training Data: Compared to popular software languages like Python or JavaScript, there are far fewer publicly available RTL codebases for training language models.

- Complexity of Hardware Semantics: Correct HDL code requires understanding of hardware-specific concepts like clock domains, metastability, and signal propagation delays.

These challenges necessitate specialized approaches to effectively model and generate RTL code, which we address in the following sections.

## 5.2 Language Agents and Environment Feedback for Code Generation

### 5.2.1 Motivation

Hardware description languages (HDLs) such as Verilog and VHDL present unique challenges for language models due to their specialized syntax, strict design requirements, and programming paradigms that differ significantly from traditional software languages. While large language models (LLMs) have achieved substantial progress in software code generation and debugging, their application to hardware design automation remains limited. This section introduces RTLFixer, an AI agent system that enhances LLM capabilities through integration with verification tool feedback to address the specialized requirements of RTL debugging.

### 5.2.2 Preliminaries

This section examines three foundational components underpinning RTLFixer: advancements in LLMs for hardware description language generation, reasoning and action synthesis capabilities in modern language models, and retrieval-augmented generation for enhancing code generation.

LLMs for Verilog Code Generation

The application of Large Language Models to code generation has evolved rapidly, with systems like Codex [20] establishing early benchmarks for capability. GitHub Copilot [34] extended this foundation to create production-grade code completion and generation tools. In the hardware design domain, DAVE [98] pioneered LLM applications specifically for hardware description languages, with VeriGen [127] later expanding both dataset coverage and model exploration. ChipChat [11] demonstrated GPT-4's capabilities in collaborative processor design, while benchmarks such as VerilogEval [71] and RTLLM [79] have established formal evaluation frameworks for LLM-based HDL generation.

Reasoning and Action Synthesis in LLMs

Modern LLMs exhibit sophisticated reasoning and planning capabilities essential for complex problem-solving. Chain-of-thought techniques [149] enable these models to decompose problems into logical sequences, significantly enhancing solution quality. In tool-based environments, models demonstrate effective decision-making and action planning, as evidenced in frameworks like ToolLLM [106].

The ReAct framework [165] represents a critical advancement by integrating reasoning traces with action planning. This integration enables LLMs to interact dynamically with external information sources, substantially improving performance on complex tasks while enhancing their reliability as autonomous agents capable of context-aware problem solving.

Retrieval-Augmented Generation

Retrieval-Augmented Generation (RAG) [60] addresses fundamental limitations in LLMs' handling of knowledge-intensive tasks. While containing extensive implicit knowledge, LLMs often struggle with efficient information access and manipulation. RAG combines parametric memory (the LLM itself) with non-parametric memory (external knowledge bases), creating a hybrid system that can retrieve relevant information based on input queries. This enriches the context available to the model, producing outputs with superior accuracy and factual consistency. In code generation specifically, systems like ReACC [78] and RepoCoder [171] have demonstrated RAG's effectiveness in enhancing LLMs' programming capabilities across diverse tasks and domains.

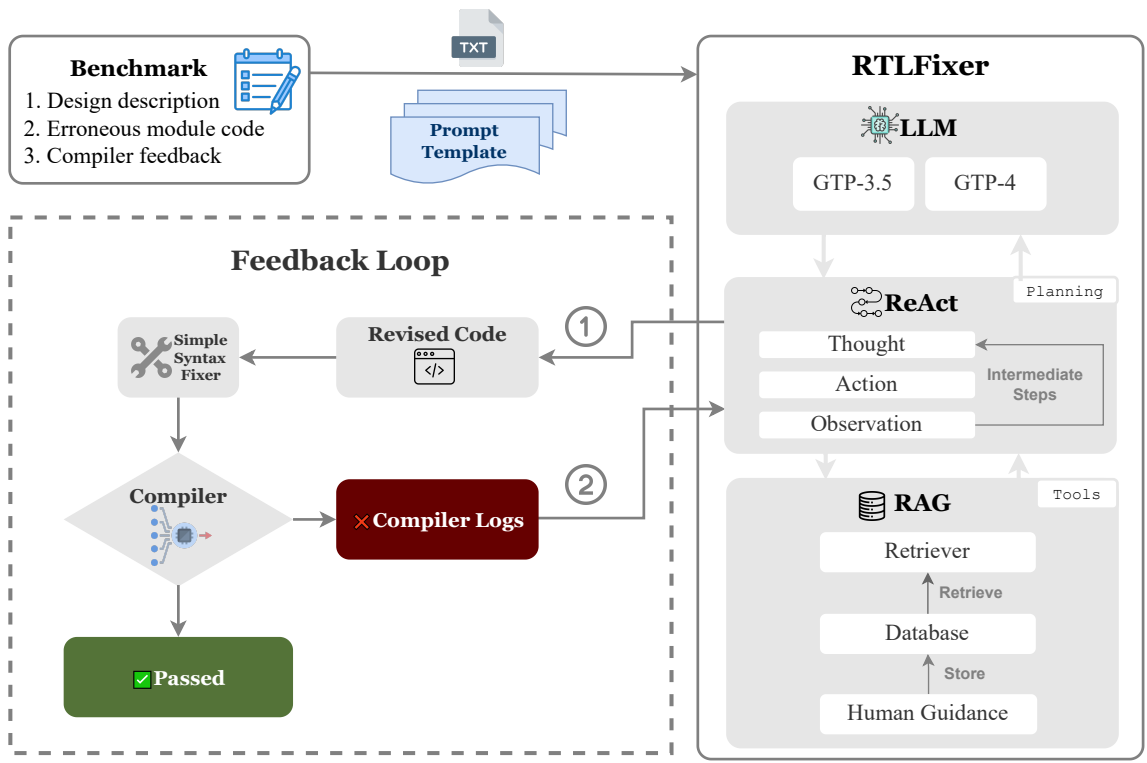

Figure 5.1: Architectural overview of the RTLFixer framework illustrating the seamless integration of language model reasoning with hardware verification tools to create a robust iterative debugging pipeline.

### 5.2.3 RTLFixer System Overview: Resolving Syntax Error with LLM Agents and Retrieval

RTLFixer is an AI agent designed to autonomously debug and fix errors in hardware description language (HDL) code, specifically focusing on Verilog RTL designs. As illustrated in Figure 5.1, the system implements a novel approach that combines the reasoning capabilities of large language models with specialized hardware verification tools.

The core components of the RTLFixer system include:

- A language model-based agent that comprehends hardware semantics

- Verification tool integration to provide diagnostic feedback

- A multi-stage debugging process that addresses progressively more complex error types

- Specialized prompting techniques that incorporate hardware domain knowledge

Table 5.1: Comparative analysis of RTLFixer with conventional RTL debugging approaches across key functional capabilities

| Feature | Manual Debug | Standard LLMs | RTLFixer |
| --- | --- | --- | --- |
| Tool feedback integration | ✓ | X | ✓ |
| Multi-stage approach | ✓ | X | ✓ |
| Hardware semantics reasoning | ✓ | Partial | ✓ |
| Automatic fix generation | X | ✓ | ✓ |
| Adaptability to new tools | Limited | ✓ | ✓ |

Reasoning and Action Planning through ReAct Iterative Prompting

The ReAct prompting mechanism [165] transforms Large Language Models into effective autonomous agents capable of sophisticated reasoning and action planning. This framework enables LLMs to generate interleaved reasoning traces and task-specific actions, creating a dynamic problem-solving process. When presented with an input prompt and ReAct instructions, the LLM generates a structured sequence of steps—each comprising Thought, Action, and Observation components. Figure H.23 illustrates a ReAct instruction prompt, while Figure H.24 demonstrates the self-prompting workflow across iterations.

In this orchestrated process, the LLM first formulates analytical thoughts about addressing errors before strategically selecting appropriate actions. These actions may include generating error explanations, querying the human expertise database, revising code, or submitting updated implementations to the compiler. The outputs from these actions serve as observations that inform subsequent reasoning steps. The agent continues this cycle until achieving successful compilation, at which point it executes the Finish action to generate the final response. If unsuccessful, the system iterates up to a configurable limit (n), allowing for comprehensive evaluation of this automated feedback-driven approach.

For comparative baseline purposes, Figure H.22 illustrates One-shot prompting—a simpler approach involving single-turn compiler feedback. Unlike ReAct, this method lacks the structured decomposition of syntax fixing through explicit reasoning and action planning, as well as the sophisticated multi-round interaction pattern with the compiler.

Retrieval Augmented Generation (RAG)

Retrieval-Augmented Generation (RAG) is an effective framework for enhancing the performance of Large Language Models (LLMs) by integrating expert-curated external knowledge during inference. In contrast to traditional RAG setups, our approach utilizes a carefully constructed repository containing human-authored instructions and examples tailored to common syntax error scenarios.

The curation workflow involves classifying syntax errors based on compiler-generated error identifiers—such as those from Quartus—and then associating each error type with specific remediation instructions and illustrative code demonstrations. Manual analysis reveals that vague or uninformative compiler diagnostics frequently obstruct LLMs from resolving issues accurately. To address this, the curated content includes not only fix strategies but also human-written interpretations of compiler messages, offering essential clarification. All data, including compiler logs, faulty code segments, and corresponding expert guidance, are indexed in a retrievable format for later use.

This knowledge base is paired with LLMs through RAG, enabling the model to fetch relevant context at inference time without the need for retraining. By retrieving pertinent human feedback and error-specific examples, the model is better positioned to handle nuanced or under-specified errors. This architecture significantly boosts the model's precision in syntax correction and debugging tasks.

Figure H.26 showcases two representative error types, pairing raw compiler output with annotated expert responses. To support this system, we experimented with several retrieval mechanisms—including pattern-based matching, approximate string search, and vector-based similarity. Due to the structured and categorical nature of compiler error codes, tag-based exact matching proved most effective. The final database includes 75 entries across 18 error classes: 30 entries for 7 categories aligned with iverilog and 45 entries spanning 11 categories tailored for the more advanced Quartus compiler.

Error Categories in RTL Design

Before discussing the approach, it's important to understand the types of errors commonly encountered in RTL design:

1. Syntax Errors: Basic grammatical errors in the HDL code that prevent successful compilation

2. Semantic Errors: Errors that are syntactically correct but violate language

| Error # | Occurance | Compiler logs / Humna feedback |
|---|---|---|
| 10161 | 297 | **Logs** object "clk" is not declared. Verify the object name is correct. If the name is correct, declare the object. |
| | | **Human** Check if clk variable exists. |
| 10161 | 43 | **Logs** index cannot fall outside the declared range for vector |
| | | **Human** Check if clk variable exists. |
| 12153 | 32 | **Logs** Can't elaborate top-level user hierarchy |
| | | **Human** Check your design to make sure every sub-module is instantiated correctly. |
| 10028 | 20 | **Logs** Can't resolve multiple constant drivers for net. |
| | | **Human** Combined output logic into a single always block. |
| 100289 | 20 | **Logs** Constant driver at Line. |
| | | **Human** Check if driver by two always blocks which is illegal for synthesis. |
| 10170 | 15 | **Logs** Verilog HDL syntax error at (46) near text: "endmodule"; expecting a description. |
| | | **Human** Check `endmodule` statement at the end. |
| 10759 | 11 | **Logs** Object declared in a list of port declarations cannot be redeclared within the module body |
| | | **Human** Check if multiple declaration on one variable. |

Figure 5.2: Frequency distribution of error categories in the RTL benchmark dataset, highlighting the relative prevalence of different error types encountered in practical hardware designs

rules (e.g., type mismatches)

3. Timing Errors: Issues related to clock domain crossings, setup/hold violations, or improper sequential logic

4. Functional Errors: Logical flaws that cause incorrect behavior, often detected through simulation failures

5. Resource Errors: Inefficient use of hardware resources that lead to excessive area or power consumption

Each error category requires different debugging approaches. RTLFixer addresses this challenge through a multi-stage debugging process.

Table 5.2: Comprehensive overview of verification tools integrated with RTLFixer and their specific data extraction capabilities

| Tool Type | Examples | Information Extracted |
| --- | --- | --- |
| Compilers | Verilator, VCS | Syntax errors, type mismatches, undefined modules |
| Static Analyzers | SpyGlass, Questa | Structural issues, coding guideline violations, potential race conditions |
| Simulators | ModelSim, VCS | Runtime errors, test failures, signal activity |
| Formal Verifiers | JasperGold | Property violations, counterexamples, coverage metrics |

Verification Tool Integration

A key innovation in RTLFixer is its integration with industry-standard hardware verification tools:

- Compilation tools (Verilator, Synopsys VCS) for syntax checking

- Static analyzers (SpyGlass, Questa) for design rule verification

- Simulation environments (ModelSim, VCS) for functional validation

- Formal verification tools (JasperGold) for property checking

Table 5.2 summarizes the verification tools integrated with RTLFixer and the information extracted from each.

The system parses outputs from these tools and converts error messages into actionable information that guides the language model's debugging process. This integration allows RTLFixer to leverage industry-standard verification techniques while benefiting from the reasoning capabilities of large language models.

Prompt Engineering for Hardware Semantics

The effectiveness of RTLFixer depends significantly on specialized prompt engineering tailored for hardware design tasks. The research developed prompt templates that incorporate hardware domain expertise and guide the model's reasoning process.

Table 5.3 shows examples of these specialized prompts for different debugging scenarios.

Table 5.3: Specialized prompt templates designed for distinct RTL debugging scenarios, demonstrating context-specific error resolution approaches

| Scenario | Prompt Template |
| --- | --- |
| Syntax Error | "The following Verilog code has a syntax error according to [tool]. Error message: '[error message]'. Please identify and fix the error while maintaining the original functionality:" |
| Timing Issue | "The following Verilog module has a timing issue related to [issue type]. Analyze the design for [specific timing concern] and propose a fix that ensures proper timing while maintaining the original functionality:" |
| Incorrect Output | "The following Verilog module is producing incorrect output values. Expected behavior: [expected]. Actual behavior: [actual]. Test vector that failed: [test vector]. Analyze the design and fix the logical error while maintaining the original interface:" |

The prompts are designed to:

- Provide the necessary context for understanding the hardware design

- Structure the debugging process in a systematic way

- Guide the model to reason about hardware-specific concepts

- Contextualize error messages within the design hierarchy

### 5.2.4 Experiment

This section begins by outlining the evaluation metrics in Section 5.2.4, followed by the main experimental results in Section 5.2.4, which highlight performance gains and the contributions of ReAct and RAG. Lastly, Section 5.2.5 presents ablation studies examining the effects of feedback quality and language model selection.

Experiment Setup

All experiments utilized GPT-3.5 as the LLM via OpenAI APIs [97], with the exception of the ablation experiment exploring different LLMs. Specifically, we employed gpt-3.5-turbo-16k-0613. Note that we also utilize the OpenAI function calling feature to restrict the LLM response into certain format. The default compiler will be Quartus. Each LLM-generated Verilog code underwent processing by a lightweight rule-based syntax fixer, effectively eliminating simple errors such as misplaced timescale derivatives. This is for avoiding simple errors and facilitate the experiment without wasting budgets for LLM sampling. A consistent sampling temperature of 0.4 was maintained across all experiments. Note that LLMs with identical prompts and the same temperature settings can still yield different outputs. For ReAct prompting, the LLM operated within a constraint of 10 maximum iterations of Thought-Action-Observation, with Actions potentially involving compiler interactions. Syntax errors were deemed resolved when any generated code passed successfully. Each experiment was repeated 10 times with results averaged to minimize test variance. For our ReAct prompting, the number of feedback iterations is capped at 10.

Evaluation Metric

Compile Fix Rate: To evaluate the effectiveness of the method in resolving syntax or compile-time errors, we compute the expected fix rate across multiple problems. For each problem, let $c$ denote the number of compilable outputs among $n = 10$ generated samples. The expected fix rate is then defined as:

$$\text{Fix Rate} = \mathbb{E}_{\text{problems}} \left[ \frac{c}{n} \right] \tag{5.1}$$

Functional Correctness: To assess semantic correctness, we follow established simulation-based evaluation protocols using the pass@k metric [20]. A task is deemed solved if at least one of the $k$ generated samples passes all test cases. We adopt the following unbiased estimator for pass@k, using $n = 20$ samples per problem to ensure reliable statistical significance:

$$\text{pass@k} = \mathbb{E}_{problems} \left[ 1 - \frac{\binom{n-c}{k}}{\binom{n}{k}} \right] \tag{5.2}$$

Main Results

Table 5.4 highlights the effectiveness of both ReAct and RAG in enhancing syntax correction performance, with each contributing notable improvements. The One-shot baseline involves a single interaction, utilizing either minimal feedback or compiler-generated diagnostics. Complementary results in Table 5.5 report signif-icant gains in pass@{1,5} on the VerilogEval benchmark following syntax repair, with corresponding visualizations shown in Figure 5.3.

| Method | RAG | Simple | iverilog | Quartus | GPT-4 |
|---|---|---|---|---|---|
| One-shot | No | 0.414 | 0.536 | 0.587 | 0.91 |
|  | Yes | – | 0.800 | 0.899 | 0.98 |
| ReAct | No | 0.671 | 0.731 | 0.799 | 0.92 |
|  | Yes | – | 0.820 | 0.985 | 0.99 |

Table 5.4: Syntax correction success rates under various prompting paradigms and compiler feedback sources, evaluated on the VerilogEval-syntax subset. Results highlight the impact of iterative prompting (ReAct) and retrieval augmentation (RAG) across simulation tools and model configurations.

An analysis of the VerilogEval-Human subset reveals that syntax errors constitute the majority (55%) of failures in Verilog code generated by GPT-3.5, surpassing simulation-related issues. Applying our correction pipeline (depicted in the inner ring of the figure) increases the pass@1 rate from 26.7% to 36.8%. Furthermore, Table 5.6 demonstrates the approach's strong generalizability, yielding consistent improvements across alternative benchmarks. Additional insights are discussed in the following sections.

| Dataset VerilogEval | Set | pass@1 | | pass@5 | |
|---|---|---|---|---|---|
|  |  | original | fixed | original | fixed |
| Human | All | 0.267 | 0.368 | 0.458 | 0.506 |
|  | easy | 0.521 | 0.666 | 0.808 | 0.847 |
|  | hard | 0.053 | 0.120 | 0.164 | 0.221 |
| Machine | All | 0.467 | 0.799 | 0.691 | 0.891 |
|  | easy | 0.568 | 0.833 | 0.782 | 0.892 |
|  | hard | 0.367 | 0.771 | 0.601 | 0.890 |

Table 5.5: Simulation pass rate improvements (pass@k) on the VerilogEval dataset before and after syntax error correction, demonstrating enhanced functional correctness

Effectiveness of ReAct: The ReAct framework significantly outperforms traditional one-shot generation methods. By leveraging iterative code refinement, guided reasoning, and structured planning, it achieves superior outcomes across all settings.

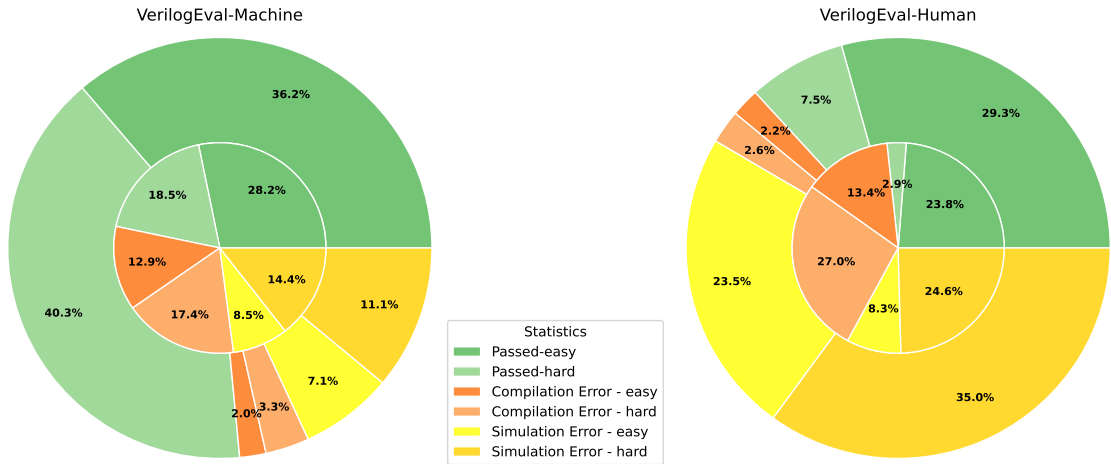

Figure 5.3: Comparative visualization of VerilogEval pass@1 results before (inner circle) and after (outer circle) syntax error correction with RTLFixer, demonstrating substantial improvement in successful compilation

Remarkably, even without access to explicit compiler diagnostics (Simple feedback), the model benefits from intermediate reasoning steps—similar to chain-of-thought prompting—improving syntax success from 41.4% to 67.1%. Compared to one-shot baselines without RAG, ReAct improves syntax success rates by 25.7%, 26.4%, and 31.2% when combined with Simple, Icarus Verilog, and Quartus feedback, respectively. These consistent gains highlight the robustness of ReAct, independent of the compiler or retrieval method used.

Effectiveness of RAG: Incorporating RAG with curated domain-specific knowledge leads to substantial gains in correction accuracy and overall solution stability. Results using the Quartus compiler (Table 5.4) demonstrate that adding RAG improves syntax fix rates by 31.2% for one-shot generation (from 58.7% to 89.9%) and by 18.6% for ReAct (from 79.9% to 98.5%). These improvements hold across different compiler feedback sources and remain effective regardless of the underlying LLM, including GPT-4.

Simulation Correctness Improvement: Previous studies [71, 79] have primarily relied

| Model | Syntax Pass Rate | pass@1 Score |
|---|---|---|
| GPT-3.5 | 73% | 11% |
| GPT-3.5 + RTLFixer | 93% | 16% |

Table 5.6: Evaluation results on the RTLLM benchmark using the Quartus compiler. Integrating ReAct and RAG notably improves syntax correctness and downstream simulation accuracy, as reflected in both syntax pass rate and pass@1 metrics.

on the pass@k metric to evaluate the performance of LLM-generated Verilog code. However, this metric often fails to account for the detrimental impact of syntax errors on functional correctness. In this study, we conduct a more rigorous evaluation using the VerilogEval benchmark, specifically targeting the remediation of syntax errors in generated code.

Table 5.5 reports the results, showing substantial improvements in pass@1 scores after syntax correction—32.3% for the Machine category and 10.1% for Human. To enable more granular analysis, the VerilogEval dataset is partitioned into two subsets: an easy subset of 71 problems and a hard subset of 85 problems, based on a pass rate threshold of 0.1 in the Human category.

For simpler tasks in Human and foundational descriptions in Machine, syntax correction yields significant improvements, with pass@1 rates approaching 80%. Notably, the improvement for easy Human problems is 14.5%, compared to 6.7% for hard problems, indicating that syntax repair has a more pronounced effect on tasks with lower reasoning complexity. These findings highlight the persistent difficulty LLMs face when solving complex Verilog problems requiring deeper semantic understanding. Further analysis and solutions for simulation-level errors are discussed in Section 5.2.6.

Generalizability: The combined ReAct and RAG framework demonstrates strong generalization across different code generation benchmarks. To assess potential overfitting in retrieval database construction, we evaluate the framework on the RTLLM benchmark [79] without incorporating any handcrafted guidance in the retrieval corpus.

As shown in Table 5.6, the proposed approach raises the syntax success rate from 73% to 93%, validating its robustness and transferability. [1]

### 5.2.5 Ablation Studies

This research encompasses two comprehensive ablation studies designed to evaluate the influence of feedback quality and language model selection on syntax error correction efficacy.

#### 5.2.5.1 Impact of Feedback Quality

The relationship between feedback quality and error correction performance was methodically investigated through varied feedback mechanisms described below.

Simple: A minimalist instruction approach utilizing only the basic prompt "Correct the syntax error in the code." without specific error identification or rectification guidance.

Icarus Verilog (iverilog) [154]: An open-source Verilog simulation environment that occasionally exhibits limitations in diagnostic capabilities, sometimes producing ambiguous messages such as "I give up." The error logs frequently lack sufficient clarity,

---

[1]Note: The reported syntax success rate differs slightly from the original RTLLM publication due to the use of their released Verilog code samples for each problem instance.

presenting interpretation challenges even for experienced practitioners.

Quartus[2]: A sophisticated commercial FPGA compiler that consistently provides well-structured and comprehensible diagnostic messages, effectively identifying error conditions while frequently offering specific correction recommendations and validation insights, resulting in superior user experience and informational content.

The feedback quality progression from Simple through iverilog to Quartus represents a distinct quality hierarchy, with comparative compiler characteristics illustrated in Figure H.28. Results presented in Table 5.4 conclusively demonstrate that compiler-generated diagnostics outperform Simple feedback, with higher quality compiler outputs directly correlating to improved LLM debugging performance. The performance differential between iverilog and Quartus becomes particularly pronounced when implementing ReAct with RAG, suggesting that superior compiler messaging enhances the model's capacity to effectively leverage retrieved expert knowledge.

#### 5.2.5.2 Impact of Different LLMs

Table 5.4 presents performance metrics obtained when deploying GPT-4 with Quartus compiler integration. A substantial enhancement in syntax error resolution efficacy is observed with GPT-4 compared to GPT-3.5, particularly evident in One-shot prompting with RAG implementation, where success rates increased dramatically from 89.9% to 98%. Comparative analysis between GPT-4's performance using One-shot versus ReAct approaches reveals modest 1% improvements, indi-cating that GPT-4 inherently possesses robust error-correction capabilities without necessarily requiring additional reasoning frameworks, action planning mechanisms, or iterative refinement processes.

---

[2]https:// www.intel.com/ content/ www/ us/ en/ products/ details/ fpga/ development-tools/ quartus-prime/resource.html

Nevertheless, it is crucial to emphasize that the ReAct and RAG augmentation methodology significantly reduces the performance differential between less sophisticated and more advanced language models. This advantage becomes particularly valuable when implementing smaller open-source models [71, 127] that may demonstrate limited effectiveness compared to GPT-4 when addressing complex programming challenges. The implementation leverages the OpenAI function call feature for formatting LLM responses, which currently limits its applicability to other open-source models. Expanding this evaluation to include more models is a prospective avenue for future research.

Table 5.7: Performance comparison of syntax error resolution capabilities across different language models, highlighting the advantages of more advanced LLMs

| Prompt | Feedback | GPT-3.5 | GPT-4 |
|---|---|---|---|
| One Shot | w/o RAG | 0.57 | 0.91 |
|  | RAG | 0.899 | 0.97 |

### 5.2.6 Analysis and Discussion

This section presents a series of analytical insights derived from key experimental findings, focusing on failure case analysis, the role of iterative code refinement, and challenges related to debugging simulation logic errors.

Failures Attributable to LLM Limitations: Most unresolved syntax errors—despite the use of ReAct and RAG techniques—stem from inherent limitations of the language model. Figure H.29 shows a representative failure case involving complex arithmetic index computations required to fix an index out-of-range error. Additional failures were observed in cases where the model exhibited high confidence in generating syntactically invalid code, often due to patterns mistakenly learned from syntactically similar constructs in other languages such as C/C++.

Iterative Code Refinement: Figure 5.4 presents a quantitative analysis of ReAct iteration requirements for syntax error resolution. Approximately 90% of issues are successfully addressed in a single revision cycle. The remaining cases necessitate multiple code revisions, primarily due to cascading errors that emerge only after initial corrections have been implemented.

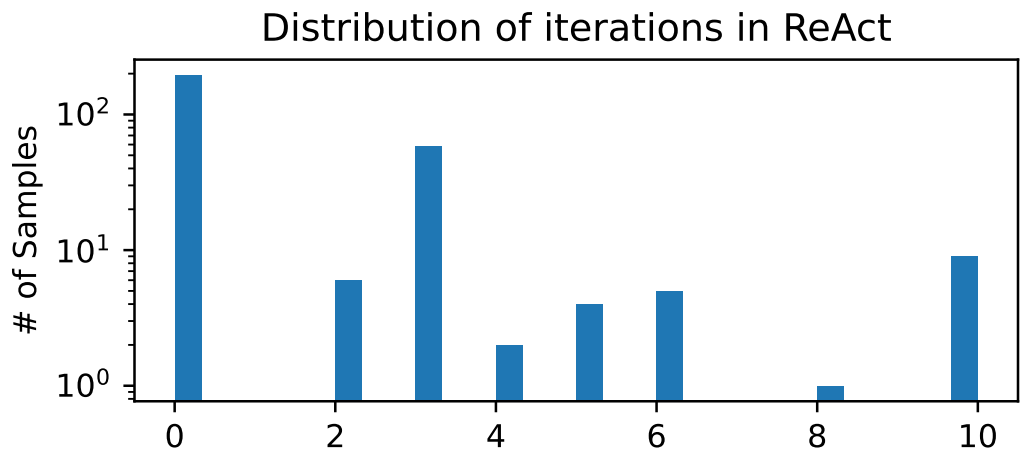

Figure 5.4: Frequency distribution of iterations required by ReAct to successfully resolve syntax errors, demonstrating the efficiency of the approach with most errors fixed in a single revision

Challenges in Debugging Simulation Errors: While this framework is readily adaptable for simulation error debugging applications, preliminary investigations revealed limited improvements beyond syntax error remediation. Despite comprehensive efforts to provide simulation error diagnostics to LLM agents—including detailed output error summaries and text-formatted waveform comparisons between erroneous and expected outputs—models demonstrated constrained capabilities in simulation feedback interpretation. They exhibited proficiency primarily in addressing logic im-plementation errors in elementary problems but encountered significant challenges with complex questions, particularly those involving high-level design functionality descriptions and advanced reasoning requirements. Resolving implementation errors in such sophisticated contexts remains an exciting frontier for future research, highlighting the critical need for enhanced LLM capabilities in hardware-specific reasoning and problem-solving domains.

### 5.2.7 Limitations and Future Directions

While RTLFixer demonstrates significant improvements over conventional approaches, several important limitations remain:

- Complex Microarchitectural Optimizations: The system currently struggles with sophisticated performance optimizations that require deep hardware domain knowledge

- Physical Design Constraints: Integration with place-and-route tools would enable addressing physical implementation challenges more effectively

- Specification Mining: Future work could focus on automatically extracting formal specifications from natural language descriptions

- Multi-language Support: Extending the framework beyond Verilog to robustly support VHDL and SystemVerilog

Table 5.8 summarizes these key limitations and proposed solutions.

Table 5.8: Identified limitations of the current RTLFixer system and proposed strategic solutions for future research and development

| Limitation | Proposed Solution |
| --- | --- |
| Complex Microarchitectural Optimizations | Integrate specialized hardware optimization knowledge bases and sophisticated performance models |
| Physical Design Constraints | Extend tool integration to include advanced place-and-route feedback and timing closure information |
| Specification Ambiguity | Develop robust specification mining techniques to extract formal specifications from natural language descriptions |
| Cross-language Support | Create unified intermediate representations for multiple HDLs to enable broader language compatibility |

### 5.2.8 RTLFixer Summary

RTLFixer convincingly demonstrates that the strategic integration of large language models with domain-specific verification tools can significantly enhance the accuracy and efficiency of RTL debugging processes. By implementing a comprehensive multi-stage debugging strategy, specialized prompt engineering techniques, and verification tool feedback mechanisms, the system achieves substantial improvements over standard language models across diverse error types and hardware designs. This innovative approach offers a promising direction for applying AI to hardware design automation, potentially reducing development cycles and improving overall design quality.

## 5.3 High-Quality Synthetic Data Generation

### 5.3.1 Motivation

While RTLFixer demonstrated significant improvements in debugging and fixing errors in RTL code through verification tool feedback, it primarily represents a reactive approach to hardware design. Through the experience with RTLFixer, fundamental limitations were identified that necessitated a more proactive strategy—specifically, the need to enhance the underlying language models' capabilities through training data that can ensure function correctness and maintain high-quality.

Hardware description languages (HDLs) like Verilog present unique challenges for language models. Despite the remarkable progress Large Language Models (LLMs) have made in generating software code, their performance on HDLs remains constrained. Analysis of fine-tuned LLMs on Verilog code revealed two critical issues: (1) models struggle with non-textual representations such as Karnaugh maps, state-transition diagrams, and waveforms, which are essential for hardware design; and (2) models exhibit significant variability in performance across benchmark problems and training checkpoints, often making "minor" yet consequential programming mistakes.

These challenges stem from two fundamental constraints. First, Verilog exists as a low-resource language with limited available training data compared to popular software languages like Python. Current state-of-the-art code models contain less than 0.3% hardware-related data in their training datasets [26], creating a severe data scarcity problem. This limited data availability restricts models' ability to learn diverse and complex hardware coding patterns. Second, verifying the correctness of HDL code is inherently more complex than verifying software code. While software code correctness can often be assessed using random test cases and automated unit tests [17], hardware code demands comprehensive testbenches and rigorous verification planning and methodologies. This additional complexity makes it challenging to ensure that synthetic Verilog code is functionally accurate [10, 107], posing a barrier to improving model performance.

RTLFixer, while effective at correcting errors, operates reactively on already-written code and struggles with complex microarchitectural optimizations requiring deep hardware knowledge. This limitation highlighted the need for a more fundamental solution: enhancing the quality of data used to train the models in the first place.

Large Language Models (LLMs) have achieved significant success across various natural language processing tasks and have extended their capabilities to code generation, leading to the development of specialized models targeting code generation. The effectiveness of these models is largely influenced by the size and quality of their training datasets, as highlighted by scaling laws [2, 169]. Moreover, these SOTA code models have less than 0.3% Hardware related data in their training datasets [26]. Prominent code LLMs have set new benchmarks records by utilizing extensive, synthetically generated datasets through methods like Self-Instruct [16, 147], Evol-Instruct [161], and OSS-Instruct [152]. These synthetic data generation techniques allow code LLMs to generate a wide range of complex code examples, enhancing their training and performance in real-world coding scenarios.

While most code LLMs concentrate on software programming languages, there is increasing interest in developing models for hardware description languages (HDLs), which are essential for chip design and hardware verification. Despite efforts to collect and synthesize more diverse Verilog code to enhance specialized code LLMs [22, 73, 100, 174], HDLs still face challenges akin to those encountered in low-resource languages [14]. These challenges are mainly due to the limited availability of high-quality instruction-following data and the constrained capability of existing LLMs to generate RTL code, which affects the models' performance and their ability to generalize across programming languages.

Developing high-quality synthetic Verilog code for training code large language models (LLMs) faces significant challenges due to two primary factors. Firstly, Verilog is considered a low-resource language [14], meaning there is a scarcity of available training data compared to high-resource software programming languages like Python. This limited data availability restricts the models' ability to learn diverse and complex coding patterns effectively. Secondly, verifying the correctness of hardware description language (HDL) code, such as Verilog, is inherently more complex than verifying software code. While software code correctness can often be assessed using random test cases and automated unit tests [17], hardware code requires comprehensive testbenches and rigorous verification planning and methodologies. This additional complexity makes it challenging to ensure that synthetic Verilog code is functionally accurate [10, 107], posing a barrier to improving model performance.

This dissertation begins with a thorough analysis of fine-tuned large language models (LLMs) applied to Verilog code, using synthetic data techniques from previ-ous works. The analysis reveals two key issues: (1) models have difficulty handling non-textual elements in problem statements, indicating challenges in interpreting complex or unconventional inputs; and (2) there is notable variability in the models' pass rates across different benchmark problems and training checkpoints, exposing inconsistencies in learning outcomes, often due to the models making "minor" programming mistakes.

Given the limitations identified in the analysis of relying solely on LLMs for generating synthetic data, focus was shifted to improving data curation to address these issues. Current LLMs frequently struggle with interpreting and processing non-textual representations and are insufficient in generating effective testbenches for evaluating solution quality. Therefore, instead of depending exclusively on LLMs to address data quality concerns, targeted fine-tuning data was developed to better mitigate these problems. Experimental results demonstrate that the models achieve state-of-the-art (SOTA) results on VerilogEval [71] and RTLLM v1.1 [80] benchmarks, outperforming prior works by large margins on problems with human-level description. The major contributions are as follows:

- A thorough analysis of fine-tuned LLMs on Verilog code using previously established synthetic data generation methods, uncovering challenges with non-textual elements and notable variability in performance across benchmark problems during training.

- The creation of correct-by-construction data to ensure solution correctness, incorporating Karnaugh Maps, state-transition diagrams, and waveforms, which significantly enhance the model's ability to handle non-textual representations.

- The development of an automated framework that utilizes LLMs to generate error reports from benchmark problems at various checkpoints, which are then injected into open-source code to create a fine-tuning dataset targeted at correcting the model's specific "minor" mistakes.

- A rigorous evaluation of the latest foundational and frontier code models, noting that recent advanced models like GPT-4o already reached competitive performance compared to previous efforts targeting Verilog code generation.

- Experimental results demonstrating that our fine-tuned models achieve state-of-the-art performance on Verilog coding tasks. Notably, the Starcoder2-15B [77] fine-tuned model surpasses previous SOTA results by 3.8%, 10.9%, and 6.6% for pass@1 on VerilogEval-Machine, VerilogEval-Human, and RTLLM benchmarks, respectively.

### 5.3.2 Examining Fine-tuned LLMs Using Synthetic Generated Data on Verilog Coding

In this section, a thorough analysis of fine-tuned large language models (LLMs) applied to Verilog code is presented. Previous approaches for generating synthetic data for general coding are adapted to focus on Verilog code. For this pilot study, results based on fine-tuning StarCoder2-15B [77] are presented. Details on experimental settings are the same as in 5.3.7. Model performance in Verilog code completion is assessed, identifying two main issues. First, the models demonstrate notably poor performance when dealing with non-textual elements in the problem statements. Second, the variability in the models' pass rates across different benchmark problems and training checkpoints suggests inconsistencies in learning outcomes and model variability.

Table 5.9: Data quantity SDG.

| Method | Quantity |
|---|---|
| Self-Instruct | 24.7k |
| OSS-Instruct | 28.4k |
| Docu-Instruct | 12.0k |
| Non-textual | 15.0k |
| SDG Total | 80.1k |

Synthetic Data Generation for Verilog Coding

Previous methods for synthetic data generation are extended by applying Self-Instruct [147] and OSS-Instruct [152] with custom prompt templates tailored for Verilog coding. To enhance data coverage and diversity, these techniques are supplemented with additional context from Wikipedia and textbooks. Models are also prompted to generate problem descriptions that include non-textual representations.

Nemotron-4-340b-instruct [96] was selected for its open license that allows commercial use. The process includes deduplication and a decontamination procedure akin to that outlined by [64]. Additionally, syntax checks are conducted to eliminate coding problems containing docstrings or solutions from Verilog benchmarks. To ensure further data quality, code solutions that fail these syntax checks are discarded and self-verification [153] is applied to remove entries where the LLM identifies errors in the solution. Table 5.9 shows the quantity of synthetic data generation (denoted as SDG) after deduplication and filtering, yielding a total of 80.1k fine-tuning examples.

Self-Instruct   The approach outlined in [147] is followed to generate synthetic Verilog coding problems. Initially, random generation from the LLM is performed and 50 questions that request Verilog coding problems without any in-context examples are curated. From these, 1 to 5 seed questions are randomly chosen to use as in-context examples.

OSS-Instruct   The process begins by processing pretraining code data to extract seed code from The Stack v2 [77], focusing on Verilog and SystemVerilog. Following the approach in [71], this data is post-processed by selecting self-contained Verilog code that passes syntax checks using Pyverilog [124]. With the refined seed code data, large language models (LLMs) are then prompted to use this code as inspiration for generating Verilog coding problems similar to [152].

Docu-Instruct   Drawing inspiration from [96] and [123], document sources from Wikipedia and textbooks are utilized for instruction generation. The process begins by filtering Wikipedia entries, prompting the LLM to classify whether the content pertains to hardware design or Verilog coding concepts. Additionally, approximately 100 relevant textbooks are manually selected. These textbooks are then segmented into chunks of paragraphs or sentences, ensuring each chunk contains fewer than 2k tokens.

Non-textual Representations   VerilogEval-Human [71] includes benchmark problems involving non-textual representations. For example, Boolean logic tables and Karnaugh maps are presented in tabular formats, state-transition diagrams for finite state machines are depicted as edge lists and sequential waveforms are described in

Table 5.10: pass@1 results on VerilogEval sampled with temperature of 0.8.

| Model | Machine | Human | NonText |
|---|---|---|---|
| GPT-4o | 63.7 | 55.4 | 27.0 |
| Starcoder2 | 57.7 | 29.1 | 10.3 |
| Starcoder2-SDG | 73.7 | 47.4 | 22.2 |

tables with signal outputs recorded at various time steps. To incorporate such representations, LLMs are encouraged to generate problems from open-source code, with instructions to utilize these tabular data structures.

### 5.3.3 Challenges with Non-Textual Representations

It is observed that models underperform on benchmark problems involving non-textual input formats, such as Karnaugh Maps, state-transition diagrams, and waveforms. Table 5.10 shows the pass@1 results for the VerilogEval [71]. Additionally, a subset of 45 questions within VerilogEval-Human that include non-textual representations has been identified, termed VerilogEval-NonText. It appears that models like GPT-4o and Starcoder2 struggle with these non-textual formats, likely due to insufficient representation of such data during both pretraining and fine-tuning. Despite efforts to generate such questions during synthetic data creation, the fine-tuned models still lag in these areas. This outcome is not entirely surprising, given that the LLMs used were also ineffective at generating problems with these representations, complicating the validation of fine-tuning data. These results suggest that merely including non-textual data is insufficient; ensuring the quality and correctness of the data, particularly that the code solutions accurately align with these representations, is crucial.

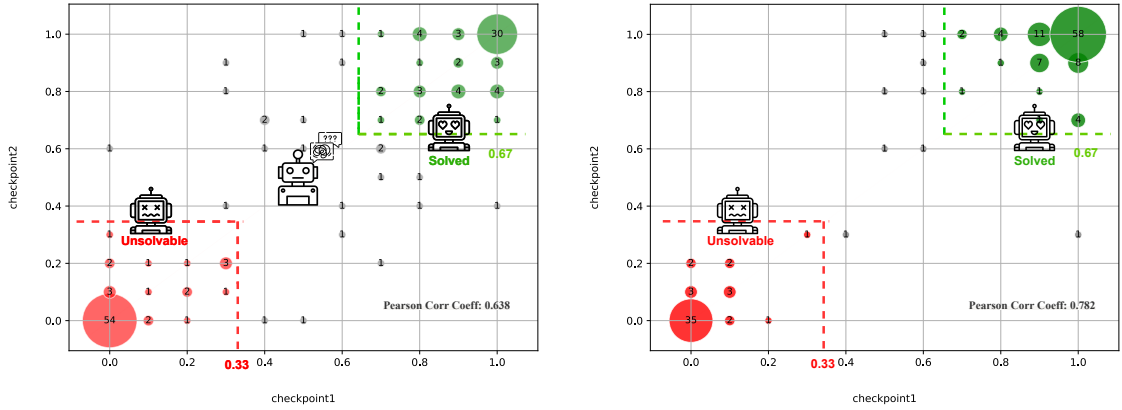

(a) Starcoder2-15B on SDG.  (b) Starcoder2-15B on SDG-CC-Repair.

Figure 5.5: The proposed methods reduce pass rate variability during training: SDG (left) shows high volatility with significant degradation on many problems, while SDG-CC-Repair (right) stabilizes learning outcomes on solvable problems (details in Appendix I.1.10).

### 5.3.4 Variability on Pass Rates During Training

During training, significant variability in the model's pass rate on specific benchmark problems across different checkpoints was observed. This variance is different from training instability [155] as a stable decrease in the training loss is observed. This variability persists even in the later stages of training, despite using a low learning rate. This variability is illustrated in Figure 5.5a. The scatter plot tracks the pass rate for each problem in VerilogEval-Human, with each point representing the pass rate for the same problem across two checkpoints. The size of each point indicates the number of problems with the same pass rates for the two model checkpoints. The region is further categorized into areas where the checkpoints agree on problem difficulty and areas where they do not.

Alarmingly, nearly 15% of the problems show significant discrepancies between these two checkpoints, with an equal number of problems demonstrating improvement and degradation. Detailed analysis of the sampled code completions for such problems when pass rate degrades suggests that the model is generally on the right track but makes "minor" errors that are small, detailed, and seemingly trivial. While it is possible that LLMs experience catastrophic forgetting during fine-tuning [? ], this is not anticipated to be a major factor due to the low learning rate and the small number of gradient updates (64 steps with 16k data samples). Instead, the primary issue appears to be the inability to ensure the quality of the data, particularly in verifying whether the sampled code solutions correctly solve the code problems.

Table 5.11: Data quantity CC.

| Method | Quantity |
|:---:|:---:|
| KMap | 12.5k |
| FSM | 8.0k |
| Waveforms | 8.0k |
| CC Total | 28.5k |

## 5.3.5 Improving Verilog Coding with Correct-by-Construction Non-Textual Representations and Targeted Code Repair

Based on the detailed analysis of the limitations of relying solely on LLMs for generating synthetic data, data curation efforts were focused to address these shortcomings. The goal is to enhance data quality and ensure the correctness of solutions for the generated problems. Current LLMs often lack the capability to understand and process non-textual representations effectively and are unable to generate satisfactory testbenches for assessing solution quality. Consequently, rather than depending entirely on LLMs to resolve data quality issues, targeted fine-tuning data was created to mitigate these problems.

Ensuring Quality Through Correct-by-Construction

Verilog code problems and solutions that are correct-by-construction were generated, with a focus on creating problems and solutions for non-textual representations. Table 5.11 shows the quantity of correct-by-construction generation data (referred to as CC). To prevent data contamination, entries that duplicate the data representations of benchmark problems were excluded.

Karnaugh Maps and Truth Tables (KMap)  The process begins by sampling random configurations, which include selecting the number of variables and their names. After determining the number of variables, valid minterms and don't-cares are randomly chosen. For $n$ variables, there are $2^n$ possible states, and each state can be assigned one of three values (0, 1, or x), leading to $3^{2^n}$ possible combinations of minterms and don't-cares. From these minterms, the sum-of-products (SOP) form is derived to represent the Boolean logic. Truth Tables and Karnaugh Maps are then created based on the chosen minterms and don't-cares. In the KMap, Gray encoding is used as default for the row and column sequences to ensure that only a single bit changes between adjacent cells. Additionally, modifications are applied by transposing the map and randomly swapping adjacent rows or columns. Samples are randomly drawn from $n = \{3, 4\}$ variables.

State Transition Graphs and Tables (FSM)  Problems for finite-state machines (FSMs) with state-transition representations are constructed with a similar approach to KMaps. The process begins by sampling random configurations, including the number of states (e.g., 4, 6, or 10) and the bit width of the input (e.g., 1 or 2).

The transition graph is then created, ensuring that it is both meaningful and legally defined. State-transition graphs for both Moore and Mealy state machines are generated. From these graphs, edge-list and transition table representations are produced. Finally, Verilog code is constructed to implement the logic for state transitions and output assignments.

---
**Algorithm 1** Generate transition graph for Moore FSM.
---
    Input: Number of states $n$, bit width of input $w$

    Output: FSM graph with transitions and states

    Initialize the number of states $n$ and bit width of input $w$

    Randomly generate a tree with $n$ nodes

    Define the root of the tree as the reset state

    for each node in the tree do

        Assign a unique state to the node

        Assign an output to the node

    end for

    for each node in the tree do

        Add additional transition edges to form a graph

        Ensure that each node has an out-degree of $2^w$

    end for
---

Algorithm 1 outlines the process for generating a Moore FSM with random transitions. State reachability is ensured by first constructing a tree. Legality for state transition is ensured by ensuring each node has an out-degree of $2^w$ with the input bit width of $w$. The result is an FSM where transitions between states are randomly assigned but conform to the specified input bit width. The algorithm can be easily modified for a Mealy FSM by assigning the output to the edges rather than nodes.

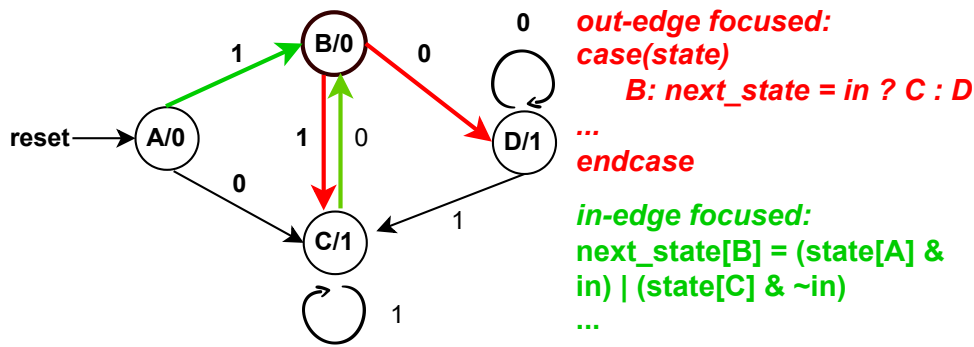

Figure 5.6: State transition logic.

Figure 5.6 illustrates the approach for generating state transition logic in Verilog from a state-transition graph. The method predominantly employs an out-edge focused strategy for state transitions. Additionally, in-edge focused transition logic is incorporated to address specific challenges encountered in benchmark problems. These benchmarks often involve states represented using one-hot encoding and require rigorous testing of non-default states.

Waveforms   Correct-by-construction code solutions for both KMaps and FSMs are utilized. Because these codes are generated using similar templates, designing corresponding testbenches is straightforward. The generated code is simulated to produce waveform Value Change Dump (VCD) files. These VCD files are then parsed and converted into waveform representations. The approach covers KMaps as combinational circuits and FSMs as sequential circuit waveforms.

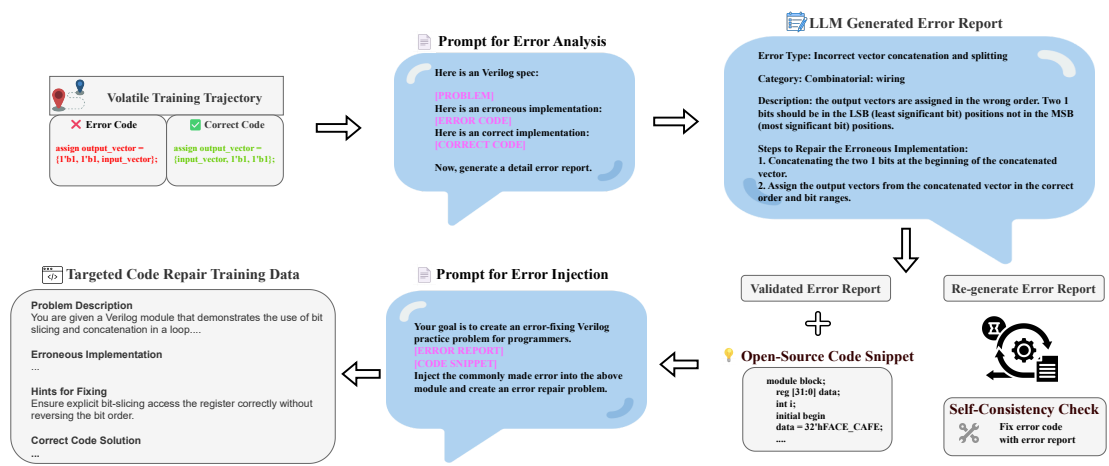

Figure 5.7: Overview of the approach for generating targeted code repair data: (1) prompting the LLM to generate detailed error reports from correct and erroneous code, (2) validating error report quality by ensuring the LLM can debug the errors based on the report, and (3) leveraging the LLM to inject similar errors into open-source code, creating a diverse training dataset.

### 5.3.6 Mitigating "Minor" Errors with targeted code repair

Analysis revealed that the models were generally on the right track to correct solutions but were making minor errors—small, detailed, and seemingly trivial. Unlike complex, unsolvable problems, these minor errors could be easily corrected by language models. This insight led to the development of a new strategy centered on targeted error code repair. The approach includes creating detailed error reports on benchmark problems, re-creating these errors on correct open-source code, and conducting rigorous validation to ensure quality. Nemotron-4-340b-instruct was used as the LLM to construct the targeted Repair data. A total of 847 error reports were generated across the three benchmarks, producing 2,736 data samples. After filtering, this resulted in a final set of 1,406 targeted code repair data points.

Error Report Construction  To systematically address the issue, a comprehensive Error Report for benchmark problems was first created using LLMs, targeting those with significant pass rate fluctuations across training checkpoints for models on SDG data. The LLM was prompted to examine the nature of the mistakes by comparing correct and erroneous code completions for each problem, categorizing the errors into common error types (details in Appendix I.1.9). This detailed report not only categorizes the errors but also highlights areas where the model consistently underperforms.

Targeted Code Repair Dataset  Building on the error report, a targeted code repair dataset was further developed to address these common errors. This dataset is constructed using two main sources: the errors identified in the Error Report and correct code snippets gathered from open-source repositories. The identified errors were introduced into correct code snippets to create repair problems, which include a problem description, erroneous code implementation, and hints about the nature of the error and how to fix it. This targeted strategy enables the model to learn how to avoid common errors and generate improved code completions, thereby enhancing model accuracy.

Quality Assurance with LLM Validation  To ensure the reliability of the error report and the code repair dataset, a two-phase validation process with LLMs was implemented. In the first phase, a self-consistency check of the Error Report was conducted by having the language model attempt to fix error code based on the report's hints. This step verifies the accuracy of the report by confirming that the model can resolve the errors using the provided guidance, whereas directly prompting the LLM without detailed error reports could resolve only 13% of the errors. In the second phase, during the generation of the code repair dataset, self-verification was applied, including deduplication, syntax filtering, and benchmark decontamination. These measures ensure the dataset's quality and uniqueness, preventing overlap with evaluation benchmarks.

### 5.3.7 Experiment

Implementation Details

Training Data  The fine-tuning training data comprises 80.1k LLM synthetic generated data using various prompting methods as described in Section 5.3.2, 28.5k data samples generated correct-by-construction aimed at non-textual representations detailed in Section 5.3.5, and 1.4k carefully filtered data for targeted code repair as outlined in Section 5.3.6. Each data set is referred to as SDG, CC, and Repair, respectively.

Pretrained Models  Following prior work, CodeLlama-7b-Instruct [111] and Deepseek-Coder-6.7b-Instruct [40] were used as the base models, with data formatted according to their default chat prompt templates. Additionally, the Starcoder2-15B [77] model was explored in the experiments.

Model Training  Models were trained using 32 NVIDIA A100-80GB GPUs with PyTorch's Distributed Data Parallel (DDP) framework. Learning rates were set to 5e-5 for CodeLlama and DeepSeek-Coder, and 1e-5 for Starcoder2. Training utilized the Adam optimizer [55] with full parameter updates, a batch size of 256, and sequences truncated at 4096 tokens. Models were fine-tuned for one epoch with cross-entropy loss applied only to response tokens.

Table 5.12: Comparison of the proposed models with various baseline models on VerilogEval [71]. Results from [174] are updated with the latest foundational and frontier code models. The best results are highlighted in bold.

| Type | Model | Size | VerilogEval [71] | | | | | |
|---|---|---|---|---|---|---|---|---|
| | | | Machine (%) | | | Human (%) | | |
| | | | pass@1 | pass@5 | pass@10 | pass@1 | pass@5 | pass@10 |
| Foundational Models | Llama-3.1 | 8B | 48.7 | 67.3 | 74.1 | 26.9 | 37.8 | 44.2 |
| | Llama-3.1 | 405B | 67.3 | 75.1 | 76.9 | 53.8 | 61.0 | 62.8 |
| | Nemotron-4 | 340B | 53.0 | 60.3 | 62.2 | 43.1 | 48.3 | 50.0 |
| | GPT-3.5-turbo | - | 58.0 | 74.0 | 77.6 | 31.2 | 44.1 | 47.4 |
| | GPT-4o | - | 65.9 | 71.4 | 72.7 | 57.1 | 63.9 | 66.7 |
| Code Models | CodeLlama | 7B | 43.1 | 47.1 | 47.7 | 18.2 | 22.7 | 24.3 |
| | CodeQwen | 7B | 46.5 | 54.9 | 56.4 | 22.5 | 26.1 | 28.0 |
| | Starcoder2 | 15B | 68.7 | 82.3 | 88.5 | 37.7 | 50.6 | 57.2 |
| | DeepSeek-Coder | 6.7B | 52.2 | 55.4 | 56.8 | 30.2 | 33.9 | 34.9 |
| | DeepSeek-Coder-V2 | 16B | 67.4 | 78.3 | 81.8 | 46.9 | 55.9 | 58.9 |
| | DeepSeek-Coder-V2 | 236B | 68.2 | 74.1 | 76.2 | 56.4 | 62.2 | 66.0 |
| RTLCoder [73] | Mistral | 7B | 62.5 | 72.2 | 76.6 | 36.7 | 45.5 | 49.2 |
| | DeepSeek-Coder | 7B | 61.2 | 76.5 | 81.8 | 41.6 | 50.1 | 53.4 |
| BetterV [100] | CodeLlama | 7B | 64.2 | 75.4 | 79.1 | 40.9 | 50.0 | 53.3 |
| | DeepSeek-Coder | 6.7B | 67.8 | 79.1 | 84.0 | 45.9 | 53.3 | 57.6 |
| | CodeQwen | 7B | 68.1 | 79.4 | 84.5 | 46.1 | 53.7 | 58.2 |
| CodeV [174] | CodeLlama | 7B | 78.1 | 86.0 | 88.5 | 45.2 | 59.5 | 63.8 |
| | DeepSeek-Coder | 6.7B | 77.9 | 88.6 | 90.7 | 52.7 | 62.5 | 67.3 |
| | CodeQwen | 7B | 77.6 | 88.2 | 90.7 | 53.2 | 65.1 | 68.5 |
| OriGen [22] | DeepSeek-Coder | 6.7B | 74.1 | 82.4 | 85.7 | 54.4 | 60.1 | 64.2 |
| Proposed SDG-CC-Repair | CodeLlama | 7B | 78.1 | 85.5 | 87.8 | 63.1 | 67.8 | 69.7 |
| | DeepSeek-Coder | 6.7B | 77.8 | 85.5 | 88.1 | 65.4 | 70.0 | 72.1 |
| | Starcoder2 | 15B | 81.9 | 86.9 | 88.1 | 68.0 | 72.4 | 74.6 |

Model Inference  For inference, we utilized vLLM [57] configured with bf16 precision, tensor parallelism (size 8), and a 4096-token limit. Each problem was sampled 20 times, and we report the best results from temperatures 0.2 and 0.8, following prior work [73, 174].

Evaluation Metric and Benchmark

Evaluation Metric Following prior work [20, 69], for each experiment the unbiased pass@k metric is used to measure the Verilog generation accuracy. The pass@k metric estimates the proportion of problems that can be solved at least once in k attempts:

$$pass@k := \mathbb{E}_{\text{Problems}} \left[ 1 - \frac{\binom{n-c}{k}}{\binom{n}{k}} \right], \qquad (5.3)$$

Table 5.13: Evaluations on RTLLM v1.1 [80] using unbiased pass@k metrics. The best results are highlighted in bold. All models are re-evaluated (see Appendix I.1 for details).

| Type | Model | Size | RTLLM v1.1 [80] | | | | | |
| --- | --- | --- | --- | --- | --- | --- | --- | --- |
| | | | Syntax (%) | | | Func. (%) | | |
| | | | pass@1 | pass@5 | pass@10 | pass@1 | pass@5 | pass@10 |
| Foundational Models | Llama-3.1 | 8B | 40.7 | 60.6 | 65.5 | 19.3 | 34.7 | 37.9 |
| | Llama-3.1 | 405B | 56.5 | 64.4 | 72.4 | 38.9 | 45.8 | 51.7 |
| | Nemotron-4 | 340B | 41.7 | 47.2 | 48.3 | 18.9 | 20.7 | 20.7 |
| | GPT-3.5-turbo | - | 50.3 | 61.2 | 65.5 | 28.3 | 36.9 | 41.4 |
| | GPT-4o | - | 50.3 | 59.9 | 62.1 | 33.8 | 44.4 | 48.3 |
| Code Models | CodeLlama | 7B | 46.6 | 62.6 | 68.9 | 17.9 | 29.9 | 34.5 |
| | CodeQwen | 7B | 45.8 | 65.8 | 72.4 | 24.1 | 34.0 | 37.9 |
| | Starcoder2 | 15B | 38.3 | 81.0 | 94.7 | 15.5 | 37.6 | 45.7 |
| | DeepSeek-Coder | 6.7B | 51.4 | 64.4 | 68.9 | 23.1 | 29.3 | 34.5 |
| | DeepSeek-Coder-V2 | 16B | 51.4 | 57.8 | 58.6 | 33.1 | 37.1 | 37.9 |
| | DeepSeek-Coder-V2 | 236B | 63.4 | 78.1 | 79.3 | 34.5 | 50.2 | 55.1 |
| RTLCoder [73] | Mistral | 7B | 64.6 | 73.7 | 78.3 | 24.5 | 37.3 | 42.3 |
| | DeepSeek-Coder | 6.7B | 73.4 | 83.9 | 86.2 | 35.8 | 40.3 | 43.1 |
| CodeV [174] | CodeLlama | 7B | 79.0 | 89.2 | 89.9 | 39.4 | 50.3 | 53.1 |
| | DeepSeek-Coder | 6.7B | 78.3 | 87.4 | 89.1 | 42.4 | 51.5 | 53.2 |
| | CodeQwen | 7B | 78.8 | 89.5 | 92.4 | 36.6 | 53.3 | 61.3 |
| OriGen [22] | DeepSeek-Coder | 6.7B | - | - | - | - | 65.5 | - |
| Proposed SDG-CC-Repair | CodeLlama | 7B | **85.7** | **93.9** | 94.8 | 42.6 | 52.9 | 58.2 |
| | DeepSeek-Coder | 6.7B | 84.3 | 92.9 | 95.4 | **53.1** | 58.8 | 62.6 |
| | Starcoder2 | 15B | 79.8 | 93.9 | **96.2** | 49.0 | **65.8** | **74.5** |

where $n \geq k$ represents the total number of trials for each problem, and $c$ represents the number of trials that pass the functional check.

VerilogEval [71] contains two subsets of problems, where VerilogEval-Human contains manually converted problem descriptions from the original HDLBits website, and VerilogEval-Machine with GPT-3.5 generated problem descriptions.

RTLLM [80] is an open-source benchmark designed for generating Register Transfer Level (RTL) code from natural language instructions. It evaluates models on syntax correctness, functional correctness, and design quality, offering a thorough analysis of model outputs.

Experiment Results

Main Results  Table 5.12 and Table 5.13 compare the proposed models with baselines on VerilogEval and RTLLM. Baseline results are mainly sourced from [174]. For RTLLM, a large variance with biased pass@5 was found, thus all models were re-evaluated and unbiased pass@k metric is reported. The latest foundational and frontier code models were rigorously evaluated, including Llama-3.1 [31], DeepSeek-Coder-V2 [26], and GPT-4o. Recent foundational and frontier code models already reached competitive performance compared to previous efforts targeting Verilog code generation.

Compared to previous approaches like CodeV [174], the proposed models achieve comparable performance on VerilogEval-Machine and show significant improvements on benchmarks with human-like descriptions. Machine descriptions often provide detailed, line-by-line coding instructions, whereas human descriptions are high-level, integrating problem-solving skills and a deeper understanding of the hardware module's functionality. Enhancing the model's ability to handle human-like descriptions is crucial, as these more accurately reflect how designers interact with the models and set expectations for Verilog generation. The fine-tuned Starcoder2-15B surpasses previous state-of-the-art results by 3.8%, 10.9%, and 6.6% in pass@1 metrics on VerilogEval-Machine, VerilogEval-Human, and RTLLM, respectively.

Table 5.14 highlights the effectiveness of the generated data fine-tuned on Starcoder2-15B. The CC data enhances the model's ability to handle non-textual representations, leading to improved scores on VerilogEval-Human. The targeted code Repair data boosts performance across all benchmarks, suggesting that the model has learned to generalize from code repair tasks and reduce similar errors during code completion.

Table 5.14: Ablation study on training data. Data quantity indicated in parentheses.

| Model | VerilogEval | | RTLLM v1.1 |
| --- | --- | --- | --- |
| | Machine | Human | Func |
| | pass@1 (%) | | pass@5 (%) |
| Starcoder2-15B | 68.7 | 37.7 | 37.6 |
| SDG (80.1k) | 75.2 | 54.7 | 62.1 |
| SDG-CC (108.6k) | 73.9 | 62.0 | 62.8 |
| SDG-CC-Repair (110.0k) | 81.9 | 68.0 | 65.8 |

Improved Variability During Training  Figure 5.5b displays the pass rates for two consecutive checkpoints of Starcoder2-SDG-CC-Repair on VerilogEval-Human problems, sampled with a temperature of 0.8. Compared to Figure 5.5a, the updated model shows significant improvements by (1) moving previously unsolved problems into the solved category, including those with non-textual representations addressed by the correct-by-construction CC data, and (2) reducing the number of problems with large pass rate discrepancies, particularly where performance had degraded. The targeted repair data has effectively mitigated the model's tendency to repeat common mistakes found in the Repair dataset, despite the noise inherent in synthetically generated SDG data.

Scaling Data for Non-textual Representations  Figure 5.8 illustrates the scaling of correct-by-construction (CC) data and the fine-tuned Starcoder2-15B pass rate on problems involving non-textual representations. Testing was expanded to include strictly in-distribution test set, with each category containing around 50 problems. The results show that the model can quickly learn and comprehend these non-textual representations with as few as 4k training data samples, with the pass rate steadily improving as more data is provided. Additionally, the model demonstrates the ability to generalize to VerilogEval-NonText benchmark problems. While the models achieve near-perfect scores on KMap and FSM problems, they perform less effectively on Waveforms, suggesting that reverse engineering circuits from waveforms pose a greater challenge.

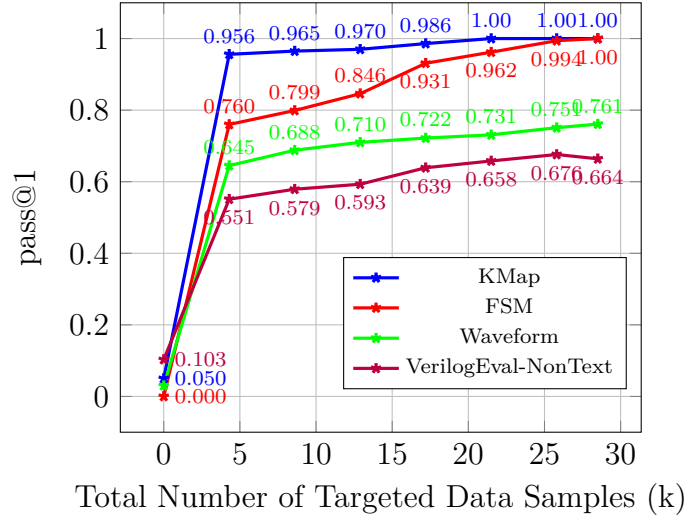

Figure 5.8: pass@1 on non-textual problems with total number of CC data with temperature 0.8.

Table 5.15: Ablation study on Repair data quality with Starcoder2-15B.

| Model | VerilogEval | | RTLLM v1.1 |
| --- | --- | --- | --- |
| | Machine | Human | Func |
| | pass@1 (%) | | pass@5 (%) |
| SDG-CC | 73.9 | 62.0 | 62.8 |
| SDG-CC-Repair | 81.9 | 68.0 | 65.8 |
| w/o self-consistency | 75.3 | 63.3 | 63.7 |
| w/o error report | 76.9 | 59.6 | 59.4 |

Ensuring Quality for Targeted Code Repair  The quality control mechanisms integrated into the data generation pipeline are crucial for improving model performance, particularly in correcting minor errors through targeted code repair. To evaluate the impact of these quality controls, an ablation study was conducted in Table 5.15, where each component of the targeted code repair generation pipeline was systemat-ically removed and the resulting model performance was assessed. Specifically, the self-consistency checks that validate whether the generated error report effectively guides the LLMs in correcting mistakes were eliminated. Additionally, the complete removal of the error report was tested, substituting it with random errors injected into the open-source code by the LLMs. The benchmark results indicate a significant performance drop when these validation processes are excluded. These findings highlight the essential role of both the self-consistency checks and the targeted error report in improving the model's ability to correct errors.

### 5.3.8 Related Work

Synthetic Data Generation for Model Fine-tuning. The performance of large language models (LLMs) hinge on the quality and diversity of their training data. To address the limitations of manual datasets, synthetic data generation methods [147, 161] have been developed to automatically create instruction-following examples from LLMs, reducing reliance on human annotations. Various techniques enhance data quality: [147] generates multiple reasoning traces and selects the most frequent output to improve robustness, while other approaches [67, 170] assess response quality based on these traces. Self-training methods utilize synthetic data for iterative fine-tuning, boosting reasoning capabilities [33, 116]. These advancements show how synthetic data can effectively scale and optimize models through iterative feedback.

Large Language Models for Code Generation. Recent breakthroughs in large language models (LLMs) have greatly enhanced their capability to tackle complex code generation tasks. Much of the research focuses on developing LLMs specialized for code by continuing their pretraining on code data [6, 26, 40, 111] from open-source repositories like GitHub [56, 77] and commit histories [91]. Further improvements to these models come from reinforcement learning [58] and more often instruction fine-tuning, which involves techniques to address more complex coding problems [81], increasing diversity with unlabeled open-source code [152, 156, 168], ensuring solution correctness through self-written tests [17], and validating and debugging code execution through interactions with LLM agents [59].

Large Language Models for Verilog Coding. While most code LLMs target software languages, there is increasing interest in models for hardware description languages like Verilog, essential for chip design and verification [70]. Previous work has addressed the challenge of limited data through various methods, including synthetic data generation [73], multi-level summarization of open-source Verilog code [174], and enhanced code augmentation with self-reflection based on compiler feedback [22, 132]. Other approaches focus on improving functional correctness and circuit performance through Monte Carlo Tree Search [27] and discriminator-guided sampling [100].

### 5.3.9 Discussions

In this work, synthetic data generation refers to methods of using large language models (LLMs) in data generation. While the approach of ensuring correctness through correct-by-construction could also be considered "synthetic" and resembles methods explored in works like AlphaGeometry [130], the problems addressed are much simpler and on a smaller scale. The observations about the variability of models on specific problems align with the findings of [89], where "the model knows how to produce the right answer, but it does not know how to select it." Instead of striving for absolute data correctness, preference learning [32, 108] or reinforcement learning [7, 58], targeted repair data is generated by analyzing errors and re-creating such scenarios by injecting similar errors into open-source code, somewhat analogous to how humans consolidate memories during sleep by integrating new information with past experiences [121, 143]. Further discussions on the generalizability and broader impact of this work are provided in Appendix I.2.

In conclusion, this work addresses key challenges in Verilog code generation through correct-by-construction data generation and targeted code repair strategies. Significant limitations are identified in existing synthetic data generation methods, including difficulties in handling non-textual representations and variability in training performance across benchmarks. To overcome these challenges, a pipeline is developed that generates correct-by-construction data and creates targeted repair data by injecting controlled errors into open-source Verilog code. Fine-tuning with the generated data achieves state-of-the-art results on the VerilogEval and RTLLM benchmarks. These advancements demonstrate the effectiveness of the proposed strategies in enhancing model robustness and performance for Verilog code generation tasks.

Reproducibility Statement  Evaluation benchmark details are provided in Appendix I.1.3, examples of targeted code repair data generation are presented in Appendix I.3, and examples of correct-by-construction data targeting non-textual representations are included in Appendix I.4. Prompt templates used for data generation are described in Appendix I.5. The full data generation pipeline is publicly available at: https://github.com/NVlabs/CraftRTL.

## 5.4 Data Pruning for Efficient Fine-Tuning

### 5.4.1 Motivation

Training dataset quality and scale profoundly impact large language model (LLM) effectiveness, as demonstrated by recent scaling law studies [2, 169]. Prominent code LLMs including CodeAlpaca [16], WizardCoder [81], and MagicCoder [152] have demonstrated remarkable capabilities by expanding their supervised fine-tuning datasets through synthetic code generation. Numerous synthetic code generation techniques have emerged, such as Self-Instruct [147], Evol-Instruct [161], and OSS-Instruct [152]. These scaling approaches, however, substantially increase computational demands and training expenses, thereby restricting widespread accessibility.

Fine-tuned models typically achieve optimal performance with comprehensive, high-quality datasets. Efficient fine-tuning methodologies for LLMs have recently garnered increasing attention. The Superficial Alignment Hypothesis [175] proposes that LLMs acquire core knowledge during pretraining and require only minimal instruction tuning data for human preference alignment. Parameter-efficient fine-tuning (PEFT) techniques present promising alternatives by minimizing trainable parameters [35, 47]. Active learning approaches offer another pathway by iteratively selecting training samples to enhance model efficiency [28, 122]. These meth-ods primarily focus on accuracy improvement through iterative processes, generally requiring multiple training and selection cycles.

Research has extensively examined data selection and pruning strategies, with evidence indicating that strategic pruning can exceed full dataset training performance [101, 144]. Pruning effectiveness, however, varies significantly based on dataset characteristics and size [120]. Many current methods involve substantial computational costs, including supervised metrics that require multiple training rounds to monitor losses and gradients [105, 159], or resource-intensive sampling techniques employing Monte Carlo methods [113], which constrain scalability. Researchers have developed effective pruning methods for large-scale datasets in LLM pretraining [24, 101], fine-tuning [19, 113], image datasets [87, 90], and vision-text training datasets [144]. These approaches demonstrate efficacy using clustering techniques with appropriate selection criteria. Nevertheless, efficient pruning strategies specifically tailored for code datasets remain relatively unexplored.

Despite significant advancements, a notable gap persists in efficient pruning approaches designed specifically for code datasets. Synthetically generated large-scale code datasets frequently exhibit substantial lexical redundancy due to consistent formatting and stylistic patterns. Synthetic datasets commonly employed for code LLM training often contain considerable redundancy and noise [144], primarily because verifying functional correctness for each generated program proves impractical. Consequently, numerous instruction-code pairs exhibit noise or redundancy. Strategic data selection and pruning to enhance efficiency becomes crucial for improving model performance without relying exclusively on extensive datasets. Furthermore, insights from AlphaCode [65] suggest that selecting representative solutions from ranked problem clusters enhances performance through underlying syntactic and semantic diversity.

This work introduces a scalable and effective data pruning methodology aimed at enhancing code generation capabilities in large language models. The proposed approach performs clustering of data samples by jointly considering problem instructions and corresponding code solutions, with dimensionality reduction applied to mitigate computational overhead. From each cluster, representative subsets are selected based on a range of pruning criteria. Empirical evaluations on large-scale synthetic datasets and downstream coding benchmarks demonstrate that the methodology significantly reduces training data volume while maintaining—or in some cases improving—model performance.

The contributions and key findings are outlined as follows:

- This work presents the first comprehensive investigation into data pruning for large-scale synthetic code fine-tuning, introducing a scalable and efficient strategy grounded in unsupervised learning techniques.

- Empirical results highlight substantial redundancy within synthetic code datasets: utilizing only 10% of the training data preserves the majority of benchmark performance, with performance reductions limited to 3.9% on HumanEval and 1.5% on MBPP compared to training on the full dataset.

- Moderate levels of pruning consistently lead to performance gains, with observed improvements of up to 2.7% on HumanEval and 3.5% on MBPP relative to models trained on the complete dataset.

- Ablation studies reveal that the choice of clustering algorithm has a pronounced effect on final performance, whereas the selection of pruning metrics has a comparatively smaller impact.

### 5.4.2 Related Work

This section examines recent advances in large language models for code generation (Section 5.4.2), reviews instructional finetuning approaches (Section 5.4.2), and explores data selection and pruning methodologies (Section 5.4.2).

Large Language Models for Code Generation

CodeAlpaca [16] extends the LLaMA architecture [128] by 20 thousands instruction-following examples generated using the Self-Instruct method [147], which aligns language models with automatically synthesized instructions. Building on this foundation, CodeLlama [111] further fine-tunes LLaMA2 [129] using 14 thousands additional Self-Instruct-style examples.

WizardCoder [81] adopts the Evol-Instruct framework [161] to enhance the CodeAlpaca dataset through progressive refinement of instruction-following samples in both depth and breadth. In contrast, Magicoder [152] applies the OSS-Instruct methodology to curate instruction-tuning data from unlabeled open-source code fragments, resulting in a dataset of 75 thousands examples derived from the StarCoder corpus [77].

Instructional Fine-tuning

Instructional dataset fine-tuning has proven to be an effective approach for alignment with human preferences. Through exploration of diverse instructional tasks, [148] revealed substantial improvements in zero-shot performance on unseen tasks. Expanding on these findings, [21] demonstrated that increasing both task diversity and model capacity yields significant performance improvements across various architectures. [102] leverage large language models to produce high-quality instruction-following data, resulting in enhanced zero-shot capabilities on novel tasks.

A seminal study [175] proposed the Superficial Alignment Hypothesis, suggesting that LLMs acquire core knowledge during pretraining while requiring minimal fine-tuning data for human preference alignment. This research demonstrated remarkable LLM performance improvements with merely 1,000 high-quality instruction examples. Consequently, numerous investigations have focused on enhancing dataset quality through various filtering techniques for general instruction following [19, 74, 163].

Data Pruning for Efficient Training

Researchers have explored diverse pruning strategies to identify informative training samples, each customized for specific applications. Data clustering has emerged as particularly effective for data pruning applications. TLDR [144] utilizes KMeans to aggregate similar data points with uniformly from every cluster, applying Image-Text Matching scores to identify suitable vision-text pairs. DEFT [24] implements unsupervised core-set selection for fine-tuning of LLMs, substantially improving efficiency in text-editing applications.

Metrics such as Hardness [120], Instruction Following Difficulty [63], and SuperFiltering [62] target challenging samples that resist learning or are susceptible to forgetting, monitoring individual data points throughout the training process. Sample influence metrics including LESS [159] and TracIn [105] monitor model gradients and individual sample impact, though these approaches incur significant computational expenses for large-scale models and datasets. Quality metrics derived from external oracles [19, 74] utilize robust language models such as ChatGPT for data selection, though this strategy may be limited by cost constraints.

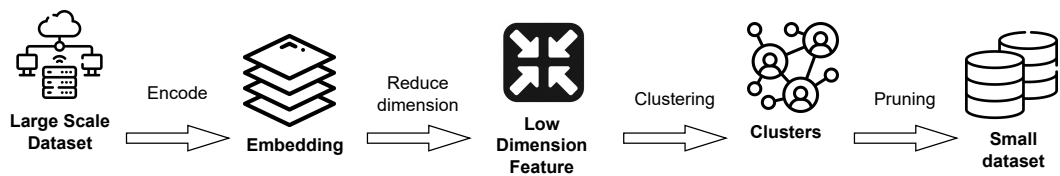

Figure 5.9: Comprehensive pipeline for efficient data pruning in LLM fine-tuning. The process consists of three key stages: (1) embedding instruction-following data and applying dimensionality reduction to the feature representations, (2) clustering to identify and group semantically similar data samples, and (3) applying pruning metrics to further reduce data size while preserving critical information.

### 5.4.3 Methodology

This research aims to identify high-quality, representative data samples that yield training performance equal to or surpassing that achieved with the complete dataset. Figure 5.9 depicts our efficient data pruning pipeline for fine-tuning LLMs with extensive datasets. The process begins with an embedding model that transforms instruction-code pairs into embedding representations. Next, reduce the dimensionality of these vectors to minimize computational demands. Clustering algorithms subsequently identify and group similar data samples, after which pruning metrics further refine the data volume. Algorithm 2 outlines the comprehensive procedure.

When processing coding datasets, two fundamental selection approaches exist: syntactical and semantic. Selecting programs that differ syntactically but share semantic equivalence, or vice versa, proves inefficient. Our methodology prioritizes the identification of syntactical differences. Detecting semantic variations between programs typically necessitates fuzzing techniques [18], which require generating numerous test samples and executing programs for behavior-based classification. Such approaches conflict with our goal of minimizing computational costs. Consequently, we emphasize syntactical analysis to achieve efficient and effective data selection.

---
**Algorithm 2** Cluster-Based Sample Selection for Data Pruning
---
1: Input: Sample representations $E$, target compression rate $r$
2: Initialize empty set $\mathcal{S} \leftarrow [\,]$
3: Apply dimensionality reduction: $Z \leftarrow \text{PCA}(E)$
4: Group samples: $\mathcal{C} \leftarrow \text{Clustering}(Z)$
5: for each cluster $(i, \mathcal{C}_i)$ in $\mathcal{C}$ do
6:     Compute selection scores: $w \leftarrow \text{PruningMetric}(\mathcal{C}_i)$
7:     Sample subset: $\mathcal{R} \leftarrow \text{WeightedSample}(\mathcal{C}_i, \text{prob} = w)$
8:     Update cluster: $\mathcal{C}_i \leftarrow \mathcal{R}$
9:     Add to final selection: $\mathcal{S} \leftarrow \mathcal{S} \cup \mathcal{R}$
10: end for
11: Return: $\mathcal{S}$
---

Dimension Reduction

Embedding models transform instruction-code pairs into vector representations, enhancing both clustering efficiency and pruning metric computation [93]. Recent studies confirm that distances derived from LLM embeddings effectively capture syntactic distinctions. Principal Component Analysis (PCA) [84] addresses computational complexity by reducing the dimensionality of vector representations, as LLM-extracted embeddings typically contain thousands of dimensions. This approach also mitigates the curse of dimensionality, preventing it from adversely affecting subsequent pruning metrics that utilize kernel methods.

Clustering

Clustering is employed to organize similar instruction-code pairs into coherent groups, facilitating the extraction of diverse and representative subsets. Prior to clustering, vector embeddings are normalized to ensure uniform contribution of each feature to the distance metrics. From each cluster, representative instruction-code pairs are sampled to construct a pruned dataset that preserves the original distribution, guided by various pruning criteria.

KMeans   The KMeans algorithm [51] partitions the dataset into $k$ disjoint clusters by minimizing the within-cluster sum of squared distances. Its scalability and computational efficiency make it particularly suitable for large-scale datasets, offering compact and well-separated cluster formations.

Agglomerative Clustering

Agglomerative Clustering [92] constructs a hierarchy of clusters through a bottom-up approach based on linkage criteria. Unlike KMeans, it does not require the number of clusters to be specified in advance, allowing for more adaptive and fine-grained grouping. This flexibility supports more effective selection of representative samples, particularly in heterogeneous datasets.

### HDBSCAN

Hierarchical Density-Based Spatial Clustering of Applications with Noise (HDB-SCAN) [109] identifies clusters based on local density, focusing on core samples located in high-density regions. This approach aligns well with the goal of extracting structurally representative examples, as it emphasizes syntactic cohesion. Additionally, HDBSCAN automatically classifies low-density points as noise or outliers, thereby offering built-in noise filtering during sample selection.

### 5.4.4 Pruning Metrics

The selection criteria for pruning metrics consistently align with identifying syntactic differences and extracting the most representative samples. The following sections detail the pruning metrics examined in our experiments. Details are in J.2

### 5.4.5 Experiments

This section presents the experimental setup in Section 5.4.5, followed by the main empirical findings in Section 5.4.5. The analysis highlights performance improvements achieved through data pruning techniques relative to training on the full dataset, evaluated across two benchmarks: MBPP(+) and HumanEval(+). In addition, $pass@1$ scores are compared against baseline methods under varying compression ratios.

Experiment Setup

All experiments utilize the DeepSeek-Coder-Base 6.7B model [41] as the base architecture, selected for its strong performance among open-source code generation models. Principal Component Analysis (PCA) [84] is applied to reduce embedding dimensionality to 10 prior to clustering. To account for variability introduced by clustering and training procedures, each experiment is conducted three times, with both the mean and standard deviation of the results reported.

Training

Datasets

Two synthetic instruction-tuning datasets are used for training: Magicoder-OSS-Instruct-75K [3] (MIT License) and Magicoder-Evol-Instruct-110K [4] (Apache-2.0 License). Combined, these datasets provide a total of 185,000 samples for large-scale model training.

The base model is fine-tuned on the merged and shuffled union of these two datasets. This training strategy differs from the original Magicoder [152] procedure, which applied two epochs of fine-tuning on OSS-Instruct followed by an additional two epochs on Evol-Instruct. Despite this difference, the performance achieved on the full dataset is comparable to that reported by the Magicoder$\mathcal{S}$-DS baseline.

Training

Training is conducted using 16 NVIDIA A100 80GB GPUs via PyTorch's

---

[3]https://huggingface.co/datasets/ise-uiuc/Magicoder-OSS-Instruct-75K
[4]https://huggingface.co/datasets/ise-uiuc/Magicoder-Evol-Instruct-110K

Distributed Data Parallel (DDP) framework. A learning rate of 5e-5 is employed with 15 warmup steps and a linear decay scheduler. The Adam optimizer [55] is used with full-parameter updates, and sequences exceeding 4096 tokens are truncated.

For experiments involving datasets with at least 10% of the original data volume, a batch size of 512 is used, following the configuration in [152]. In experiments with aggressively pruned datasets (as shown in Figure 5.10), a reduced batch size of 32 is applied in accordance with [175]. All models are fine-tuned for two epochs, regardless of dataset size.

Evaluation

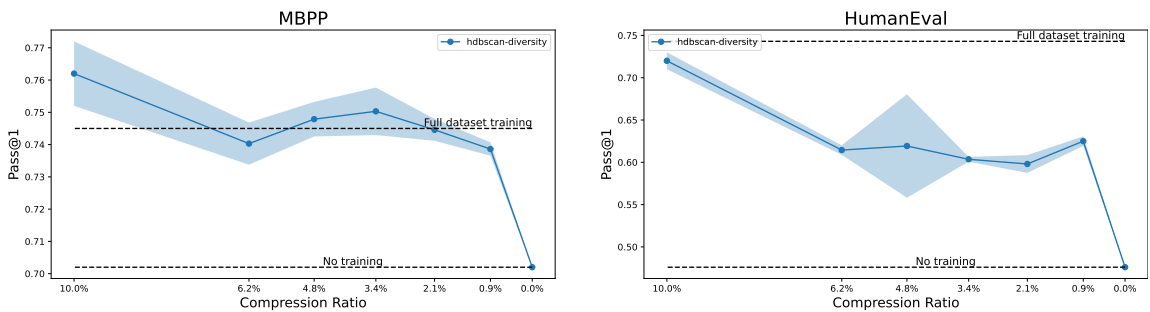

Figure 5.10: Performance comparison under extreme data pruning on the MBPP (left) and HumanEval (right) benchmarks. Using only 1% of the training data, our pruning method retains performance close to that of full-data training on MBPP, achieving a 4.1% gain over the base model. On HumanEval, although some performance drop is observed relative to full-data training, the 1% pruned set still delivers a notable 17.0% improvement over the base model.

Datasets

HumanEval [20] and MBPP [5] are two widely adopted benchmarks for evaluating code generation performance. HumanEval consists of 164 problems, while MBPP includes 1,401 problems. Each problem provides a task description (e.g., a docstring) as a prompt, and models are required to generate code solutions, which are then evaluated against predefined test cases.

To address the limited coverage of the original test suites, we use HumanEval+ and MBPP+, which are augmented versions created with EvalPlus [69]. These enhanced benchmarks offer significantly more test cases—approximately 80× more for HumanEval and 35× more for MBPP—enabling a more rigorous and comprehensive assessment of code correctness.

Metric In accordance with established methodologies [20, 69], I employ the unbiased pass@k estimator for each experiment, formulated as follows, with primary focus on the $pass@1$ metric:

$$pass@k := \mathbb{E}_{\text{Problems}} \left[ 1 - \frac{\binom{n-c}{k}}{\binom{n}{k}} \right]. \tag{5.4}$$

Inference I implement the EvalPlus [69] inference script with sanitation postprocessing. For all code generation tasks, I utilize the vLLM [57] framework with greedy decoding. The inference engine operates with bf16 precision, tensor parallel size of 2, and accommodates a maximum sequence length of 4096.

Implementation Details

In our experimental framework, PCA dimensionality reduction is first calibrated on the benchmark dataset and subsequently applied to the instruction data. Data embeddings are obtained using OpenAI's text-embedding-ada-002 model. All clustering and kernel density estimation procedures use default hyperparameters as provided by scikit-learn [99]. For clustering algorithms that require specifying the number of clusters (e.g., KMeans), we employ the Elbow method [110] to determine the point at which additional clusters yield diminishing returns in explained variance.

For pruning metrics, we adopt Scott's Rule [114]—a normal-reference rule—for selecting the Gaussian kernel bandwidth in kernel density estimation. Additionally, we randomly sample 10% of the dataset as the query set ($K$) for computing the diversity-based pruning metric.

Experiment Results

Table J.17 presents the $pass$@1 results of advanced code generation models evaluated on the HumanEval and MBPP benchmarks under greedy decoding. The reported numbers are averaged across three independent trials to mitigate the effects of variance introduced by clustering and training. All configurations utilize HDBSCAN in conjunction with the diversity-based pruning metric (HDBSCAN-diversity).

Our results show that applying moderate pruning can enhance model performance—yielding gains of up to 2.7% on HumanEval and 3.5% on MBPP relative to training on the complete dataset. Even when training with only 10% of the original data, model accuracy remains largely stable, showing minor performance degradation of 3.9% and 1.5% on HumanEval and MBPP, respectively. Under highly aggressive pruning conditions (e.g., 1% of data, approximately 700 samples), the models still achieve competitive accuracy and outperform the baseline, affirming the efficacy of our pruning pipeline.

Figure J.57 provides a detailed performance comparison across four datasets: MBPP, MBPP+, HumanEval, and HumanEval+. Each subplot contrasts the $pass$@1 accuracy of the HDBSCAN-diversity approach against a random baseline that does not apply clustering. Notably, the HDBSCAN-diversity strategy consistently deliv-

ers better results, with performance peaking when retaining 10–20% of training data, and remaining strong even when 90% of the data is pruned. These trends underline the robustness of the approach across a wide range of compression settings.

To explore the lower bounds of data efficiency, we further evaluate performance on MBPP under extreme pruning. As illustrated in Figure 5.10, using only 1% of the dataset results in a 4.1% performance gain over the baseline model, nearly matching the score achieved with the full dataset. This demonstrates that our pruning method is capable of preserving key learning signals even with minimal supervision, offering a compelling path toward reducing training overhead without compromising quality.

In summary, the experiments confirm that the proposed pruning strategy is both effective and efficient. It maintains high accuracy while drastically reducing training set size, making it a scalable and resource-conscious solution for training code generation models.

### 5.4.6  Ablation Studies

We conduct four ablation studies to evaluate the impact of key design choices on data pruning effectiveness: (1) clustering algorithms, (2) pruning metrics, (3) dimensionality reduction, and (4) input vector representations. These analyses primarily focus on the MBPP benchmark, given its stable and representative performance characteristics.

Comparison of Clustering Strategies

Figure J.58 illustrates the impact of various clustering techniques on pruning effectiveness, evaluated without the use of additional pruning metrics. The methods tested include Agglomerative Clustering, HDBSCAN, KMeans, and a control setup that omits clustering (denoted as nocluster).

Overall, the use of clustering consistently leads to better performance compared to the nocluster baseline, particularly at higher compression levels. HDBSCAN stands out as the most effective approach, delivering the highest *pass*@1 scores and showing strong capability in preserving valuable data points. While KMeans and Agglomerative Clustering also yield competitive results, they exhibit slightly more variability. These observations emphasize the importance of clustering in improving data pruning quality for code model training.

Compare Pruning Metrics

This research investigates the influence of various pruning metrics on model efficacy. Utilizing the HDBSCAN clustering algorithm, we evaluate metric performance as the dataset size decreases, illustrated in Figure J.59. Our analysis reveals that pruning metric effectiveness fluctuates across different compression ratios. Though Diversity metrics demonstrate subtle advantages over other approaches, these improvements remain modest and are primarily beneficial within the 10-40

|              | No PCA       | PCA         |
|--------------|--------------|-------------|
| Dimension    | 1536         | 10          |
| Runtime      | 1307 sec     | 183 sec     |
| MBPP (+)     | 74.4 (63.3)  | 73.8 (62.4) |
| HumanEval (+)| 71.8 (65.0)  | 67.5 (62.5) |

Table 5.16: Trade-off analysis between computational efficiency and performance when applying PCA dimensionality reduction. While PCA reduces runtime by approximately 12x (from 1307 to 183 seconds) and has minimal impact on MBPP performance (0.6% degradation), it shows a more noticeable effect on HumanEval (4.3% degradation). Results represent *pass*@1 scores at 50% compression using KMeans clustering, averaged over 3 runs.

Effect of PCA

Table 5.16 examines the influence of Principal Component Analysis on KMeans clustering with the Density metric at 50The results demonstrate that implementing PCA typically diminishes *pass*@1 scores by under 0.6

However, reducing dimensionality from 1536 to 10 accelerates KMeans processing approximately 12-fold. HDBSCAN clustering without PCA fails to complete within 4 hours, preventing statistical reporting.

Embeddings for Instruction or Code

Table J.18 investigates the performance impact of different embedding model inputs. Our analysis evaluates three input configurations: instructions only, code solutions only, and a combination of both for embedding generation. Results demonstrate that integrating both instructions and code as embedding inputs yields markedly better performance than using either type individually. We observed minimal performance differences between using only instructions versus only code inputs. These findings indicate that despite the typical correlation between instructions and their corresponding code implementations, preserving input diversity by selecting representative samples from both categories remains crucial for effective data pruning.

### 5.4.7 Limitations

The investigation faces a significant challenge due to inherent stochasticity in clustering algorithms and training frameworks. Computational constraints limited our experiments to three iterations per condition, with results subsequently averaged. Although this methodology provides valuable performance indicators, it may insufficiently capture statistical variability, potentially leading to imprecise conclusions. Additional iterations would substantially strengthen the reliability of our findings.

Our experiments adhered to hyperparameter specifications from [152], employing a batch size of 512 samples across 2 training epochs. This substantial batch size, however, yields relatively few gradient updates when applied to smaller datasets. To address this limitation, we reduced the batch size to 32 for pruned datasets containing less than 10% of the original data, following recommendations from [175]. Various hyperparameter configurations may significantly impact training outcomes, and determining optimal settings across different training data volumes represents a promising direction for future investigation.

### 5.4.8 Summary

This study proposes a resource-efficient data pruning framework aimed at optimizing the fine-tuning process of large language models on programming-related datasets. Experimental results demonstrate that integrating advanced clustering with selective pruning can notably boost data utilization efficiency, enabling reduced training cost without sacrificing—and in some cases, surpassing—baseline performance levels. Future work may explore the strategic generation of high-value training samples, particularly from clusters identified as low-utility by pruning metrics. The outcomes of this work offer meaningful contributions toward building scalable, high-performance training pipelines for code-oriented LLMs, while also informing improvements in both synthetic data curation and manual annotation workflows.

# Chapter 6  Reinforcement Learning for RTL Structured Logic Reasoning

This chapter explores the development of reinforcement learning (RL) methods tailored for structured logic reasoning in register-transfer level (RTL) hardware design. While previous work has scaled up RTL code generation datasets using synthetic and correct-by-construction techniques, a fundamental gap remains: the ability of LLMs to perform structured, logic-driven reasoning necessary for RTL correctness. Inspired by recent successes in RL-enhanced reasoning for mathematical and programmatic domains, this chapter proposes a novel RTL-oriented reasoning framework that integrates simplified logic games with RL training, achieving significant improvements in RTL code generation benchmarks.

## 6.1  Motivation

Previous efforts to enhance Verilog code generation have largely focused on improving data quality. Techniques such as correct-by-construction synthetic data generation and expert-guided targeted code repair have demonstrated notable improvements, leading to stronger functional correctness and better model robustness across RTL benchmarks. By scaling up high-quality RTL datasets and curating instruction-code pairs with domain knowledge, these approaches have pushed the performance boundaries of existing large language models.

However, despite these advancements, a fundamental challenge persists: achieving 100% functional correctness in generated RTL code remains elusive. Even models fine-tuned on carefully curated datasets exhibit inconsistencies, often producing subtle logical errors that compromise circuit functionality. The underlying difficulty lies not merely in generating syntactically correct Verilog, but in ensuring logically consistent, hardware-valid reasoning across complex design scenarios.

Meanwhile, chain-of-thought (CoT) prompting has shown that explicitly modeling intermediate reasoning steps significantly improves performance in mathematical reasoning and programming tasks. Building on this, reinforcement learning (RL)-enhanced reasoning has further demonstrated the ability to generalize across tasks and domains, offering a principled way to refine model outputs based on multi-step correctness signals rather than simple token-level supervision.

Motivated by these insights, this chapter proposes extending the strengths of chain-of-thought and reinforcement learning to RTL logic reasoning. The goal is to move beyond pure dataset scaling and toward structured logic training—teaching models not only to produce correct Verilog code, but to reason through intermediate symbolic steps that ensure functional correctness. By leveraging simplified logic tasks, reward-based optimization, and iterative refinement, it becomes possible to significantly enhance the reliability and quality of hardware-oriented code generation.

## 6.2 Background: Reasoning Models and RL Training Methods

Recent advancements in reasoning-oriented large language models (LLMs) have demonstrated substantial progress in tasks such as mathematics, programming, and multi-step decision making. This section reviews existing reasoning models and reinforcement learning (RL) training methods that have influenced the development of structured reasoning frameworks, highlighting key techniques and research gaps when applied to RTL and hardware logic domains.

### 6.2.1 State-of-the-Art Reasoning Models

Several large-scale reasoning models have been developed to improve the multi-step logical capabilities of LLMs:

- OpenAI-O1[49]: Focuses on structured chain-of-thought reasoning and alignment with human evaluators for complex reasoning tasks.

- DeepSeek-R1[39]: Emphasizes advanced multi-hop reasoning abilities, integrating synthetic reasoning chains during fine-tuning.

- Claude-3.7-thinking[4]: Specializes in coherent multi-turn reasoning and explanation generation across diverse domains.

- Gemini 2.0 Flash Thinking[25]: Combines reasoning and retrieval capabilities to improve accuracy on open-ended reasoning questions.

- Qwen/QwQ-32B [125]: Focuses on efficient scaling of reasoning abilities with lightweight architectural enhancements.

While these models achieve impressive results on general domains such as mathematics and code generation, minimal exploration has been conducted into domain-specific reasoning frameworks tailored for RTL and hardware logic tasks. The unique structured and symbolic nature of hardware designs demands specialized reasoning that differs significantly from generic math or programming challenges.

### 6.2.2 LLM Reinforcement Learning Training Methods

In addition to architectural and data improvements, reinforcement learning (RL) techniques have played a crucial role in enhancing LLM reasoning performance. Key RL-based methods include:

- RLHF (Reinforcement Learning with Human Feedback)[7]: Aligns model outputs with human preferences by using reward models trained on pairwise comparisons of outputs.

- GRPO (Guided Reward Policy Optimization)[115]: Introduces group-contrastive rewards that explicitly optimize differences between correct and incorrect reasoning traces, providing stronger alignment signals than standard RLHF.

- SWE-RL (RL on Open Software Evolution) [151]: Applies RL techniques to software evolution tasks, using code comparison and behavioral differences as reward signals for guiding model updates.

These methods demonstrate the potential of reinforcement learning to fine-tune models toward better multi-step reasoning, moving beyond simple supervised learning on static datasets.

### 6.2.3 Research Gap

Despite the progress in general-purpose reasoning and RL training, a significant gap remains in applying these methods to RTL and hardware logic domains. Existing RL reward designs are typically based on subjective human preference judgments or generic task success, which are not sufficient for verifying the structural and functional correctness required in RTL reasoning.

Specifically, there is a need for:

- Domain-specific reward signals: Reward mechanisms that explicitly verify logical correctness in hardware-oriented tasks, such as correct Karnaugh map simplifications or FSM transition validity.

- Structured reasoning supervision: Intermediate validation of logic steps rather than evaluating only final outputs, enabling more robust policy learning.

Addressing these gaps is critical for developing LLMs capable of reliable, step-by-step reasoning in RTL code generation and hardware logic design tasks.

## 6.3  Simplified RTL Reasoning Tasks

To bridge the gap between general reasoning models and hardware-oriented logic design, this work proposes a set of simplified RTL logic games inspired by the structure of symbolic logic puzzles. Drawing from the successes of logic-enhanced reasoning models such as Logic-RL [160], the proposed tasks decompose complex hardware design into small, verifiable reasoning steps.

The core idea is that mastering symbolic sub-tasks—such as Karnaugh map minimization or state transition synthesis—enhances the model's ability to reason through real-world RTL design problems, similar to how solving logic puzzles like Knights and Knaves can generalize to mathematical and programming reasoning.

### 6.3.1  Design Principles

Each simplified RTL task is designed to:

- Represent Fundamental Logic Skills: Cover symbolic reasoning steps that underlie hardware design.

- Be Sequential and Verifiable: Allow intermediate step validation, enabling reward assignment at each sub-step.

- Transfer to Real-World RTL: Improve generalization toward practical Verilog synthesis tasks.

### 6.3.2 Simplified Logic Sub-Tasks

Karnaugh Map Reasoning

Objective: Derive minimized Boolean equations from visual Karnaugh maps and generate corresponding Verilog logic.

Reasoning Steps:

1. Convert the Karnaugh map into a structured truth table.

2. Identify minterms corresponding to outputs equal to 1.

3. Derive the minimized Boolean logic expression.

4. Translate the logic into synthesizable Verilog code.

Example Reasoning Trace:

Given the input variables: a, b, c, d

The truth table generated from the Karnaugh map is: truthtable

The minterms (where output = 1) are: minterms

The corresponding minimized logic equation is: logic_equation

Finally, the Verilog code can be written as:

```verilog
module top_module(
    input a, b, c, d,
    output out
);
assign out = logic_equation;
endmodule
```

Finite State Machine (FSM) Reasoning

Objective: Construct correct next-state and output logic based on FSM state transition tables.

Reasoning Steps:

1. Determine binary encodings for each state.

2. Derive next-state transition logic from the table.

3. Derive output logic based on current state and inputs.

4. Implement the complete FSM logic in Verilog.

Boolean Logic and Waveform Reasoning

Objective: Reason over Boolean logic expressions and validate functional correctness through waveform analysis.

Reasoning Steps:

1. Expand Boolean expressions into complete truth tables.

2. Enumerate exhaustive input-output mappings in testbenches.

3. Simulate and visualize corresponding output waveforms.

### 6.3.3 Task Characteristics

These tasks share three key properties:

- Sequential Structure: Each reasoning trace follows a clear, ordered sequence, enabling stepwise validation.

- Symbolic Logic Focus: Emphasizes symbolic and structural reasoning rather than surface text pattern matching.

- Hardware Alignment: Directly aligned with RTL circuit design principles, making the learned reasoning transferable to practical Verilog code generation.

By training models on these simplified, verifiable tasks and rewarding correct intermediate reasoning steps, it becomes possible to substantially enhance hardware-specific reasoning capabilities without requiring full formal verification engines.

## 6.4 Reinforcement Learning Approaches

Building upon the simplified RTL reasoning tasks described in Section 6.3, this section details the reinforcement learning (RL) strategies employed to further enhance the structured reasoning abilities of large language models in hardware design contexts.

Rather than relying solely on supervised learning, which requires extensive labeled reasoning traces, reinforcement learning enables models to optimize multi-step decision-making policies based on reward signals tied to intermediate reasoning correctness. This paradigm is particularly well-suited for symbolic and logic-intensive domains like RTL design, where the final correctness often depends on a series of structured transformations.

### 6.4.1 Training Pipeline Overview

The overall training framework consists of the following stages:

1. Distill on Structured Reasoning Data with SFT: Models are initialized with supervised pretraining on synthetic chain-of-thought (CoT) reasoning traces derived from RTL-related tasks. This phase exposes models to the desired reasoning patterns and intermediate logic steps.

2. Reinforcement Learning Fine-tuning: Starting from the pretrained checkpoints, models are further optimized using reinforcement learning, where rewards are assigned based on the correctness of intermediate and final reasoning outputs.

3. Evaluation and Iterative Refinement: Performance is evaluated on held-out RTL reasoning tasks. If necessary, reward functions or task distributions are adjusted to better guide model improvement.

The base model used for experiments is a distilled checkpoint termed Distill-R1-llama-70B, derived from a 70B model pretrained on large-scale code datasets and further adapted to RTL-specific domains.

### 6.4.2 Reward Design

Effective reward design is critical for successful RL training in structured logic domains. In this work, reward functions are constructed to reflect:

- Intermediate Step Correctness: Partial rewards are granted for correctly completing substeps, such as generating accurate truth tables, deriving correct minterms, or correctly encoding FSM states.

- Final Output Consistency: Additional rewards are given if the final Verilog code implementation passes unit test suites or functional simulation checks.

- Symbolic Consistency: Penalties are applied for semantic mistakes such as invalid Boolean expressions, inconsistent state transitions, or waveform mismatches.

This fine-grained reward structure encourages models not only to produce correct final outputs, but also to follow logically valid intermediate reasoning paths, reducing reliance on shortcut memorization.

### 6.4.3 Training Details

Models are fine-tuned with a proximal policy optimization (PPO) variant adapted for LLMs.

Reward scaling and entropy regularization are tuned to balance exploration and stability.

Deepseek-R1 Chain-of-thought style outputs are preserved during RL fine-tuning to encourage transparent intermediate reasoning steps.

## 6.5 Experimental Evaluation

### 6.5.1 Experimental Setup

Dataset.

The training dataset consists of synthesized chain-of-thought (CoT) reasoning traces covering a wide range of simplified RTL logic tasks, including Karnaugh map minimization, FSM construction, and Boolean waveform reasoning. The CoT data is generated from multiple pretrained reasoning models, including DeepSeek-R1, OpenAI-O1, and Claude 3.7, to maximize reasoning style diversity. Additional samples specifically targeting hardware logic sub-tasks were generated to augment domain coverage.

Model.

The base model used is Distill-R1-llama-70B, a distilled checkpoint fine-tuned from a 70B parameter model optimized for programming and structured reasoning tasks.

Evaluation Metrics.

The models are evaluated using the pass@k metric, which measures whether at least one out of $k$ sampled completions is functionally correct based on simulation or output matching. We primarily report pass@1 and pass@10 accuracies across benchmarks.

## 6.5.2 Results

Table 6.1: Comparison of Verilog code generation performance across VerilogEval-Human, VerilogEval-Machine, and RTLLM benchmarks.

| Base Model | Data | VerilogEval-Human | VerilogEval-Machine | RTLLM |
|---|---|---|---|---|
| Llama-3.3-70B-Instruct | | 48.6% (51.6%) | 58.8% (60.8%) | (41.9%) |
| DeepSeek-R1-Distill-Llama-70B | | 52.5% | 62.7% | - |
| | Code-8k | 58.5% (64.3%) | 75.8% (80.8%) | (45.5%) |
| | CoT-8k | 72.2% | 79.3% (82.3%) | (48.3%) |
| | CoT-24k | 74.2% (85.3%) | 80.9% (88.8%) | (73.7%) |
| CoT-24k-CoT RL-8k | RL-8k | 76.3% (86.1%) | 81.1% (89.9%) | (74.3%) |
| | RL-24k | 79.6% (87.9%) | 83.3% (91.1%) | (78.5%) |
| R1 (full) | | 79.4% (85.8%) | 79.7% (88.8%) | - |

Key Observations:

Models fine-tuned with RL on RTL logic sub-tasks significantly outperform models trained only with CoT traces or code datasets. RL-8k achieves the highest overall accuracy across VerilogEval-Human, VerilogEval-Machine, and RTLLM, improving by +3.3% to +5.0% absolute compared to CoT-only fine-tuning. The performance gains are consistent across both human-curated and automatically evaluated Verilog datasets, demonstrating generalization.

## 6.5.3 Takeaways and Contributions

This work leads to several key findings and contributions in the intersection of reinforcement learning and RTL structured logic reasoning.

Main Findings.

Reinforcement learning on symbolic RTL sub-tasks significantly enhances structured logic reasoning and hardware-oriented code generation. Simplified logic "games"—such as Karnaugh map minimization, FSM synthesis, and waveform reasoning—serve as robust and scalable proxies for real hardware design challenges. By framing hardware logic tasks as sequential reasoning processes, it becomes possible to improve model robustness without requiring complex formal verification frameworks during training.

Contributions.

This dissertation presents the first comprehensive RTL-oriented reasoning framework built upon reinforcement learning-enhanced training methods. It systematically integrates symbolic sub-task design, chain-of-thought decomposition, and reward modeling tailored for hardware logic correctness. The approach establishes a new pathway for bridging the gap between generic reasoning models and domain-specific, high-stakes logic tasks such as RTL synthesis and verification.

Future Directions.

Several avenues are promising for further advancing this line of research. Enhancing reward modeling with coverage-driven feedback or lightweight formal verification signals could improve reasoning reliability under complex constraints. Extending the symbolic reasoning methodology to encompass more intricate RTL components—such as pipelined datapaths, dynamic state machines, and timing-critical control logic—offers the opportunity to generalize beyond basic synthesis tasks and toward end-to-end hardware design automation.

# Chapter 7  Conclusion

This dissertation explores how to enhance the generalization and usability of large language models (LLMs) under practical resource constraints—limited data availability, compute budget, or imperfect data quality. Across modality alignment, inference-time optimization, and low-resource code generation, the works presented form a unified effort toward building more adaptable, robust, and resource-efficient LLM systems for real-world deployment.

## 7.1  Summary of Contributions

This work contributes a coherent set of strategies addressing critical challenges across different aspects of LLM generalization:

First, a robust in-context multimodal alignment framework is developed, enabling LLMs to generalize across unseen modality combinations without retraining. By adopting a lightweight text-centric interface and introducing adversarial prompting techniques, the approach overcomes modality mismatch and robustness issues.

Second, the dissertation demonstrates that inference-time optimization—through uncertainty calibration, prompt perturbation, and chain-of-thought augmentation— can substantially improve model robustness without costly retraining. These methods offer a scalable alternative when computational resources are constrained.

Third, low-resource code generation is advanced through correct-by-construction data generation and targeted code repair strategies. These approaches address the significant redundancy and noise found in synthetic code datasets, achieving state-of-the-art results on Verilog benchmarks with far more efficient data use.

Finally, a new RTL-oriented reasoning framework is introduced, leveraging reinforcement learning on structured symbolic tasks such as Karnaugh map synthesis and FSM logic. This work represents the first systematic effort to bring reasoning model advances into the domain of hardware design and logic verification.

Together, these contributions provide complementary solutions for enabling LLMs to generalize under diverse and challenging resource conditions.

## 7.2 Lessons Learned

Several overarching lessons emerge from the research:

In-context learning is a powerful and underutilized avenue for generalization, especially when full fine-tuning is infeasible. Careful data design—through synthetic augmentation, symbolic structuring, and domain-specific prompting—can yield performance gains that rival or surpass scaling dataset size alone.

In the realm of symbolic reasoning, reinforcement learning techniques require careful adaptation. Standard RLHF pipelines are insufficient; instead, reward signals must be explicitly designed to reflect task-specific logical correctness.

Finally, a recurring theme across all domains is the trade-off between maximal performance and scalability. Methods like text-centric alignment and inference-time optimization deliberately sacrifice a small margin of accuracy to achieve greater flexibility, adaptability, and real-world applicability—an essential trade in practical LLM deployment.

## 7.3 Limitations

While significant progress has been made, several limitations remain:

The current text-centric multimodal framework is best suited for structured or classification tasks; generalization to more complex generation settings, such as dense video captioning or cross-modal synthesis, remains an open challenge. In reinforcement learning for RTL reasoning, sparse and binary rewards limit the full optimization potential, particularly in verifying nuanced hardware properties.

Despite efforts in correct-by-construction generation, a distribution gap persists between synthetic RTL tasks and organically developed hardware modules, potentially affecting generalization to highly complex designs.

Moreover, inference-time optimization techniques, though effective, introduce additional latency and token consumption, posing challenges for latency-sensitive deployments.

## 7.4 Future Directions

Several future research directions build naturally upon the foundations laid in this dissertation.

First, adaptive inference strategies that dynamically adjust computational effort based on task difficulty or input uncertainty could further improve the efficiency-robustness trade-off in real-time systems.

In reinforcement learning for RTL, integrating lightweight formal verification feedback into reward design holds promise for driving models toward stronger functional guarantees without requiring exhaustive simulation.

Extending text-centric multimodal alignment to open-world and long-tailed modality distributions—where inputs may be noisy, ambiguous, or sparsely labeled—remains a critical frontier for true multimodal reasoning.

Additionally, meta-learning approaches could enable LLMs to rapidly adapt to shifting resource constraints, training themselves to generalize from low-data, low-compute scenarios with minimal intervention.

Finally, blending symbolic reasoning with continuous optimization, particularly for hardware verification and automated code repair, offers a powerful new hybrid paradigm that could redefine both code generation and broader symbolic AI tasks.

By continuing to prioritize generalization, robustness, and resource efficiency as first-class objectives, future research can move closer to building LLMs that are not just powerful, but truly usable—across domains, across constraints, and across the real-world variability that defines practical intelligence.

# References

[1] Inside airbnb : Hawaii, 2023. Accessed on: 10 September, 2023.

[2] J. Achiam, S. Adler, S. Agarwal, L. Ahmad, I. Akkaya, F. L. Aleman, D. Almeida, J. Altenschmidt, S. Altman, S. Anadkat, et al. Gpt-4 technical report. arXiv preprint arXiv:2303.08774, 2023.

[3] J.-B. Alayrac, J. Donahue, P. Luc, A. Miech, I. Barr, Y. Hasson, K. Lenc, A. Mensch, K. Millican, M. Reynolds, et al. Flamingo: a visual language model for few-shot learning. Advances in Neural Information Processing Systems, 35:23716–23736, 2022.

[4] Anthropic. Claude 3.5 and claude 3.7 models. https://www.anthropic.com/news/claude-3-5-and-3-7, 2024. Accessed: 2025-04-27.

[5] J. Austin, A. Odena, M. Nye, M. Bosma, H. Michalewski, D. Dohan, E. Jiang, C. Cai, M. Terry, Q. Le, et al. Program synthesis with large language models. arXiv preprint arXiv:2108.07732, 2021.

[6] J. Bai, S. Bai, Y. Chu, Z. Cui, K. Dang, X. Deng, Y. Fan, W. Ge, Y. Han, F. Huang, et al. Qwen technical report. arXiv preprint arXiv:2309.16609, 2023.

[7] Y. Bai, A. Jones, K. Ndousse, A. Askell, A. Chen, N. DasSarma, D. Drain, S. Fort, D. Ganguli, T. Henighan, et al. Training a helpful and harmless assistant with reinforcement learning from human feedback. arXiv preprint arXiv:2204.05862, 2022.

[8] C. Batten, N. Pinckney, M. Liu, H. Ren, and B. Khailany. Pyhdl-eval: An llm evaluation framework for hardware design using python-embedded dsls. In Proceedings of the 2024 ACM/IEEE International Symposium on Machine Learning for CAD, MLCAD '24, New York, NY, USA, 2024. Association for Computing Machinery.

[9] I. Beltagy, M. E. Peters, and A. Cohan. Longformer: The long-document transformer. arXiv preprint arXiv:2004.05150, 2020.

[10] J. Bhandari, J. Knechtel, R. Narayanaswamy, S. Garg, and R. Karri. Llm-aided testbench generation and bug detection for finite-state machines, 2024.

[11] J. Blocklove, S. Garg, R. Karri, and H. Pearce. Chip-chat: Challenges and opportunities in conversational hardware design. arXiv preprint arXiv:2305.13243, 2023.

[12] T. Brown, B. Mann, N. Ryder, M. Subbiah, J. D. Kaplan, P. Dhariwal, A. Neelakantan, P. Shyam, G. Sastry, A. Askell, et al. Language models are few-shot learners. Advances in neural information processing systems, 33:1877–1901, 2020.

[13] F. Cassano, J. Gouwar, F. Lucchetti, C. Schlesinger, A. Freeman, C. J. Anderson, M. Q. Feldman, M. Greenberg, A. Jangda, and A. Guha. Knowledge

transfer from high-resource to low-resource programming languages for code llms, 2024.

[14] F. Cassano, J. Gouwar, D. Nguyen, S. Nguyen, L. Phipps-Costin, D. Pinckney, M.-H. Yee, Y. Zi, C. J. Anderson, M. Q. Feldman, A. Guha, M. Greenberg, and A. Jangda. Multipl-e: A scalable and extensible approach to benchmarking neural code generation, 2022.

[15] K. Chang, Z. Chen, Y. Zhou, W. Zhu, kun wang, H. Xu, C. Li, M. Wang, S. Liang, H. Li, Y. Han, and Y. Wang. Natural language is not enough: Benchmarking multi-modal generative ai for verilog generation, 2024.

[16] S. Chaudhary. Code alpaca: An instruction-following llama model for code generation. https://github.com/sahil280114/codealpaca, 2023.

[17] B. Chen, F. Zhang, A. Nguyen, D. Zan, Z. Lin, J.-G. Lou, and W. Chen. Codet: Code generation with generated tests, 2022.

[18] C. Chen, B. Cui, J. Ma, R. Wu, J. Guo, and W. Liu. A systematic review of fuzzing techniques. Computers & Security, 75:118–137, 2018.

[19] L. Chen, S. Li, J. Yan, H. Wang, K. Gunaratna, V. Yadav, Z. Tang, V. Srinivasan, T. Zhou, H. Huang, and H. Jin. Alpagasus: Training a better alpaca with fewer data, 2024.

[20] M. Chen, J. Tworek, H. Jun, Q. Yuan, H. P. d. O. Pinto, J. Kaplan, H. Edwards, Y. Burda, N. Joseph, G. Brockman, et al. Evaluating large language models trained on code. arXiv preprint arXiv:2107.03374, 2021.

[21] H. W. Chung, L. Hou, S. Longpre, B. Zoph, Y. Tay, W. Fedus, Y. Li, X. Wang,

M. Dehghani, S. Brahma, et al. Scaling instruction-finetuned language models. Journal of Machine Learning Research, 25(70):1–53, 2024.

[22] F. Cui, C. Yin, K. Zhou, Y. Xiao, G. Sun, Q. Xu, Q. Guo, D. Song, D. Lin, X. Zhang, et al. Origen: Enhancing rtl code generation with code-to-code augmentation and self-reflection. arXiv preprint arXiv:2407.16237, 2024.

[23] Y. Da Tsai and S. De Lin. Fast online inference for nonlinear contextual bandit based on generative adversarial network. arXiv preprint arXiv:2202.08867, 2022.

[24] D. Das and V. Khetan. Deft: Data efficient fine-tuning for large language models via unsupervised core-set selection. arXiv preprint arXiv:2310.16776, 2023.

[25] G. DeepMind. Introducing gemini 2.0: Our next-generation ai models. https://blog.google/technology/google-deepmind/google-gemini-ai-update-december-2024/#gemini-2-0-flash, 2024. Accessed: 2025-04-27.

[26] DeepSeek-AI, Q. Zhu, D. Guo, Z. Shao, D. Yang, P. Wang, R. Xu, Y. Wu, Y. Li, H. Gao, S. Ma, W. Zeng, X. Bi, Z. Gu, H. Xu, D. Dai, K. Dong, L. Zhang, Y. Piao, Z. Gou, Z. Xie, Z. Hao, B. Wang, J. Song, D. Chen, X. Xie, K. Guan, Y. You, A. Liu, Q. Du, W. Gao, X. Lu, Q. Chen, Y. Wang, C. Deng, J. Li, C. Zhao, C. Ruan, F. Luo, and W. Liang. Deepseek-coder-v2: Breaking the barrier of closed-source models in code intelligence, 2024.

[27] M. DeLorenzo, A. B. Chowdhury, V. Gohil, S. Thakur, R. Karri, S. Garg, and J. Rajendran. Make every move count: Llm-based high-quality rtl code generation using mcts, 2024.

[28] S. Diao, P. Wang, Y. Lin, and T. Zhang. Active prompting with chain-of-thought for large language models. arXiv preprint arXiv:2302.12246, 2023.

[29] X. Dong, Y. He, Z. Zhu, and J. Caverlee. Promptattack: Probing dialogue state trackers with adversarial prompts. arXiv preprint arXiv:2306.04535, 2023.

[30] A. Dosovitskiy, L. Beyer, A. Kolesnikov, D. Weissenborn, X. Zhai, T. Unterthiner, M. Dehghani, M. Minderer, G. Heigold, S. Gelly, et al. An image is worth 16x16 words: Transformers for image recognition at scale. arxiv 2020. arXiv preprint arXiv:2010.11929, 2010.

[31] A. Dubey, A. Jauhri, A. Pandey, A. Kadian, A. Al-Dahle, A. Letman, A. Mathur, A. Schelten, A. Yang, A. Fan, A. Goyal, A. Hartshorn, A. Yang, A. Mitra, A. Sravankumar, A. Korenev, A. Hinsvark, A. Rao, A. Zhang, A. Rodriguez, A. Gregerson, A. Spataru, B. Roziere, B. Biron, B. Tang, B. Chern, C. Caucheteux, C. Nayak, C. Bi, C. Marra, C. McConnell, C. Keller, C. Touret, C. Wu, C. Wong, C. C. Ferrer, C. Nikolaidis, D. Allonsius, D. Song, D. Pintz, D. Livshits, D. Esiobu, D. Choudhary, D. Mahajan, D. Garcia-Olano, D. Perino, D. Hupkes, E. Lakomkin, E. AlBadawy, E. Lobanova, E. Dinan, E. M. Smith, F. Radenovic, F. Zhang, G. Synnaeve, G. Lee, G. L. Anderson, G. Nail, G. Mialon, G. Pang, G. Cucurell, H. Nguyen, H. Korevaar, H. Xu, H. Touvron, I. Zarov, I. A. Ibarra, I. Kloumann, I. Misra, I. Evtimov, J. Copet, J. Lee, J. Geffert, J. Vranes, J. Park, J. Mahadeokar, J. Shah, J. van der Linde, J. Billock, J. Hong, J. Lee, J. Fu, J. Chi, J. Huang, J. Liu, J. Wang, J. Yu, J. Bitton, J. Spisak, J. Park, J. Rocca, J. Johnstun, J. Saxe, J. Jia, K. V. Alwala, K. Upasani, K. Plawiak, K. Li, K. Heafield, K. Stone, K. El-Arini,

K. Iyer, K. Malik, K. Chiu, K. Bhalla, L. Rantala-Yeary, L. van der Maaten, L. Chen, L. Tan, L. Jenkins, L. Martin, L. Madaan, L. Malo, L. Blecher, L. Landzaat, L. de Oliveira, M. Muzzi, M. Pasupuleti, M. Singh, M. Paluri, M. Kardas, M. Oldham, M. Rita, M. Pavlova, M. Kambadur, M. Lewis, M. Si, M. K. Singh, M. Hassan, N. Goyal, N. Torabi, N. Bashlykov, N. Bogoychev, N. Chatterji, O. Duchenne, O. Çelebi, P. Alrassy, P. Zhang, P. Li, P. Vasic, P. Weng, P. Bhargava, P. Dubal, P. Krishnan, P. S. Koura, P. Xu, Q. He, Q. Dong, R. Srinivasan, R. Ganapathy, R. Calderer, R. S. Cabral, R. Stojnic, R. Raileanu, R. Girdhar, R. Patel, R. Sauvestre, R. Polidoro, R. Sumbaly, R. Taylor, R. Silva, R. Hou, R. Wang, S. Hosseini, S. Chennabasappa, S. Singh, S. Bell, S. S. Kim, S. Edunov, S. Nie, S. Narang, S. Raparthy, S. Shen, S. Wan, S. Bhosale, S. Zhang, S. Vandenhende, S. Batra, S. Whitman, S. Sootla, S. Collot, S. Gururangan, S. Borodinsky, T. Herman, T. Fowler, T. Sheasha, T. Georgiou, T. Scialom, T. Speckbacher, T. Mihaylov, T. Xiao, U. Karn, V. Goswami, V. Gupta, V. Ramanathan, V. Kerkez, V. Gonguet, V. Do, V. Vogeti, V. Petrovic, W. Chu, W. Xiong, W. Fu, W. Meers, X. Martinet, X. Wang, X. E. Tan, X. Xie, X. Jia, X. Wang, Y. Goldschlag, Y. Gaur, Y. Babaei, Y. Wen, Y. Song, Y. Zhang, Y. Li, Y. Mao, Z. D. Coudert, Z. Yan, Z. Chen, Z. Papakipos, A. Singh, A. Grattafiori, A. Jain, A. Kelsey, A. Shajnfeld, A. Gangidi, A. Victoria, A. Goldstand, A. Menon, A. Sharma, A. Boesenberg, A. Vaughan, A. Baevski, A. Feinstein, A. Kallet, A. Sangani, A. Yunus, A. Lupu, A. Alvarado, A. Caples, A. Gu, A. Ho, A. Poulton, A. Ryan, A. Ramchandani, A. Franco, A. Saraf, A. Chowdhury, A. Gabriel, A. Bharambe, A. Eisenman, A. Yazdan, B. James, B. Maurer, B. Leonhardi,

B. Huang, B. Loyd, B. D. Paola, B. Paranjape, B. Liu, B. Wu, B. Ni, B. Hancock, B. Wasti, B. Spence, B. Stojkovic, B. Gamido, B. Montalvo, C. Parker, C. Burton, C. Mejia, C. Wang, C. Kim, C. Zhou, C. Hu, C.-H. Chu, C. Cai, C. Tindal, C. Feichtenhofer, D. Civin, D. Beaty, D. Kreymer, D. Li, D. Wyatt, D. Adkins, D. Xu, D. Testuggine, D. David, D. Parikh, D. Liskovich, D. Foss, D. Wang, D. Le, D. Holland, E. Dowling, E. Jamil, E. Montgomery, E. Presani, E. Hahn, E. Wood, E. Brinkman, E. Arcaute, E. Dunbar, E. Smothers, F. Sun, F. Kreuk, F. Tian, F. Ozgenel, F. Caggioni, F. Guzmán, F. Kanayet, F. Seide, G. M. Florez, G. Schwarz, G. Badeer, G. Swee, G. Halpern, G. Thattai, G. Herman, G. Sizov, Guangyi, Zhang, G. Lakshminarayanan, H. Shojanazeri, H. Zou, H. Wang, H. Zha, H. Habeeb, H. Rudolph, H. Suk, H. Aspegren, H. Goldman, I. Damlaj, I. Molybog, I. Tufanov, I.-E. Veliche, I. Gat, J. Weissman, J. Geboski, J. Kohli, J. Asher, J.-B. Gaya, J. Marcus, J. Tang, J. Chan, J. Zhen, J. Reizenstein, J. Teboul, J. Zhong, J. Jin, J. Yang, J. Cummings, J. Carvill, J. Shepard, J. McPhie, J. Torres, J. Ginsburg, J. Wang, K. Wu, K. H. U, K. Saxena, K. Prasad, K. Khandelwal, K. Zand, K. Matosich, K. Veeraraghavan, K. Michelena, K. Li, K. Huang, K. Chawla, K. Lakhotia, K. Huang, L. Chen, L. Garg, L. A, L. Silva, L. Bell, L. Zhang, L. Guo, L. Yu, L. Moshkovich, L. Wehrstedt, M. Khabsa, M. Avalani, M. Bhatt, M. Tsimpoukelli, M. Mankus, M. Hasson, M. Lennie, M. Reso, M. Groshev, M. Naumov, M. Lathi, M. Keneally, M. L. Seltzer, M. Valko, M. Restrepo, M. Patel, M. Vyatskov, M. Samvelyan, M. Clark, M. Macey, M. Wang, M. J. Hermoso, M. Metanat, M. Rastegari, M. Bansal, N. Santhanam, N. Parks, N. White, N. Bawa, N. Singhal, N. Egebo, N. Usunier, N. P. Laptev, N. Dong, N. Zhang,

N. Cheng, O. Chernoguz, O. Hart, O. Salpekar, O. Kalinli, P. Kent, P. Parekh, P. Saab, P. Balaji, P. Rittner, P. Bontrager, P. Roux, P. Dollar, P. Zvyagina, P. Ratanchandani, P. Yuvraj, Q. Liang, R. Alao, R. Rodriguez, R. Ayub, R. Murthy, R. Nayani, R. Mitra, R. Li, R. Hogan, R. Battey, R. Wang, R. Maheswari, R. Howes, R. Rinott, S. J. Bondu, S. Datta, S. Chugh, S. Hunt, S. Dhillon, S. Sidorov, S. Pan, S. Verma, S. Yamamoto, S. Ramaswamy, S. Lindsay, S. Lindsay, S. Feng, S. Lin, S. C. Zha, S. Shankar, S. Zhang, S. Zhang, S. Wang, S. Agarwal, S. Sajuyigbe, S. Chintala, S. Max, S. Chen, S. Kehoe, S. Satterfield, S. Govindaprasad, S. Gupta, S. Cho, S. Virk, S. Subramanian, S. Choudhury, S. Goldman, T. Remez, T. Glaser, T. Best, T. Kohler, T. Robinson, T. Li, T. Zhang, T. Matthews, T. Chou, T. Shaked, V. Vontimitta, V. Ajayi, V. Montanez, V. Mohan, V. S. Kumar, V. Mangla, V. Albiero, V. Ionescu, V. Poenaru, V. T. Mihailescu, V. Ivanov, W. Li, W. Wang, W. Jiang, W. Bouaziz, W. Constable, X. Tang, X. Wang, X. Wu, X. Wang, X. Xia, X. Wu, X. Gao, Y. Chen, Y. Hu, Y. Jia, Y. Qi, Y. Li, Y. Zhang, Y. Zhang, Y. Adi, Y. Nam, Yu, Wang, Y. Hao, Y. Qian, Y. He, Z. Rait, Z. DeVito, Z. Rosnbrick, Z. Wen, Z. Yang, and Z. Zhao. The llama 3 herd of models, 2024.

[32] K. Ethayarajh, W. Xu, N. Muennighoff, D. Jurafsky, and D. Kiela. Kto: Model alignment as prospect theoretic optimization. arXiv preprint arXiv:2402.01306, 2024.

[33] X. Feng, Z. Wan, M. Wen, Y. Wen, W. Zhang, and J. Wang. Alphazero-like tree-search can guide large language model decoding and training. arXiv preprint arXiv:2309.17179, 2023.

[34] N. Friedman. Introducing github copilot: your ai pair programmer. 2021.

[35] Z. Fu, H. Yang, A. M.-C. So, W. Lam, L. Bing, and N. Collier. On the effectiveness of parameter-efficient fine-tuning. In Proceedings of the AAAI Conference on Artificial Intelligence, volume 37, pages 12799–12807, 2023.

[36] W. Gao, Z. Deng, Z. Niu, F. Rong, C. Chen, Z. Gong, W. Zhang, D. Xiao, F. Li, Z. Cao, et al. Ophglm: Training an ophthalmology large language-and-vision assistant based on instructions and dialogue. arXiv preprint arXiv:2306.12174, 2023.

[37] R. Girdhar, A. El-Nouby, Z. Liu, M. Singh, K. V. Alwala, A. Joulin, and I. Misra. Imagebind: One embedding space to bind them all. In Proceedings of the IEEE/CVF conference on computer vision and pattern recognition, pages 15180–15190, 2023.

[38] Y. Gorishniy, I. Rubachev, V. Khrulkov, and A. Babenko. Revisiting deep learning models for tabular data. Advances in Neural Information Processing Systems, 34:18932–18943, 2021.

[39] D. Guo, D. Yang, H. Zhang, J. Song, R. Zhang, R. Xu, Q. Zhu, S. Ma, P. Wang, X. Bi, et al. Deepseek-r1: Incentivizing reasoning capability in llms via reinforcement learning. arXiv preprint arXiv:2501.12948, 2025.

[40] D. Guo, Q. Zhu, D. Yang, Z. Xie, K. Dong, W. Zhang, G. Chen, X. Bi, Y. Wu, Y. Li, et al. Deepseek-coder: When the large language model meets programming–the rise of code intelligence. arXiv preprint arXiv:2401.14196, 2024.

[41] D. Guo, Q. Zhu, D. Yang, Z. Xie, K. Dong, W. Zhang, G. Chen, X. Bi, Y. Wu, Y. K. Li, F. Luo, Y. Xiong, and W. Liang. Deepseek-coder: When the large language model meets programming – the rise of code intelligence, 2024.

[42] P.-F. Guo, Y.-H. Chen, Y.-D. Tsai, and S.-D. Lin. Towards optimizing with large language models. arXiv preprint arXiv:2310.05204, 2023.

[43] P.-F. Guo, Y.-D. Tsai, and S.-D. Lin. Benchmarking large language model uncertainty for prompt optimization. arXiv preprint arXiv:2409.10044, 2024.

[44] I. Guz, J. Elliott, M. Konstantin, S. Dane, V. Kassym, and W. Kan. Avito demand prediction challenge. https://kaggle.com/competitions/avito-demand-prediction, 2018. Accessed: 2025-02-01.

[45] T. Henighan, J. Kaplan, M. Katz, M. Chen, C. Hesse, J. Jackson, H. Jun, T. B. Brown, P. Dhariwal, S. Gray, et al. Scaling laws for autoregressive generative modeling. arXiv preprint arXiv:2010.14701, 2020.

[46] J. Ho, A. Jain, and P. Abbeel. Denoising diffusion probabilistic models. Advances in neural information processing systems, 33:6840–6851, 2020.

[47] E. J. Hu, Y. Shen, P. Wallis, Z. Allen-Zhu, Y. Li, S. Wang, L. Wang, and W. Chen. Lora: Low-rank adaptation of large language models. arXiv preprint arXiv:2106.09685, 2021.

[48] B. Hui, H. Yuan, N. Gong, P. Burlina, and Y. Cao. Pleak: Prompt leaking attacks against large language model applications. arXiv preprint arXiv:2405.06823, 2024.

[49] A. Jaech, A. Kalai, A. Lerer, A. Richardson, A. El-Kishky, A. Low, A. Helyar,

A. Madry, A. Beutel, A. Carney, et al. Openai o1 system card. arXiv preprint arXiv:2412.16720, 2024.

[50] R. Just, D. Jalali, and M. D. Ernst. Defects4j: a database of existing faults to enable controlled testing studies for java programs. In Proceedings of the 2014 International Symposium on Software Testing and Analysis, ISSTA 2014, page 437–440, New York, NY, USA, 2014. Association for Computing Machinery.

[51] T. Kanungo, D. M. Mount, N. S. Netanyahu, C. D. Piatko, R. Silverman, and A. Y. Wu. An efficient k-means clustering algorithm: Analysis and implementation. IEEE transactions on pattern analysis and machine intelligence, 24(7):881–892, 2002.

[52] J. Kaplan, S. McCandlish, T. Henighan, T. B. Brown, B. Chess, R. Child, S. Gray, A. Radford, J. Wu, and D. Amodei. Scaling laws for neural language models. arXiv preprint arXiv:2001.08361, 2020.

[53] T. Karras, S. Laine, and T. Aila. A style-based generator architecture for generative adversarial networks. In Proceedings of the IEEE/CVF conference on computer vision and pattern recognition, pages 4401–4410, 2019.

[54] B. Kawar, M. Elad, S. Ermon, and J. Song. Denoising diffusion restoration models. Advances in Neural Information Processing Systems, 35:23593–23606, 2022.

[55] D. P. Kingma and J. Ba. Adam: A method for stochastic optimization, 2017.

[56] D. Kocetkov, R. Li, L. B. Allal, J. Li, C. Mou, C. M. Ferrandis, Y. Jernite, M. Mitchell, S. Hughes, T. Wolf, D. Bahdanau, L. von Werra, and H. de Vries. The stack: 3 tb of permissively licensed source code, 2022.

[57] W. Kwon, Z. Li, S. Zhuang, Y. Sheng, L. Zheng, C. H. Yu, J. E. Gonzalez, H. Zhang, and I. Stoica. Efficient memory management for large language model serving with pagedattention. In Proceedings of the ACM SIGOPS 29th Symposium on Operating Systems Principles, 2023.

[58] H. Le, Y. Wang, A. D. Gotmare, S. Savarese, and S. C. H. Hoi. Coderl: Mastering code generation through pretrained models and deep reinforcement learning, 2022.

[59] B. Lei, Y. Li, and Q. Chen. Autocoder: Enhancing code large language model with AIEV-Instruct, 2024.

[60] P. Lewis, E. Perez, A. Piktus, F. Petroni, V. Karpukhin, N. Goyal, H. Küttler, M. Lewis, W. tau Yih, T. Rocktäschel, S. Riedel, and D. Kiela. Retrieval-augmented generation for knowledge-intensive nlp tasks, 2021.

[61] K. Li, Y. He, Y. Wang, Y. Li, W. Wang, P. Luo, Y. Wang, L. Wang, and Y. Qiao. Videochat: Chat-centric video understanding. arXiv preprint arXiv:2305.06355, 2023.

[62] M. Li, Y. Zhang, S. He, Z. Li, H. Zhao, J. Wang, N. Cheng, and T. Zhou. Superfiltering: Weak-to-strong data filtering for fast instruction-tuning. arXiv preprint arXiv:2402.00530, 2024.

[63] M. Li, Y. Zhang, Z. Li, J. Chen, L. Chen, N. Cheng, J. Wang, T. Zhou, and J. Xiao. From quantity to quality: Boosting llm performance with self-guided data selection for instruction tuning. arXiv preprint arXiv:2308.12032, 2023.

[64] R. Li, L. B. Allal, Y. Zi, N. Muennighoff, D. Kocetkov, C. Mou, M. Marone, C. Akiki, J. Li, J. Chim, Q. Liu, E. Zheltonozhskii, T. Y. Zhuo, T. Wang,

O. Dehaene, M. Davaadorj, J. Lamy-Poirier, J. Monteiro, O. Shliazhko, N. Gontier, N. Meade, A. Zebaze, M.-H. Yee, L. K. Umapathi, J. Zhu, B. Lipkin, M. Oblokulov, Z. Wang, R. Murthy, J. Stillerman, S. S. Patel, D. Abulkhanov, M. Zocca, M. Dey, Z. Zhang, N. Fahmy, U. Bhattacharyya, W. Yu, S. Singh, S. Luccioni, P. Villegas, M. Kunakov, F. Zhdanov, M. Romero, T. Lee, N. Timor, J. Ding, C. Schlesinger, H. Schoelkopf, J. Ebert, T. Dao, M. Mishra, A. Gu, J. Robinson, C. J. Anderson, B. Dolan-Gavitt, D. Contractor, S. Reddy, D. Fried, D. Bahdanau, Y. Jernite, C. M. Ferrandis, S. Hughes, T. Wolf, A. Guha, L. von Werra, and H. de Vries. Starcoder: may the source be with you!, 2023.

[65] Y. Li, D. Choi, J. Chung, N. Kushman, J. Schrittwieser, R. Leblond, T. Eccles, J. Keeling, F. Gimeno, A. Dal Lago, et al. Competition-level code generation with alphacode. Science, 378(6624):1092–1097, 2022.

[66] P. P. Liang, Y. Lyu, X. Fan, Z. Wu, Y. Cheng, J. Wu, L. Chen, P. Wu, M. A. Lee, Y. Zhu, et al. Multibench: Multiscale benchmarks for multimodal representation learning. arXiv preprint arXiv:2107.07502, 2021.

[67] H. Lightman, V. Kosaraju, Y. Burda, H. Edwards, B. Baker, T. Lee, J. Leike, J. Schulman, I. Sutskever, and K. Cobbe. Let's verify step by step. arXiv preprint arXiv:2305.20050, 2023.

[68] H. Liu, C. Li, Q. Wu, and Y. J. Lee. Visual instruction tuning, 2023.

[69] J. Liu, C. S. Xia, Y. Wang, and L. Zhang. Is your code generated by chatGPT really correct? rigorous evaluation of large language models for code genera-

tion. In Thirty-seventh Conference on Neural Information Processing Systems, 2023.

[70] M. Liu, T.-D. Ene, R. Kirby, C. Cheng, N. Pinckney, R. Liang, J. Alben, H. Anand, S. Banerjee, I. Bayraktaroglu, B. Bhaskaran, B. Catanzaro, A. Chaudhuri, S. Clay, B. Dally, L. Dang, P. Deshpande, S. Dhodhi, S. Halepete, E. Hill, J. Hu, S. Jain, A. Jindal, B. Khailany, G. Kokai, K. Kunal, X. Li, C. Lind, H. Liu, S. Oberman, S. Omar, G. Pasandi, S. Pratty, J. Raiman, A. Sarkar, Z. Shao, H. Sun, P. P. Suthar, V. Tej, W. Turner, K. Xu, and H. Ren. Chipnemo: Domain-adapted llms for chip design, 2024.

[71] M. Liu, N. Pinckney, B. Khailany, and H. Ren. Verilogeval: Evaluating large language models for verilog code generation. arXiv preprint arXiv:2309.07544, 2023.

[72] M. Liu, Y.-D. Tsai, W. Zhou, and H. Ren. Craftrtl: High-quality synthetic data generation for verilog code models with correct-by-construction non-textual representations and targeted code repair. arXiv preprint arXiv:2409.12993, 2024.

[73] S. Liu, W. Fang, Y. Lu, Q. Zhang, H. Zhang, and Z. Xie. Rtlcoder: Outperforming gpt-3.5 in design rtl generation with our open-source dataset and lightweight solution. arXiv preprint arXiv:2312.08617, 2023.

[74] W. Liu, W. Zeng, K. He, Y. Jiang, and J. He. What makes good data for alignment? a comprehensive study of automatic data selection in instruction tuning. arXiv preprint arXiv:2312.15685, 2023.

[75] Y. Liu, G. Deng, Y. Li, K. Wang, Z. Wang, X. Wang, T. Zhang, Y. Liu,

H. Wang, Y. Zheng, et al. Prompt injection attack against llm-integrated applications. arXiv preprint arXiv:2306.05499, 2023.

[76] Y. Liu, G. Deng, Z. Xu, Y. Li, Y. Zheng, Y. Zhang, L. Zhao, T. Zhang, K. Wang, and Y. Liu. Jailbreaking chatgpt via prompt engineering: An empirical study. arXiv preprint arXiv:2305.13860, 2023.

[77] A. Lozhkov, R. Li, L. B. Allal, F. Cassano, J. Lamy-Poirier, N. Tazi, A. Tang, D. Pykhtar, J. Liu, Y. Wei, et al. Starcoder 2 and the stack v2: The next generation. arXiv preprint arXiv:2402.19173, 2024.

[78] S. Lu, N. Duan, H. Han, D. Guo, S. won Hwang, and A. Svyatkovskiy. Reacc: A retrieval-augmented code completion framework, 2022.

[79] Y. Lu, S. Liu, Q. Zhang, and Z. Xie. Rtllm: An open-source benchmark for design rtl generation with large language model. arXiv preprint arXiv:2308.05345, 2023.

[80] Y. Lu, S. Liu, Q. Zhang, and Z. Xie. Rtllm: An open-source benchmark for design rtl generation with large language model. In 2024 29th Asia and South Pacific Design Automation Conference (ASP-DAC), pages 722–727. IEEE, 2024.

[81] Z. Luo, C. Xu, P. Zhao, Q. Sun, X. Geng, W. Hu, C. Tao, J. Ma, Q. Lin, and D. Jiang. Wizardcoder: Empowering code large language models with evol-instruct. In The Twelfth International Conference on Learning Representations, 2024.

[82] J. Ma, A. Cao, Z. Xiao, J. Zhang, C. Ye, and J. Zhao. Jailbreaking prompt at-

tack: A controllable adversarial attack against diffusion models. arXiv preprint arXiv:2404.02928, 2024.

[83] M. Ma, J. Ren, L. Zhao, D. Testuggine, and X. Peng. Are multimodal transformers robust to missing modality? In Proceedings of the IEEE/CVF Conference on Computer Vision and Pattern Recognition, pages 18177–18186, 2022.

[84] A. Maćkiewicz and W. Ratajczak. Principal components analysis (pca). Computers & Geosciences, 19(3):303–342, 1993.

[85] A. Madry, A. Makelov, L. Schmidt, D. Tsipras, and A. Vladu. Towards deep learning models resistant to adversarial attacks. arXiv preprint arXiv:1706.06083, 2017.

[86] L. McInnes, J. Healy, and J. Melville. Umap: Uniform manifold approximation and projection for dimension reduction. arXiv preprint arXiv:1802.03426, 2018.

[87] K. Meding, L. M. S. Buschoff, R. Geirhos, and F. A. Wichmann. Trivial or impossible–dichotomous data difficulty masks model differences (on imagenet and beyond). arXiv preprint arXiv:2110.05922, 2021.

[88] C. Meng, Y. He, Y. Song, J. Song, J. Wu, J.-Y. Zhu, and S. Ermon. Sdedit:guided image synthesis and editing with stochastic differential equations. International Conference on Learning Representations, 2022.

[89] Meta AI. Introducing meta llama 3: The most capable openly available llm to date, 2024. Accessed: 2024-09-10.

[90] B. B. Moser, F. Raue, and A. Dengel. A study in dataset pruning for image super-resolution. arXiv preprint arXiv:2403.17083, 2024.

[91] N. Muennighoff, Q. Liu, A. Zebaze, Q. Zheng, B. Hui, T. Y. Zhuo, S. Singh, X. Tang, L. von Werra, and S. Longpre. Octopack: Instruction tuning code large language models. arXiv preprint arXiv:2308.07124, 2023.

[92] D. Müllner. Modern hierarchical, agglomerative clustering algorithms. arXiv preprint arXiv:1109.2378, 2011.

[93] A. Naik. On the limitations of embedding based methods for measuring functional correctness for code generation. arXiv preprint arXiv:2405.01580, 2024.

[94] R. Nakano, J. Hilton, S. Balaji, J. Wu, L. Ouyang, C. Kim, C. Hesse, S. Jain, V. Kosaraju, W. Saunders, et al. Webgpt: Browser-assisted question-answering with human feedback. arXiv preprint arXiv:2112.09332, 2021.

[95] D. Nichols, J. H. Davis, Z. Xie, A. Rajaram, and A. Bhatele. Can large language models write parallel code? In Proceedings of the 33rd International Symposium on High-Performance Parallel and Distributed Computing, HPDC '24, page 281–294, New York, NY, USA, 2024. Association for Computing Machinery.

[96] Nvidia, :, B. Adler, N. Agarwal, A. Aithal, D. H. Anh, P. Bhattacharya, A. Brundyn, J. Casper, B. Catanzaro, S. Clay, J. Cohen, S. Das, A. Dattagupta, O. Delalleau, L. Derczynski, Y. Dong, D. Egert, E. Evans, A. Ficek, D. Fridman, S. Ghosh, B. Ginsburg, I. Gitman, T. Grzegorzek, R. Hero, J. Huang, V. Jawa, J. Jennings, A. Jhunjhunwala, J. Kamalu, S. Khan, O. Kuchaiev, P. LeGresley, H. Li, J. Liu, Z. Liu, E. Long, A. S.

Mahabaleshwarkar, S. Majumdar, J. Maki, M. Martinez, M. R. de Melo, I. Moshkov, D. Narayanan, S. Narenthiran, J. Navarro, P. Nguyen, O. Nitski, V. Noroozi, G. Nutheti, C. Parisien, J. Parmar, M. Patwary, K. Pawelec, W. Ping, S. Prabhumoye, R. Roy, T. Saar, V. R. N. Sabavat, S. Satheesh, J. P. Scowcroft, J. Sewall, P. Shamis, G. Shen, M. Shoeybi, D. Sizer, M. Smelyanskiy, F. Soares, M. N. Sreedhar, D. Su, S. Subramanian, S. Sun, S. Toshniwal, H. Wang, Z. Wang, J. You, J. Zeng, J. Zhang, J. Zhang, V. Zhang, Y. Zhang, and C. Zhu. Nemotron-4 340b technical report, 2024.

[97] OpenAI. Openai models api. 2023.

[98] H. Pearce, B. Tan, and R. Karri. Dave: Deriving automatically verilog from english. In *Proceedings of the 2020 ACM/IEEE Workshop on Machine Learning for CAD*, pages 27–32, 2020.

[99] F. Pedregosa, G. Varoquaux, A. Gramfort, V. Michel, B. Thirion, O. Grisel, M. Blondel, P. Prettenhofer, R. Weiss, V. Dubourg, J. Vanderplas, A. Passos, D. Cournapeau, M. Brucher, M. Perrot, and E. Duchesnay. Scikit-learn: Machine learning in Python. *Journal of Machine Learning Research*, 12:2825–2830, 2011.

[100] Z. Pei, H.-L. Zhen, M. Yuan, Y. Huang, and B. Yu. Betterv: Controlled verilog generation with discriminative guidance. *arXiv preprint arXiv:2402.03375*, 2024.

[101] G. Penedo, H. Kydlíček, L. von Werra, and T. Wolf. Fineweb, April 2024.

[102] B. Peng, C. Li, P. He, M. Galley, and J. Gao. Instruction tuning with gpt-4. *arXiv preprint arXiv:2304.03277*, 2023.

[103] Z. Peng, W. Wang, L. Dong, Y. Hao, S. Huang, S. Ma, and F. Wei. Kosmos-2: Grounding multimodal large language models to the world. arXiv preprint arXiv:2306.14824, 2023.

[104] F. Perez and I. Ribeiro. Ignore previous prompt: Attack techniques for language models. arXiv preprint arXiv:2211.09527, 2022.

[105] G. Pruthi, F. Liu, S. Kale, and M. Sundararajan. Estimating training data influence by tracing gradient descent. Advances in Neural Information Processing Systems, 33:19920–19930, 2020.

[106] Y. Qin, S. Liang, Y. Ye, K. Zhu, L. Yan, Y. Lu, Y. Lin, X. Cong, X. Tang, B. Qian, et al. Toolllm: Facilitating large language models to master 16000+ real-world apis. arXiv preprint arXiv:2307.16789, 2023.

[107] R. Qiu, G. L. Zhang, R. Drechsler, U. Schlichtmann, and B. Li. Autobench: Automatic testbench generation and evaluation using llms for hdl design, 2024.

[108] R. Rafailov, A. Sharma, E. Mitchell, C. D. Manning, S. Ermon, and C. Finn. Direct preference optimization: Your language model is secretly a reward model. Advances in Neural Information Processing Systems, 36, 2024.

[109] M. F. Rahman, W. Liu, S. B. Suhaim, S. Thirumuruganathan, N. Zhang, and G. Das. Hdbscan: Density based clustering over location based services. arXiv preprint arXiv:1602.03730, 2016.

[110] S. N. Roy. On a heuristic method of test construction and its use in multivariate analysis. The Annals of Mathematical Statistics, 24(2):220–238, 1953.

[111] B. Roziere, J. Gehring, F. Gloeckle, S. Sootla, I. Gat, X. E. Tan, Y. Adi,

J. Liu, T. Remez, J. Rapin, et al. Code llama: Open foundation models for code. arXiv preprint arXiv:2308.12950, 2023.

[112] T. Schick, J. Dwivedi-Yu, R. Dessì, R. Raileanu, M. Lomeli, L. Zettlemoyer, N. Cancedda, and T. Scialom. Toolformer: Language models can teach themselves to use tools. arXiv preprint arXiv:2302.04761, 2023.

[113] S. Schoch, R. Mishra, and Y. Ji. Data selection for fine-tuning large language models using transferred shapley values. arXiv preprint arXiv:2306.10165, 2023.

[114] D. W. Scott. Scott's rule. Wiley Interdisciplinary Reviews: Computational Statistics, 2(4):497–502, 2010.

[115] Z. Shao, P. Wang, Q. Zhu, R. Xu, J. Song, X. Bi, H. Zhang, M. Zhang, Y. Li, Y. Wu, et al. Deepseekmath: Pushing the limits of mathematical reasoning in open language models. arXiv preprint arXiv:2402.03300, 2024.

[116] A. Singh, J. D. Co-Reyes, R. Agarwal, A. Anand, P. Patil, P. J. Liu, J. Harrison, J. Lee, K. Xu, A. Parisi, et al. Beyond human data: Scaling self-training for problem-solving with language models. arXiv preprint arXiv:2312.06585, 2023.

[117] C. Snell, J. Lee, K. Xu, and A. Kumar. Scaling llm test-time compute optimally can be more effective than scaling model parameters. arXiv preprint arXiv:2408.03314, 2024.

[118] J. Song, C. Meng, and S. Ermon. Denoising diffusion implicit models. arXiv:2010.02502, October 2020.

[119] Y. Song, C. Lothritz, D. Tang, T. F. Bissyandé, and J. Klein. Revisiting code similarity evaluation with abstract syntax tree edit distance, 2024.

[120] B. Sorscher, R. Geirhos, S. Shekhar, S. Ganguli, and A. Morcos. Beyond neural scaling laws: beating power law scaling via data pruning. Advances in Neural Information Processing Systems, 35:19523–19536, 2022.

[121] R. Stickgold. Sleep-dependent memory consolidation. Nature, 437(7063):1272–1278, 2005.

[122] H. Su, J. Kasai, C. H. Wu, W. Shi, T. Wang, J. Xin, R. Zhang, M. Ostendorf, L. Zettlemoyer, N. A. Smith, et al. Selective annotation makes language models better few-shot learners. arXiv preprint arXiv:2209.01975, 2022.

[123] S. Sudalairaj, A. Bhandwaldar, A. Pareja, K. Xu, D. D. Cox, and A. Srivastava. Lab: Large-scale alignment for chatbots, 2024.

[124] S. Takamaeda-Yamazaki. Pyverilog: A python-based hardware design processing toolkit for verilog hdl. In Applied Reconfigurable Computing, volume 9040 of Lecture Notes in Computer Science, pages 451–460. Springer International Publishing, Apr 2015.

[125] Q. Team. Qwq-32b: Embracing the power of reinforcement learning, March 2025.

[126] A. TehraniJamsaz, A. Bhattacharjee, L. Chen, N. K. Ahmed, A. Yazdanbakhsh, and A. Jannesari. Coderosetta: Pushing the boundaries of unsupervised code translation for parallel programming, 2024.

[127] S. Thakur, B. Ahmad, H. Pearce, B. Tan, B. Dolan-Gavitt, R. Karri, and

S. Garg. Verigen: A large language model for verilog code generation. arXiv preprint arXiv:2308.00708, 2023.

[128] H. Touvron, T. Lavril, G. Izacard, X. Martinet, M.-A. Lachaux, T. Lacroix, B. Rozière, N. Goyal, E. Hambro, F. Azhar, et al. Llama: Open and efficient foundation language models. arXiv preprint arXiv:2302.13971, 2023.

[129] H. Touvron, L. Martin, K. Stone, P. Albert, A. Almahairi, Y. Babaei, N. Bashlykov, S. Batra, P. Bhargava, S. Bhosale, et al. Llama 2: Open foundation and fine-tuned chat models. arXiv preprint arXiv:2307.09288, 2023.

[130] T. H. Trinh, Y. Wu, Q. V. Le, H. He, and T. Luong. Solving olympiad geometry without human demonstrations. Nature, 625(7995):476–482, 2024.

[131] T.-H. Tsai, Y.-D. Tsai, and S.-D. Lin. lil'hdoc: an algorithm for good arm identification under small threshold gap. In Pacific-Asia Conference on Knowledge Discovery and Data Mining, pages 78–89. Springer, 2024.

[132] Y. Tsai, M. Liu, and H. Ren. Rtlfixer: Automatically fixing rtl syntax errors with large language models. arXiv preprint arXiv:2311.16543, 2023.

[133] Y.-D. Tsai and S.-D. Lin. Handling concept drift in non-stationary bandit through predicting future rewards. In Pacific-Asia Conference on Knowledge Discovery and Data Mining, pages 161–173. Springer, 2024.

[134] Y.-D. Tsai, C. Liow, Y. S. Siang, and S.-D. Lin. Toward more generalized malicious url detection models. In Proceedings of the AAAI Conference on Artificial Intelligence, volume 38, pages 21628–21636, 2024.

[135] Y.-D. Tsai, M. Liu, and H. Ren. Code less, align more: Efficient llm fine-tuning for code generation with data pruning. arXiv preprint arXiv:2407.05040, 2024.

[136] Y.-D. Tsai, T.-H. Tsai, and S.-D. Lin. Differential good arm identification. arXiv preprint arXiv:2303.07154, 2023.

[137] Y.-D. Tsai, Y.-C. Tsai, B.-W. Huang, C.-P. Yang, and S.-D. Lin. Automl-gpt: Large language model for automl. arXiv preprint arXiv:2309.01125, 2023.

[138] Y.-D. Tsai, T.-Y. Yen, P.-F. Guo, Z.-Y. Li, and S.-D. Lin. Text-centric alignment for multi-modality learning. arXiv preprint arXiv:2402.08086, 2024.

[139] Y.-D. Tsai, T.-Y. Yen, K.-T. Liao, and S.-D. Lin. Enhance modality robustness in text-centric multimodal alignment with adversarial prompting. arXiv preprint arXiv:2408.09798, 2024.

[140] M. Tufano, C. Watson, G. Bavota, M. D. Penta, M. White, and D. Poshyvanyk. An empirical study on learning bug-fixing patches in the wild via neural machine translation. ACM Trans. Softw. Eng. Methodol., 28(4), Sept. 2019.

[141] E. Tzeng, J. Hoffman, K. Saenko, and T. Darrell. Adversarial discriminative domain adaptation. In Proceedings of the IEEE conference on computer vision and pattern recognition, pages 7167–7176, 2017.

[142] O. Vinyals, A. Toshev, S. Bengio, and D. Erhan. Show and tell: A neural image caption generator. In Proceedings of the IEEE conference on computer vision and pattern recognition, pages 3156–3164, 2015.

[143] M. P. Walker and R. Stickgold. Sleep-dependent learning and memory consolidation. Neuron, 44(1):121–133, 2004.

[144] A. J. Wang, K. Q. Lin, D. J. Zhang, S. W. Lei, and M. Z. Shou. Too large; data reduction for vision-language pre-training. In Proceedings of the IEEE/CVF International Conference on Computer Vision, pages 3147–3157, 2023.

[145] H. Wang, Y. Zhang, and X. Yu. An overview of image caption generation methods. Computational intelligence and neuroscience, 2020(1):3062706, 2020.

[146] S. Wang, Z. Zhao, X. Ouyang, Q. Wang, and D. Shen. Chatcad: Interactive computer-aided diagnosis on medical image using large language models. arXiv preprint arXiv:2302.07257, 2023.

[147] Y. Wang, Y. Kordi, S. Mishra, A. Liu, N. A. Smith, D. Khashabi, and H. Hajishirzi. Self-instruct: Aligning language models with self-generated instructions. arXiv preprint arXiv:2212.10560, 2022.

[148] J. Wei, M. Bosma, V. Y. Zhao, K. Guu, A. W. Yu, B. Lester, N. Du, A. M. Dai, and Q. V. Le. Finetuned language models are zero-shot learners. arXiv preprint arXiv:2109.01652, 2021.

[149] J. Wei, X. Wang, D. Schuurmans, M. Bosma, B. Ichter, F. Xia, E. Chi, Q. Le, and D. Zhou. Chain-of-thought prompting elicits reasoning in large language models, 2023.

[150] J. Wei, X. Wang, D. Schuurmans, M. Bosma, F. Xia, E. Chi, Q. V. Le, D. Zhou, et al. Chain-of-thought prompting elicits reasoning in large language models. Advances in neural information processing systems, 35:24824–24837, 2022.

[151] Y. Wei, O. Duchenne, J. Copet, Q. Carbonneaux, L. Zhang, D. Fried, G. Synnaeve, R. Singh, and S. I. Wang. Swe-rl: Advancing llm reasoning via reinforcement learning on open software evolution. arXiv preprint arXiv:2502.18449, 2025.

[152] Y. Wei, Z. Wang, J. Liu, Y. Ding, and L. Zhang. Magicoder: Source code is all you need. arXiv preprint arXiv:2312.02120, 2023.

[153] Y. Weng, M. Zhu, F. Xia, B. Li, S. He, S. Liu, B. Sun, K. Liu, and J. Zhao. Large language models are better reasoners with self-verification, 2023.

[154] S. Williams and M. Baxter. Icarus verilog: open-source verilog more than a year later. Linux Journal, 2002(99):3, 2002.

[155] M. Wortsman, P. J. Liu, L. Xiao, K. Everett, A. Alemi, B. Adlam, J. D. Co-Reyes, I. Gur, A. Kumar, R. Novak, J. Pennington, J. Sohl-dickstein, K. Xu, J. Lee, J. Gilmer, and S. Kornblith. Small-scale proxies for large-scale transformer training instabilities, 2023.

[156] Y. Wu, D. Huang, W. Shi, W. Wang, L. Gao, S. Liu, Z. Nan, K. Yuan, R. Zhang, X. Zhang, Z. Du, Q. Guo, Y. Pu, D. Yin, X. Hu, and Y. Chen. Inversecoder: Unleashing the power of instruction-tuned code llms with inverse-instruct, 2024.

[157] Y.-A. Wu, Y.-D. Tsai, and S.-D. Lin. Linearapt: An adaptive algorithm for the fixed-budget thresholding linear bandit problem. arXiv preprint arXiv:2403.06230, 2024.

[158] C. S. Xia, Y. Wei, and L. Zhang. Automated program repair in the era of

large pre-trained language models. In 2023 IEEE/ACM 45th International Conference on Software Engineering (ICSE), pages 1482–1494, 2023.

[159] M. Xia, S. Malladi, S. Gururangan, S. Arora, and D. Chen. Less: Selecting influential data for targeted instruction tuning. arXiv preprint arXiv:2402.04333, 2024.

[160] T. Xie, Z. Gao, Q. Ren, H. Luo, Y. Hong, B. Dai, J. Zhou, K. Qiu, Z. Wu, and C. Luo. Logic-rl: Unleashing llm reasoning with rule-based reinforcement learning. arXiv preprint arXiv:2502.14768, 2025.

[161] C. Xu, Q. Sun, K. Zheng, X. Geng, P. Zhao, J. Feng, C. Tao, and D. Jiang. Wizardlm: Empowering large language models to follow complex instructions. arXiv preprint arXiv:2304.12244, 2023.

[162] Y. Xu and W. Wang. Linkprompt: Natural and universal adversarial attacks on prompt-based language models. In Proceedings of the 2024 Conference of the North American Chapter of the Association for Computational Linguistics: Human Language Technologies (Volume 1: Long Papers), pages 6473–6486, 2024.

[163] Y. Xu, Y. Yao, Y. Huang, M. Qi, M. Wang, B. Gu, and N. Sundaresan. Rethinking the instruction quality: Lift is what you need, 2023.

[164] Y. Yang, P. Huang, J. Cao, J. Li, Y. Lin, and F. Ma. A prompt-based approach to adversarial example generation and robustness enhancement. Frontiers of Computer Science, 18(4):184318, 2024.

[165] S. Yao, J. Zhao, D. Yu, N. Du, I. Shafran, K. Narasimhan, and Y. Cao. Re-188

act: Synergizing reasoning and acting in language models. arXiv preprint arXiv:2210.03629, 2022.

[166] T.-Y. Yen, Y.-D. Tsai, K.-T. Liao, and S.-D. Lin. Enhance the robustness of text-centric multimodal alignments. arXiv preprint arXiv:2407.05036, 2024.

[167] P. Young, A. Lai, M. Hodosh, and J. Hockenmaier. From image descriptions to visual denotations: New similarity metrics for semantic inference over event descriptions. Transactions of the Association for Computational Linguistics, 2:67–78, 2014.

[168] Z. Yu, X. Zhang, N. Shang, Y. Huang, C. Xu, Y. Zhao, W. Hu, and Q. Yin. Wavecoder: Widespread and versatile enhancement for code large language models by instruction tuning, 2024.

[169] B. Zhang, Z. Liu, C. Cherry, and O. Firat. When scaling meets llm finetuning: The effect of data, model and finetuning method. arXiv preprint arXiv:2402.17193, 2024.

[170] D. Zhang, S. Zhoubian, Y. Yue, Y. Dong, and J. Tang. Rest-mcts*: Llm self-training via process reward guided tree search. arXiv preprint arXiv:2406.03816, 2024.

[171] F. Zhang, B. Chen, Y. Zhang, J. Liu, D. Zan, Y. Mao, J.-G. Lou, and W. Chen. Repocoder: Repository-level code completion through iterative retrieval and generation. arXiv preprint arXiv:2303.12570, 2023.

[172] H. Zhang, P.-N. Kung, M. Yoshida, G. Van den Broeck, and N. Peng. Adaptable logical control for large language models. Advances in Neural Information Processing Systems, 37:115563–115587, 2024.

[173] K. Zhang, G. Li, Y. Dong, J. Xu, J. Zhang, J. Su, Y. Liu, and Z. Jin. Codedpo: Aligning code models with self generated and verified source code, 2024.

[174] Y. Zhao, D. Huang, C. Li, P. Jin, Z. Nan, T. Ma, L. Qi, Y. Pan, Z. Zhang, R. Zhang, et al. Codev: Empowering llms for verilog generation through multi-level summarization. arXiv preprint arXiv:2407.10424, 2024.

[175] C. Zhou, P. Liu, P. Xu, S. Iyer, J. Sun, Y. Mao, X. Ma, A. Efrat, P. Yu, L. Yu, S. Zhang, G. Ghosh, M. Lewis, L. Zettlemoyer, and O. Levy. Lima: Less is more for alignment, 2023.

[176] B. Zhu, B. Lin, M. Ning, Y. Yan, J. Cui, H. Wang, Y. Pang, W. Jiang, J. Zhang, Z. Li, et al. Languagebind: Extending video-language pretraining to n-modality by language-based semantic alignment. arXiv preprint arXiv:2310.01852, 2023.

[177] J. Zhu, Y. Shen, D. Zhao, and B. Zhou. In-domain gan inversion for real image editing. In European conference on computer vision, pages 592–608. Springer, 2020.

# Appendix A — Multimodal Mismatch Experiment Detail Setup

## A.1 Model Checkpoints

We conduct all experiments GPT-4-vision as the image caption model through OpenAI APIs [2], except for the analysis experiment that compares different image caption models.

## A.2 Hyperparameters

## A.3 Dataset

### A.3.1 PetFinder.my Adoption Prediction [? ]

examines what factors predict how quickly a pet is adopted after being listed. The dataset is a composite of the following modalities:

- Text: contains the description of the status of the pet

- Image: contains a profile photo of the pet

| Model | Hyperparameters |
| --- | --- |
| GPT-3.5-turbo | temperature=1, max_tokens=4096 |
| GPT-4-vision | temperature=0.8, max_tokens=300 |
| BLIP2 | default parameter |
| Kosmos2 | default parameter |
| Vision Transformer | default parameter |
| Flamingo | default parameter |
| Longformer | max_length=2048 |
| LLAMA-2-7b-chat | temperature=1, max_tokens=4096 |
| LLAMA-2-13b-chat | temperature=1, max_tokens=4096 |
| LLAMA-2-70b-chat | temperature=1, max_tokens=4096 |
| Mixtral | temperature=1, max_tokens=4096 |
| SDEdit | batch_size=1, sample_step=3, noise_scale=150 |
| DDRM | batch_size=1, degradation_type=deno, noise=1.5 |
| Idinvert | batch_size=64, gradient_accumulate=8, network_capacity=32 |

Table A.1: Hyper parameters.

- Tabular: contains basic information, such as gender and breed.

### A.3.2 Airbnb Pricing Prediction [1]

is composed of the following modalities used for making a regression prediction of housing prices:

- Text: contains the human-written description of the homestay, the neighborhood description, and the host's profile.

- Image: contains images of the homestay

- Tabular: delivers essential details such as location, rating score, and review counts.

### A.3.3 Avito Demand Prediction [44]

predicts the likelihood of an ad selling something based on user item and context features:

- Text: contains the ad title and description.

- Image: contains a profile photo of the item.

- Tabular: contains basic information, such as region, city, item category, etc.

| PetFinder | |
|---|---|
| Field | Value |
| url | https://www.kaggle.com/competitions/petfinder-adoption-prediction |
| # instances | 13453 |
| tabular columns | 23 |

| Airbnb | |
|---|---|
| Field | Value |
| url | http://insideairbnb.com/get-the-data/ |
| # instances | 12184 |
| tabular columns | 30 |

| Avito | |
|---|---|
| Field | Value |
| url | https://www.kaggle.com/competitions/avito-demand-prediction/data |
| # instances | 7000 |
| tabular columns | 18 |

Table A.2: Dataset Meta Info

## A.4 Foundation Models

For image modality, we utilize the embedding layer and tokenization method of the Vision Transformer [30]. This process splits the image into fixed-size patches and then projects each patch to obtain embeddings. For tabular modality, we employ the FT-Transformer[38] method to encode, dividing tabular features into numeric and categorical with separate projection layers for dimension enhancement. For text modality, the embedding layer of Longformer[9] is used for projection.

# Appendix B — Multimodal Mismatch Detailed Prompt

In this appendix, we provide detailed information about the prompts used in our experiments across different tasks and the implementation details that support our methodology.

We illustrate the summarization prompt in Figure B.2 and the reasoning prompts in Figure B.3. Figure B.1 further presents the detailed prompts used for image-to-tabular translation.

> Background
> The following is an dataset about Airbnb Rating Prediction. We will predict the rating score of the homestays, based on the information of homestays' listing on the dataset.
> Input Modality Related
> I would like you to rewrite all items in the input as text that represent tabular modality. That is, transfer image text data start with "The image shows" into format "The (column meaning) is (item)", with each column separated by a period.
> Output Modality Related
> I would like you to rewrite all items in the input as text that represent tabular modality. That is, transfer image data start with "The image shows" into format "The (column meaning) is (item)", with each column separated by a period.
> Ensure that:
> 1. All information should naturally blend together from different modalities, just like the contents of the same paragraph, and the order should be switched and blended, but not missing any information.
> 2. The tabluar column has : [host_response_time, host_response_rate,⋯]
> 3. Please start with "The host's response time to booking inquiries is".
> The following is the meaning of each column, brackets represent the options, you must include each column. If there are brackets, you must select one of the options, if there are no brackets and for numeric information, is only able to answer the number, if there are no brackets and for string information, please returns string.
> host_response_time - host's response time to booking inquiries
> host_response_rate - host's response rate to booking inquiries⋯
> Examples
> The host's response time to booking inquiries is within an hour. The host's response rate to booking inquiries is 96
> The host's response time to booking inquiries is within a few hours. The host's response rate to booking inquiries is 100
> The host's response time to booking inquiries is within a day. The host's response rate to booking inquiries is 90

Figure B.1: This example shows the prompt of translation. The blue part is the information of table modality format.

Background
The following is an dataset about Airbnb Rating Prediction. We will predict the pricing of the homestays, based on the information of homestays' listing on the dataset.
Input Modality Related
Input contains multiple modalities which are tabular, and image modality. We previously convert these modality into text and we try to use the combined text to do regression task. The tabular modality follows the format "The (column meaning) is (item)", with each column separated by a period.
The image modality start with "The image shows."
Output Modality Related
Ensure that:
1. If the original table contains columns that are marked as "Unknown," the conversion into natural text will omit any mention of those particular attributes.
2. Avoid fabricating or inventing any content!
3. All information should naturally blend together from different modalities, just like the contents of the same paragraph, and the order should be switched and blended, but not missing any information.
4. Please start with "This homestay is" and summarize into 1 to 2 paragraph.
Examples
Input:
The host's response time to booking inquiries is within an hour. The host's response rate to booking inquiries is 96%.⋯
Answer:
This homestay is a modern and inviting entire condo located in Honolulu's Primary Urban Center. Managed by a host who is⋯

Figure B.2: This example shows the prompt of summarization

Background
Airbnb, revolutionized travel by connecting guests with unique accommodations worldwide. Its diverse offerings, from cozy homes to exotic locales, cater to individual preferences. Airbnb offers a range of accommodation options with varying prices, allowing users to find lodging that suits their budget and preferences. Prices are determined by hosts based on factors such as location, amenities, and property type. Following are examples of different Airbnb data and their price.
Output Related
Now you are given an information of Airbnb data
What price range do you think the Airbnb falls within? Give me your answer and elaborate your reason.
Answer in the following format.
Ans:
Predicted Price Range:
Reason:(Explain why? Let's think step by step)
Examples
Following are examples of different pets and their adoption rate.
Airbnb 1 Information : The host's response time is within a week. The host's response rate is 70%. ⋯Predicted Price Range: 20 40 Airbnb 2 Information : The host's response time is within a month. The host's response rate is 78%. ⋯Predicted Price Range: 40 80
Airbnb 3 Information : The host's response time is within an hour. The host's response rate is 96%. ⋯Predicted Price Range: 100 140
Airbnb 4 Information : The host's response time is within an hour. The host's response rate is 99%. ⋯Predicted Price Range: 140 180
Airbnb 4 Information : The host's response time is within a few hours. The host's response rate is 100%. ⋯Predicted Price Range: 140 180

Figure B.3: This example shows the prompt of reasoning

# Appendix C — Multimodal Mismatch Implementation Details

## C.1 Implementation Details

### C.1.1 Model Configuration

For our experiments with Large Language Models, we utilized the following configuration parameters:

- Temperature: 0.7
- Top-p: 0.95
- Maximum token length: 4096
- Context window: 16K tokens

### C.1.2 Multimodal Integration Technique

Our multimodal integration approach follows these key steps:

1. Each modality is first processed by a modality-specific encoder

2. Representations are aligned to a common embedding space

3. The LLM performs cross-modal attention to integrate information

4. A text-centric decoder generates the final output

### C.1.3 Perturbation Methods

For robustness testing, we applied various types of perturbations:

- Text: Character/word/sentence deletion, substitution, and reordering

- Images: Gaussian noise, blur, cropping, and color distortion

- Tabular: Column deletion, value corruption, and row shuffling

The severity of perturbations was controlled using a scale from 1 (minor) to 5 (severe), as per standard practice in robustness benchmarks.

# Appendix D — Multimodal Mismatch Examples

In this section, we provide concrete examples that illustrate the application of our text-centric multimodal alignment approach in real-world scenarios.

## D.1  Modality Mismatch Intermediate Examples

This section presents step-by-step examples of the transformation process for handling modality mismatch, demonstrating each component of our pipeline.

### D.1.1  Text Transformation

A Real-world Example: When predicting diseases, we often have access to patients' pathology table reports, medical imaging, and audio of patient narration. First, we will perform text transformation on these data. For the images, we transfer it into captions such as "The patient has sigmoid colon cancer causing an obstruction, which has led to dilation in the descending colon." For the tables, we transform it into statements like "Histologic Type is Adenocarcinoma" and "Histologic Grade is Moderately differentiated." For the audio files, we perform speech recognition and acquire descriptions such as "I've been a little bloated for two weeks, and I have had only three bowel movements."

### D.1.2 Text-style Translation across Modality

A Real-world Example: When the training combination for disease prediction includes table modality, and only video modality data is available at inference, we will perform text-style translation on the textual representation of audio data. Continuing the example from Section 3.2.3, the textual representation of the audio, "I've been a little bloated for two weeks, and I have had only three bowel movements,", is translated as "Symptom is Bloating. The symptom is difficulty with bowel movements. Duration is Two weeks."

### D.1.3 Modality Summarization

A Real-world Example: Building on the example from Section 3.2.5, now the input includes two modalities: image and table. We summarize the textual representations from these modalities. Here is how the summarization looks: "The patient has moderately differentiated adenocarcinoma of the sigmoid colon, causing an obstruction and dilation of the descending colon."

### D.1.4 LLM Reasoning Augmentation

A Real-world Example: Building on the example from Section 3.2.6, now the input includes two modalities: image and table. The current goal is to determine whether a patient requires hospital observation. The results after augmentation are as follows:

"The obstruction in the sigmoid colon can lead to increased risks of bowel perforation, where the colon wall might rupture due to increased pressure. This complication is serious and requires immediate medical intervention."

## D.2 Modality Robustness Qualitative Analysis and Findings

In this section, we delve into the qualitative aspects and explore how LLMs can effectively handle scenarios where information is either incomplete or lost across different modalities and how LLMs compensate these lost information.

> Text with noise
> The type of pet is Dog. The name of pet is Filo. The age of pet when listed (in months) is 78. The color 1 of pet is Brown. The fur length of pet is Short. The primary breed of pet is Mixed Breed. The secondary breed of pet (if pet is of mixed breed) is Unknown.
> Summarization
> This pet is a gentle and handsome 6-year-old mixed breed dog named Filo. He has a light brown, short coat and soft brown eyes.

Figure D.4: Leveraging multimodal context to recover missing tabular data columns through LLM-based synthesis.

> Text with noise
> Jack experienced it Jack dog even is He recovering but special care skin which hair to Vet this was to Hopefully lover who take good dog a
> Summarization
> Although the profile write-up for Jack Jack is somewhat fragmented, it indicates that he has experienced some challenges but is on the road to recovery
> Reasoning
> 3. **Health Condition**: Jack Jack has a minor injury and is recovering. Potential adopters might be hesitant to take on a pet that requires special care, even if the injury is minor.

Figure D.5: Alignment module refining noisy text input into coherent and structured descriptions.

**LLMs recover lost data from other modalities** Figure D.4 illustrates that even when critical information is lost from one modality, our approach effectively leverages data from other available modalities to reconstruct the missing content. This ability highlights the strength of multimodal learning, where the complementary information across different modalities compensates for gaps, ensuring robust data recovery. Detailed input examples and reconstructions are provided in Appendix.

LLMs compensate missing information with knowledge and transform implicit relations into explicit text description by reasoning. In scenarios where input text is fragmented due to word dropout, and no relevant data is available from other modalities, Figure E.9 demonstrates how our method utilizes the extensive knowledge embedded in LLMs. The model not only reconstructs the missing words but also enhances the coherence of the text by drawing on contextual understanding and reasoning capabilities. This allows the LLM to infer and explicitly articulate underlying meanings that were only implicit in the original input.

| Train | Test | no finetuning: Emb | no finetuning: TAMML | unsupervised domain adaptation | supervised training (all modalities) |
|---|---|---|---|---|---|
| text | image | 0.288 | 0.374 | 0.195 | 0.338 |
| text | tabular | 0.289 | 0.357 | 0.281 | 0.359 |
| image | text | 0.270 | 0.319 | 0.276 | 0.306 |
| image | tabular | 0.273 | 0.348 | 0.276 | 0.359 |
| tabular | text | 0.289 | 0.364 | 0.195 | 0.306 |
| tabular | image | 0.279 | 0.364 | 0.195 | 0.338 |

Table D.3: Comparative evaluation of zero-shot and non-zero-shot adaptation methods under modality mismatch conditions. Results show that TAMML, despite requiring no fine-tuning, achieves competitive or superior performance compared to methods that require adaptation to target modalities.

| Training | Testing | PetFinder \| Accuracy ↑ | | | | |
| --- | --- | --- | --- | --- | --- | --- |
| | | SDEdit | Text Transformation | +Modality Summarization | +Reasoning Augmentation | +Text-style Translation |
| text+image | tabular | 0.282 | 0.310 | 0.321 | 0.338 | <u>0.348</u> |
| text+tabular | image | 0.289 | 0.329 | 0.365 | 0.363 | <u>0.380</u> |
| image+tabular | text | 0.281 | 0.305 | 0.295 | 0.321 | <u>0.355</u> |
| text | image+tabular | 0.291 | 0.282 | 0.296 | 0.343 | <u>0.344</u> |
| text | image | 0.289 | 0.293 | 0.298 | 0.341 | <u>0.374</u> |
| text | tabular | 0.293 | 0.297 | 0.318 | 0.315 | <u>0.357</u> |
| image | text+tabular | 0.290 | 0.314 | 0.289 | 0.325 | <u>0.341</u> |
| image | text | 0.288 | 0.306 | 0.330 | <u>0.336</u> | 0.319 |
| image | tabular | 0.291 | 0.300 | 0.307 | 0.303 | <u>0.348</u> |
| tabular | text+image | 0.290 | 0.194 | <u>0.366</u> | 0.341 | 0.360 |
| tabular | text | 0.289 | 0.193 | 0.306 | 0.327 | <u>0.364</u> |
| tabular | image | 0.289 | 0.196 | 0.357 | 0.353 | <u>0.364</u> |
| Average ± Variance ($\times 10^{-4}$) | | 0.289 ± 0.12 | 0.277 ± 25.91 | 0.321 ± 7.2 | 0.334 ± 2.5 | 0.355 ± 2.4 |

Table D.4: Ablation studies on various components of TAMML . Text transformations improve performance across all modalities except tabular data, which uses fixed-format text. Adding a summarization module effectively addresses this formatting issue, significantly enhancing performance. The integration of both translation and reasoning augmentation modules further improves overall results.

> Text transformation - text
>
> The type of pet is Dog. The name of pet is Jack Jack. The age of pet when listed (in months) is 36. The gender of pet is Male. The color 1 of pet is Brown. The color 2 of pet is Unknown. The color 3 of pet is Unknown. The size of pet at maturity is Small. The fur length of pet is Short. The vaccination status of pet is Yes. The deworming status of pet is Yes. The sterilisation status of pet is No. The health condition of pet is Minor-Injury. The number of pets represented in profile is 1. The pet adoption fee is Free. The state location of pet in Malaysia is Selangor. The total uploaded videos for this pet is 0. The total uploaded photos for this pet is 3. The primary breed of pet is Chihuahua. The secondary breed of pet (if pet is of mixed breed) is Unknown.
>
> The image presents a small, fawn-colored dog with large ears that stand upright, creating an alert and inquisitive expression. The dog is looking up towards the camera with its head slightly tilted, which may indicate curiosity or attentiveness. It is wearing a light blue collar with a bell, signifying that it is a pet likely kept indoors. The dog is positioned on a plain tiled floor, with the low angle perspective of the shot highlighting its petite stature and endearingly disproportionate ear-to-head size ratio.
>
> <span style="color:red">The profile write-up for this pet is : Jack experienced it Jack dog even is He recovering but special care skin which hair to Vet this was to Hopefully lover who take good dog a</span>

> After Summarization and Reasoning - text
>
> This pet is a charming 36-month-old male Chihuahua named Jack Jack, who has a short brown fur coat. Despite a minor injury, he is in good health and has been both vaccinated and dewormed, although he has not been sterilized. Jack Jack is a small-sized dog who currently resides in Selangor, Malaysia, and is available for adoption free of charge. Jack Jack's profile includes three photos, one of which captures him as a small, fawn-colored dog with large, upright ears that give him an alert and inquisitive look. He is seen wearing a light blue collar with a bell, suggesting he is an indoor pet. The image shows him on a plain tiled floor, with his head slightly tilted up towards the camera, highlighting his petite stature and the endearing size ratio of his ears to his head. <span style="color:blue">Although the profile write-up for Jack Jack is somewhat fragmented, it indicates that he has experienced some challenges but is on the road to recovery. He requires special care for his skin and fur, and hopefully, a dog lover will come forward to provide him with the good care he deserves.</span> For more information or photos, interested adopters can reach out via Whatsapp. Several factors influence the adoption rate for Jack Jack, and while he has some positive attributes, there are also some challenges that may slow down his adoption.
>
> 1. **Breed and Size**: Jack Jack is a Chihuahua, a breed that is generally popular due to their small size and suitability for apartment living. This is a positive factor that could help in his adoption.
> 2. **Age**: At 36 months (3 years old), Jack Jack is no longer a puppy. Puppies tend to get adopted faster than adult dogs, which may slow down his adoption rate.
> 3. <span style="color:lightblue">**Health Condition**: Jack Jack has a minor injury and is recovering. Potential adopters might be hesitant to take on a pet that requires special care, even if the injury is minor.</span>
> 4. **Vaccination and Deworming**: The fact that Jack Jack is vaccinated and dewormed is a positive aspect and can reassure potential adopters about his health.
> 5. **Sterilisation Status**: Jack Jack is not sterilized, which could be a concern for some adopters. Many adopters prefer pets that are already neutered to avoid the cost and responsibility of the procedure.
> 6. **Profile Write-Up**: The profile write-up for Jack Jack is somewhat unclear and lacks detailed information that could appeal to potential adopters. A well-written profile can significantly impact adoption rates.
> 7. **Photos**: There are only 3 photos uploaded. More photos and possibly videos could help showcase Jack Jack's personality and increase interest.
> 8. **Location**: Jack Jack is located in Selangor, Malaysia, which has a significant population and potentially a larger pool of adopters. However, the adoption rate can still vary based on local preferences and the number of available pets.

Figure D.6: LLM extracting semantic meaning from noise-corrupted text and converting implicit relationships to explicit descriptions. Red: noisy input; blue: summarized output; light blue: reasoning-derived meaning.

Text transformation - table

The type of pet is Dog. The name of pet is Filo. <span style="color:red">The age of pet when listed (in months) is 78.</span> The gender of pet is Male. The color 1 of pet is Brown. The color 2 of pet is Unknown. The color 3 of pet is Unknown. The size of pet at maturity is Large. The fur length of pet is Short. The vaccination status of pet is Yes. The deworming status of pet is Yes. The sterilisation status of pet is Unsure. The health condition of pet is Healthy.<span style="color:red">The number of pets represented in profile is 1.</span> The pet adoption fee is Free. The state location of pet in Malaysia is Kuala Lumpur. The total uploaded videos for this pet is 0.<span style="color:red">The total uploaded photos for this pet is 5.</span> The primary breed of pet is Mixed Breed. <span style="color:red">The secondary breed of pet (if pet is of mixed breed) is Unknown</span>

<span style="color:purple">The image shows a dog with a light brown, short coat lying comfortably on a blue plastic crate.</span> The dog has a gentle expression, with soft brown eyes, and appears to be calmly resting inside what looks like a tiled room, likely a part of its home. The dog's collar with an identification tag suggests it is well-loved and cared for. The environment looks clean and provides a sense of a safe and secure space for the pet to relax.

The profile write-up for this pet is : Meet Filo, the cute and handsome 6 year old mixed breed boy. Filo is a lucky stray saved by a kind owner earlier this year. <span style="color:purple">This kind owner brought Filo to the vet to treat some external wounds.</span> However, the kind owner has 2 dogs and don't have room for Filo. Filo continues to stay in the cage with the vet from that day till now while his wounds are treated. When the vet is available, Filo gets to go out for a short walk. If the clinic is packed, Filo stays in the cage for the whole day. Its not anybody's fault. Filo is still grateful for this kind owner and the kind vet for their hospitality. But Filo's future is still very uncertain and he is hoping that he's lucky to be granted a permanent loving home. I met Filo when I brought my poodle to visit the clinic. I learn of Filo's story from this kind vet. If you can give Filo a loving home or if you need more information, please call me.

---

After Summarization and Reasoning - table

<span style="color:blue">This pet is a gentle and handsome 6-year-old mixed breed dog named Filo. He has a light brown, short coat and soft brown eyes,</span> giving him a calm and serene expression.

...

Filo's profile features five photos, one of which shows him lying comfortably on a blue plastic crate inside a tiled room, likely part of his temporary home at the clinic. The clean and secure environment, along with his collar and identification tag, suggests that he is well-loved and cared for. If you can provide Filo with the forever home he deserves or need more information, please call the number provided. Filo's adoption rate is likely to be more than 3 months for several reasons:

1. **Age**: At 78 months (6.5 years old), Filo is significantly older than the other pets listed. Younger pets, especially puppies and kittens, generally have higher adoption rates because they are perceived as more adaptable and have a longer potential lifespan.

<span style="color:blue">2. **Health and Background**: While Filo has been treated for his external wounds, his background as a stray and his extended stay in a cage might raise concerns for potential adopters about his health and behavior. The write-up mentions his gratefulness and calm demeanor, but it does not provide detailed information about his current health status or behavior, which could be crucial for potential adopters.</span>

3. **Emotional Connection**: The profile write-up is heartfelt and provides a touching backstory, but it lacks the emotional appeal seen in the profiles of younger pets. The language used to describe Filo's situation is more factual and less engaging compared to descriptions of other pets, which emphasize their cuteness and playful nature.

4. **Visual Appeal**: Although there are 5 photos, the description of the image shows Filo in a resting position, which may not be as engaging as images of playful or interactive behavior. Potential adopters often respond more positively to images that show the pet's personality and energy.

5. **Competition**: Filo is competing with younger, more visually appealing pets that are often adopted faster. His profile needs to stand out more to attract potential adopters who are specifically looking for an older, more mature dog.

6. **Location and Accessibility**: The profile does not specify the exact location beyond being at a vet clinic, which might make it less accessible for potential adopters who prefer to know more about where the pet is currently staying.

Figure D.7: Cross-modal information synthesis for reconstructing missing tabular data. Gray: dropped columns; red: retained columns; purple: referenced image data; blue: generated information from visual context.

# Appendix E — Modality Robustness Analysis

This section examines how large language models (LLMs) qualitatively respond to incomplete or missing information across input modalities, with a focus on their ability to compensate for such deficits through cross-modal reasoning.

Cross-Modal Information Recovery

As illustrated in Figure E.8, the proposed method demonstrates the ability to recover critical information missing from one modality by effectively utilizing cues from the remaining modalities. This behavior underscores the advantages of multi-modal learning, where complementary information sources contribute to resilience against missing or degraded input, thereby enabling more robust and complete inference.

Compensation for Missing Information

When input text is fragmented due to word dropout and no relevant data is available from other modalities, Figure E.9 illustrates how the proposed method utilizes the knowledge embedded in LLMs. The model reconstructs missing words

> **Text with noise**
> The type of pet is Dog. The name of pet is Filo. The age of pet when listed (in months) is 78. The color 1 of pet is Brown. The fur length of pet is Short. The primary breed of pet is Mixed Breed. The secondary breed of pet (if pet is of mixed breed) is Unknown.
> **Summarization**
> This pet is a gentle and handsome 6-year-old mixed breed dog named Filo. He has a light brown, short coat and soft brown eyes.

Figure E.8: Cross-modal information recovery from dropped columns. When tabular color and fur length data (gray) contains missing information (red), the alignment module successfully reconstructs this information (blue) by leveraging visual data from the input image.

and enhances text coherence by drawing on contextual understanding and reasoning capabilities, allowing it to infer and explicitly articulate meanings that were merely implicit in the original input.

> **Text with noise**
> Jack experienced it Jack dog even is He recovering but special care skin which hair to Vet this was to Hopefully lover who take good dog a
> **Summarization**
> Although the profile write-up for Jack Jack is somewhat fragmented, it indicates that he has experienced some challenges but is on the road to recovery
> **Reasoning**
> 3. **Health Condition**: Jack Jack has a minor injury and is recovering. Potential adopters might be hesitant to take on a pet that requires special care, even if the injury is minor.

Figure E.9: Noise compensation in textual data. The alignment module transforms highly fragmented text input (red) into coherent descriptions (blue) by inferring semantic meaning and reconstructing the narrative flow despite substantial information loss.

#### E.0.0.1 Perturbation Mismatch: Training vs. Inference-Time Perturbations

An investigation was conducted into how models trained under different perturbation regimes perform when confronted with potentially mismatched attack types during inference. Table E.5 presents a detailed analysis of model performance when trained on one type of adversarial input but tested on another. The experiments compared four training configurations: (1) no perturbation, (2) random perturbations (including paraphrasing and token editing), (3) adversarial prompting (LLM-based rewriting, noise injection, and order permutation), and (4) both perturbation types

| Training | Evaluation | Precision | Deviation |
|---|---|---|---|
| No Perturbation | Combined Attack | 0.383 | 0.014 |
| Random Augmentation | Random Aug. | 0.386 | 0.011 |
| | Adversarial | 0.384 | 0.013 |
| | Combined | 0.384 | 0.013 |
| Adversarial Strategy | Random Aug. | <u>0.391</u> | <u>0.006</u> |
| | Adversarial | 0.387 | 0.010 |
| | Combined | 0.386 | 0.011 |
| Hybrid Approach | Combined Attack | 0.397 | 0.000 |

Table E.5: Cross-strategy generalization effectiveness under different training-evaluation configurations. The hybrid approach yields optimal results with perfect retention of performance, while adversarial strategies demonstrate superior generalization to random augmentation techniques (underlined values) compared to the inverse transfer scenario.

combined.

The results demonstrate that training on any single adversarial type yields performance improvements even for previously unseen attacks, with the highest accuracy (0.397) and zero performance drop achieved when training incorporates both perturbation types. Notably, models trained with adversarial prompting show stronger generalization (0.391 accuracy when tested on random perturbations) than vice versa, confirming that adversarial prompting represents a more challenging and comprehensive attack surface than random perturbations.

Corruption of Input Modalities before Text Transformation

The resilience of the proposed approach was examined when input modalities are corrupted before undergoing text transformation. To test this, significant corruption was introduced across different input channels: heavy Gaussian noise for images, randomly dropped columns for tabular data, and random token insertions for text.

| Corruption Condition | Accuracy | Drop from Baseline |
|---|---|---|
| No Corruption (All Modalities) | 0.397 | 0.000 |
| Image Corrupted | 0.385 | 0.012 |
| Text Corrupted | 0.392 | 0.005 |
| Tabular Corrupted | 0.394 | 0.003 |
| All Modalities Corrupted | 0.327 | 0.070 |
| All Modalities Corrupted w/o adv prompting | 0.293 | 0.104 |

Table E.6: Performance impact of modality corruption at different levels. Single modality corruption causes minimal degradation (0.003-0.012 drop), while simultaneous corruption of all modalities results in significant performance reduction (0.070 drop), which is partially mitigated by adversarial prompting.

Table E.6 presents performance metrics under various corruption conditions.

The results demonstrate two key findings. First, when only a single modality is corrupted (while others remain intact), the model maintains approximately 95% of its original performance. This minimal degradation (drops of 0.012, 0.005, and 0.003 for image, text, and tabular corruptions respectively) indicates robust recovery capabilities leveraging uncorrupted modalities. Second, when all modalities are simultaneously corrupted, performance decreases substantially (0.070 reduction in accuracy), though the model still performs better than random guessing. Importantly, adversarial prompting provides approximately 4% additional resistance against complete corruption compared to models without adversarial training.

Qualitative analysis reveals that this "recovery" mechanism is primarily driven by the model's text-centric alignment process. When certain modalities are corrupted, the textual summaries from uncorrupted modalities effectively compensate for missing or garbled content. The multimodal summarization stage integrates information across modalities, allowing intact information to address knowledge gaps. This cross-modal compensation highlights a significant advantage of text-centric transformations: they create a unified representation space where information from one modality can supplement deficiencies in another, providing inherent robustness against partial corruption.

# Appendix F — Multimodal Robustness Qualitative Examples

This appendix presents additional qualitative examples that demonstrate our method's capabilities in handling incomplete or corrupted information across different modalities. We provide detailed instances showing how the language model (LLM) recovers missing data and compensates for information loss by leveraging cross-modal knowledge.

## F.1  Recovery Across Modalities

Figure F.10 illustrates how our method recovers tabular information using image data. In this example, the LLM processes a pet adoption profile with corrupted tabular data but manages to reconstruct the missing information by analyzing the visual content.

The transformation shows how the LLM compensated for missing information in the pet profile for Filo, a 78-month-old mixed-breed dog. Despite corruption in the tabular data regarding Filo's age, breed, and health status, the model accurately inferred these details by analyzing visual features in the image. The LLM identified

> After Summarization and Reasoning - table
>
> <span style="color:blue">This pet is a gentle and handsome 6-year-old mixed breed dog named Filo. He has a light brown, short coat and soft brown eyes,</span> giving him a calm and serene expression.
>
> ...
>
> Filo's profile features five photos, one of which shows him lying comfortably on a blue plastic crate inside a tiled room, likely part of his temporary home at the clinic. The clean and secure environment, along with his collar and identification tag, suggests that he is well-loved and cared for. If you can provide Filo with the forever home he deserves or need more information, please call the number provided. Filo's adoption rate is likely to be more than 3 months for several reasons:
>
> 1. **Age**: At 78 months (6.5 years old), Filo is significantly older than the other pets listed. Younger pets, especially puppies and kittens, generally have higher adoption rates because they are perceived as more adaptable and have a longer potential lifespan.
>
> 2. <span style="color:blue">**Health and Background**: While Filo has been treated for his external wounds, his background as a stray and his extended stay in a cage might raise concerns for potential adopters about his health and behavior. The write-up mentions his gratefulness and calm demeanor, but it does not provide detailed information about his current health status or behavior, which could be crucial for potential adopters.</span>
>
> 3. **Emotional Connection**: The profile write-up is heartfelt and provides a touching backstory, but it lacks the emotional appeal seen in the profiles of younger pets. The language used to describe Filo's situation is more factual and less engaging compared to descriptions of other pets, which emphasize their cuteness and playful nature.
>
> 4. **Visual Appeal**: Although there are 5 photos, the description of the image shows Filo in a resting position, which may not be as engaging as images of playful or interactive behavior. Potential adopters often respond more positively to images that show the pet's personality and energy.
>
> 5. **Competition**: Filo is competing with younger, more visually appealing pets that are often adopted faster. His profile needs to stand out more to attract potential adopters who are specifically looking for an older, more mature dog.
>
> 6. **Location and Accessibility**: The profile does not specify the exact location beyond being at a vet clinic, which might make it less accessible for potential adopters who prefer to know more about where the pet is currently staying.

Figure F.10: Cross-modal information recovery for tabular data. Gray areas indicate dropped columns that were reconstructed using visual information from accompanying images.

Filo's approximate age based on physical characteristics and recognized key breed traits consistent with a mixed-breed dog with terrier features.

## F.2 Knowledge-Based Compensation for Missing Information

Figure F.12 demonstrates how the LLM reconstructs meaning from fragmented text data. In this example, portions of a pet description were deliberately corrupted, but the model successfully restored the narrative by applying domain knowledge and contextual understanding.

---
Text with noise
The type of pet is Dog. The name of pet is Filo. The age of pet when listed (in months) is 78. The color 1 of pet is Brown. The fur length of pet is Short. The primary breed of pet is Mixed Breed. The secondary breed of pet (if pet is of mixed breed) is Unknown.
Summarization
This pet is a gentle and handsome 6-year-old mixed breed dog named Filo. He has a light brown, short coat and soft brown eyes.

---

Figure F.11: The tabular data has dropped the color and fur length column (gray). However, it was recovered (blue) after applying alignment module with LLM that compensate the information from input image.

---
Text with noise
Jack experienced it Jack dog even is He recovering but special care skin which hair to Vet this was to Hopefully lover who take good dog a
Summarization
Although the profile write-up for Jack Jack is somewhat fragmented, it indicates that he has experienced some challenges but is on the road to recovery
Reasoning
3. **Health Condition**: Jack Jack has a minor injury and is recovering. Potential adopters might be hesitant to take on a pet that requires special care, even if the injury is minor.

---

Figure F.12: Compensation for noisy text input through language model reasoning. Despite severely degraded text (red), the alignment module successfully transforms fragmented content into coherent descriptions (blue) with meaningful inferences (cyan).

This example demonstrates the reconstruction of a fragmented profile for Jack Jack, a 36-month-old domestic shorthair cat. Despite significant corruption in the textual description of Jack Jack's personality and behavior, the LLM was able to infer likely characteristics based on the remaining context and general knowledge about cat behavior patterns. The reconstructed narrative maintains consistency with typical domestic shorthair traits while preserving the emotional appeal that is crucial for adoption profiles.

## F.3 Factors Influencing Adoption Outcomes

Our analysis of these and similar examples reveals several key factors that influence the successful adoption outcomes:

- Age representation: Both profiles (Filo at 78 months and Jack Jack at 36 months) required appropriate age representation that balanced precision with emotional appeal.

- Health condition clarity: The LLM correctly emphasized manageable health conditions for both pets, presenting them in a way that is informative without deterring potential adopters.

- Personality portrayal: The reconstructed profiles maintained consistent personality traits that matched visual cues and partial textual information.

- Breed-specific characteristics: For Filo, the LLM accurately identified mixed-breed characteristics, while for Jack Jack, domestic shorthair traits were appropriately emphasized.

- Emotional resonance: Both reconstructed profiles preserved an emotionally appealing narrative that connects with potential adopters despite the information loss.

These examples demonstrate our method's robust capability to maintain meaningful and accurate information even when significant portions of multimodal data are corrupted or missing.

# Appendix G — Prompt Templates and Additional Analysis for Optimization and Uncertainty

## G.1 Question Prompt Templates

> You will receive a question and your goal is to generate a new version of it that convey the same meaning as the original.
> Q1.Original Question: Would a dog respond to bell before Grey seal?
> New-Version: Would a dog react to a bell sooner than a grey seal?
> Q2.Original Question: The perimeter of a rectangle is the sum of all its sides.
> New-Version: A rectangle's perimeter is obtained by summing the lengths of its sides.
> Q3. Original Question: <Question>
> New-Version:

Figure G.13: Example prompt for question perturbation.

> Q: Was ethanol beneficial to Jack Kerouac's health?
> A: Jack Kerouac died from internal bleeding due to long-term alcohol abuse. Thus, ethanol was not beneficial to Jack Kerouac's health. So the answer is no.
> Q: If Goofy were a pet, would he need heartworm prevention?
> A: Goofy is a dog, and dogs require regular heartworm prevention. Thus, if Goofy were a pet, he would need heartworm prevention. So the answer is yes.
> Q : <Question>
> A :

Figure G.14: Example prompt for StrategyQA with few-shot samples.

> Question 1: Mark has a garden with flowers. He planted plants of three different colors in it. Ten of them are yellow, and there are 80% more of those in purple. There are only 25% as many green flowers as there are yellow and purple flowers. How many flowers does Mark have in his garden?
> Answer: There are 80% more purple flowers than yellow flowers, so there are 10 * 1.8 = 18 purple flowers. There are 10 yellow flowers and 18 purple flowers, so there are 10 + 18 = 28 yellow and purple flowers. There are 25% as many green flowers as there are yellow and purple flowers, so there are 28 * 0.25 = 7 green flowers. Mark has 10 yellow flowers, 18 purple flowers, and 7 green flowers, so he has 10 + 18 + 7 = 35 flowers in his garden. The answer to the question is 35.
> Question 2: Albert is wondering how much pizza he can eat in one day. He buys 2 large pizzas and 2 small pizzas. A large pizza has 16 slices and a small pizza has 8 slices. If he eats it all, how many pieces does he eat that day?
> Answer: He buys 2 large pizzas, so he has 2 * 16 = 32 slices. He buys 2 small pizzas, so he has 2 * 8 = 16 slices. There are 32 slices from the large pizzas and 16 slices from the small pizzas, so he eats 32 + 16 = 48 pieces that day. The answer to the question is 48.
> Question 3: <Question>
> Answer:

Figure G.15: Example prompt for GSM8K with few-shot samples.

> User Prompt:
> Q:
> Given the data points (y1, y2, ...) = {data}, what is the MSE loss function with respect to the ŷs for a hypothetical set of predicted ŷs values?
> A:
> The MSE loss function for the given data points (y1, y2, ...) = {data} with respect to ŷs is:...

Figure G.16: Example prompt for getting objective function.

## G.2  Optimization Task Prompts

## G.3  All Model Dataset Pairs Uncertainty Metrics Scatter Plots

> User Prompt:
> Q:
> Please minimize the loss function using gradient descent with learning rate of 0.1 at point $(\hat{y}1, \hat{y}2, \hat{y}3, .....) = \{point\}$. What is the point we eventually end up after one update? Your answer includes two parts an explanation with calculation and a short answer of result.
> A:
> Explanation: Let's think step by step ...
> Short Answer: After calculation, the next update point is $(\hat{y}1_{new}, \hat{y}2_{new}, \hat{y}3_{new}, .....) = ...$

Figure G.17: Example prompt for Gradient-Descent.

> User Prompt:
> Q:
> I want to do grid search on the ŷs and the range of them are the integers of $\{low\_bound\}$ to $\{high\_bound\}$. Generate all possible combinations of ŷs values from the specify range. What are the combinations? Your answer includes two parts an explanation with calculation and a list containing all the combinations.
> A:
> Explanation: Let's think step by step ...
> List: [write all the combinations here]

Figure G.18: Example prompt for Grid-Search (Create Grid Points)

> User Prompt:
> Q:
> You want to minimize an unknown MSE loss function by guessing the values of the ŷs. When you guess, you should take consider of the past guessing result so that your new guess will have smaller loss than the past results. Pass guessing result are $\{pass\_result\}$. Base on the previous guesses, what is your next guess?
> A:
> $(\hat{y}1, \hat{y}2,....) = $ [your answer]

Figure G.19: Example prompt for Black-Box Optimization

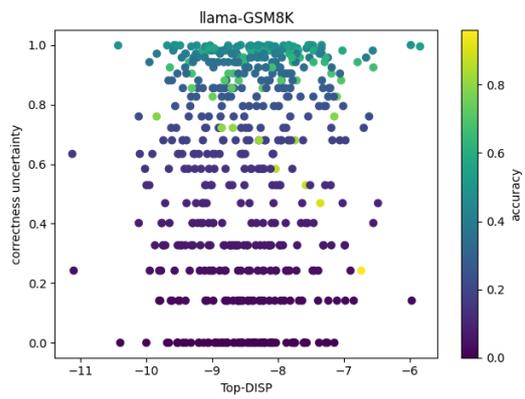
(a) Top-DISP metric for Llama

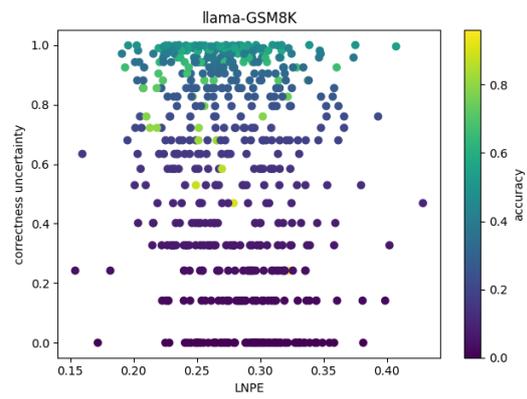
(b) LNPE metric for Llama

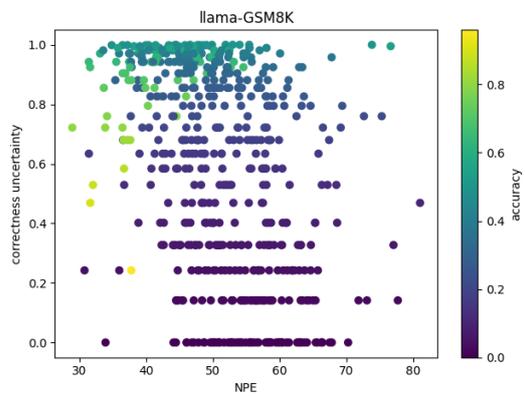
(c) NPE metric for Llama

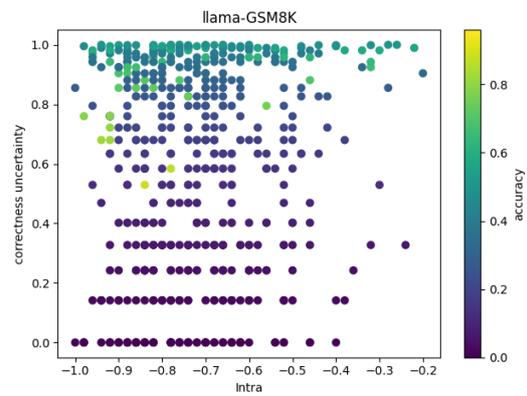
(d) Intra-Sample Similarity for Llama

Figure G.20: Uncertainty metrics visualization for Llama model on StrategyQA dataset. Each plot shows a different uncertainty metric (Top-DISP, LNPE, NPE, and Intra-Sample Similarity) and its relationship with model accuracy, revealing patterns in uncertainty distribution.

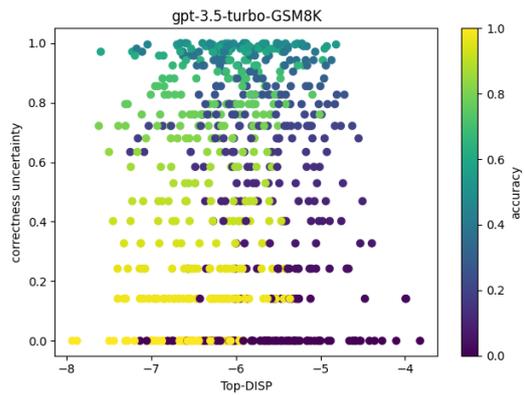

(a) Top-DISP metric for GPT-3.5-Turbo

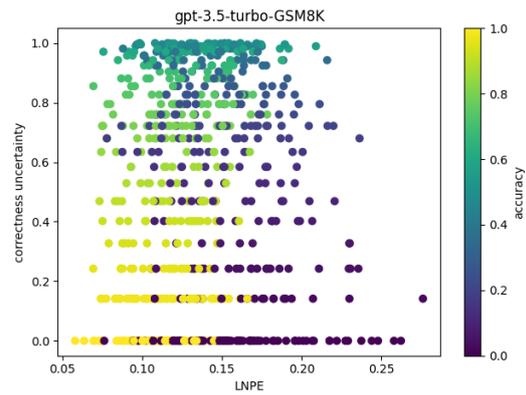

(b) LNPE metric for GPT-3.5-Turbo

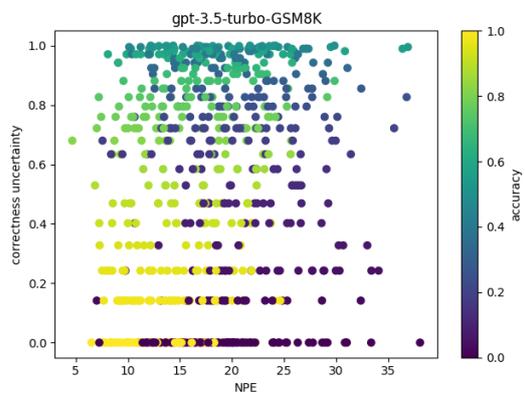

(c) NPE metric for GPT-3.5-Turbo

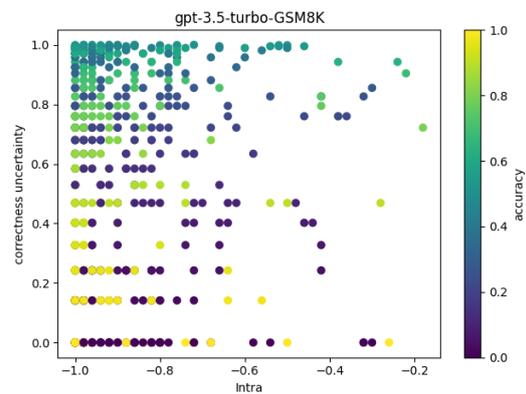

(d) Intra-Sample Similarity for GPT-3.5-Turbo

Figure G.21: Uncertainty metrics analysis for GPT-3.5-Turbo on StrategyQA dataset. The four plots display different metrics (Top-DISP, LNPE, NPE, and Intra-Sample Similarity) and their correlation with model accuracy, providing insights into uncertainty characteristics.

# Appendix H — RTLFixer Appendix

---

System Prompt:
Implement the Verilog module based on the following description. Assume that signals are positive clock/clk edge triggered unless otherwise stated.

Problem Description:
Implement the Verilog module based on the following description. Assume that signals are positive clock/clk edge triggered unless otherwise stated.
Given an 8-bit input vector [7:0], reverse its bit ordering.

module top_module (
input [7:0] in,
output [7:0] out
);

Erroneous Implementation:

assign {out[0],out[1],out[2],out[3],out[4],out[5],out[6],out[8]} = in;
endmodule

Feedback:
main.v:5: error: Index out[8] is out of range.
1 error(s) during elaboration.

---

Figure H.22: Template structure for one-shot prompting with compiler feedback for error resolution

---

Solve a task with interleaving Thought, Action, Observation steps.
Thought can reason about the current situation, and Action can be the following types:
(1) Compiler[code], which compiles the input code and provide error message if there is syntax error.
(2) Finish[answer], which returns the answer and finished the task.
(3) RAG[logs], input the compiler log and retrieve expert solutions to fix the syntax error.

---

Figure H.23: System prompt for ReAct instruction demonstrating the framework for iterative error resolution

> Question:
> What is the syntax error in the given Verilog module implementation and how to fix it?
>
> Thought 1:
> Implement a finite state machine that searches for message boundaries in a byte stream.
> Action 1: Compiler
>
> always @ (state or in) begin
> case (state)
>  4'b0001: begin
>
> Observation 1:
> main.v:15: error: out is not a valid l-value in top_module.
> Thought 2:
> The code has a syntax error because the "out" signal was declared as a wire instead of a register. I need to correct this and run the compilation.
> Action 2: RAG[..not a valid l-value..]
> Use assign statements instead of always block if possible.
>                                ...
> Action n: Finish
> The revised code is as follow: ...

Figure H.24: Illustration of ReAct's thought-action-observation cycle, demonstrating the step-by-step reasoning process for error resolution

Figure H.25: The ReAct methodology prompt framework: (a) one-shot prompting with compiler feedback, (b)-(c) autonomous agent reasoning and action planning for systematic syntax error resolution

> Compiler Logs:
> Object 'clk' is not declared. Verify the object name is correct. If the name is correct, declare the object.
> Human Expert Guidance:
> Check if 'clk' is an input. If not, and if 'clk' is used within the module, make sure the name is correct. If it's meant to trigger an 'always' block, replace 'posedge clk' with '*'.
>
> Compiler Logs:
> Index cannot fall outside the declared range for vector
> Human Expert Guidance:
> Carefully examine the index values to prevent encountering 'index out of bound' errors in your code. When utilizing parameters for indexing, try to use binary strings for performing the indexing operation instead.

Figure H.26: Representative examples of error categories with corresponding human expert guidance in the retrieval database, demonstrating the knowledge enhancement through RAG

| Task id | Columns | Content |
|---|---|---|
| popcount255 | Problem description | A "population count" circuit counts the number of '1's in an input vector. Build a population count circuit for a 255-bit input vector. |
| | Erroneous implementation | always @(posedge clk) begin<br>reg [7:0] count;<br>count = 0;<br>for (int i = 0; i < 255; i = i + 1) begin<br>if (in[i] == 1'b1) begin<br>count = count + 1;<br>end<br>end<br>out <= count;<br>end<br>endmodule |
| | Compiler log | main.v:1: error: Variable declaration in unnamed block requires SystemVerilog.<br>main.v:9: syntax error<br>main.v:9: error: Incomprehensible for loop. |
| Task id | Columns | Content |
| vector100r | Problem description | Given a 100-bit input vector [99:0], reverse its bit ordering. |
| | Erroneous implementation | always @(posedge clk) begin<br>for (int i = 0; i < 100; i = i + 1) begin<br>out[i] <= in[99 - i];<br>end<br>end<br>endmodule |
| | Compiler log | main.v:7: syntax error<br>main.v:7: error: Incomprehensible for loop. |

Figure H.27: Analysis of persistent error categories that challenge LLM-based resolution and the corresponding human guidance strategies developed to address them

```
Task ID: vector100r

Erroneous Implementation
1  module top_module (
2    input [99:0] in,
3    output reg [99:0] out
4  );
5  always @(posedge clk) begin
6    for (int i = 0; i < 100; i = i +1) begin
7      out[i] <= in[99 - i];
8    end
9  end
10 endmodule
```

```
iverilog:
vector100r.sv:5: error: Unable to bind wire/reg/memory 'clk' in 'top_module'
vector100r.sv:5: error: Failed to evaluate event expression 'posedge clk'.
2 error(s) during elaboration.
```

```
Quartus:
Error (10161): Verilog HDL error at vector100r.sv(5): object "clk" is not declared. Verify the object name is correct. If the name is correct, declare the object. File: /tmp/tmp4u6ib9ig/vector100r.sv Line: 5 Error: Quartus Prime Analysis & Synthesis was unsuccessful. 1 error, 1 warning
```

Figure H.28: Comparison of compiler feedback quality between iverilog and Quartus, illustrating the enhanced error diagnosis capabilities of commercial HDL tools

| Task id | Implementation | compiler error message |
|---------|----------------|------------------------|
| vector100r | 1 module top_module (<br>2 input [99:0] in,<br>3 output reg [99:0] out<br>4 );<br>5 always @(posedge clk) begin<br>6 for (int i = 0; i < 100; i = i + 1) begin<br>7 out[i] <= in[99 - i];<br>8 end<br>9 end<br>10 endmodule | **iverilog**<br>vector100r.sv:5: error: Unable to bind wire/reg/memory `clk' in `top_module'<br>vector100r.sv:5: error: Failed to evaluate event expression 'posedge clk'.<br>2 error(s) during elaboration.<br><br>**ModelSim**<br>** Error: vector100r.sv(5): (vlog-2730) Undefined variable: 'clk'.<br><br>**Quartus**<br>Error (10161): Verilog HDL error at vector100r.sv(5): object "clk" is not declared. Verify the object name is correct. If the name is correct, declare the object. File: /tmp/tmp4u6ib9ig/vector100r.sv Line: 5<br>Error: Quartus Prime Analysis & Synthesis was unsuccessful. 1 error, 1 warning |

```
Erroneous Implementation (Partial):

for (i = 0; i < 16; i = i + 1) begin : ROW
    for (j = 0; j < 16; j = j + 1) begin : COLUMN
        neighbors[0] = q[(i-1)*16 + (j-1)];
        row_above = q[((i-1) & 15)*16 + j];
        ...

Compile Error:
Error (10232): Verilog HDL error at conwaylife.sv(23): index -17 cannot fall outside the declared range [255:0] for vector "q"
```

Figure H.29: Case study of a syntax error that challenged the AI agent's capabilities, demonstrating limitations in array index calculation and boundary recognition

# Appendix I — CraftRTL Details

## I.1 Detailed Results

### I.1.1 CraftRTL Models

Results for the models on Verilog benchmarks are presented with temperatures 0.2 and 0.8. Ablation studies are conducted across different data blends, with SDG indicating the use of LLM synthetic generated data in 5.3.2, CC indicating correct-by-construction data targeting non-textual representations in 5.3.5, and Repair representing the targeted code repair dataset in 5.3.6.

Results for RTLLM use the open-source Icarus Verilog simulator[1] to check syntax and functional pass rates. This might lead to lower pass rate scores compared to previous work that used Synopsys VCS, as Icarus Verilog does not support all syntax.

---

[1] https://github.com/steveicarus/iverilog

Table I.7: Results for our models, across different dataset and temperature on VerilogEval.

| Model | Dataset | Temperature | VerilogEval [71] | | | | | |
|---|---|---|---|---|---|---|---|---|
| | | | Machine (%) | | | Human (%) | | |
| | | | pass@1 | pass@5 | pass@10 | pass@1 | pass@5 | pass@10 |
| Starcoder2-15b | SDG | 0.2 | 75.2 | 79.2 | 80.1 | 54.7 | 60.1 | 61.2 |
| | | 0.8 | 73.7 | 84.0 | 86.1 | 47.4 | 61.9 | 64.8 |
| | SDG-CC | 0.2 | 73.9 | 78.1 | 79.5 | 62.0 | 65.6 | 67.0 |
| | | 0.8 | 72.9 | 84.1 | 87.1 | 58.5 | 70.3 | 73.7 |
| | SDG-CC-Repair | 0.2 | 81.9 | 84.2 | 85.0 | 68.0 | 71.7 | 72.0 |
| | | 0.8 | 78.1 | 86.9 | 88.1 | 64.1 | 72.4 | 74.6 |
| DeepSeek-6.7b-Instruct | SDG | 0.2 | 73.4 | 77.8 | 78.9 | 48.3 | 53.2 | 54.5 |
| | | 0.8 | 71.4 | 82.5 | 85.4 | 44.0 | 58.1 | 62.3 |
| | SDG-CC | 0.2 | 72.6 | 78.2 | 79.3 | 58.5 | 62.6 | 63.5 |
| | | 0.8 | 70.2 | 83.1 | 85.4 | 56.3 | 67.0 | 70.7 |
| | SDG-CC-Repair | 0.2 | 77.8 | 82.7 | 83.4 | 65.4 | 67.7 | 68.2 |
| | | 0.8 | 75.2 | 85.5 | 88.1 | 61.6 | 70.0 | 72.1 |
| CodeLlama-7b-Instruct | SDG | 0.2 | 74.5 | 77.9 | 78.8 | 45.3 | 50.3 | 51.5 |
| | | 0.8 | 71.2 | 82.6 | 85.1 | 42.6 | 55.6 | 59.0 |
| | SDG-CC | 0.2 | 74.2 | 77.4 | 78.1 | 55.1 | 61.0 | 62.4 |
| | | 0.8 | 70.0 | 81.2 | 83.7 | 51.6 | 64.4 | 67.7 |
| | SDG-CC-Repair | 0.2 | 78.1 | 81.5 | 81.7 | 63.1 | 66.2 | 66.8 |
| | | 0.8 | 73.7 | 85.5 | 87.8 | 58.1 | 67.8 | 69.7 |

Table I.8: Results for our models, across different dataset and temperature on RTLLM.

| Model | Dataset | Temperature | RTLLM v1.1 [80] | | | | | |
|---|---|---|---|---|---|---|---|---|
| | | | Syntax (%) | | | Func. (%) | | |
| | | | pass@1 | pass@5 | pass@10 | pass@1 | pass@5 | pass@10 |
| Starcoder2-15b | SDG | 0.2 | 78.1 | 86.5 | 90.1 | 49.0 | 60.4 | 66.3 |
| | | 0.8 | 77.1 | 89.0 | 94.1 | 43.8 | 62.1 | 68.0 |
| | SDG-CC | 0.2 | 78.3 | 89.3 | 92.7 | 45.5 | 58.3 | 62.0 |
| | | 0.8 | 76.9 | 92.6 | 95.5 | 38.4 | 62.8 | 70.4 |
| | SDG-CC-Repair | 0.2 | 79.8 | 87.9 | 90.5 | 49.0 | 59.1 | 62.6 |
| | | 0.8 | 79.3 | 93.9 | 96.2 | 45.3 | 65.8 | 74.5 |
| DeepSeek-6.7b-Instruct | SDG | 0.2 | 79.3 | 86.8 | 90.5 | 40.3 | 45.9 | 49.6 |
| | | 0.8 | 76.6 | 92.5 | 96.2 | 40.0 | 53.8 | 63.6 |
| | SDG-CC | 0.2 | 73.6 | 84.5 | 86.0 | 44.3 | 52.2 | 54.3 |
| | | 0.8 | 76.7 | 90.5 | 93.8 | 39.5 | 56.4 | 63.1 |
| | SDG-CC-Repair | 0.2 | 84.3 | 92.2 | 93.0 | 53.1 | 58.8 | 60.3 |
| | | 0.8 | 80.0 | 92.9 | 95.4 | 45.5 | 57.9 | 62.6 |
| CodeLlama-7b-Instruct | SDG | 0.2 | 74.0 | 82.5 | 86.8 | 30.0 | 33.9 | 35.8 |
| | | 0.8 | 70.9 | 89.1 | 94.5 | 34.0 | 47.2 | 52.8 |
| | SDG-CC | 0.2 | 75.0 | 90.2 | 94.6 | 39.7 | 44.4 | 47.2 |
| | | 0.8 | 76.4 | 93.9 | 96.3 | 35.5 | 47.6 | 52.7 |
| | SDG-CC-Repair | 0.2 | 85.7 | 93.9 | 94.8 | 42.6 | 49.4 | 51.2 |
| | | 0.8 | 80.3 | 93.9 | 94.8 | 36.9 | 52.9 | 58.2 |

### I.1.2 Foundational and Frontier Code Models

Detailed results on recent foundational and frontier code models are presented.

All models are re-evaluated on RTLLM using unbiased pass@k metric.

Table I.9: Results on foundational and code models on VerilogEval.

| Type | Model | Size | Temp | VerilogEval [71] | | | | | |
|---|---|---|---|---|---|---|---|---|---|
| | | | | Machine (%) | | | Human (%) | | |
| | | | | pass@1 | pass@5 | pass@10 | pass@1 | pass@5 | pass@10 |
| Foundational Models | Llama-3.1 | 8B | 0.2 | 48.7 | 66.2 | 70.6 | 26.9 | 36.9 | 40.4 |
| | | | 0.8 | 42.1 | 67.3 | 74.1 | 23.0 | 37.8 | 44.2 |
| | Llama-3.1 | 70B | 0.2 | 66.7 | 73.8 | 76.9 | 48.7 | 53.6 | 55.1 |
| | | | 0.8 | 64.5 | 77.7 | 80.4 | 48.0 | 57.0 | 60.9 |
| | Llama-3.1 | 405B | 0.2 | 67.3 | 72.8 | 74.1 | 51.9 | 57.0 | 58.9 |
| | | | 0.8 | 66.4 | 75.1 | 76.9 | 53.8 | 61.0 | 62.8 |
| | Nemotron-4 | 340B | 0.2 | 53.0 | 59.1 | 61.5 | 43.1 | 43.9 | 44.9 |
| | | | 0.8 | 50.8 | 60.3 | 62.2 | 40.8 | 48.3 | 50.0 |
| | GPT-3.5-turbo | - | 0.2 | 58.0 | 66.4 | 68.5 | 31.2 | 39.4 | 41.7 |
| | | | 0.8 | 56.6 | 74.0 | 77.6 | 28.9 | 44.1 | 47.4 |
| | GPT-4 | - | 0.2 | 53.2 | 63.7 | 66.4 | 36.1 | 43.5 | 46.2 |
| | | | 0.8 | 35.3 | 53.4 | 58.9 | 35.2 | 53.4 | 58.9 |
| | GPT-4-turbo | - | 0.2 | 57.8 | 66.7 | 70.6 | 54.1 | 61.2 | 62.8 |
| | | | 0.8 | 56.9 | 69.5 | 73.4 | 53.6 | 63.6 | 66.7 |
| | GPT-4o | - | 0.2 | 65.9 | 68.9 | 69.2 | 57.1 | 61.3 | 62.2 |
| | | | 0.8 | 62.9 | 71.4 | 72.7 | 55.4 | 63.9 | 66.7 |
| Code Models | Starcoder2 | 15B | 0.2 | 68.7 | 76.7 | 78.6 | 37.7 | 48.3 | 51.1 |
| | | | 0.8 | 57.7 | 82.3 | 88.5 | 29.1 | 50.6 | 57.2 |
| | DeepSeek-Coder-V2 | 16B | 0.2 | 67.4 | 74.6 | 76.2 | 46.9 | 53.3 | 54.5 |
| | | | 0.8 | 65.6 | 78.3 | 81.8 | 46.3 | 55.9 | 58.9 |
| | DeepSeek-Coder-V2 | 236B | 0.2 | 68.2 | 72.7 | 75.0 | 56.4 | 60.7 | 64.3 |
| | | | 0.8 | 66.5 | 74.1 | 76.2 | 54.8 | 62.2 | 66.0 |

Table I.10: Results on foundational and code models on RTLLM.

| Type | Model | Size | Temp | RTLLM v1.1 [80] | | | | | |
|---|---|---|---|---|---|---|---|---|---|
| | | | | Syntax (%) | | | Func. (%) | | |
| | | | | pass@1 | pass@5 | pass@10 | pass@1 | pass@5 | pass@10 |
| Foundational Models | Llama-3.1 | 8B | 0.2 | 39.7 | 53.1 | 55.2 | 19.3 | 25.8 | 27.6 |
| | | | 0.8 | 40.7 | 60.6 | 65.5 | 17.6 | 34.7 | 37.9 |
| | Llama-3.1 | 70B | 0.2 | 47.9 | 51.7 | 55.2 | 34.1 | 34.5 | 34.5 |
| | | | 0.8 | 48.9 | 57.6 | 58.6 | 29.6 | 31.0 | 31.0 |
| | Llama-3.1 | 405B | 0.2 | 56.5 | 63.9 | 65.5 | 38.9 | 45.0 | 48.3 |
| | | | 0.8 | 52.1 | 64.4 | 72.4 | 35.8 | 45.8 | 51.7 |
| | Nemotron-4 | 340B | 0.2 | 41.7 | 47.2 | 48.3 | 14.1 | 15.5 | 17.2 |
| | | | 0.8 | 41.7 | 46.3 | 48.3 | 18.9 | 20.7 | 20.7 |
| | GPT-3.5-turbo | - | 0.2 | 50.3 | 58.2 | 58.6 | 28.3 | 36.9 | 41.4 |
| | | | 0.8 | 48.2 | 61.2 | 65.5 | 24.1 | 36.9 | 41.4 |
| | GPT-4 | - | 0.2 | 49.3 | 65.9 | 68.9 | 30.0 | 44.4 | 48.3 |
| | | | 0.8 | 42.8 | 61.2 | 65.5 | 25.9 | 40.0 | 44.8 |
| | GPT-4-turbo | - | 0.2 | 38.9 | 44.8 | 48.3 | 27.2 | 35.1 | 37.9 |
| | | | 0.8 | 40.3 | 48.8 | 51.7 | 27.5 | 40.2 | 44.8 |
| | GPT-4o | - | 0.2 | 50.3 | 59.9 | 62.1 | 33.8 | 44.4 | 48.3 |
| | | | 0.8 | 47.5 | 63.2 | 66.7 | 31.3 | 44.1 | 48.3 |
| Code Models | CodeLlama | 7B | 0.2 | 46.6 | 62.6 | 68.9 | 17.9 | 29.9 | 34.5 |
| | | | 0.8 | 34.8 | 59.7 | 68.9 | 13.4 | 25.9 | 31.0 |
| | CodeQwen | 7B | 0.2 | 45.8 | 55.8 | 58.6 | 24.1 | 33.1 | 37.9 |
| | | | 0.8 | 45.5 | 65.7 | 72.4 | 22.4 | 34.0 | 37.9 |
| | Starcoder2 | 15B | 0.2 | 38.3 | 77.5 | 86.3 | 15.5 | 37.6 | 44.6 |
| | | | 0.8 | 31.6 | 81.0 | 94.7 | 11.0 | 34.2 | 45.7 |
| | DeepSeek-Coder | 6.7B | 0.2 | 51.4 | 62.6 | 65.5 | 23.1 | 26.8 | 27.6 |
| | | | 0.8 | 49.7 | 64.4 | 68.9 | 21.0 | 29.3 | 34.5 |
| | DeepSeek-Coder-V2 | 16B | 0.2 | 51.4 | 51.7 | 51.7 | 33.1 | 34.5 | 34.5 |
| | | | 0.8 | 51.4 | 57.8 | 58.6 | 30.0 | 37.1 | 37.9 |
| | DeepSeek-Coder-V2 | 236B | 0.2 | 63.4 | 73.0 | 79.3 | 34.5 | 44.9 | 52.9 |
| | | | 0.8 | 61.8 | 78.1 | 79.3 | 32.9 | 50.2 | 55.1 |

### I.1.3 Details on Evaluations

The prompt input is formatted as follows for VerilogEval, where the detail_description is the problem description (Machine or Human) and prompt field is the problem mod-ule header. Module headers are included to avoid confusion on the signals naming.

```
prompt = f"{task['detail_description'].strip()}\n\n{task['prompt'].
   strip()}"
```

An example of mux2to1 in VerilogEval-Human:

```
Create a 2-1 multiplexer. When sel=0, choose a. When sel=1, choose b.

module top_module (
        input a,
        input b,
        input sel,
        output out
);
```

Similar templates are used for RTLLM v1.1, where the top module header is extracted from the reference solution and provided as input. Below is an example of adder_8bit:

```
Please act as a professional verilog designer.

Implement a module of an 8-bit adder with multiple bit-level adders in
    combinational logic.

Module name:
    adder_8bit
Input ports:
    a[7:0]: 8-bit input operand A.
    b[7:0]: 8-bit input operand B.
    cin: Carry-in input.
Output ports:
    sum[7:0]: 8-bit output representing the sum of A and B.
    cout: Carry-out output.

Implementation:
The module utilizes a series of bit-level adders (full adders) to
```

```
    perform the addition operation.

Give me the complete code.

module adder_8bit(
    input [7:0] a, b,
    input cin,
    output [7:0] sum,
    output cout);
```

Default chat templates and default system prompts are used for open-source models tested. For GPT models from OpenAI, the following system prompt is used:

```
Please act as a professional verilog designer.
```

Model responses are post-processed to extract code. Content enclosed by triple backticks is extracted and the language identifier (Verilog) is removed. Code enclosed in module and endmodule keywords is then extracted with response.find('module ') and response.rfind('endmodule'). If the extracted code does not include a module header, the reference solution's module header is prepended. The code is then tested with the provided testbenches with the Icarus Verilog (iverilog) simulator to evaluate for syntax and functional correctness. This might lead to lower pass rate scores for RTLLM compared to previous work that used Synopsys VCS, as Icarus Verilog does not support all syntax.

### I.1.4 VerilogEval-NonText

The following 45 problems from VerilogEval-Human are selected as they consist of non-textual representations in their problem descriptions:

2012_q1g, 2012_q2b, 2012_q2fsm, 2013_q2afsm, 2014_q3bfsm, 2014_q3c, always_nolatches, circuit1, circuit10, circuit2, circuit3, circuit4, circuit5, circuit6, circuit7, circuit8, circuit9, ece241_2013_q7, ece241_2014_q3, ece241_2014_q5b, fsm1, fsm1s, fsm2, fsm2s, fsm3, fsm3comb, fsm3onehot, fsm3s, fsm_onehot, fsm_ps2data, kmap1, kmap2, kmap3, kmap4, m2014_q3, m2014_q6, m2014_q6b, m2014_q6c, mt2015_q4, mt2015_q4a, mt2015_q4b, review2015_fsmonehot, rule110, rule90, truthtable1

## I.1.5 Template Problems for Correct-by-Construction Data

When generating correct-by-construction CC data, 11 problems from VerilogEval-NonText are selected to use as representative templates for constructing prompts. To prevent contamination, benchmark problems are excluded from the data. While the prompts resemble those of the selected problems, the non-textual representations and solutions differ. Additionally, to prevent overfitting to specific prompt templates, LLMs are used to rewrite the problem instructions for 20% of the data. Furthermore, validation test problems that are strictly in-distribution are created, based on the chosen problems.

Karnaugh Maps and Truth Tables: kmap1, m2014_q3, truthtable1.

State Transition Graphs and Tables: 2012_q2b, 2014_q3c, ece241_2014_q5b, fsm3comb, fsm3onehot, fsm_onehot, m2014_q6b, m2014_q6c.

Waveforms: No specific benchmark problems are used as a base for this data.

Table I.11: Performance impact of scaling repair data quantity. A carefully filtered 1.4k sample dataset achieves similar performance to a much larger 7.8k dataset, demonstrating that quality rather than quantity is the determining factor.

|                  | VerilogEval       |          | RTLLM v1.1      |
|                  | Machine  | Human  | Func            |
| Model            | pass@1 (%)         |          | pass@5 (%)      |
|---|---|---|---|
| SDG-CC           | 73.9     | 62.0   | 62.8            |
| SDG-CC-Repair 1k | 81.9     | 68.0   | 65.8            |
| SDG-CC-Repair 7k | 82.2     | 67.4   | 64.5            |

Table I.12: Performance comparison of iterative code repair approaches. Results show that a single iteration achieves most of the potential gains, with minimal additional improvement from a second iteration targeting approximately 2.7k repairs.

|                       | VerilogEval       |          | RTLLM v1.1      |
|                       | Machine  | Human  | Func            |
| Model                 | pass@1 (%)         |          | pass@5 (%)      |
|---|---|---|---|
| SDG-CC                | 73.9     | 62.0   | 62.8            |
| SDG-CC-Repair Iter 1  | 81.9     | 68.0   | 65.8            |
| SDG-CC-Repair Iter 2  | 81.3     | 68.1   | 65.6            |

### I.1.6  Scaling Repair Data

As shown in Table I.11, a carefully filtered dataset of 1.4k samples achieves comparable performance to a 7.8k dataset. This suggests that merely increasing the dataset size by injecting the same types of errors does not contribute meaningfully to improving model performance.

### I.1.7  Iterative Code Repair

A second iteration is conducted by generating 2.7k repair data for the model based on the Repair data from the first iteration. As shown in I.12, performance mostly saturates after this initial iteration. The remaining issues are likely due to significant errors that are challenging to correct.

## I.1.8 Diversity of Generated Code

The diversity of the code generated by the models is assessed using BLEU score, Jaccard similarity, and abstract tree edit distance (TSED) in [119]. The VerilogEval-Human problems are categorized into NonText and Text, as described in I.1.4. For each problem, the average code diversity score is computed across sampled codes for the same problem and the mean score for all problems is reported. For TSED, PyVerilog [124] is used to extract the abstract syntax tree, and codes that fail syntax checks are excluded from the analysis.

I.13 presents the results on code diversity. 20 solutions are sampled with temperature of 0.8 for each model. Fine-tuned models generally show a decrease in code diversity for both Text and NonText problems. This reduction is expected, as BLEU and Jaccard metrics account for both correct and incorrect code solutions, and there are often multiple ways to implement a correct solution. When comparing the fine-tuned models with GPT-4o, code diversity is similar for Text problems, but the models exhibit poor diversity for NonText problems. This is anticipated, given that the CC training dataset for NonText problems is generated using correct-by-construction methods and follows similar templates for Verilog code. However, the models demonstrate comparable diversity to GPT-4o for Text problems, particularly in TSED metric.

Table I.13: Diversity of generated code solutions on VerilogEval-Human sampled with temperature of 0.8. Lower scores indicate higher diversity.

| Type | Models | Text | | | NonText | | |
|---|---|---|---|---|---|---|---|
| | | Jaccard | BLEU | TSED | Jaccard | BLEU | TSED |
| Pretrained Models | CodeLlama | 0.5330 | 0.3808 | 0.4255 | 0.4707 | 0.2507 | 0.3521 |
| | DeepSeek-Coder | 0.6606 | 0.5454 | 0.5956 | 0.6548 | 0.3797 | 0.3847 |
| | Starcoder2 | 0.7724 | 0.5084 | 0.5520 | 0.7212 | 0.3607 | 0.4020 |
| | GPT-4o | 0.6798 | 0.6633 | 0.6906 | 0.7390 | 0.6376 | 0.6137 |
| Ours SDG-CC-Repair | CodeLlama | 0.6848 | 0.5992 | 0.6354 | 0.8583 | 0.7242 | 0.7158 |
| | DeepSeek-Coder | 0.6828 | 0.6040 | 0.6319 | 0.8308 | 0.6866 | 0.6598 |
| | Starcoder2 | 0.7018 | 0.6381 | 0.6721 | 0.8799 | 0.7750 | 0.7740 |

| Type | Models | VerilogEval-Human (Overall) | | |
|---|---|---|---|---|
| | | Jaccard | BLEU | TSED |
| Pretrained Models | CodeLlama | 0.5155 | 0.3441 | 0.4156 |
| | DeepSeek-Coder | 0.6590 | 0.4987 | 0.5505 |
| | Starcoder2 | 0.7580 | 0.4667 | 0.5198 |
| | GPT-4o | 0.6965 | 0.6561 | 0.6802 |
| Ours SDG-CC-Repair | CodeLlama | 0.7333 | 0.6345 | 0.6515 |
| | DeepSeek-Coder | 0.7246 | 0.6273 | 0.6379 |
| | Starcoder2 | 0.7512 | 0.6767 | 0.6942 |

## I.1.9 Error Types of LLM generated Error reports

Table I.14: Error types of LLM generated error reports.

| Error Type | #Errors | One-line Description |
|---|---|---|
| Vector Concatenation | 15.3% | Errors during vector concatenation or bit slicing. |
| Incorrect Initialization | 13.1% | Missing or faulty initialization of registers or signals. |
| Boolean Logic Flaws | 12.4% | Logical inconsistencies or errors in combinational logic expressions. |
| Shift Operation Faults | 10.2% | Misaligned or unintended behavior during shift operations. |
| Timing Violations | 10.2% | Errors where signal propagation violates timing requirements. |
| KMap Misinterpretation | 8.8% | Incorrect derivation of Boolean expressions from Karnaugh maps. |
| Latch Hazards | 6.5% | Unintended latches caused by missing or faulty conditions. |
| Bit Manipulation Bugs | 7.3% | Errors in operations like masking, flipping, or extracting specific bits. |
| Casez Priority Conflicts | 4.4% | Ambiguities or conflicts in casez or case statements. |
| Nested Loop Design Flaws | 3.7% | Incorrect or inefficient nested loop designs. |
| Others | 8.1% | Miscellaneous errors not covered above. |

Table I.15: Checkpoints of 5.5.

| Model | checkpoint1 | | checkpoint2 | |
|---|---|---|---|---|
| | Steps | Epoch | Steps | Epoch |
| SDG | 256 | 0.82 | 320 | 1.0 |
| SDG-CC-Repair | 386 | 0.86 | 448 | 1.0 |

I.1.9 shows the distribution of common error types in LLM-generated error reports, along with brief one-line descriptions. Most of these "minor" errors occur in solvable problems and stem from hardware-specific concepts (e.g., shift operations, timing violations) and Verilog related issues uncommon in software languages (e.g., latch hazards, casez priority conflicts). When generating targeted repair training data, detailed error reports and open-source code snippets are randomly sampled, ensuring the error type distribution in training aligns with their natural occurrences.

### I.1.10 Details on 5.5

In 5.3.4, findings on training variability in learning outcomes for specific benchmark problems are discussed. To analyze this, checkpoints were saved every 64 gradient steps during training and the pass rates of specific benchmarks were tracked. The training process was limited to a single epoch, as further training was found to be not helpful. Problems with pass rates exceeding 67

Table I.16: Pass rates for selected benchmark problems from VerilogEval-Human throughout the training progression. Each entry shows the pass rate for SDG-CC-Repair (SDG), with SDG results in parentheses.

| Problem | Step 64 | Step 128 | Step 256 | Step 320 | Step 386 | Step 448 |
|---|---|---|---|---|---|---|
| m2014_q4h | 1.0 (1.0) | 1.0 (0.9) | 1.0 (0.967) | 1.0 (0.875) | 1.0 (-) | 1.0 (-) |
| always_nolatches | 1.0 (0.867) | 1.0 (0.9) | 1.0 (0.6) | 1.0 (0.833) | 1.0 (-) | 1.0 (-) |
| vectorr | 1.0 (0.633) | 1.0 (0.925) | 1.0 (0.467) | 0.95 (0.925) | 1.0 (-) | 1.0 (-) |
| fsm2s | 1.0 (0.8) | 1.0 (0.8334) | 0.8 (0.775) | 1.0 (0.967) | 1.0 (-) | 1.0 (-) |
| fsm3comb | 1.0 (0.0) | 0.95 (1.0) | 0.5 (0.533) | 1.0 (0.233) | 1.0 (-) | 1.0 (-) |

Such volatility primarily is due to noise in SDG data where solution correctness cannot be verified. Because of the difficulties of verifying coding solutions in hardware descriptive languages, targeted repair data is instead generated for LLMs to learn to mitigate common errors which have shown to generalize to writing correct code during completion. To the best of current knowledge, this work is the first to describe such findings and provide an effective solution.

## I.2  Further Discussions and Broader Impacts

In this section, further discussions are provided to address concerns regarding the novelty, generalizability, and significance of the proposed methods. Clarifications are offered to highlight the relevance and broader impact of this work, underscoring its value to the broad research community.

### I.2.1  Generalizability of Correct-by-Construction Data Generation

The approach to curating correct-by-construction data is largely inspired by [130], who introduced a mathematically rigorous method utilizing symbolic deduction engines to construct synthetic training data, significantly improving LLM capabilities in solving Olympiad geometry problems. Similarly, this method ensures the correctness of problems and solutions through a custom-designed data generation pipeline, leveraging custom-designed solvers to generate accurate solutions to their corresponding problems. In contrast to methods distilling LLM responses like Self-Instruct [147], the correct-by-construction approach ensures data quality and solution accuracy without relying on strong LLM performance on downstream tasks.

This mathematically rigorous approach to generating synthetic data can further inspire future work on improving LLMs general capabilities in areas such as math, coding, and symbolic reasoning. Moreover, adapting these methods to other domains may require human tuning to identify the best data generation method, and automating this process for scalability could be a promising future research direction.

### I.2.2 Novelty and Generalizability of Targeted Code Repair

Analysis shows that LLMs frequently make "minor" errors in Verilog coding, often correctable within few lines of code. This is primarily attributed to the LLMs' insufficient training in comprehending problem descriptions and instructions alongside their correct solutions. Prior research has tackled this challenge by improving data quality. For instance, [17] filters incorrect code using tests generated by LLMs, while [173] creates preference learning datasets by ranking code through self-validation. [59] focus on generating fine-tuning data through code completion, test validation, and debugging with LLM agents, while [58] trained reward models based on compilation and unit test outcomes to enhance LLM performance via reinforcement learning. However, low-resource languages face additional obstacles due to limited data availability, making it particularly difficult to synthesize unit tests directly in these languages. To address this issue, [13] introduced lightweight compilers to translate test cases from source to target languages.

Verilog coding encounters challenges typical of low-resource languages, compounded by additional domain-specific challenges as a hardware description language rather than a conventional programming language. Its unique characteristics pose significant barriers to knowledge transfer from high-resource languages, as highlighted in studies on execution performance in parallel programming [95] and high-performance computing extensions [126]. To address these challenges, a novel pipeline for generating targeted code repair data is proposed. While automatic code repair has been extensively studied, most existing methods focus on widely-used programming languages [158], relying on data of buggy code and fixes from open-source repositories [50, 140]. In contrast, the pipeline presented here utilizes a small set of well-curated benchmarks and testbench to automate the generation of error reports, quality assurance, and augmentation of training datasets by injecting similar errors into open-source code. The results highlight the effectiveness of this approach, which is language agnostic and can be adapted to other low-resource and domain-specific programming languages.

### I.2.3 Significance of Non-Textual Data Representations in Hardware Design

In this work, the significance of non-textual data representations, specifically Karnaugh maps, state-transition diagrams, and waveforms, for accurately capturing hardware functionality is emphasized. These representations are widely utilized by hardware designers to mitigate the ambiguity and verbosity inherent in natural language descriptions. While they may be specific to hardware design, they are not limited to Verilog and can be applied to various domain-specific languages (DSLs) for hardware design. This is supported by [8], who leveraged similar non-textual representations from VerilogEval-Human to evaluate the performance of LLMs on several Python-embedded hardware design DSLs.

This study focuses exclusively on limited representations, which constitute a sig-nificant portion (45 problems, approximately 30%) of all problems in the VerilogEval-Human benchmark (details in I.1.4). Other types of non-textual representations are excluded due to the lack of a suitable benchmark for evaluating LLMs in Verilog coding. [15] emphasize the importance of non-textual representations, particularly visual representations, in describing hardware designs. While their work targets visual-language models and is therefore beyond the scope of this study, similar methodologies such as correct-by-construction methods could be employed to generate training data for visual representations, such as circuit schematics, data flow diagrams, and state transition graphs.

## I.3 Examples of Targeted Code Repair Data

> **Problem description: vector3 from VerilogEval-Machine**
>
> The top module contains 6 ports that are fed into a concatenation which outputs 4 ports. The first input "a" is fed into the first port in the concatenation, the second input "b" is fed into the second port of the concatenation, the third input "c" is fed into the third port of the concatenation, the fourth input "d" is fed into the fourth port of the concatenation, the fifth input "e" is fed into the fifth port of the concatenation, the sixth input "f" is fed into the sixth port of the concatenation. The last bit of the concatenation is 2'b11. This value is fed in to the last port of the concatenation.
>
> ```verilog
> module top_module (
>   input [4:0] a,
>   input [4:0] b,
>   input [4:0] c,
>   input [4:0] d,
>   input [4:0] e,
>   input [4:0] f,
>   output [7:0] w,
>   output [7:0] x,
>   output [7:0] y,
>   output [7:0] z
> );
> ```

```verilog
                                  // Concatenate the inputs with the
  assign {w, x, y, z} = {a, b, c,     last bit as 2'b11
    d, e, f, 2'b11};                assign {w, x, y, z} = {2'b11, a,
                                      b, c, d, e, f};

        endmodule
                                         endmodule
```

          Correct Code                                        Error Code

> **LLM generated Error Report**
>
> Error Type: Incorrect vector concatenation and splitting
>
> Category: Combinatorial: wiring
>
> Description: Steps to repair the erroneous implementation:
>
> 1. Concatenating the two 1 bits at the beginning of the concatenated vector.
>
> 2. Assign the output vectors from the concatenated vector in the correct order and bit ranges
>
> For example:
>
> ```verilog
> assign {w, x, y, z} = {2'b11, a, b, c, d, e, f};
> ```
>
> To correct the implementation, you should concatenate the last bit as '2'b11' on the right, as shown in the correct implementation:
>
> ```verilog
> assign {w, x, y, z} = {a, b, c, d, e, f, 2'b11};
> ```

> **LLM constructed Repair data from open-source code**
>
> Input:
>
> #### 1. Problem Description
>
> You are given a Verilog module that demonstrates the use of bit slicing and concatenation in a loop. The goal is to correctly display specific portions of a 32-bit register using a for-loop and explicit slicing. However, there is a commonly made error in the concatenation and bit slicing order that causes incorrect results to be displayed. Your task is to identify and fix the error in the code.

#### 2. Erroneous Implementation

```verilog
module block;
    reg [31:0]   data;
    int          i;
    initial begin
        data = 32'hFACE_CAFE;
        // Loop to display 8-bit segments of the data register
        for (i = 0; i < 4; i++) begin
            // Incorrectly reversing the bit order in the concatenation
            $display ("data[8*%0d +: 8] = 0x%0h", i, data[8*(3-i) +: 8]);
        end
        // Incorrect slicing order, placing higher bits in lower indexes
        $display ("data[7:0]   = 0x%0h", data[31:24]);
        $display ("data[15:8]  = 0x%0h", data[23:16]);
        $display ("data[23:16] = 0x%0h", data[15:8]);
        $display ("data[31:24] = 0x%0h", data[7:0]);
    end
endmodule
```

In this erroneous implementation:

- The loop erroneously reverses the order of the 8-bit slices.

- The explicit bit-slicing outside of the loop has the slices ordered incorrectly, assigning the MSB to the least significant part and vice versa.

#### 3. Hints for Fixing

1. **Correct the loop:**

The loop currently reverses the order of the 8-bit slices. Instead of accessing 'data[8*(3-i) +: 8]', you should access 'data[8*i +: 8]' to display the correct slices of the register.

2. **Fix the slicing outside of the loop:**

Ensure that the LSB ('data[7:0]') corresponds to the lower bits of the 'data' register, and the MSB ('data[31:24]') corresponds to the higher bits. Correct the slicing order for accurate display.

The corrected implementation should ensure that both the loop and explicit bit-slicing access the register correctly, without reversing the bit order.

Output:

```verilog
module block;
   reg [31:0]   data;
   int          i;
   initial begin
      data = 32'hFACE_CAFE;
      for (i = 0; i < 4; i++) begin
           $display ("data[8*%0d +: 8] = 0x%0h", i, data[8*i +: 8]);
      end
      $display ("data[7:0]   = 0x%0h", data[7:0]);
      $display ("data[15:8]  = 0x%0h", data[15:8]);
      $display ("data[23:16] = 0x%0h", data[23:16]);
      $display ("data[31:24] = 0x%0h", data[31:24]);
   end
endmodule
```

Figure I.30: An example demonstrating the process for targeted code repair. In this example, training checkpoints during training would have the model sometimes generated correct and error solutions. We use LLMs to first summarize the errors for a detailed Error Report and then inject the errors to open-source code to construct Repair data.

---

**Problem description: shift4 from VerilogEval-Human**

Build a 4-bit shift register (right shift), with asynchronous positive edge triggered areset, synchronous active high signals load, and enable.

// (1) areset: Resets shift register to zero.

// (2) load: Loads shift register with data[3:0] instead of shifting.

// (3) ena: Shift right (q[3] becomes zero, q[0] is shifted out and disappears).

// (4) q: The contents of the shift register. If both the load and ena inputs are asserted (1), the load input has higher priority.

```verilog
module top_module(
  input clk,
  input areset,
  input load,
  input ena,
  input [3:0] data,
  output reg [3:0] q);
```

```verilog
   always @(posedge clk or              always @(posedge clk or
posedge areset) begin              posedge areset) begin
    if (areset) begin                  if (areset) begin
        // Asynchronous                   // Asynchronous positive
positive edge triggered areset     edge triggered areset
        q <= 4'b0000;                      q <= 4'b0;
    end else if (load) begin           end else if (load) begin
        // Synchronous active              // Synchronous active high
high signals load                  signals load
        q <= data;                         q <= data;
    end else if (ena) begin            end else if (ena) begin
        // Shift right                     // Shift right
        q <= {1'b0, q[3:1]};               q <= {q[2:0], 1'b0};
    end                                end
  end                                end

endmodule                          endmodule
```

Correct Code                       Error Code

> **LLM generated Error Report**
>
> Error Type: shifting operation
>
> Category: Sequential: shift registers
>
> Description:
>
> The error in the erroneous implementation is related to the shifting operation in the shift register.
>
> ```
> q <= {q[2:0], 1'b0};
> ```
>
> This line is intended to shift the bits in the register to the right, but it incorrectly connects the least significant bit (q[0]) to the most significant bit (q[3]) instead of connecting a zero to the most significant bit. This results in a circular shift instead of a right shift.
>
> To correct this error, the line should be changed to:
>
> ```
> q <= {1'b0, q[3:1]};
> ```
>
> Here are the steps to repair the erroneous implementation
>
> 1. Identify the line of code that performs the shifting operation
>
> 2. Notice that the least significant bit (q[0]) is connected to the most significant bit (q[3]) instead of a zero.
>
> 3. Change the line of code to connect a zero to the most significant bit and shift the remaining bits to the right
>
> 4. Test the implementation to ensure that it behaves as expected.

## LLM constructed Repair data from open-source code

Input:

You are provided with a Verilog module that implements a shift register. However, the implementation contains a common error related to shifting operations. Your task is to identify the error, which involves a mistake in the way bits are shifted within the register. Specifically, the problem results in an incorrect circular shift rather than the intended right shift operation. You need to fix the error to ensure that the register shifts correctly.

Erroneous Implementation:

```verilog
module shift_registers_0 (clk, clken, SI, SO);
    parameter WIDTH = 32;
    input clk, clken, SI;
    output SO;
    reg [WIDTH-1:0] shreg;
    always @(posedge clk)
    begin
        if (clken)
            // Erroneous shifting operation: circular shift instead of right shift
            shreg = {shreg[WIDTH-2:0], shreg[WIDTH-1]};
    end
    assign SO = shreg[WIDTH-1];
endmodule
```

In this erroneous implementation:

- The shift register incorrectly shifts the most significant bit (MSB) into the least significant bit (LSB), creating a circular shift.

- The correct behavior should be shifting in the new input bit ('SI') while moving existing bits to the right.

Hints for Fixing:

1. **Identify the shifting operation error:**

The line 'shreg = {shreg[WIDTH-2:0], shreg[WIDTH-1]};' causes a circular shift because the MSB is incorrectly fed back into the LSB position.

2. **Fix the shifting logic:**

Replace the erroneous MSB assignment with the correct new input bit ('SI'). The correct line should be 'shreg = {shreg[WIDTH-2:0], SI};' to shift in the new input bit from the 'SI' input and ensure the bits shift to the right.

By fixing the shifting operation, the register will now shift correctly, with the new input bit filling the LSB and the existing bits shifting to the right.

Output:

```verilog
module shift_registers_0 (clk, clken, SI, SO);
    parameter WIDTH = 32;
    input clk, clken, SI;
    output SO;
    reg [WIDTH-1:0] shreg;
    always @(posedge clk)
    begin
    if (clken)
        shreg = {shreg[WIDTH-2:0], SI};
    end
    assign SO = shreg[WIDTH-1];
endmodule
```

Figure I.31: Another example based on shift4 from VerilogEval-Human.

---

**Problem description: multi_booth_8bit from RTLLM**

Please act as a professional verilog designer.

Implement an 8-bit Radix-4 booth multiplier that performs the multiplication of two 8-bit inputs (a and b) using the Booth algorithm. It utilizes a clock signal (clk), and a reset signal (reset), and provides the product output (p) and a ready signal (rdy). The ready signal (rdy) is set to 1 to indicate the completion of the multiplication process.

Module name:
    multi_booth_8bit

Input ports:
    clk: Clock signal used for synchronous operation.
    reset: Reset signal used to initialize the multiplier module.
    a: 8-bit input representing the multiplicand.
    b: 8-bit input representing the multiplier.

Output ports:
    p: 16-bit output representing the product of the multiplication.
    rdy: Ready signal indicating the completion of the multiplication operation.

Implementation:
    On the positive edge of the clock signal (clk) or the positive edge of the reset signal (reset), the module performs the multiplication process.

    If the reset signal (reset) is high, two 16-bit registers multiplier <= 8a[7], a and multiplicand <= 8b[7], b.

    If the reset signal (reset) is low, indicating normal operation, the module checks if the counter (5bit ctr) is less than 16.

    If the counter (ctr) is less than 16, the multiplicand register (multiplicand) is left-shifted by 1 to simulate the Booth algorithm's shifting operation.

    If the current bit of the multiplier register (multiplier[ctr]) is 1, indicating a positive Booth encoding, the product register (p) accumulates the value of the multiplicand register (multiplicand).

    The counter (ctr) is incremented by 1. Once the counter (ctr) reaches 16, indicating the completion of the multiplication process, the ready signal (rdy) is set to 1.

Give me the complete code.

```verilog
module multi_booth_8bit (p, rdy, clk,
    reset, a, b);
  input clk, reset;
  input [7:0] a, b;
  output reg [15:0] p;
  output reg rdy;
  reg [15:0] multiplicand, multiplier;
  reg [4:0] ctr;
  always @(posedge clk or posedge reset)
    begin
      if (reset) begin
        multiplier <= {{8{a[7]}}, a};
        multiplicand <= {{8{b[7]}}, b};
        p <= 0;
        ctr <= 0;
        rdy <= 0;
      end else begin
        if (ctr < 16) begin
          multiplicand <= multiplicand
   << 1;
          if (multiplier[ctr]) begin
            p <= p + multiplicand;
          end
          ctr <= ctr + 1;
        end else begin
          rdy <= 1;
        end
      end
    end
endmodule
```
Correct Code

```verilog
module multi_booth_8bit (p, rdy, clk,
    reset, a, b);
  input clk, reset;
  input [7:0] a, b;
  output reg [15:0] p;
  output reg rdy;
  reg [15:0] multiplier, multiplicand;
  reg [4:0] ctr;
  always @(posedge clk or posedge reset)
    begin
      if (reset) begin
        multiplier <= {{8{a[7]}}, a};
        multiplicand <= {{8{b[7]}}, b};
        ctr <= 0;
        rdy <= 0;
      end else begin
        if (ctr < 16) begin
          multiplicand <= multiplicand << 1;
          if (multiplier[ctr])
            p <= p + multiplicand;
          ctr <= ctr + 1;
        end else begin
          rdy <= 1;
        end
      end
    end
endmodule
```
Error Code

> **LLM generated Error Report**
>
> Error Type: Incorrect initialization
>
> Category: Sequential: counters
>
> Description:
>
> In the erroneous implementation, the p register is not initialized to zero during reset. This can lead to incorrect multiplication results, as the initial value of p is unknown and may contain random values. The correct implementation initializes the p register to zero during reset, ensuring that the multiplication result is accurate.
>
> To repair the erroneous implementation, follow these steps:
>
> 1. Locate the always block that handles the reset and clock edge. In this case, it's the block that starts with always @(posedge clk or posedge reset).
>
> 2. Within the if (reset) block, add a line to initialize the p register to zero: p <= 0;.

> **LLM constructed Repair data from open-source code**
>
> Input:
>
> Problem Description:
>
> You are given a Verilog module that implements a 4-bit up counter. The counter should reset to zero when the reset signal is active and increment with each positive clock edge. However, there is a common error in this implementation: the counter_up register is not correctly initialized during reset, which may lead to unpredictable behavior after reset. Your task is to identify and fix this initialization error. Erroneous Implementation:
>
> ```verilog
> module up_counter(
>     input clk,
>     input reset,
>     output [3:0] counter
>     );
>     reg [3:0] counter_up;
>     // up counter
>     always @(posedge clk or posedge reset)
>     begin
>         if (reset)
>             counter_up <= 4'd3;   // Incorrect initialization value
>         else
>             counter_up <= counter_up + 4'd1;
>     end
>     assign counter = counter_up;
> endmodule
> ```
>
> In this erroneous implementation:
>
> * The counter_up register is incorrectly initialized to 4'd3 during reset, instead of being reset to zero (4'd0), which can cause incorrect counter behavior when reset is activated.
>
> Hints for Fixing:
>
> 1. Identify the initialization issue: The counter should reset to zero, but currently, it is being set to 4'd3 during reset. This causes incorrect counting behavior after a reset.
>
> 2. Fix the reset logic: Modify the if (reset) block in the always block to properly initialize the counter to zero. Change counter_up <= 4'd3; to counter_up <= 4'd0;.
>
> 3. Ensure proper reset behavior: After fixing the initialization, the counter will correctly reset to zero when the reset signal is active and increment correctly on each clock cycle thereafter.
>
> Output:
>
> ```verilog
> module up_counter(
>     input clk,
>     input reset,
>     output [3:0] counter
>     );
>     reg [3:0] counter_up;
>     // up counter
>     always @(posedge clk or posedge reset)
>     begin
>         if(reset)
>             counter_up <= 4'd0;
>         else
>             counter_up <= counter_up + 4'd1;
>     end
>     assign counter = counter_up;
> endmodule
> ```

Figure I.32: Another example based on multi_booth_8bit from RTLLM.

## I.4 Examples of Correct-by-Construction Data for Non-Textual Representations

### I.4.1 Karnaugh Maps and Truth Tables

Problem

```
Implement the circuit described by the Karnaugh map below.
//      c
// ab   0   1
// 00 | 1 | 0
// 01 | 0 | 0
// 11 | 0 | 0
// 10 | 0 | 0

module top_module(
        input a,
        input b,
        input c,
        output out
);
```

Solution

```
The input variables are: ['a', 'b', 'c'].
Based on the Karnaugh map, I can transform in to the following truth table:
 a | b | c | f
 0 | 0 | 0 | 1
 0 | 0 | 1 | 0
 0 | 1 | 0 | 0
 0 | 1 | 1 | 0
 1 | 0 | 0 | 0
 1 | 0 | 1 | 0
 1 | 1 | 0 | 0
 1 | 1 | 1 | 0

The minterms (when output is 1) are:
(0,0,0) => (~a & ~b & ~c)
This corresponds to the following minterms logic:
`(~a & ~b & ~c)`

Finally, based on the above logic equation, I can now write the Verilog code that could be described by the
     Karnaugh map:
```
```
module top_module(
        input a,
        input b,
        input c,
        output out
);
        assign out = (~a & ~b & ~c);
endmodule
```

Figure I.33: Example of a generated problem with Karnaugh Map representation in the problem description. We include detailed step-by-step analysis for reaching the correct solution.

## I.4.2 State Transition Graphs and Tables

**Problem**

Given the state–assigned table shown below, implement the logic functions Y[0] and z.
// Present state y[2:0] | Next state Y[2:0] x=0, Next state Y[2:0] x=1 | Output z
// 000 | 010, 011 | 1
// 001 | 100, 010 | 0
// 010 | 001, 100 | 1
// 011 | 011, 100 | 0
// 100 | 100, 001 | 0

**Solution**

The state transition is as follows:
// state | Next state in=0, Next state in=1 | Output
// A | C, D | 1
// B | E, C | 0
// C | B, E | 1
// D | D, E | 0
// E | E, B | 0

The transition logic is then:
A: next = x ? D : C;
B: next = x ? C : E;
C: next = x ? E : B;
D: next = x ? E : D;
E: next = x ? B : E;

The output is 1 for states: A, C.
Thus the output logic is: assign z = ( y == A || y == C );
Y0 corresponds to 001 (A), 011 (D).

Finally, below is the Verilog code for the finite state machine:
```
module top_module (
 input clk,
 input x,
 input [2:0] y,
 output reg Y0,
 output reg z
);
    reg [2:0] next_state;
    parameter A=0, B=1, C=2, D=3, E=4;
    always_comb begin
        case(y)

            A: next_state = x ? D : C;
            B: next_state = x ? C : E;
            C: next_state = x ? E : B;
            D: next_state = x ? E : D;
            E: next_state = x ? B : E;
            default: next_state = 'x;
        endcase
    end
    assign z = ( y == A || y == C );
    assign Y0 = ( next_state == B || next_state == D );
endmodule
```

Figure I.34: Example of problem with state transition tables.

## Problem

This is a Moore state machine with four states, four inputs, and one output. Implement this state machine in Verilog. Reset is an active–high synchronous reset to state D.

```
// D (out=0) ---in0=1---> D
// D (out=0) ---in0=0---> A
// C (out=0) ---in1=1---> D
// C (out=0) ---in1=0---> B
// B (out=1) ---in2=1---> D
// B (out=1) ---in2=0---> B
// A (out=0) ---in3=1---> B
// A (out=0) ---in3=0---> C
```

## Solution

The finite state machine has four inputs, and the state transition logic is as follows:
D: next = in0 ? D : A;
C: next = in1 ? D : B;
B: next = in2 ? D : B;
A: next = in3 ? B : C;

The output is 1 for states: B.
Thus the output logic is: `assign out = ( state == B );`.

Finally, below is the Verilog code for the finite state machine:
```
module top_module (
  input clk,
  input reset,
  input in0,
  input in1,
  input in2,
  input in3,
  output out
);
    parameter D=0, C=1, B=2, A=3;
    reg state;
    reg next;
    always_comb begin
        case(state)
            D: next = in0 ? D : A;
            C: next = in1 ? D : B;
            B: next = in2 ? D : B;
            A: next = in3 ? B : C;
            default: next = 'x;
        endcase
    end
    always @(posedge clk) begin
        if (reset) state <= D;
        else state <= next;
    end
    assign out = ( state == B );
endmodule
```

Figure I.35: Example of problem with Moore FSM.

## Problem

The following diagram is a Mealy machine. Implement in Verilog using one–hot encoding. Resets into state A and reset is asynchronous active–high.

// A ––x=0 (z=0)––> D
// A ––x=1 (z=1)––> C
// B ––x=0 (z=1)––> C
// B ––x=1 (z=0)––> B
// C ––x=0 (z=0)––> C
// C ––x=1 (z=0)––> D
// D ––x=0 (z=1)––> C
// D ––x=1 (z=0)––> B

## Solution

From the transition diagram, we have the following transition logic:
// state | next state in=0, next state in=1
// A | D, C
// B | C, B
// C | C, D
// D | C, B
Thus the state transition logic is as follows:
A: next = x ? C : D;
B: next = x ? B : C;
C: next = x ? D : C;
D: next = x ? B : C;
The output is 1 for states: (A, x), (B, ~x), (D, ~x).
Thus the output logic is: `assign z = ( ( state == A & x ) || ( state == B & ~x ) || ( state == D & ~x ) );`.

Finally, below is the Verilog code for the finite state machine:
```
module top_module (
        input clk,
        input areset,
        input x,
        output z
);
        parameter A=2'b00, B=2'b01, C=2'b10, D=2'b11;
        reg [1:0] state;
        reg [1:0] next_state;
        always_comb begin
                case(state)

                        A: next_state = x ? C : D;
                        B: next_state = x ? B : C;
                        C: next_state = x ? D : C;
                        D: next_state = x ? B : C;
                        default: next_state = 'x;
                endcase
        end
        always @(posedge clk, posedge areset) begin
                if (areset) state <= A;
                else state <= next_state;
        end
        assign z = ( ( state == A & x ) || ( state == B & ~x ) || ( state == D & ~x ) );
endmodule
```

Figure I.36: Example of problem with Mealy FSM.

---

**Problem**

The following is the state transition table for a Moore state machine with one input, one output, and four states. Use the following one-hot state encoding: A=4'b0001, B=4'b0010, C=4'b0100, D=4'b1000. Derive state transition and output logic equations by inspection assuming a one-hot encoding. Implement only the state transition logic and output logic (the combinational logic portion) for this state machine.
// state | Next state in=0, Next state in=1 | Output
// A | B, A | 0
// B | B, C | 1
// C | D, A | 1
// D | B, C | 0

**Solution**

Based on the state transition table, we can obtain the next state from observing the row (previous state) and column (input).
Next state is A on the following (row, column): (A, in=1) (C, in=1). This correspond to the following logic: `state[A] & in || state[C] & in`.
Next state is B on the following (row, column): (A, in=0) (B, in=0) (D, in=0). This correspond to the following logic: `state[A] & ~in || state[B] & ~in || state[D] & ~in`.
Next state is C on the following (row, column): (B, in=1) (D, in=1). This correspond to the following logic: `state[B] & in || state[D] & in`.
Next state is D on the following (row, column): (C, in=0). This correspond to the following logic: `state[C] & ~in`.

The output is 1 for states: B, C.
Thus the output logic is: `assign out = ( state[B] || state[C] );`.

Finally, below is the Verilog code for the finite state machine:
```
module top_module (
 input in,
 input [3:0] state,
 output reg [3:0] next_state,
 output out
);

    parameter A=0, B=1, C=2, D=3;

    assign next_state[A] = state[A] & in || state[C] & in;
    assign next_state[B] = state[A] & ~in || state[B] & ~in || state[D] & ~in;
    assign next_state[C] = state[B] & in || state[D] & in;
    assign next_state[D] = state[C] & ~in;

    assign out = ( state[B] || state[C] );

endmodule
```

Figure I.37: Example of problem with in-edge focused solution shown in Figure 5.6.

## I.4.3 Waveforms

### Problem

This is a combinational circuit. Read the simulation waveforms to determine what the circuit does, then implement it.

```
// time     a       b       c       d       q
// 0ns      0       0       0       0       0
// 5ns      0       0       0       0       0
// 10ns     0       0       0       0       0
// 15ns     0       0       0       0       0
// 20ns     0       0       0       1       0
// 25ns     0       0       1       0       0
// 30ns     0       0       1       1       0
// 35ns     0       1       0       0       0
// 40ns     0       1       0       1       0
// 45ns     0       1       1       0       0
// 50ns     0       1       1       1       0
// 55ns     1       0       0       0       1
// 60ns     1       0       0       1       1
// 65ns     1       0       1       0       0
// 70ns     1       0       1       1       1
// 75ns     1       1       0       0       0
// 80ns     1       1       0       1       1
// 85ns     1       1       1       0       0
// 90ns     1       1       1       1       0
```

### Solution

Based on the simulation waveform, I can transform in to the following truth table:

```
a | b | c | d | f
0 | 0 | 0 | 0 | 0
0 | 0 | 0 | 1 | 0
0 | 0 | 1 | 0 | 0
0 | 0 | 1 | 1 | 0
0 | 1 | 0 | 0 | 0
0 | 1 | 0 | 1 | 0
0 | 1 | 1 | 0 | 0
0 | 1 | 1 | 1 | 0
1 | 0 | 0 | 0 | 1
1 | 0 | 0 | 1 | 1
1 | 0 | 1 | 0 | 0
1 | 0 | 1 | 1 | 1
1 | 1 | 0 | 0 | 0
1 | 1 | 0 | 1 | 1
1 | 1 | 1 | 0 | 0
1 | 1 | 1 | 1 | 0
```

The minterms (when output is 1) are:
(1,0,0,0) ⇒ (a & ~b & ~c & ~d)
(1,0,0,1) ⇒ (a & ~b & ~c & d)
(1,0,1,1) ⇒ (a & ~b & c & d)
(1,1,0,1) ⇒ (a & b & ~c & d)

This corresponds to the following minterms logic:
`(a & ~b & ~c & ~d) | (a & ~b & ~c & d) | (a & ~b & c & d) | (a & b & ~c & d)`

Finally, based on the above logic equation, I can now write the Verilog code:
```
module top_module(
        input a,
        input b,
```

```
        input c,
        input d,
        output q
);
        assign q = (a & ~b & ~c & ~d) | (a & ~b & ~c & d) | (a & ~b & c & d) | (a & b
    & ~c & d);
endmodule
```

Figure I.38: Example of a waveform problem for combinational circuits.

## Problem

This is a sequential circuit. Read the simulation waveforms to determine what the circuit does, then implement it.

```
// time            clk             reset           in              out
// 0ns             0               1               0               x
// 5ns             1               1               0               1
// 10ns            0               1               0               1
// 15ns            1               0               0               1
// 20ns            0               0               0               1
// 25ns            1               0               1               0
// 30ns            0               0               1               0
// 35ns            1               0               0               1
// 40ns            0               0               0               1
// 45ns            1               0               0               0
// 50ns            0               0               0               0
// 55ns            1               0               0               1
// 60ns            0               0               0               1
// 65ns            1               0               1               0
// 70ns            0               0               1               0
// 75ns            1               0               1               1
// 80ns            0               0               1               1
// 85ns            1               0               0               1
// 90ns            0               0               1               1
```

... Simulation waveform redacted ...

## Solution

```
From the waveform, we have the following transition logic and output logic:
// state | Next state in=0, Next state in=1 | Output
// D | C, B | 1
// C | B, D | 0
// B | D, A | 1
// A | C, B | 0

Thus the state transition logic is as follows:
D: next = in ? B : C;
C: next = in ? D : B;
B: next = in ? A : D;
A: next = in ? B : C;

The output is 1 for states: D, B.
Thus the output logic is: `assign out = ( state == D || state == B );`.

Finally, below is the Verilog code for the finite state machine:
```

```
module top_module (
 input clk,
 input in,
 input reset,
 output out
);

        parameter D=0, C=1, B=2, A=3;
        reg state;
        reg next;

        always_comb begin
                case(state)

                        D: next = in ? B : C;
                        C: next = in ? D : B;
                        B: next = in ? A : D;
                        A: next = in ? B : C;
                        default: next = 'x;
                endcase
        end

        always @(posedge clk) begin
                if (reset) state <= D;
                else state <= next;
        end

        assign out = ( state == D || state == B );

endmodule
```

Figure I.39: Example of a waveform problem for sequential circuits.

## I.5  Prompt Templates

### I.5.1  Synthetic Data Generation

#### I.5.1.1  Self-Instruct

> Your goal is to create a high-quality Verilog problem.
>
> * Guidelines for designing the problem description:
>
> 1. This should be **completely self-contained**, providing all the contextual information one needs to understand and solve the problem.
>
> 2. Assume common verilog knowledge, but ensure that any specific context, variables, or code snippets pertinent to this problem are explicitly included.
>
> 3. Do not include the code snippet in the problem.
>
> 4. The problem should be desinged for the programmers to solve it with one verilog module.
>
> 5. The problem description section should be enclosed within <PROBLEM> </PROBLEM> tags.
>
> Now, Please use your creativity to create a brand new high-quality Verilog problem.

Figure I.40: Prompt used to generate initial 50 seed problems for Self-Instruct.

Your goal is to create a high-quality Verilog problem.

* Guidelines for designing the problem description:

1. This should be **completely self-contained**, providing all the contextual information one needs to understand and solve the problem.

2. Assume common verilog knowledge, but ensure that any specific context, variables, or code snippets pertinent to this problem are explicitly included.

3. Do not include the code snippet in the problem.

4. The problem should be desinged for the programmers to solve it with one verilog module.

5. The problem description section should be enclosed within <PROBLEM> </PROBLEM> tags.

Below shows some examples:

<PROBLEM>

{seed problems}

</PROBLEM>

Now, Please use your creativity to create a brand new high-quality Verilog problem.

Figure I.41: Prompt used for Self-Instruct.

### I.5.1.2 OSS-Instruct

---

Your goal is to create a high-quality Verilog problem.

* Guidelines for designing the problem description:

1. This should be **completely self-contained**, providing all the contextual information one needs to understand and solve the problem.
2. Assume common verilog knowledge, but ensure that any specific context, variables, or code snippets pertinent to this problem are explicitly included.
3. Do not include the code snippet in the problem.
4. The problem should be designed for the programmers to solve it with one Verilog module.

* Guidelines for the problem description format: The problem description section should be enclosed within <PROBLEM> </PROBLEM> tags.

Please increase the difficulty of the given programming test question a bit. You can increase the difficulty using, but not limited to, the following methods:

1. Your new problem should not be directly solved by the original code snippet.
2. You can also change the bit-width requiremnt, how to reset internal signals (if applicable), and whether the solution needs a clock signal (combinatorial versus sequential logic). If you do have a reset method that is synchronous to a clock, make sure to add the clock signal to the problem module input.
3. Add new constraints and requirements to the original problem, adding approximately 10 additional words.
4. Replace a commonly used requirement in the programming task with a less common and more specific one.
5. If the original problem can be solved with only a few logical steps, please add more reasoning steps.

Now, Please gain inspiration from the following random code snippet to create a high-quality Verilog problem.

Code snippet for inspiration:
```

{code snippet}
```

Output:

Figure I.42: Prompt used for OSS-Instruct. We also include prompts inspired from Evol-Instruct [81] to increase problem difficulty.

### I.5.1.3 Docu-Instruct

---

Your goal is to create a high-quality Verilog problem.

* Guidelines for designing the problem description:

1. This should be **completely self-contained**, providing all the contextual information one needs to understand and solve the problem.

2. Assume common verilog knowledge, but ensure that any specific context, variables, or code snippets pertinent to this problem are explicitly included.

3. Do not include the code snippet in the problem.

4. The problem should be designed for the programmers to solve it with one Verilog module.

* Guidelines for the problem description format: The problem description section should be enclosed within <PROBLEM> </PROBLEM> tags.

Now, Please gain inspiration from the following textbook or wikipedia snippet to create a high-quality Verilog problem. The information might not be directly related to Verilog, but try to be make the problem as relevant as possible to the textbook issue discussed.

Textbook snippet for inspiration:
```
{document snippet}
```

Output:

---

Figure I.43: Prompt used for Docu-Instruct with Wikipedia and textbooks.

I am going to give you a concept and some descriptions about that concept. Based on the descriptions and concept name, determine if the concept belongs to one of the following categories:

- Hardware description and modeling in Verilog.

- Fundamental constructs such as modules, ports, and wires specific to Verilog.

- Synthesis and optimization techniques employed in hardware design using Verilog.

- Simulation tools and methodologies for verifying Verilog-based hardware designs.

- Common design patterns and best practices in Verilog for efficient hardware implementation.

- Programming concepts like loops, functions related to Verilog.

- Hardware related concepts such as finite state machines that could be implemented in Verilog.

- Algorithms that could be implemented in hardware, such as Fourier Transforms.

Concept: {Wikipedia title}

Description: {Wikipedia content}

Do not make assumptions and only respond "Yes" if you are certain that the {Wikipedia title} is related to hardware design or Verilog coding language.

Your answer should start with "Yes" or "No".

Figure I.44: Prompt used to filter Verilog related Wikipedia pages.

### I.5.1.4 Non-textual Representations

Your goal is to create a high-quality Verilog problem. Specifically, we would like to test the skills of understanding Karnaugh maps and state transition diagrams. The problem description section should be enclosed within <PROBLEM> </PROBLEM> tags.

Now, please gain inspiration from the following random code snippet to create a high-quality Verilog problem. Remember that the problem you generated must include Karnaugh maps in the format above. The random code snippet MUST be related to the solution. Your problem statement should be short and succinct (no more than 5 sentences) and you MUST generate a Karnaugh map in the problem description. Your problem description should not describe the Karnaugh map in words and should assume that the student need to decipher the Karnaugh map to solve the problem.

Code snippet for inspiration:

```

{code snippet}

```

Below are two examples on how to represent Karnaugh map related questions in purely textual format. You should NOT use the following to generate the problem but only consider the style.

```
<PROBLEM>
Given the state-assigned table shown below, implement the finite-state machine. Reset should synchronous active
    high reset the FSM to state 000.
// Present state y[2:0] | Next state y[2:0] x=0, Next state y[2:0] x=1, Output z
// 000 | 000, 001 | 0
// 001 | 001, 100 | 0
// 010 | 010, 001 | 0
// 011 | 001, 010 | 1
// 100 | 011, 100 | 1
</PROBLEM>
<PROBLEM>
Implement the circuit described by the Karnaugh map below.
//         a
// bc   0 1
// 00 | 0 | 1 |
// 01 | 1 | 1 |
// 11 | 1 | 1 |
// 10 | 1 | 1 |
```

</PROBLEM>

Figure I.45: An prompt example to encourage LLMs to generate questions with Karnaugh Maps.

## I.5.1.5 Prompts for Sampling Solutions with LLM Generated Problems

Your goal is to solve a Verilog problem by completing one verilog module.

* Guidelines for writing the solution (verilog module) and formatting:

1. The module should be **completely self-contained**, fulfilling all the requirements needed to solve the problem.
2. Ensure the correctness of the syntax and functionality.
3. The module name must be "top_module".
4. Do not add blackslash in front of underscore symbol.
5. The solution must be valid Verilog code.
6. The solution section should be enclosed within <SOLUTION> </SOLUTION> tags.
7. The reasoning section should be enclosed within <REASON> </REASON> tags.
8. Write comments in the solution section. Try to embed the reasoning in comments. Always try to write the corresponding comments before the code.
9. The module should be **completely self-contained**, do not include or import outside the module and define everything inside the module.

Below shows an example:

Problem description:

```
Build a counter that counts from 0 to 999, inclusive, with a period of 1000 cycles. The reset input is active high
    synchronous, and should reset the counter to 0. Solve the problem by completing the following module.
```

Output:

```
<REASON>
To solve the problem of building a counter that counts from 0 to 999 with a reset input that is active high
    synchronous, you will need a 10-bit register (since 2^10=1024 and 999 < 1024). The counter should increment
    on each clock cycle and wrap around back to 0 when it reaches 1000. The reset input, when active, should
    reset the counter to 0 on the next clock edge.
</REASON>
```

Here is the Verilog code for the module:
```
<SOLUTION>
module top_module(
    input clk,
    input reset,
    output reg [9:0] q);

    // This block executes on the positive edge of the clock signal.
    always @(posedge clk) begin
        // Checks if the reset is active. This is synchronous with the clock.
        if (reset) begin
            // Reset the counter to 0 when reset is active
            q <= 0;
        // If the counter has reached 999, it needs to wrap around to 0 on the next cycle.
        end else if (q == 999) begin
            // Reset to 0 after reaching 999
            q <= 0;
        // In all other cases, just increment the counter.
        end else begin
            // Increment the counter
            q <= q + 1;
        end
    end
endmodule
</SOLUTION>
```

> Now, please solve the following Verilog problem. I will also attach a reference code snippet which was used as an inspiration to generate the problem. The provided code may not directly solve the problem so you should use it only as a reference.
>
> Reference code:
>
> ```
>
> {code snippet}
>
> ```
>
> Problem description:
>
> ```
>
> {in context examples}
>
> ```
>
> Output:

Figure I.46: Prompt used for sampling solutions for synthetic data generation. We include a in context example to encourage models to include reasoning traces. Prompts in blue are only included for problems generated from a code snippet.

### I.5.1.6 Prompts for Verifying Solutions

> Check if the given Verilog module is a valid solution to the problem. The output should be in "True" or "False" and be enclosed within <VALID> </VALID> tags and the explanation in <REASON></REASON> tags.
>
> Now check the following:
>
> <PROBLEM>
>
> {problem}
>
> <PROBLEM>
>
> <SOLUTION>
>
> {solution}
>
> </SOLUTION>

Figure I.47: Prompt used for verifying solutions.

## I.5.2 Prompts for Targeted Code Repair

### I.5.2.1 Error Report

---

Here is a Verilog problem description:
```
{problem description}
```

Here is an erroneous implementation:
```
{error code}
```

Here is a correct implementation:
```
{correct code}
```

Generate a detail error report.
The error report should describe the common error type and output the code category.
The error report should also be detailed enough to let beginners to repair the erroneous implementation step by step.

Output:

---

Figure I.48: Prompt for Error Report generation.

> Here is a Verilog problem description:
> ```
> {problem description}
> ```
>
> Here is an erroneous implementation:
> ```
> {error code}
> ```
>
> Here is the error report:
> ```
> {error report}
> ```
>
> Now fix the erroneous implementation and give me the correct code.
>
> Output:

Figure I.49: Prompt for Error Report self-consistency validation. The generated code fix will be evaluated for functional correctness. Error reports whose code fixes do not pass will be filtered.

## I.5.2.2  Error Injection

Your goal is to create an error-fixing Verilog practice problem for programmers. You will demonstrate a type of error that is commonly made by programmers.

Create an error repair practice problem with three components:

1. Problem description
2. Erroneous implementation
3. Hints for fixing

Here is an example:

<EXAMPLE>

The following Verilog module is intended to implement the specification below. However, there is a bug in the code which causes incorrect results. Please fix the bug to make the module work as intended.

Erroneous Implementation:

```verilog
// Verilog code with the injected error
module example_module (
    input wire clk,
    input wire reset,
    output reg [3:0] counter
);

// Intended functionality:
// This module should count from 0 to 15 and then wrap around.

always @(posedge clk or posedge reset) begin
    if (reset) begin
        counter <= 4'b0000;
    end else begin
        counter <= counter + 1'b1; // Error injected: Should be 4'b1
    end
end

endmodule
```

Hints for Fixing:

1. Verify the bit-width of the counter and the increment operation.
2. Check the initialization and wrapping condition of the counter.
3. Ensure that the addition operation correctly handles the 4-bit counter.

</EXAMPLE>

Now, here is the commonly made error:

```

{error report}

```

Inject the above error into the following module and create an error repair practice problem. Check if it is possible to inject the error. If not, create the problem with the given error alone and ignore the module in the code snippet.

```

{code snippet}

```

Output:

Figure I.50: Prompt used to inject targeted errors to open-source code in code Repair data. We also prompt the LLM to self-verify if the error could be injected to the code snippet.

**Karnaugh Maps and
Truth Tables**

**Step1. Sample
Configurations**

**Sample random minterms**
variables=['a','b','c'], minterms=[1, 2, 5], don't_cares=[7]
SOP form: (~a & ~b & c) | (~a & b & ~c) | (a & ~b & c)

**Step2. Construct
Representations
and Problems**

**Truth table**
a | b | c | f
0 | 0 | 0 | 0
0 | 0 | 1 | 1
0 | 1 | 0 | 1
0 | 1 | 1 | 0
1 | 0 | 0 | 0
1 | 0 | 1 | 1
1 | 1 | 0 | 0
1 | 1 | 1 | x

**Karnaugh map**
  bc
a 00  01 11 10
0 | 0 | 1 | 0 | 1
1 | 0 | 1 | x | 0

**Step3. Construct
Solution**

```
module top_module(
    input a,
    input b,
    input c,
    output f
);
    assign f = (~a & ~b & c) | (~a & b & ~c) | (a & ~b & c)
endmodule
```

Figure I.51: Karnaugh map and truth table generation using the correct-by-construction methodology. The approach ensures mathematical consistency between problems and their solutions.

**State Transition Graphs
and Tables**

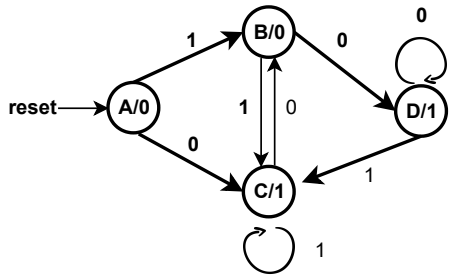

**Step1. Sample
Configurations**

Construct random legal transition graphs

**Step2. Construct
Representations
and Problems**

State transition graph.

// A (out=0) --in=0--> C
// A (out=0) --in=1--> B
// B (out=0) --in=0--> D
// B (out=0) --in=1--> C
// C (out=1) --in=0--> B
// C (out=1) --in=1--> C
// D (out=1) --in=0--> D
// D (out=1) --in=1--> C

State transition table.

// state | in=0, in=1 | Output
// A | C, B | 0
// B | D, C | 0
// C | B, C | 1
// D | D, C | 1

**Step3. Construct
Solution**

```
always_comb begin
    case (state)
        A: next = in ? B : C;
        B: next = in ? C : D;
        C: next = in ? C : B;
        D: next = in ? C : D;
    endcase
end

assign out = (state==C) | (state==D);
```

Figure I.52: Finite-state machine representation generated with correct-by-construction techniques. The methodology ensures consistent state transitions and output relationships in generated problems.

**Waveforms**

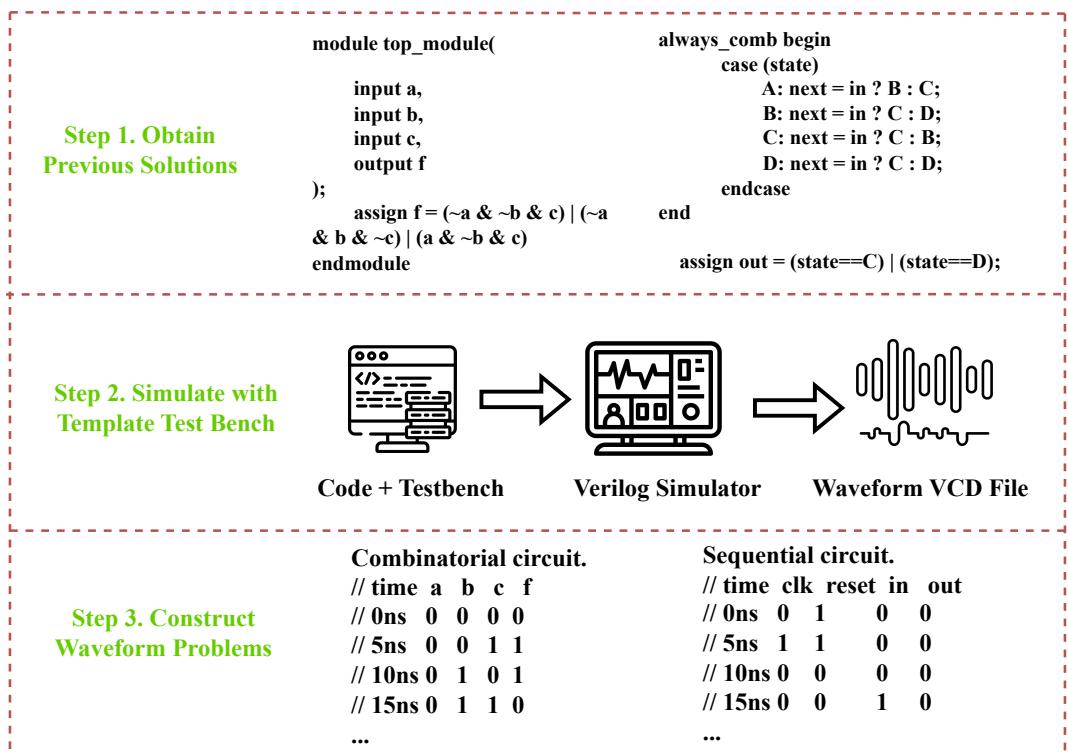

Figure I.53: Waveform diagrams constructed using correct-by-construction methodology. The approach generates signal patterns with guaranteed timing relationships and logical consistency.

# Appendix J — Details of Efficient Code Data Pruning

## J.1 Code Samples from Data Pruning

We show examples from our data pruning. The selected data samples adheres closely to the downstream coding tasks, from English problem description to code generation. We also provide an example of removed data from our pruning strategy.

| Model | Training Tokens | Benchmark | | Improvement Over Base | |
|---|---|---|---|---|---|
| | | HumanEval (+) | MBPP (+) | HumanEval (+) | MBPP (+) |
| GPT-3.5 Turbo | - | 72.6 (65.9) | 81.7 (69.4) | - | - |
| GPT-4 Turbo | - | 85.4 (81.7) | 83.0 (70.7) | - | - |
| DeepSeek-Coder-Base | - | 47.6 (39.6) | 70.2 (56.6) | - | - |
| DeepSeek-Coder-Instruct | 2B | 73.8 (70.1) | 72.7 (63.4) | 26.2 (30.5) | 2.5 (6.8) |
| Magicoder-DS | 90M | 66.5 (60.4) | 75.4 (61.9) | 18.9 (20.8) | 5.2 (5.3) |
| Magicoder$\mathcal{S}$-DS | 240M | 76.8 (70.7) | 75.7 (64.4) | 29.2 (31.1) | 5.5 (7.8) |
| Ours (full data) | 234M | 74.3 (70.8) | 74.5 (62.3) | 26.7 (31.2) | 4.3 (5.7) |
| Ours (90%) | 192M | 77.0 (71.6) | 76.9 (64.0) | 29.4 (32.0) | 6.7 (7.4) |
| Ours (50%) | 106M | 71.0 (64.0) | 78.0 (64.0) | 23.4 (24.4) | 7.8 (7.4) |
| Ours (10%) | 21M | 70.4 (65.0) | 73.0 (60.2) | 22.8 (25.4) | 2.8 (3.6) |
| Ours (1%) | 2M | 64.6 (58.0) | 74.3 (61.9) | 17.0 (18.4) | 4.1 (5.3) |

Table J.17: pass@1 (%) performance on HumanEval (+) and MBPP (+) benchmarks using greedy decoding. Results for baseline models are obtained from prior studies [40, 152]. The reported values for this work are derived using the HDB-SCAN clustering algorithm combined with a diversity-based pruning metric. All results represent the mean of three independent evaluation runs using EvalPlus [69], accounting for variability in both clustering and fine-tuning processes.

---

Input
You are given a Python function that takes in a list of integers and a target integer. The function is intended to find two distinct elements in the list that sum up to the target integer. However, the function is incomplete and needs to be filled in with the appropriate code to achieve this functionality.
Complete the Python function 'find_two_sum' to find and return the indices of two distinct elements in the input list that sum up to the target integer. If no such pair exists, return an empty list. The function signature is:
def find_two_sum(nums: List[int], target: int) -> List[int]:

For example, given the input 'nums = [2, 7, 11, 15]' and 'target = 9', the function should return '[0, 1]' since 'nums[0] + nums[1] = 2 + 7 = 9'.

Output

```python
from typing import List

def find_two_sum(nums: List[int], target: int) -> List[int]:
    num_indices = {}
    for i, num in enumerate(nums):
        complement = target - num
        if complement in num_indices:
            return [num_indices[complement], i]
        num_indices[num] = i
    return []
```

---

Figure J.54: Example of a selected data sample.

## J.2 Pruning Metrics

Diversity Metric

A distance-based method is adopted to quantify diversity, where each sample is assigned a diversity score defined as:

$$d_i = \min_{x \in \mathcal{K} \setminus \{x_i\}} \text{dist}(x_i, x), \tag{J.1}$$

in which $x_i$ denotes the embedding of the current sample, $\mathcal{K}$ represents the selected query set within a given cluster, and dist is the distance function. The score $d_i$ reflects the dissimilarity between $x_i$ and its nearest neighbor in the set. Given that embeddings are normalized prior to pruning, the dot product is used as the distance metric.

Density Metric

Kernel Density Estimation (KDE) is employed to measure the density of each sample within the embedding space, providing a non-parametric estimate of the underlying probability distribution. For a sample $x_i$, the density score is computed as:

$$\rho(x_i) = \frac{1}{nh^d} \sum_{j=1}^{n} K\left(\frac{x_i - x_j}{h}\right), \quad (J.2)$$

where $K$ is the kernel function (typically Gaussian), $h$ denotes the bandwidth, $d$ is the dimensionality of the feature space, and $n$ is the total number of samples. Higher density scores indicate that a sample lies in a region populated by many similar instances, suggesting lower value for preserving dataset diversity.

Random Selection

As a baseline, a random selection strategy is implemented. Samples are drawn uniformly at random either from individual clusters or from the entire training dataset (in the absence of clustering) for instruction tuning purposes.

| Embedding Input Type | MBPP (+) | HumanEval (+) |
|---|---|---|
| Instruction + Code | 76.3 (62.5) | 73.1 (69.6) |
| Instruction Only | 74.0 (63.7) | 69.1 (63.6) |
| Code Only | 74.1 (62.7) | 69.2 (63.3) |

Table J.18: Effect of input feature type on *pass@1* scores at a 50% compression ratio using KMeans clustering. Combined instruction and code embeddings yield the best performance across both MBPP and HumanEval benchmarks. Scores in parentheses indicate results on the original benchmarks, while main values correspond to their augmented versions (MBPP+ and HumanEval+).

Input
You are given a list of integers representing the daily stock prices of a company. You need to write a function to calculate the maximum profit that can be obtained by buying and selling the stock at most once. If it is not possible to make any profit, return 0.
Write a function 'maxProfit' that takes in a list of integers 'prices' as input and returns the maximum profit that can be obtained.
Function signature:

```python
def maxProfit(prices: List[int]) -> int:
```

Example:

Input: prices = [7, 1, 5, 3, 6, 4]
Output: 5
Explanation: Buy on day 2 (price = 1) and sell on day 5 (price = 6), profit = 6 - 1 = 5.

Input: prices = [7, 6, 4, 3, 1]
Output: 0
Explanation: In this case, no transaction is done, i.e., max profit = 0.

Assumptions:
- The input list 'prices' will have at least one element.
- The stock prices are given in chronological order.

Output

```python
from typing import List

def maxProfit(prices: List[int]) -> int:
    if not prices:
        return 0

    min_price = prices[0]
    max_profit = 0

    for price in prices:
        if price < min_price:
            min_price = price
        else:
            max_profit = max(max_profit, price - min_price)

    return max_profit
```

The 'maxProfit' function iterates through the list of prices, keeping track of the minimum price encountered so far and the maximum profit that can be obtained. It updates these values as it iterates through the list and returns the maximum profit at the end.

Figure J.55: Another example of a selected data sample.

```
Input
Background:
You are a Product Manager, named Alice, your goal is Efficiently create a successful product, and the
↪  constraint is .
Question:

# Context
## Original Requirements
[BOSS: Write a podcast search engine which search for all index podcast. Each podcast belongs to a
↪  channel and each podcast contain title, description, published, audio link, transcript text]

## Search Information
### Search Results
### Search Summary

## mermaid quadrantChart code syntax example. DONT USE QUOTO IN CODE DUE TO INVALID
↪  SYNTAX. Replace the <Campain X> with REAL COMPETITOR NAME

quadrantChart
    title Reach and engagement of campaigns
    x-axis Low Reach --> High Reach
    y-axis Low Engagement --> High Engagement
...

Output

## Original Requirements
We are tasked with creating a podcast search engine that indexes all podcasts. Each podcast is associated
↪  with a channel and contains a title, description, publication date, audio link, and transcript text.

## Product Goals
```python
[
    "Create a user-friendly podcast search engine that indexes all podcasts",
    "Ensure each podcast contains necessary information: title, description, publication date, audio link,
    ↪   and transcript text",
    "Develop a system that efficiently updates and maintains the podcast index"
]
```

## User Stories
```python
[
    "As a user, I want to search for podcasts by title or description so that I can find the content I'm
    ↪   interested in",
    "As a user, I want to see all the necessary information about a podcast (title, description, publication
    ↪   date, audio link, transcript text) so that I can decide whether to listen to it",
    "As a user, I want the search engine to suggest podcasts based on my search history so that I can
    ↪   discover new content",
    "As a user, I want the podcast search engine to be updated regularly so that I can access the latest
    ↪   podcasts",
    "As a user, I want to filter search results by publication date, channel, or transcript text so that I
    ↪   can narrow down my search"
]
```

## Competitive Analysis
...
```

Figure J.56: Example of a removed data sample (outlier).

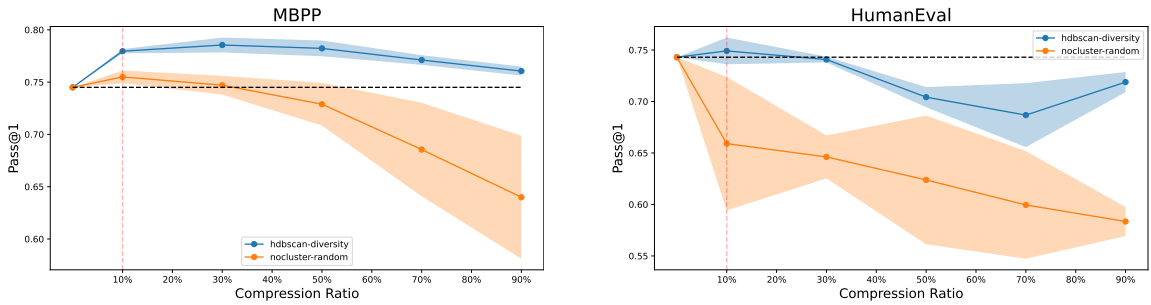

Figure J.57: Comparison of performance between the HDBSCAN-diversity strategy and the nocluster-random baseline across multiple benchmarks. The proposed method consistently achieves superior results across all datasets. On MBPP, it matches or exceeds the performance of models trained on the full dataset, despite utilizing only 10% of the data. The *pass*@1 scores, plotted against different compression ratios, highlight the robustness and effectiveness of the pruning approach. Greater variance observed on HumanEval is attributed to its limited number of problem instances.

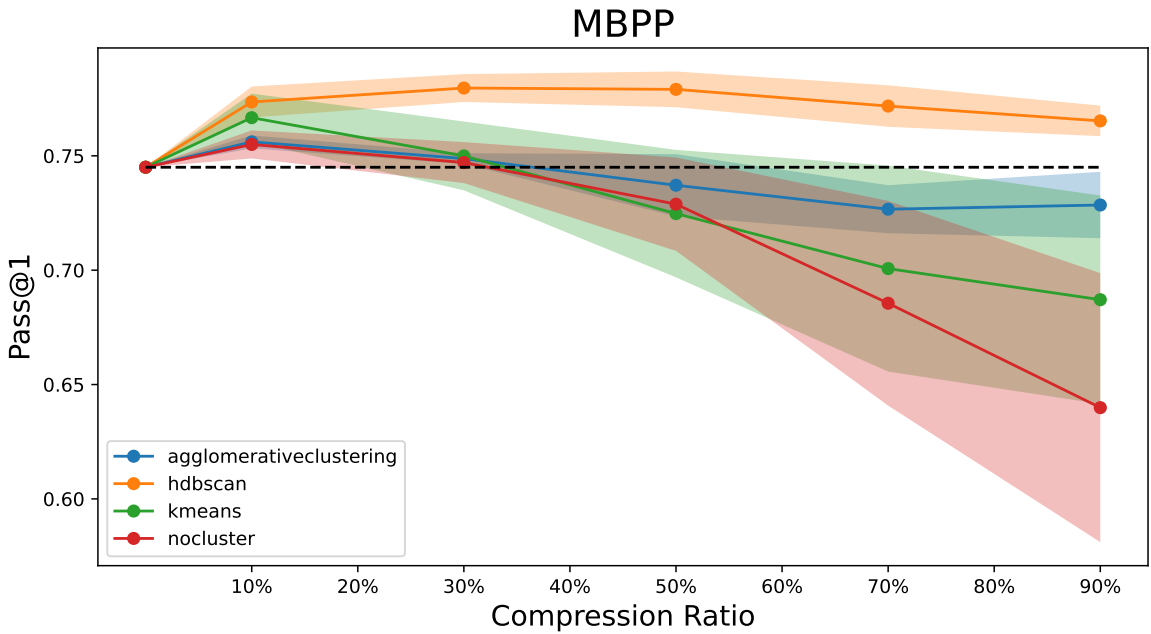

Figure J.58: Performance comparison of different clustering algorithms on the MBPP benchmark across various compression ratios. HDBSCAN demonstrates superior robustness, consistently maintaining higher *pass*@1 scores compared to other algorithms, particularly at high compression ratios where it outperforms even the full dataset baseline at 90% compression.

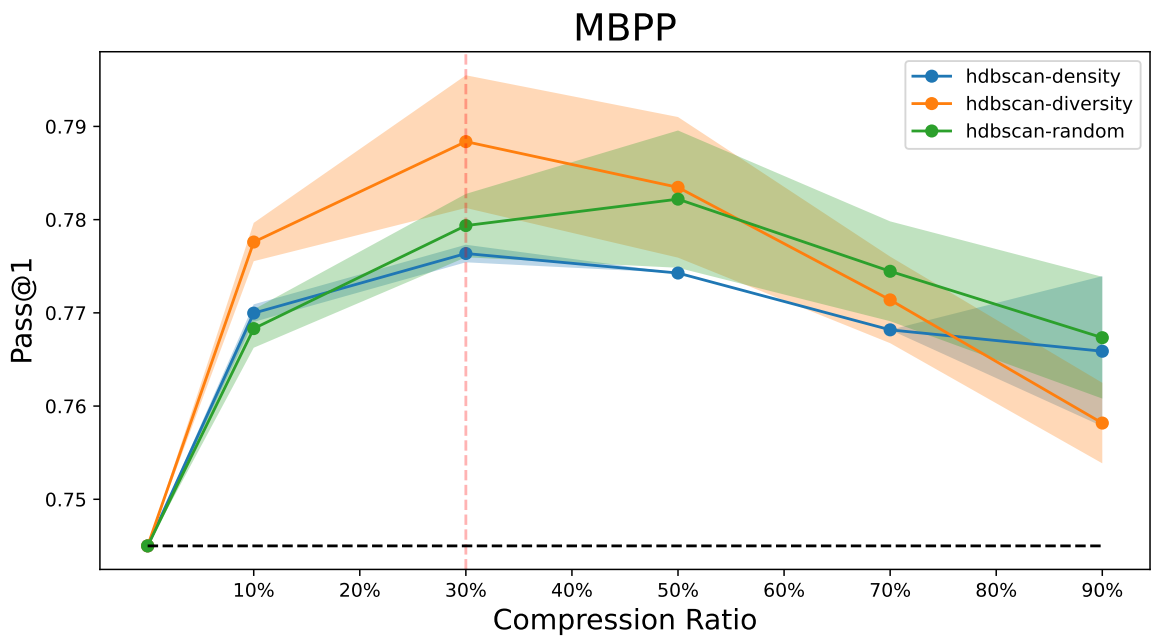

Figure J.59: Comparative analysis of different pruning metrics using HDBSCAN clustering on the MBPP benchmark. While the Diversity metric shows slight advantages over Random and Density metrics in the 10-40% compression range, performance differences between metrics are minimal, suggesting that clustering algorithm selection has greater impact than the specific pruning metric employed.

# 國立臺灣大學研究所博、碩士學位考試試卷(口試紀錄表)

__113__ 學年度第 __2__ 學期　　　　　　　　考試日期：114 年 4 月 28 日

系所組別：　__資訊工程學系　博士班__　　　考試地點：線上

學號：__F08946007__

姓名：__蔡　昀　達__

**學位考試成績：■A+ □A □A- □B+ □B □B- □C+ □C □C- □F □X**
(請勾選成績，塗改請簽章。成績評量定義詳見下列說明，研究生及格標準為 B-)

論文題目：　__賦能大型語言模型多領域資源挑戰__

※本委員會確認學位論文是否符合專業領域　■是　□否

考試委員簽章：

| Signature: 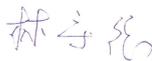 | Signature: 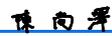 |
| --- | --- |
| Email: sdlin@csie.ntu.edu.tw | Email: stchen@csie.ntu.edu.tw |
| Signature: 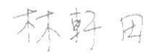 | Signature: 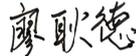 |
| Email: htlin@csie.ntu.edu.tw | Email: d05922001@ntu.edu.tw |
| Signature: 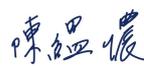 | |
| Email: yvchen@csie.ntu.edu.tw | |

※成績評量定義：Definition of Grades

```
A+：所有目標皆達成且超越期望（All goals achieved beyond expectation）
A ：所有目標皆達成（All goals achieved）
A-：所有目標皆達成，但需一些精進（All goals achieved, but need some polish）
B+：達成部分目標，且品質佳（Some goals well achieved）
B ：達成部分目標，但品質普通（Some goals adequately achieved）
B-：達成部分目標，但有些缺失（Some goals achieved with minor flaws）
C+：達成最低目標（Minimum goals achieved）
C ：達成最低目標，但有些缺失（Minimum goals achieved with minor flaws）
C-：達成最低目標但有重大缺失（Minimum goals achieved with major flaws）
F ：未達成最低目標（Minimum goals not achieved）
X ：因故不核予成績（Not graded due to unexcused absences or other reasons）
```

附表一　等第制與百分制單科成績對照表

| 等第制成績 | 等第績分 | 百分制分數區間 | 百分制分數 |
| --- | --- | --- | --- |
| A+ | 4.3 | 90-100 | 95 |
| A | 4.0 | 85-89 | 87 |
| A- | 3.7 | 80-84 | 82 |
| B+ | 3.3 | 77-79 | 78 |
| B | 3.0 | 73-76 | 75 |
| B-（研究生及格標準） | 2.7 | 70-72 | 70 |
| C+ | 2.3 | 67-69 | 68 |
| C | 2.0 | 63-66 | 65 |
| C-（學士班及格標準） | 1.7 | 60-62 | 60 |
| F | 0 | 59(含)以下 | 50 |
| X | 0 | 0 | 0 |